\documentclass[11pt]{article}

\usepackage[preprint]{acl}

\usepackage{times}
\usepackage{latexsym}
\usepackage{tabularray}
\usepackage{booktabs}
\UseTblrLibrary{booktabs}
\usepackage{longtable}
\usepackage[table,xcdraw]{xcolor}
\usepackage{hyperref}

\usepackage{graphicx}
\usepackage{subcaption}

\usepackage[T1]{fontenc}

\usepackage[utf8]{inputenc}

\usepackage{microtype}

\usepackage{inconsolata}

\usepackage{graphicx}

\usepackage{amssymb,amsmath,xspace}
\usepackage{cleveref}

\newcommand{\eg}{\emph{e.g.}\xspace}
\newcommand{\ie}{\emph{i.e.}\xspace}

\newcommand{\etc}{\emph{etc.}\xspace}
\newcommand{\wrt}{wrt\xspace}

\title{A Tale of Two Error Categories: Exploring Concealed Trade-Offs in the Errors of Automated Judges in Evaluation of Uncertainty Quantifiers}

\author{Evgenia Ilia \\
  University of Amsterdam \\
  \texttt{e.ilia@uva.nl} \\\And
  Wilker Aziz \\
  University of Amsterdam \\
  \texttt{w.aziz@uva.nl} }

\begin{document}
\maketitle
\begin{abstract}
The wide adoption of LLMs across broad NLG applications heightens the importance of providing users with the means to  
avert errors and hallucinations.
Uncertainty quantification is poised to fill that gap; with low uncertainty (high confidence), as a proxy for correctness, 
allowing users to be selective (\eg, reject low-confidence, likely incorrect responses).
Correlation between confidence and correctness then serves as a useful criterion for 
evaluation of uncertainty quantifiers (UQs). 
But in NLG, where diverse responses can be adequate to a prompt, obtaining reliable correctness judgements is not simple, especially without human intervention.
Errors in automated judgement are hardly avoidable and  known to diminish the reliability of evaluation protocols \citep{santilli-etal-2025-revisiting, ielanskyi2025addressing}. 
In a meta-analysis of published work, we show that automated judgement is the present norm. Besides, automated judgements are rarely validated against human ones, and the validation of the UQ evaluation they automate is even rarer.
With experiments in question answering, using 4 LLMs, human and automated judgements and 7 popular UQs, 
 we find that i) a judge's performance can only coarsely predict the observed impact of its errors on the reliability of UQ evaluation, and that ii) judgement errors tend to misrepresent informative UQs most. We link these observations to patterns of correlation between confidence and categories of judgement error.

\end{abstract}

\begin{figure}[h!]
\centering
\includegraphics[width=\linewidth]{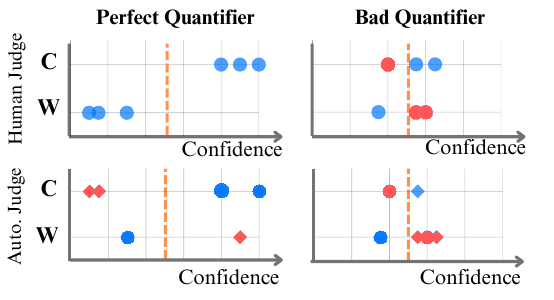}

\caption{An informative UQ (top-left) assigns higher confidence to responses that humans deem correct (C) than to those that humans deem wrong (W), allowing a decision maker to be selective (\eg, by rejecting lower-confidence responses \wrt a threshold (dashed line)). 
If assigned confidences do not neatly separate  
Cs from Ws (top-right), the UQ leads to selection errors (shown in red). 
The degree of separation 
gives us a criterion to choose amongst UQs. But when judgement is automated (bottom), as routinely done in UQ evaluation, judgement errors (C$\to$W or W$\to$C; shown as diamonds) distort our view of the relative merits of the UQs. This affects informative UQs most: on the left, we go from 0/6 to 3/6 selection errors (2 rejected Cs and 1 accepted W), while on the right we go from 3/6 to 4/6 errors.} 

\label{fig:explanation}
\end{figure}

\section{Introduction}

While modern LMs have undeniably advanced the state-of-the-art of NLG applications, even large LMs still err and hallucinate \citep{singh2025openai, comanici2025gemini, yang2025qwen3,  manakul-etal-2023-selfcheckgpt, ji2023survey}, often with no appreciable disclaimers \citep{yona-etal-2024-large, wang2026llm}. Uncertainty quantifiers (UQs), which qualify generated responses with a notion of uncertainty (or confidence), have been introduced to close this gap \citep{fomicheva-etal-2020-unsupervised, malinin2021uncertainty, lin-etal-2024-contextualized, kadavath2022language}. An informative UQ is predictive of correctness, assigning low uncertainty (high confidence) to responses that are likely to be correct and high uncertainty (low confidence) otherwise. It therefore provides information that a selective decision maker can use to avoid errors (\eg, by rejecting low-confidence responses)---see \Cref{fig:explanation} (top).

To establish the relative merits of uncertainty quantifiers---the goal of UQ evaluation---we measure one or another notion of `correlation' between correctness and uncertainty in generation. The most common strategies quantify errors in selective prediction \citep{varshney-etal-2022-towards, zablotskaia-etal-2023-uncertainty}. 
In NLG, however, a plethora of factors (\eg, paraphrastic variation, task open-endedness, input under-specification, user perspectives, \etc) render the variability in responses that are deemed plausible \wrt a given prompt impossible to represent in a reference set, making \emph{correctness}  difficult to ascertain automatically \citep{ novikova-etal-2017-need, plank-2022-problem, celikyilmaz2020evaluation}. 
In effect, human judgement is the most reliable evaluation signal in NLG, but, in practice, it is routinely replaced by automated systems \citep{zhou-etal-2022-deconstructing}. 
Such proxies to NLG evaluation distort our views of errors in selective prediction (SP)---see \Cref{fig:explanation} (compare top and bottom)---and diminish the reliability of our practical protocols for UQ evaluation \citep{santilli-etal-2025-revisiting, ielanskyi2025addressing}.
In this work, we document  the widespread neglect of human validation of these automated protocols, study the complex relationship between judgement errors and selective prediction errors, and seek to vindicate the role of human validation in UQ evaluation.

We begin by mapping out the UQ evaluation practices that have become common within the NLP community in recent years, particularly with regards to the automation of correctness judgement and the validation of this practice in the specific context of UQ evaluation. 
We perform a meta-analysis of published ACL papers within the uncertainty in NLG field and find that a variety of judges are employed to approximate correctness (\eg, LLM as a judge, overlap-based and semantic-similarity-based metrics) but these are rarely validated against human judgement ($\sim23\%$ of studied papers, as per our meta-analysis). 
In experiments situated within question answering (TriviaQA; \citet{joshi-etal-2017-triviaqa} and AmbigQA; \citet{min-etal-2020-ambigqa}) using 4 LLMs (Llama-8B/3B; \citet{grattafiori2024llama} and Qwen-7B/0.5B; \citet{hui2024qwen2}), we assess a representative range of automated judges by comparing their predictions to those of human judges. This reveals that judgement quality varies
 across settings and is highly sensitive to hyperparameters (\eg, similarity cutoffs). 
We then characterise how the errors of the automated judges impact UQ evaluation: we use each one of 7 UQs to parameterise a selective decision maker, and rank these UQs in terms of AUROC (powered by either human or automated judgement).
We observe how judges' quality only coarsely relates with the quality of the UQ evaluation they induce, 
probing us to dig deeper into the relationship between the error patterns of automated judges and the patterns UQs' scores exhibit.
We indeed identify a non-trivial impact in the relationship between judges' error types and the distribution of confidences---implying that an optimal choice of judge  requires its validation \emph{within} the UQ evaluation protocol.

Overall, judgement errors have non-trivial effects on UQ evaluation, often misrepresenting informative quantifiers. We caution against the careless adoption of judges without validation of their effectiveness within UQ evaluation. To that end, we introduce data analysis tools that help identify high quality judges and analyse their errors’ tradeoffs, to support more informed decision making.\footnote{Code, generations and annotations at \url{https://github.com/evgeniael/auroc_eval}.}

\section{Background}

\subsection{Uncertainty Quantifiers}

An LM can be regarded as a tool to represent probabilistic beliefs about responses given a prompt \citep{du-etal-2023-measure,du-etal-2024-language}. These beliefs can be assessed in at least two ways: i) by assuming that an underlying probability mass function (pmf) is prescribed by the autoregressive factorisation of the model, 
or ii) by letting the token-level distributions parameterise an arbitrary choice of sampling algorithm \citep{pmlr-v97-kool19a,Holtzman2020The,meister-etal-2023-locally}, in which case a pmf exists but is unknown. 
Sampling is tractable in both views, making Monte Carlo (MC) estimation of probabilities (as well as entropy and other quantities) viable.\looseness=-1 

The vast majority of UQs is powered by one of these two views. 
Quantities derived from the probability assigned to a response are seen as the model's confidence in that prediction (\eg, log-probability of response  \citep{jiang-etal-2021-know, xiao-wang-2021-hallucination} or its sequence-length-normalised version  \citep{manakul-etal-2023-selfcheckgpt, bakman-etal-2024-mars, flores-etal-2025-improving}).
Degrees of variation in the conditional distributions (at sequence- or token-level), such as Shannon entropy, are viewed as uncertainty signals \citep{fomicheva-etal-2020-unsupervised,xiao-wang-2021-hallucination, duan-etal-2024-shifting}.
Others, first push the LM distribution forward through a transformation (\eg, by analysing sampled outcomes in terms of some of their attributes) and assess notions of confidence or uncertainty \wrt this transformed distribution: 
a popular choice being semantic variation \citep{kuhn2023semantic, cheng-vlachos-2024-measuring, kunitomo-jacquin-etal-2026-evidential} or consistency \citep{gao-etal-2024-spuq, zhang-etal-2024-luq, bhattacharjya-etal-2025-simba, nguyen-etal-2025-beyond, wang-etal-2025-semantic, joo-cho-2025-cleanse, vazhentsev-etal-2025-token}, where the transformation groups outcomes that are semantically equivalent. 
`Verbalised' confidence quantifiers, which are estimates that are similar in essence but communicated differently, prompt LLMs to generate their confidence in a response. Some directly sample from the next-token distributions and extract the relevant confidence values from the sampled strings \citep{tian-etal-2023-just, ishii-etal-2025-fine}, or assess the probability of tokens semantically related with correctness or confidence values \citep{kadavath2022language,lin-etal-2024-contextualized, khanmohammadi-etal-2025-calibrating}.
Recently, some works focus on adapting LLMs to predict some of the above UQs, \eg via some form of teacher-student distillation
\citep{chen-etal-2023-adaptation, liu-etal-2024-llms-learn-uncertainty, xu-etal-2024-sayself, shelmanov-etal-2025-head, eikema2025teaching}.

\subsection{Uncertainty Quantification Evaluation}

An informative UQ should indicate when we will generate correct responses (with high confidence/low uncertainty) and incorrect responses (low confidence/ high uncertainty).
Under this lens of informativeness, 
we can validate various UQs; with selective prediction comprising a useful paradigm for this purpose --- it accepts or rejects responses depending on whether a confidence score surpasses a threshold \citep{varshney-etal-2022-towards, zablotskaia-etal-2023-uncertainty}. 
At a given threshold $t$, we can measure various performance trade-offs: for example, risk/error against coverage (percentage of responses answered/accepted), or precision against recall, or the true positive rate (TPR) against the false positive rate (FPR). 
Existing metrics can help summarise information about trade-offs under various thresholds, by measuring the area under the curves plotting the assessed trade-offs of interest across two axes: \emph{e.g.}, the area under the precision-recall curve (AUCPR) or the area under the receiver operating characteristic curve (AUROC), assessing TPR and FPR trade offs \citep{flach2016roc}.\footnote{$\mathrm{AUROC} = \int_{0}^{1} \mathrm{TPR}(t)\, d(\mathrm{FPR}(t))$ has a probabilistic interpretation, which enables efficient implementation: the probability that a randomly chosen correct response has higher confidence than a randomly chosen incorrect one.}

To compute such performance metrics (\emph{e.g.}, TPR, FPR, precision, recall) at varying thresholds, it is necessary to judge whether generated responses are correct or incorrect. As human judgement is expensive to obtain at large scale, in practice this correctness judgement is delegated to an automated judge, \emph{e.g.}, a quality metric or a judge model.
Unavoidably, automated judges err. Their errors were found to distort AUROC values and rankings \citep{santilli-etal-2025-revisiting, ielanskyi2025addressing}. \citet{santilli-etal-2025-revisiting} identify sequence length as a common bias affecting both UQs and judges, and partly a source of the distortion---with judges that align well with humans being impacted to a lesser extent; while \citet{ielanskyi2025addressing} propose combining judges' predictions as a means to mitigate biases. 
We dissect even further the complex relationships between UQs and automated judges of varying quality, as assessed by their ability to predict human judgement, so as to understand how they interact within UQ evaluation. We discover a previously unidentified impact of errors that is disproportionately affecting higher quality quantifiers; and point to error categories as a factor that can explain such impacts on UQs.

\section{Correctness Judgements}

\subsection{Meta-analysis of Current Practices}

To the best of our knowledge, no extensive analysis of current practices for the choice of automated judges exists. Here we report on our own meta-analysis of such choices in recent literature.

\paragraph{Methodology.} We survey the ACL anthology from 2020 to 2026 for papers relevant to uncertainty in NLG (we start with an automatic keyword-matching filter and proceed with careful, manual inspection). 
This results in 67 papers, whose choices with regards to correctness judgements we summarise next (further details in Appendix \ref{sec:appendix_meta_analysis_methods}).

\paragraph{Results.} 
\Cref{tab:correctness_functions} summarises the main findings. The 2nd column documents the number of times we encounter different types of judge, and the 3rd column lists the proportion of matches amongst the 67 papers---as in some papers more than one  judge is used (\eg, because different tasks use different judges, or because hybrid methods combine the predictions of multiple judges to reach a final decision) the cells are normalised independently and hence do not add up to 100\%. 
\begin{table}[ht]
\centering
\begin{tabular}{lrr}
\hline
\textbf{Correctness judge} & \textbf{\# Papers} & \textbf{Proportion} \\
\hline
LLM judge & 26 & 38.8\% \\
Overlap based  & 19 & 28.3\% \\
Exact/fuzzy match & 14 & 20.9\% \\
Semantic similarity & 11 & 16.4\% \\
Unidentifiable & 7 & 10.4\% \\
Human labelling & 6 & 9.0\% \\
\hline
\end{tabular}
\caption{Correctness judges used in surveyed papers.}
\label{tab:correctness_functions}
\end{table}
First, it is worth to remark how human labelling is rare, with $91\%$ of papers delegating correctness decisions to an automated judge. When it comes to automated judges, a broad range of methods is used, with the most popular being an LLM as a judge (\eg, GPT-4). However, an appreciable number of papers uses other methods such as overlap-based metrics (\eg, Rouge-L), exact or fuzzy matching, semantic-similarity-based metrics (\eg, cosine of BERT embeddings). Even within one judge type we see high variation among implementation choices (metrics, thresholds, LLMs and their prompts, \etc; see Appendix \ref{sec:appendix_meta_analysis_experiments} for an in-depth analysis and a complete list of the analysed papers and their associated correctness judges). 
When the judge is automated (61 papers), only 23\% of times (14 papers) the judge is validated against human judgements.
We also tracked the choices made \wrt the UQ evaluation metric (noting that some papers employ multiple). The majority of papers employ AUROC ($\approx 60\%$), while other SP metrics (\ie, AUCPR, AURAC) and calibration metrics (\ie, ECE, Brier) remain popular (see \Cref{tab:evaluation_metrics} of Appendix \ref{sec:appendix_meta_analysis_experiments}).

\subsection{Evaluating Correctness Judges}
\label{sec:decision_judge}

Our meta-analysis reveals that automated judges are rarely validated against human judgements. Hence, we gather a representative set of such judges and collect human annotation to assess their classification performance (\emph{e.g.}, accuracy, F1, etc.).

\paragraph{Data Collection.} 
We focus on question answering (QA): we sample 100 questions from \texttt{TriviaQA} \citep{joshi-etal-2017-triviaqa}, which contains factual questions, and 200 questions from \texttt{AmbigQA} \citep{min-etal-2020-ambigqa}, which contains ambiguous and unambiguous factual questions (100 non-ambiguous and 100 ambiguous questions). 
We employ 4 generator models: \texttt{Llama-3.1-8B-Instruct}, \texttt{Llama-3.2-3B-Instruct},\texttt{Qwen2.5-7B-Instruct} and \texttt{Qwen2.5-0.5B-Instruct} \citep{grattafiori2024llama, hui2024qwen2}.
We generate greedy responses from these models (\emph{i.e.}, $\mathrm{temp}=0$). A human judge annotates the correctness of each greedy response with respect to a set of references. For  Llama 8B/TriviaQA and Llama 8B/AmbigQA (ambiguous), 2 judges annotate each response, so that we can appreciate the extent to which they agree (Cohen's Kappa score between annotators 0.88 for the former and 0.75 for the latter; see Appendix \ref{sec:appendix:annotators} for details).
We analyse various judges, representative of the categories of judges found in our meta-analysis (see \Cref{tab:correctness_functions}): exact matching, thresholded token F1, thresholded BLEU \citep{reiter-2018-structured}, thresholded ROUGE-L \citep{lin-2004-rouge}, thresholded embedding similarity with \texttt{all-MiniLM-L6-v2}, thresholded BERT score \citep{Zhang*2020BERTScore:}, and LLM as judge (using \texttt{gpt-5.4-mini} and \texttt{Llama-3.1-8B-Instruct}; an API and a Hugging Face model, resp.). Implementation details in Appendix \ref{sec:appendix:implement_details}. \Cref{fig:example_datapoint} shows one annotated data point, for one TriviaQA question.

\paragraph{Methodology.} Our goal in this section is to use the annotated data (human and automated judgements) to infer our judges' classification performances, but we also need to cope with various sources of uncertainty (\eg, adequacy judgements are subject to variability due to annotator perspective, prompt under-specification and ambiguity, annotation errors, \etc). 
To accommodate them, as well as the inevitable scarcity of human data, we model available data as draws from a hierarchical Bayesian model whose latent parameters capture the probability of correctness of each prompt-response pair, as well as each judge's sensitivity (related to TPR) and specificity (related to TNR). Together, these parameters allow us to infer posterior distributions of a judge's performance in terms of TPR, TNR, FPR, FNR and transformations thereof (\eg F1).  
See complete model in Appendix \ref{sec:bda-judges}.

\begin{figure}[t!]
    \centering
    \includegraphics[width=0.9\linewidth, trim={3.5cm 0 1.5cm 1.5cm}, clip]{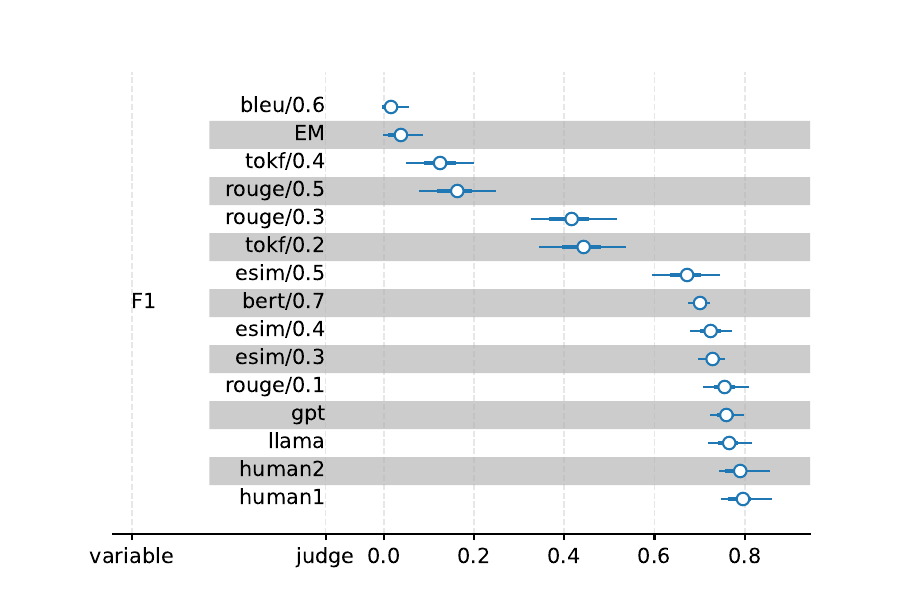}
    \caption{Posterior F1 (horizontal axis) for a subset of judges for TriviaQA-Llama8b. Rows are sorted by posterior median (circle) and display two high density intervals (50\% and 95\%; thick/thin line segments).}
    \label{fig:placeholder_bda_f1}
\end{figure}

\begin{figure}[h]
\begin{center}
\includegraphics[width=\linewidth]{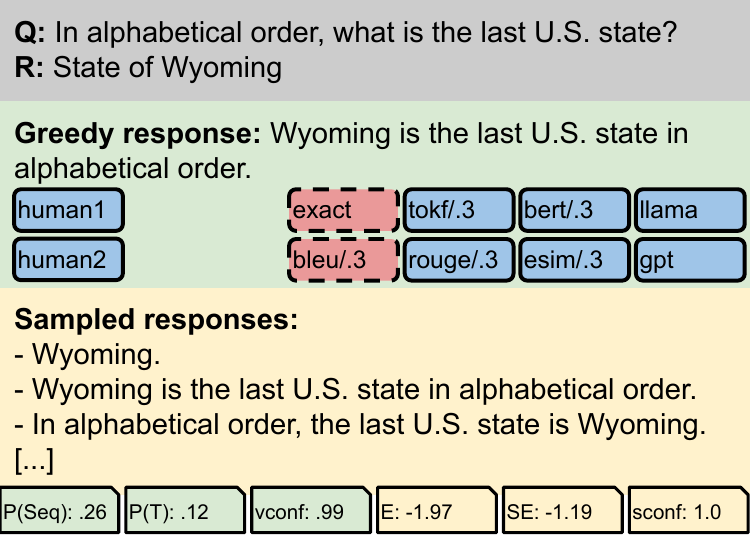}

\caption{An example data point of question (Q) and references (R) from TriviaQA and responses by Llama 8B. The greedy response is judged manually and automatically (\eg, via exact match, BLEU, BertScore, LLM judges, \etc); positive judgements are blue/solid, negative are red/dashed. We compute various UQs: some (\eg, P(True)) qualify the greedy response specifically (in green), others (\eg, entropy) qualify the entire conditional distribution (estimated using samples; in yellow).}
\label{fig:example_datapoint}
\end{center}
\end{figure}

\paragraph{Results.} 
\Cref{fig:placeholder_bda_f1} plots the inferred posterior F1 for a subset of the judges in the setting Llama 8B/TriviaQA. These judges cover a representative range of F1 medians.\footnote{A complete list of distributions for all judges and model-generator pairs can be seen in Appendix \ref{appendix:f1_dists}.} 
Such a plot allows us to group bad (\emph{e.g.}, BLEU@0.6, exact match), medium (\emph{e.g.} Rouge-L@0.3, Token-F1@0.2) and good judges (\emph{e.g.}, Rouge-L@0.1, Llama8B and GPT5.4-mini). 
We generally observe across dataset-generator pairs similar distribution spreads; with human F1 distributions concentrating towards higher F1 scores. The order of the judges varies across pairs; but across plots, distributions for LLM and semantic similarity judges (for one threshold or another) are skewed towards higher F1 values.

\section{Evaluating UQ Evaluation}

UQ evaluation by AUROC sorts competing UQs from least to most informative for selective prediction.
Here, we evaluate whether this strategy is at all reliable when correctness judgements are automated. 
For a collection of UQs being compared, we characterise their AUROC using human correctness judgements to establish an `oracle' or `gold-standard'.
We then analyse the impact of automated correctness by assessing changes in AUROC when replacing human with automated judgement.

\begin{table*}[h!]
\centering
 \resizebox{0.85\textwidth}{!}{
\begin{tabular}{llrrrlrrrlrrr}
 & \multicolumn{4}{c}{\cellcolor[HTML]{6AA84F}\textbf{Best Quantifier (Rank \#1)}} & \multicolumn{4}{c}{\cellcolor[HTML]{E6B8AF}\textbf{Medium Quantifier (Rank \#4)}} & \multicolumn{4}{c}{\cellcolor[HTML]{CC0000}\textbf{Worst Quantifier (Rank \#7)}} \\
\textbf{Model/Dataset pair} & \textbf{Actual} & \multicolumn{1}{l}{\textbf{Rank G}} & \multicolumn{1}{l}{\textbf{Rank M}} & \multicolumn{1}{l}{\textbf{Rank B}} & \textbf{Actual} & \multicolumn{1}{l}{\textbf{Rank G}} & \multicolumn{1}{l}{\textbf{Rank M}} & \multicolumn{1}{l}{\textbf{Rank B}} & \textbf{Actual} & \multicolumn{1}{l}{\textbf{Rank G}} & \multicolumn{1}{l}{\textbf{Rank M}} & \multicolumn{1}{l}{\textbf{Rank B}} \\
TriviaQA/Llama8b & {\color[HTML]{333333} VC} & \cellcolor[HTML]{B6D7A8}2.3 & \cellcolor[HTML]{EA9999}4.3 & \cellcolor[HTML]{EA9999}4.4 & {\color[HTML]{333333} SE} & \cellcolor[HTML]{EA9999}4.2 & \cellcolor[HTML]{E6B8AF}3.5 & \cellcolor[HTML]{E6B8AF}3.6 & {\color[HTML]{333333} SC} & \cellcolor[HTML]{CC0000}6.8 & \cellcolor[HTML]{CC0000}6.0 & \cellcolor[HTML]{E06666}5.8 \\
TriviaQA/Llama3b & {\color[HTML]{333333} P(Seq.)} & \cellcolor[HTML]{B6D7A8}2.2 & \cellcolor[HTML]{6AA84F}1.4 & \cellcolor[HTML]{B6D7A8}2.1 & {\color[HTML]{333333} VC} & \cellcolor[HTML]{B6D7A8}2.5 & \cellcolor[HTML]{E06666}5.6 & \cellcolor[HTML]{E06666}5.2 & {\color[HTML]{333333} SC} & \cellcolor[HTML]{CC0000}7.0 & \cellcolor[HTML]{E06666}5.6 & \cellcolor[HTML]{CC0000}6.1 \\
TriviaQA/Qwen7b & {\color[HTML]{333333} SC} & \cellcolor[HTML]{6AA84F}1.5 & \cellcolor[HTML]{CC0000}6.0 & \cellcolor[HTML]{CC0000}6.3 & {\color[HTML]{333333} N.P(Seq.)} & \cellcolor[HTML]{E06666}5.0 & \cellcolor[HTML]{B6D7A8}2.0 & \cellcolor[HTML]{B6D7A8}2.8 & {\color[HTML]{333333} E} & \cellcolor[HTML]{CC0000}6.0 & \cellcolor[HTML]{E6B8AF}3.3 & \cellcolor[HTML]{6AA84F}1.3 \\
TriviaQA/Qwen0.5b & {\color[HTML]{333333} SC} & \cellcolor[HTML]{6AA84F}1.0 & \cellcolor[HTML]{E6B8AF}3.7 & \cellcolor[HTML]{E06666}5.2 & {\color[HTML]{333333} N.P(Seq.)} & \cellcolor[HTML]{E6B8AF}3.0 & \cellcolor[HTML]{EA9999}4.2 & \cellcolor[HTML]{E6B8AF}3.5 & {\color[HTML]{333333} P(True)} & \cellcolor[HTML]{CC0000}6.0 & \cellcolor[HTML]{EA9999}4.6 & \cellcolor[HTML]{EA9999}4.8 \\
AmbigQA\_non/Llama8b & {\color[HTML]{333333} VC} & \cellcolor[HTML]{6AA84F}1.4 & \cellcolor[HTML]{B6D7A8}2.8 & \cellcolor[HTML]{EA9999}4.7 & {\color[HTML]{333333} P(True)} & \cellcolor[HTML]{E6B8AF}3.9 & \cellcolor[HTML]{E06666}5.8 & \cellcolor[HTML]{E06666}5.1 & {\color[HTML]{333333} E} & \cellcolor[HTML]{E06666}5.0 & \cellcolor[HTML]{EA9999}4.7 & \cellcolor[HTML]{B6D7A8}2.2 \\
AmbigQA\_non/Llama3b & {\color[HTML]{333333} N.P(Seq.)} & \cellcolor[HTML]{6AA84F}1.5 & \cellcolor[HTML]{6AA84F}1.5 & \cellcolor[HTML]{6AA84F}1.2 & {\color[HTML]{333333} SC} & \cellcolor[HTML]{CC0000}6.7 & \cellcolor[HTML]{CC0000}7.0 & \cellcolor[HTML]{CC0000}6.8 & {\color[HTML]{333333} E} & \cellcolor[HTML]{E06666}5.5 & \cellcolor[HTML]{E6B8AF}3.5 & \cellcolor[HTML]{EA9999}4.0 \\
AmbigQA\_non/Qwen7b & {\color[HTML]{333333} P(True)} & \cellcolor[HTML]{B6D7A8}2.0 & \cellcolor[HTML]{B6D7A8}2.8 & \cellcolor[HTML]{EA9999}4.5 & {\color[HTML]{333333} VC} & \cellcolor[HTML]{E06666}5.5 & \cellcolor[HTML]{E06666}5.0 & \cellcolor[HTML]{EA9999}4.0 & {\color[HTML]{333333} E} & \cellcolor[HTML]{CC0000}7.0 & \cellcolor[HTML]{CC0000}6.4 & \cellcolor[HTML]{EA9999}4.5 \\
AmbigQA\_non/Qwen0.5b & {\color[HTML]{333333} N.P(Seq.)} & \cellcolor[HTML]{B6D7A8}2.0 & \cellcolor[HTML]{6AA84F}1.6 & \cellcolor[HTML]{6AA84F}1.5 & {\color[HTML]{333333} P(Seq.)} & \cellcolor[HTML]{E6B8AF}3.0 & \cellcolor[HTML]{B6D7A8}2.1 & \cellcolor[HTML]{6AA84F}1.5 & {\color[HTML]{333333} E} & \cellcolor[HTML]{CC0000}6.5 & \cellcolor[HTML]{EA9999}4.9 & \cellcolor[HTML]{E06666}5.0 \\
AmbigQA\_am/Llama8b & {\color[HTML]{333333} P(True)} & \cellcolor[HTML]{6AA84F}1.4 & \cellcolor[HTML]{E6B8AF}3.3 & \cellcolor[HTML]{CC0000}6.4 & {\color[HTML]{333333} SE} & \cellcolor[HTML]{E06666}5.6 & \cellcolor[HTML]{CC0000}6.0 & \cellcolor[HTML]{E6B8AF}3.9 & {\color[HTML]{333333} E} & \cellcolor[HTML]{CC0000}6.0 & \cellcolor[HTML]{CC0000}6.0 & \cellcolor[HTML]{EA9999}4.3 \\
AmbigQA\_am/Llama3b & {\color[HTML]{333333} VC} & \cellcolor[HTML]{6AA84F}1.9 & \cellcolor[HTML]{E6B8AF}3.0 & \cellcolor[HTML]{E06666}5.5 & {\color[HTML]{333333} N.P(Seq.)} & \cellcolor[HTML]{E6B8AF}3.4 & \cellcolor[HTML]{B6D7A8}2.2 & \cellcolor[HTML]{EA9999}4.5 & {\color[HTML]{333333} E} & \cellcolor[HTML]{E06666}5.3 & \cellcolor[HTML]{EA9999}4.4 & \cellcolor[HTML]{E6B8AF}3.2 \\
AmbigQA\_am/Qwen7b & {\color[HTML]{333333} P(True)} & \cellcolor[HTML]{6AA84F}1.5 & \cellcolor[HTML]{E6B8AF}3.6 & \cellcolor[HTML]{E06666}5.8 & {\color[HTML]{333333} N.P(Seq.)} & \cellcolor[HTML]{B6D7A8}2.5 & \cellcolor[HTML]{EA9999}4.3 & \cellcolor[HTML]{E6B8AF}3.4 & {\color[HTML]{333333} E} & \cellcolor[HTML]{CC0000}7.0 & \cellcolor[HTML]{EA9999}4.6 & \cellcolor[HTML]{B6D7A8}2.8 \\
AmbigQA\_am/Qwen0.5b & {\color[HTML]{333333} SC} & \cellcolor[HTML]{6AA84F}1.5 & - & \cellcolor[HTML]{E06666}5.2 & {\color[HTML]{333333} N.P(Seq.)} & \cellcolor[HTML]{E6B8AF}3.0 & - & \cellcolor[HTML]{B6D7A8}2.2 & {\color[HTML]{333333} E} & \cellcolor[HTML]{CC0000}6.5 & \multicolumn{1}{l}{{\color[HTML]{333333} -}} & \cellcolor[HTML]{EA9999}4.7
\end{tabular}
}
\caption{We separate the correctness judges as `good' (G; $F1 > 0.7$), `medium' (M; $0.3 < F1 < 0.7$) or `bad' (B; $F1 < 0.3$) in terms of their F1 score, and  identify the best UQ (ranked 1st), medium-quality UQ (ranked 4th) and worst UQ (ranked 7th), according to their oracle AUROC values. We show how the oracle rankings of UQs are affected on average.}
\label{tab:rankings}
\end{table*}

\begin{table*}[h!]
 \resizebox{\textwidth}{!}{
\begin{tabular}{llrrrrlrrrrlrrrr}
 & \multicolumn{5}{c}{\textbf{Best Quantifier (Rank \#1)}} & \multicolumn{5}{c}{\textbf{Medium Quantifier (Rank \#4)}} & \multicolumn{5}{c}{\textbf{Worst Quantifier (Rank \#7)}} \\
\textbf{Model/Dataset pair} & \textbf{Actual} & \multicolumn{1}{l}{\textbf{AUROC}} & \multicolumn{1}{l}{\textbf{Diff. G}} & \multicolumn{1}{l}{\textbf{Diff. M}} & \multicolumn{1}{l}{\textbf{Diff. B}} & \textbf{Actual} & \multicolumn{1}{l}{\textbf{AUROC}} & \multicolumn{1}{l}{\textbf{Diff. G}} & \multicolumn{1}{l}{\textbf{Diff. M}} & \multicolumn{1}{l}{\textbf{Diff. B}} & \textbf{Actual} & \multicolumn{1}{l}{\textbf{AUROC}} & \multicolumn{1}{l}{\textbf{Diff. G}} & \multicolumn{1}{l}{\textbf{Diff. M}} & \multicolumn{1}{l}{\textbf{Diff. B}} \\
TriviaQA/Llama8b & {\color[HTML]{333333} VC} & {\color[HTML]{333333} 0.92} & \cellcolor[HTML]{E06666}-0.11 & \cellcolor[HTML]{E06666}-0.22 & \cellcolor[HTML]{E06666}-0.19 & {\color[HTML]{333333} SE} & {\color[HTML]{333333} 0.76} & \cellcolor[HTML]{E6B8AF}-0.03 & \cellcolor[HTML]{E6B8AF}-0.03 & \cellcolor[HTML]{E6B8AF}0.02 & {\color[HTML]{333333} SC} & {\color[HTML]{333333} 0.69} & \cellcolor[HTML]{EA9999}-0.07 & \cellcolor[HTML]{E6B8AF}-0.05 & \cellcolor[HTML]{E6B8AF}-0.01 \\
TriviaQA/Llama3b & {\color[HTML]{333333} P(Seq.)} & {\color[HTML]{333333} 0.88} & \cellcolor[HTML]{EA9999}-0.09 & \cellcolor[HTML]{EA9999}-0.10 & \cellcolor[HTML]{E6B8AF}0.03 & {\color[HTML]{333333} VC} & {\color[HTML]{333333} 0.79} & \cellcolor[HTML]{E6B8AF}0.00 & \cellcolor[HTML]{E06666}-0.12 & \cellcolor[HTML]{E6B8AF}-0.02 & {\color[HTML]{333333} SC} & 0.67 & \cellcolor[HTML]{EA9999}-0.06 & \cellcolor[HTML]{E6B8AF}-0.02 & \cellcolor[HTML]{E6B8AF}0.06 \\
TriviaQA/Qwen7b & {\color[HTML]{333333} SC} & {\color[HTML]{333333} 0.82} & \cellcolor[HTML]{EA9999}-0.08 & \cellcolor[HTML]{E06666}-0.18 & \cellcolor[HTML]{E06666}-0.11 & {\color[HTML]{333333} N.P(Seq.)} & 0.72 & \cellcolor[HTML]{EA9999}-0.08 & \cellcolor[HTML]{B6D7A8}0.08 & \cellcolor[HTML]{93C47D}0.23 & {\color[HTML]{333333} E} & {\color[HTML]{333333} 0.67} & \cellcolor[HTML]{E6B8AF}-0.05 & \cellcolor[HTML]{B6D7A8}0.08 & \cellcolor[HTML]{93C47D}0.29 \\
TriviaQA/Qwen0.5b & {\color[HTML]{333333} SC} & {\color[HTML]{333333} 0.73} & \cellcolor[HTML]{E6B8AF}0.02 & \cellcolor[HTML]{E06666}-0.14 & \cellcolor[HTML]{E6B8AF}0.03 & {\color[HTML]{333333} N.P(Seq.)} & {\color[HTML]{333333} 0.56} & \cellcolor[HTML]{E6B8AF}0.03 & \cellcolor[HTML]{E6B8AF}0.05 & \cellcolor[HTML]{6AA84F}0.32 & {\color[HTML]{333333} P(True)} & {\color[HTML]{333333} 0.49} & \cellcolor[HTML]{E6B8AF}0.02 & \cellcolor[HTML]{B6D7A8}0.07 & \cellcolor[HTML]{6AA84F}0.31 \\
AmbigQA\_non/Llama8b & {\color[HTML]{333333} VC} & {\color[HTML]{333333} 0.78} & \cellcolor[HTML]{E6B8AF}-0.02 & \cellcolor[HTML]{EA9999}-0.07 & \cellcolor[HTML]{CC0000}-0.31 & {\color[HTML]{333333} P(True)} & 0.68 & \cellcolor[HTML]{E6B8AF}-0.01 & \cellcolor[HTML]{EA9999}-0.10 & \cellcolor[HTML]{E06666}-0.14 & {\color[HTML]{333333} E} & {\color[HTML]{333333} 0.58} & \cellcolor[HTML]{E6B8AF}0.02 & \cellcolor[HTML]{B6D7A8}0.06 & \cellcolor[HTML]{93C47D}0.25 \\
AmbigQA\_non/Llama3b & {\color[HTML]{333333} N.P(Seq.)} & {\color[HTML]{333333} 0.79} & \cellcolor[HTML]{E6B8AF}-0.02 & \cellcolor[HTML]{E6B8AF}0.02 & \cellcolor[HTML]{93C47D}0.11 & {\color[HTML]{333333} SC} & {\color[HTML]{333333} 0.68} & \cellcolor[HTML]{E06666}-0.11 & \cellcolor[HTML]{E06666}-0.12 & \cellcolor[HTML]{EA9999}-0.09 & {\color[HTML]{333333} E} & 0.59 & \cellcolor[HTML]{B6D7A8}0.06 & \cellcolor[HTML]{93C47D}0.13 & \cellcolor[HTML]{93C47D}0.19 \\
AmbigQA\_non/Qwen7b & {\color[HTML]{333333} P(True)} & {\color[HTML]{333333} 0.84} & \cellcolor[HTML]{EA9999}-0.09 & \cellcolor[HTML]{E06666}-0.11 & \cellcolor[HTML]{E06666}-0.13 & {\color[HTML]{333333} VC} & {\color[HTML]{333333} 0.70} & \cellcolor[HTML]{EA9999}-0.07 & \cellcolor[HTML]{E6B8AF}-0.03 & \cellcolor[HTML]{E6B8AF}0.00 & {\color[HTML]{333333} E} & {\color[HTML]{333333} 0.58} & \cellcolor[HTML]{E6B8AF}0.01 & \cellcolor[HTML]{E6B8AF}0.05 & \cellcolor[HTML]{E6B8AF}0.05 \\
AmbigQA\_non/Qwen0.5b & {\color[HTML]{333333} N.P(Seq.)} & {\color[HTML]{333333} 0.66} & \cellcolor[HTML]{E6B8AF}0.05 & \cellcolor[HTML]{B6D7A8}0.09 & \cellcolor[HTML]{93C47D}0.15 & {\color[HTML]{333333} P(Seq.)} & {\color[HTML]{333333} 0.61} & \cellcolor[HTML]{B6D7A8}0.09 & \cellcolor[HTML]{93C47D}0.12 & \cellcolor[HTML]{93C47D}0.20 & {\color[HTML]{333333} E} & {\color[HTML]{333333} 0.53} & \cellcolor[HTML]{E6B8AF}-0.01 & \cellcolor[HTML]{E6B8AF}0.00 & \cellcolor[HTML]{E6B8AF}-0.04 \\
AmbigQA\_am/Llama8b & {\color[HTML]{333333} P(True)} & {\color[HTML]{333333} 0.79} & \cellcolor[HTML]{EA9999}-0.06 & \cellcolor[HTML]{E06666}-0.17 & \cellcolor[HTML]{CC0000}-0.37 & {\color[HTML]{333333} SE} & {\color[HTML]{333333} 0.68} & \cellcolor[HTML]{EA9999}-0.09 & \cellcolor[HTML]{E06666}-0.13 & \cellcolor[HTML]{E6B8AF}0.04 & {\color[HTML]{333333} E} & {\color[HTML]{333333} 0.52} & \cellcolor[HTML]{E6B8AF}0.03 & \cellcolor[HTML]{E6B8AF}0.03 & \cellcolor[HTML]{B6D7A8}0.10 \\
AmbigQA\_am/Llama3b & {\color[HTML]{333333} VC} & {\color[HTML]{333333} 0.70} & \cellcolor[HTML]{E6B8AF}0.00 & \cellcolor[HTML]{EA9999}-0.06 & \cellcolor[HTML]{E06666}-0.27 & {\color[HTML]{333333} N.P(Seq.)} & {\color[HTML]{333333} 0.66} & \cellcolor[HTML]{E6B8AF}-0.05 & \cellcolor[HTML]{B6D7A8}0.09 & \cellcolor[HTML]{E06666}-0.15 & {\color[HTML]{333333} E} & {\color[HTML]{333333} 0.55} & \cellcolor[HTML]{E6B8AF}0.00 & \cellcolor[HTML]{B6D7A8}0.06 & \cellcolor[HTML]{B6D7A8}0.06 \\
AmbigQA\_am/Qwen7b & {\color[HTML]{333333} P(True)} & {\color[HTML]{333333} 0.68} & \cellcolor[HTML]{E6B8AF}0.01 & \cellcolor[HTML]{EA9999}-0.08 & \cellcolor[HTML]{E06666}-0.14 & {\color[HTML]{333333} N.P(Seq.)} & {\color[HTML]{333333} 0.65} & \cellcolor[HTML]{E6B8AF}0.03 & \cellcolor[HTML]{EA9999}-0.06 & \cellcolor[HTML]{93C47D}0.14 & {\color[HTML]{333333} E} & {\color[HTML]{333333} 0.57} & \cellcolor[HTML]{E6B8AF}0.01 & \cellcolor[HTML]{E6B8AF}0.01 & \cellcolor[HTML]{93C47D}0.21 \\
AmbigQA\_am/Qwen0/5b & {\color[HTML]{333333} SC} & {\color[HTML]{333333} 0.71} & \cellcolor[HTML]{EA9999}-0.07 & \multicolumn{1}{l}{-} & \cellcolor[HTML]{E06666}-0.28 & {\color[HTML]{333333} N.P(Seq.)} & {\color[HTML]{333333} 0.60} & \cellcolor[HTML]{E6B8AF}-0.02 & \multicolumn{1}{l}{-} & \cellcolor[HTML]{93C47D}0.15 & {\color[HTML]{333333} E} & 0.50 & \cellcolor[HTML]{E6B8AF}0.00 & \multicolumn{1}{l}{{\color[HTML]{333333} -}} & \cellcolor[HTML]{E6B8AF}0.00
\end{tabular}
}
\caption{We separate the correctness judges as `good' (G; $F1 > 0.7$), `medium' (M; $0.3 < F1 < 0.7$) or `bad' (B; $F1 < 0.3$) in terms of their F1 score and  identify the best UQ (ranked 1st), medium-quality UQ (ranked 4th) and worst UQ (ranked 7th), according to their oracle AUROC values. We show how the oracle AUROC values of UQs are affected on average.}
\label{tab:auroc_diffs}
\end{table*}

\paragraph{Experiments.} We continue analysing the prompt-response pairs from \texttt{TriviaQA} and \texttt{AmbigQA}, and the generators we have used in the previous analysis (Llama 3B/8B, Qwen 0.5B/7B); whose greedy responses we have generated and annotated.
We analyse 7 different UQs: log probability \citep{jiang-etal-2021-know}, normalised log probability \citep{manakul-etal-2023-selfcheckgpt}, predictive entropy \citep{fomicheva-etal-2020-unsupervised}, semantic entropy (\emph{i.e.}, entropy over semantic clusters of sampled responses; \citet{kuhn2023semantic}), semantic confidence (\emph{i.e.}, rate at which a response is not contradicted by sampled responses; \citet{yona-etal-2024-large})), P(True)  (\emph{i.e.}, the  probability a model assigns to the token associated with correctness of a response; \citet{kadavath2022language}) and verbalised confidence (\emph{i.e.}, extracted confidence sampled from a model when prompted for the task; \citet{tian-etal-2023-just}). 
To compute these UQs, we sample 10 responses from our generators (with $\mathrm{temp}\!=\!1$).
We use the same correctness judges as before. See \Cref{fig:example_datapoint} for an illustration of a datapoint (and  Appendix \ref{sec:appendix:implement_details} for more details).

\paragraph{Results.} 
\Cref{tab:rankings} and \ref{tab:auroc_diffs} present the main results. For each dataset-generator pair we separate the correctness judges as `good' (G; $F1 > 0.7$), `medium' (M; $0.3 < F1 < 0.7$) or `bad' (B; $F1 < 0.3$) in terms of their F1 score, pointed-estimated against a human judge. Then, for each case we identify the best UQ (ranked 1st), medium-quality UQ (ranked 4th) and worst UQ (ranked 7th), according to their oracle AUROC values. We then check how the oracle evaluation is distorted by automated evaluation at different levels of judgement quality (G/M/B); by tracking the change in oracle figures, for the best, medium and worst UQs, due to automation. In \Cref{tab:rankings} we track the position at which the automated judge places these UQs and in \Cref{tab:auroc_diffs} the change in automated AUROC relative to the oracle. 
These are averages, as each group (G/M/B) contributes multiple measurements (one per judge).

It is clear that automated protocols vary widely, echoing previous findings 
\citep{santilli-etal-2025-revisiting, ielanskyi2025addressing}. 
But the oracle allows for further insight: we can now see that automated protocols can perform very poorly, with AUROC values and rankings distorted away from the oracle. 
We also see that judgement quality has a non-uniform impact on how the merits of different UQs are depicted, with informative UQs being harshly downplayed.
The best UQs are often seriously harmed 
(\emph{e.g.} the actual best UQ often ranks 4th-6th under a bad correctness judge). 
On the other hand, the medium-quality UQs are affected somewhat less (see medium columns of \Cref{tab:auroc_diffs}). The worst UQs are affected minimally or even appear better in terms of AUROC values (see right-most columns, in green), with their absolute rankings also appearing improved (bad judges place them in rankings 2-4, when they should rank last).

In Appendix \ref{sec:appendix:extensive}, for transparency, we present a complete list of AUROC values and rankings per UQ and judge. For the dataset-generator pairs with two sets of human annotations, we analyse the AUROC values and rankings they induce side by side (see Appendix \ref{sec:appendix:annotators}); verifying their similar patterns. 
 We also perform a bootstrapping analysis, where we repeatedly subsample from available prompts and compute key metrics (\ie, AUROC values  and rankings; Appendix \ref{sec:appendix:bootstrat}), and observe similar patterns as in the main study. 
Lastly, we investigate the use of other SP metrics in Appendix \ref{sec:appendix:other_metrics} and show that the issues discussed here persist.

\section{Error Analysis}

\subsection{How Does NLG Evaluation and UQ Evaluation Interact?}

Here we dig deeper into the relation between the quality of automated judges and their (in)ability to power a reliable proxy to oracle UQ evaluation.

\paragraph{Experiments.} We rank UQs by oracle AUROC (with human judgements) and by proxy AUROC (with automated judgements) and measure the similarity of the rankings these protocols induce. For that, we use Rank-Biased Overlap (RBO; \citet{10.1145/1852102.1852106}), which penalises disagreement at top positions more (we use patience $0.9$).\footnote{As the field progresses,  more UQs earn their spot in comparisons, with the list growing long and  uninformative---naturally demanding for stronger agreement at top positions.}
To grasp the relationship with judgement performance, we study how RBO co-varies with judges' F1 scores.

\paragraph{Results.}
\Cref{fig:placeholder_rbo_f1} shows results for TriviaQA-Llama 8B and TriviaQA-Qwen 7B (other pairs in Appendix \ref{appendix:rbo_plots}). 
Even within a narrow range of F1 values (\ie,  similar-quality judges), there’s considerable variability in their UQ evaluation performance.
This is not only prominent in low-quality judges, even high-quality judges ($F1 \simeq0.8-0.9$) exhibit this variability (judges are distributed across the vertical axis). We also inspect Kendall $\tau$, instead of RBO, and observe similar patterns.

\begin{figure}[t!]
    \centering
    \includegraphics[width=\linewidth, trim={0cm 0 0cm 0}, clip]{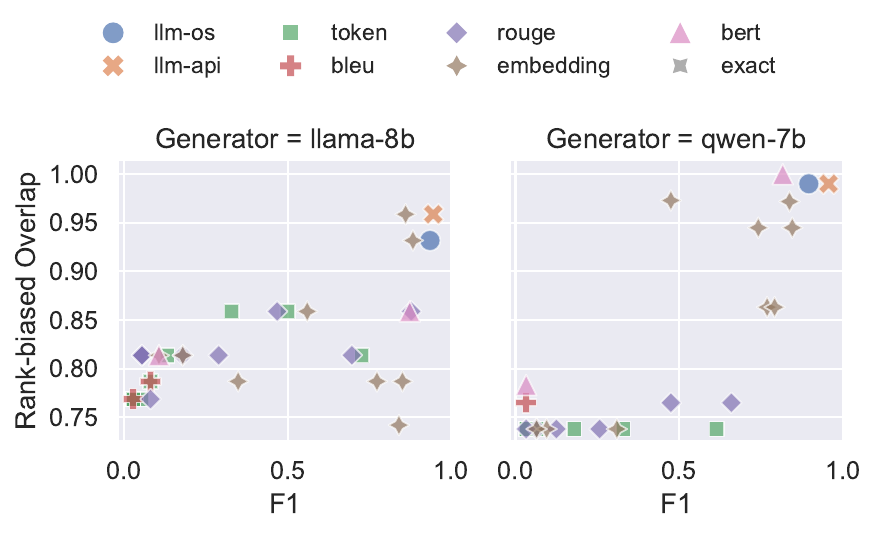}
    \caption{F1 scores ($\uparrow$)  for judges against the RBO score ($\uparrow$) of human- vs judge-powered UQ rankings. Even within a narrow range of F1 values (\ie,  similar-quality judges), there’s considerable variability in their UQ evaluation performance.}
    \label{fig:placeholder_rbo_f1}
\end{figure}

This suggests that F1 may be hiding information that is relevant to UQ evaluation.  
Connecting with our analysis of \Cref{sec:decision_judge}, where we observed different error distributions across judges' TPRs and TNRs, we speculate that the ways a judge's errors distribute across these two categories play an important role in UQ evaluation. This role is elusive until we relate them to confidences values.

\subsection{Understanding Trade-offs}
\label{sec:error_trade-offs}
Automated judges err, predicting that a response is correct when it is incorrect (FP), or vice versa (FN); we analyse how the confidences assigned by UQ methods  distribute across these error categories.

\paragraph{Methodology.} 
With a heterogenous set of UQs being compared, it is hopeless to expect that the confidences they assign are comparable (many do not even lie in the same interval). For the tools we develop here, having all UQs communicate comparable confidences facilitates analysis. For that, we learn to regress from the raw scores assigned by UQs to a (normalised) confidence using a Bayesian logistic regressor (details in Appendix \ref{sec:bda-quantifiers}).

\begin{figure}[t!]
    \centering
    \includegraphics[width=\linewidth]{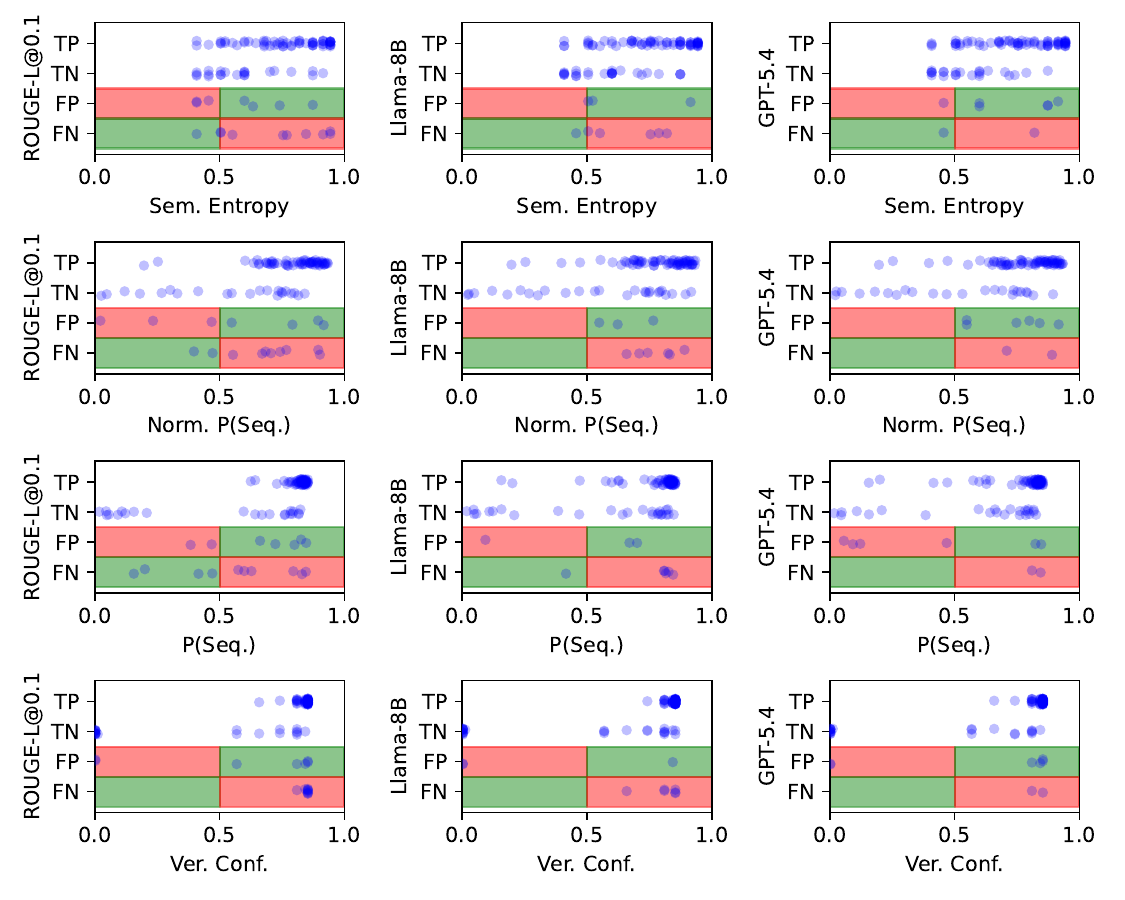}
    \caption{Error trade-offs in SP (with threshold $0.5$) for various UQs under best automated judges (TriviaQA-Llama8B).  For a detailed guide on how to interpret this figure, see \Cref{sec:error_trade-offs}.}
    \label{fig:errors_vs_conf_bda}
\end{figure}

Let's start with the simplest form of SP: a quantifier Q assigns confidence $c$ to a response $r$; if $c$ exceeds a fixed threshold (\eg, $t=0.5$), the response is accepted, else it is rejected. An automated judge categorises the responses as correct (positive/P) or incorrect (negative/N), but, relative to human judgement,  the judge may be right (\ie, then a P is a \emph{true} positive/TP and an N is a true negative/TN) or wrong (\ie, then a P is a \emph{false} positive/FP and an N is a false negative/FN). 
In \Cref{fig:errors_vs_conf_bda} (for TriviaQA-Llama8B), we scatter the confidences assigned by different UQs, separated by categories of classification error in automated judgement by different judges.
When the judge is right (TP, TN), a metric for SP will gather reliable information about the quantifier's merits in separating Ps from Ns (\ie, Ps to the right and Ns to the left of the threshold). When the judge is wrong (FP, FN), the UQ is unfairly rewarded (\ie, it accepts a response that appears P but is not \emph{or} it rejects a response that appears N, but is not; green region in plots) or penalised (\ie, it rejects a response that appears P but is not \emph{or} it accepts a response that appears N but is not; red region in plots) depending on how confidence distributes.
In this way to view the data, it is clear that an \emph{imbalance in concentration} of points across error categories (FP, FN) amounts to a net impact that will misrepresent the informativeness of the UQ; this is will be true even for judges that are often (but not always) right.

The figure displays the three best automated judges (as per the analysis of  \Cref{fig:placeholder_bda_f1}). 
While all judges have similar F1 distributions, their errors are not always similarly distributed. 
For instance, Verb. Conf. seems to have a more dense region of unfair penalties (bottom, right red region) when the judge is ROUGE-L@0.1 compared to the LLM judges, indicating a higher negative impact on that UQ by the specific judge. 
However, GPT5.4 judge's errors may be impacting P(seq.) more harshly than ROUGE-L@0.1, since the former's negative impact regions seem to be more densely populated than positive ones. ROUGE-L@0.1  contains instances in positive and negative regions, possibly cancelling each other's effects.

\begin{figure*}[h!]
    \centering
    \includegraphics[width=0.9\linewidth]{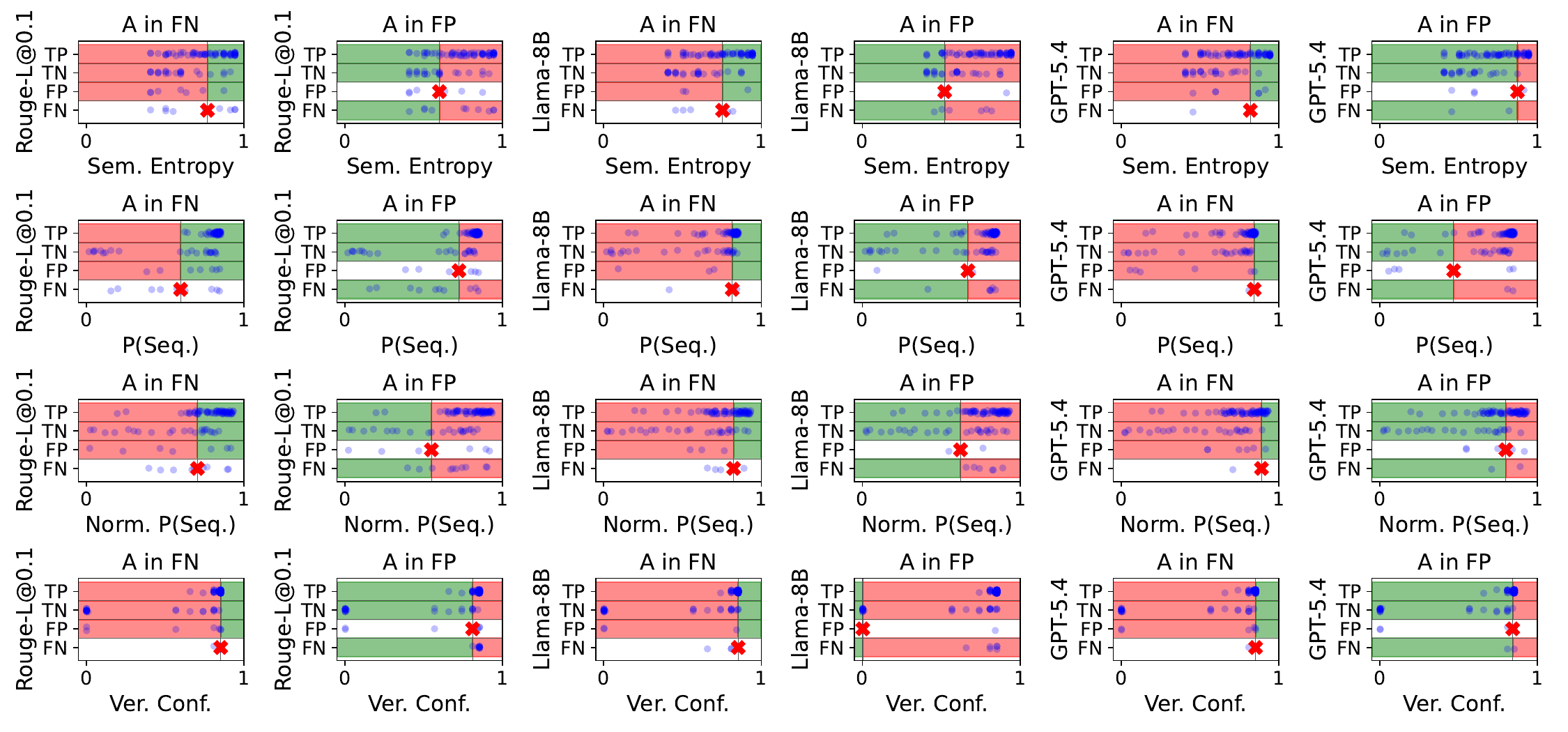}
    \caption{Error impacts on AUROC for various quantifiers under best automated judges, for TriviaQA-Llama8B. For a detailed guide on how to interpret this figure, see \Cref{sec:error_trade-offs}.}
    \label{fig:errors_vs_conf_auroc}
\end{figure*}

Let's now turn to evaluation by AUROC, which is based on rate of `successful' pairwise comparisons among correct and incorrect responses: the metric rewards the UQ when a correct response (P) has a higher confidence than an incorrect one (N), and penalises it otherwise; same-label comparisons are discarded.
In \Cref{fig:errors_vs_conf_auroc} (for TriviaQA-Llama 8B) we outline regions of unfair reward/penalty relative to the confidence value of an instance A (red cross in the plot), rather than a fixed threshold. If the judge is right for this instance (A is TP or TN), AUROC gathers  reliable information about the UQ; when the judge is wrong (A is FP or FN), AUROC is distorted. Because AUROC depends on pairs of instances and discards same-label comparisons, we have more regions to consider than in SP with fixed thresholds. For A in FN, every other instance B not in FN will lead to unfair reward, if assigned confidence higher than A (green in the plots), or to unfair penalty, if assigned confidence lower than A (red in the plots): when B is TP, the UQ collects reward for ranking a P over an apparent N (under human judgement this would have been a tie); when B is TN, A and B appear to tie, but that would not be the case under human judgement, so the UQ undeservedly foregoes a penalty; when B is FP, the UQ collects rewards for ranking an apparent P over an apparent N  (under human judgement this would have been a penalty). For A in FP, penalties and rewards are reversed in each case. To make each plot visually representative of an actual trend, we place A (the red cross) in the \emph{median} of the scattered points. As before, imbalance in concentration of points in FP and FN amounts to a net impact that will misrepresent the informativeness of the UQ.

GPT5.4 judge's errors may be impacting P(seq.) more profoundly than ROUGE-L@0.1. Focusing on FN instances: GPT5.4's negatively impactful (red) regions are densely populated; while ROUGE-L@0.1 seems to contain instances that trade-off unfair rewards and penalties, when we iterate across FN instances. 
In another example, when observing the trade-off regions of the FP instances of different judges for Norm. P(Seq.), ROUGE-L@0.1 appears to be negatively impacted (with green and red regions populated, with a dense cluster of points in the red region). Simultaneously, LLama 8B, whose FP error distribution is skewed towards higher values, will more often benefit from unfair rewards of the instances in the TP and TN green regions; with a possibly lower negative impact on the quantifier compared to ROUGE-L@0.1. 
Even the same judge can impact different UQs in distinct ways. Rouge-L@0.1 has a similar distribution of FP and FN errors across semantic entropy scores; with an overall minimal impact on that UQ, as the positive and negative impacts balance each other out. 
Considering the FP and FN distributions across Verb. Conf. scores, errors do not cancel each other out (since TP and TN errors distribute very differently; with the judge harming the UQ).

The way the judges err has a non-trivial impact on UQ evaluation, with plots such as the ones we just analysed demonstrating precisely that.
Hence, the choice of judge should be informed by its use in UQ evaluation. Simply opting for a judge that is aligned with human judgment (\emph{e.g.}, high F1), while agnostic to the application of UQ evaluation, could lead to suboptimal decisions. 
Data analyses of the kind we use can help in a more thorough evaluation, which however require human annotations for the quality of the judge's predictions. Future work could investigate whether we could determine insufficient judges before the collection of human data, by analysing patterns of the data, models, judges and UQs, that underlie the different error categories.

\section{Recommendations}
We conclude this section by gathering some suggestions and recommendations for practitioners working within uncertainty quantification.
To begin with, in settings we have tested, we note that even a relatively small number of annotations suffices to discriminate judges reliably. 
Hence, practitioners should not be discouraged if practical constraints allow for gathering only a small number of annotations. Of course, this observation may well be task/generator-sensitive, but with Bayesian analysis, of the kind we use here, we get reassurances (or warnings) in the form of posterior intervals for the metrics in consideration.
When choosing an automated judge in a new experimental setup, it is necessary to validate it. At first, one might think that it is sufficient to validate the judge as a tool for NLG evaluation (\eg, correctness classification performance assessed against a human judge). That, however, turns out to be insufficient, for judgement errors and selection errors are not typically independent. Instead, validation needs to be performed \emph{within} a UQ evaluation protocol (\eg, correlation between human-powered and machine-powered SP or AUROC).
It must also be noted that, while validation of a judge as tool for NLG evaluation makes the conclusions dependent on task and generators, validation of a judge as a tool for UQ evaluation makes the conclusions further dependent on the specific uncertainty quantifiers being compared and the evaluation protocol in use (\eg, SP at fixed threshold, AUROC, \etc). All of that serves to emphasise that borrowing decisions from published work while transferring them to different settings is a gamble, and reference to published work in such cases cannot be safely regarded as a substitute for validation.

\section{Conclusion}
We delve into an in-depth analysis of errors caused by automated correctness judgement during UQ evaluation. We document current practices in literature, 
and reveal that automated judges are rarely validated.
We uncover that judges' errors impact UQs differently; with some unfairly rewarded and others harmed, in experiments we conduct using two QA datasets, four LLMs, seven popular UQs and seven judges. 
This impact is not fully explained by the judges' ability to simply align with human judgement. In fact, in analyses we show that \emph{how} and \emph{when} judges err plays an important role in explaining such impacts.
Hence, employing a judge without understanding \emph{how} it errs, may cloud insights and conclusions.

\section*{Limitations}

Our meta analysis only considers papers that belong to the ACL anthology from 2020 up until March 2026 (leaving other possibly relevant venues out). As such, not all published research papers that are concerned with UQ in NLG tasks that use selective prediction as part of their evaluation protocol is part of our analysis. Nevertheless, from a small scale inspection of papers published at other venues, the findings of our meta-analysis seem to hold across publication venues.

Due to the expensive nature of manual annotations, we were only able to annotate 100 responses per dataset-generator pair. However, even with a limited number of annotations, our BDA analyses piece together trends and patterns found in the data.
Ideally, we would be able to have multiple annotators judge the correctness of responses for all dataset-generator pairs; due to budget constraints we only do so for two.For those, we assess the alignment among annotators and the effect of annotations from different human judges on our results. We find that human judgements are well aligned and results hold similar across annotators, showing that our results are robust to different annotators. 

Lastly, our experiments are focused on short-form QA, since most UQ techniques are grounded within this task. We leave expanding to long form QA as possible future work. Nonetheless, evaluating the correctness of a longer response can be at least as difficult, if not more, as a short response; hence, it remains just as important, if not more, to validate the judges' quality within UQ evaluation.

\section*{Acknowledgements}
We would like to thank the anonymous reviewers for their feedback.

\bibliography{custom}

\appendix

\section{Meta-analysis}

\subsection{Methodology}
\label{sec:appendix_meta_analysis_methods}

We describe in more detail below the methodology we followed in order to conduct our meta-analysis. 

\begin{itemize}
    \item We gather titles of papers from the entire ACL ontology from 2020-2026
    \item We create a list of relevant to the topic key words: \emph{uncertainty},
    \emph{variability}, \emph{variation}, \emph{aleatoric}, \emph{epistemic}, \emph{evidential}, \emph{confidence}, \emph{calibration}, \emph{selective prediction}, \emph{selective generation}
    \item Filter papers whose title contains any of the key words. This resulted in 600 papers. We manually go through the titles of the papers and classify them as possibly relevant or irrelevant, wrt confidence or uncertainty quantifiers within language generative tasks. This resulted in 206 papers.
    \item We manually go through the abstracts of these papers and if necessary, the methods and experiments of those papers to identify ones that are actually relevant to uncertainty quantification within a selective prediction framework or calibration.
    \item We filtered irrelevant papers, papers that only concern themselves with classification problems, or have a very specific, unique or unconventional evaluation method. The final set is comprised of 67 papers. 
    \item We identify the correctness criterion used in each of these papers and their evaluation metrics. We then analyse and summarise information and statistics from gathered information.
\end{itemize}

\subsection{Experiments}
\label{sec:appendix_meta_analysis_experiments}

The complete table containing including all retrieved papers and their associated judges and evaluation metrics can be found in \Cref{table:meta-analysis}.

We observe a variety of correctness judge types, and even variation within a single type, revealing that there is no one universally employed method within the field. Our later analysis reveals trade-offs of various of these. 

Additionally, we gather statistics of the UQ evaluation metrics used in practice (see \Cref{tab:evaluation_metrics}). AUROC is employed most often ($\simeq 60\% $of papers employ AUROC). Other SP metrics, such as AUCPR (which assesses precision-recall trade-offs), AURAC (which assesses accuracy or risk vs coverage tradeoffs) and Prediction Rejection Ratio (PRR) are popular. Calibration metrics such as Expected Calibration Error (ECE) and Brier Score are also often employed. Less often, but still noteworthy, correlation scores (\eg Pearson) among confidence values and errors are used.

\onecolumn

\begin{center}
\begin{longtable}{p{0.2\textwidth}
    p{0.22\textwidth}
    p{0.22\textwidth}
    p{0.28\textwidth}}
\caption{An overview of all tables} \label{table:meta-analysis} \\
    \toprule
    Paper & Judge Type & Judge & Evaluation method \\
    \midrule
    \endfirsthead
    \toprule
    Paper & Judge Type & Judge & Evaluation method \\
    \midrule
    \endhead
    \midrule
    \multicolumn{2}{r}{\emph{Continued on the next page}} \\
    \endfoot
    \bottomrule
    \endlastfoot

\citet{khanmohammadi-etal-2026-reliable}	&	Exact/fuzzy matching, LLM judge	&	String match, LLM (GPT-5-nano)	&	AUROC, AUCPR, ECE, Brier	\\ \\
\citet{testoni-calixto-2026-mind}	&	Exact/fuzzy matching, LLM judge, Human validation	&	String match, LLM (LLaMA-3.1-8B-Instruct), Human validation	&	AUROC, ECE, Brier	\\ \\
\citet{kunitomo-jacquin-etal-2026-evidential}	&	LLM judge	&	LLM (gpt4o-mini)	&	AUROC, AUARC	\\ \\
\citet{cao-etal-2026-semantic}	&	LLM judge	&	LLM (GPT-4.1)	&	AUROC	\\ \\
\citet{kunitomo-jacquin-etal-2025-role}	&	LLM judge	&	LLM ( Meta-Llama-3-8B-Instruct)	&	AUROC	\\ \\
\citet{huang-etal-2025-confidence}	&	Human labelling	&	Human labeling	&	AUROC	\\ \\
\citet{cecere-etal-2025-monte}	&	LLM judge	&	LLM (Claude Haiku 3.5)	&	AUROC, PR-AUC, AURAC	\\ \\
\citet{bodhwani-etal-2025-calibrated}	&	Human labelling	&	Human judges (from datasets)	&	AUPRC, AUROC, ECE, Brier	\\ \\
\citet{vashurin-etal-2025-benchmarking}	&	LLM judge, Overlap based &	Align-Score, COMET, Rouge-L	&	Prediction Rejection Ratio (PRR)	\\ \\
\citet{vazhentsev-etal-2025-token}	&	LLM judge, Overlap based	&	RougeL, FactScore	&	Prediction Rejection Ratio (PPR), AUROC, PR-AUC	\\ \\
\citet{v-etal-2025-sunar}	&	Exact/fuzzy matching	&	CoverEM	&	nDCG@k, R@k	\\ \\
\citet{chen-etal-2025-llms-recognize}	&	Human labelling	&	Human judges (authors)	&	Predictive Rate Ratio (PRR)	\\ \\
\citet{ishii-etal-2025-fine}	&	LLM judge, Human validation	&	LLM judge (GPT-4.1), Human validation	&	ECE, Brier, AUROC, AUPRC	\\ \\
\citet{joo-cho-2025-cleanse}	&	Overlap based	&	Rouge-L>0.7	&	AUROC, Pearson Correlation Coefficient	\\ \\
\citet{yaldiz-etal-2025-design}	&	LLM judge, Human validation	&	LLM judge (GPT-3.5-turbo), Human validation	&	AUROC, Prediction Rejection Ratio (PRR)	\\ \\
\citet{wang-etal-2025-semantic}	&	Overlap based, Semantic similarity	&	BLEU, BertScore	&	Pearson Corr. Coefficient, RCE, Factuality vs Coverage curve, Uncertainty Precision Alignment	\\ \\
\citet{yang-etal-2025-maqa}	&	Exact/fuzzy matching	&	Exact match, precision (for multiple correct answers)	&	AUROC, PR-AUC	\\ \\
\citet{xue-etal-2025-mlingconf}	&	Exact/fuzzy matching, Human validation	&	positive-recall exact matching, human validation	&	AUROC, ECE	\\ \\
\citet{nguyen-etal-2025-beyond}	&	Overlap based, Semantic similarity	&	F1, RougeL, BertScore	&	AUROC, AURAC, PRR	\\ \\
\citet{xia-etal-2025-survey}	&	LLM judge	&	LLM judge ( LLaMA3.1-8B-Instruct)	&	AUROC, AUARC	\\ \\
\citet{zhang-zhang-2025-cot}	&	Unidentifiable	&	Unidentifiable	&	AUROC	\\ \\
\citet{feng-etal-2025-rethinking}	&	Unidentifiable	&	Unidentifiable	&	AUROC	\\ \\
\citet{bhattacharjya-etal-2025-simba}	&	Overlap based	&	RougeL>0.5	&	AUROC, ACE, ATS	\\ \\
\citet{wang-etal-2025-genuine}	&	Overlap based	&	Rouge-1>=0.3	&	AUROC, ECE, NLL	\\ \\
\citet{zhou-etal-2025-evaluating}	&	Unidentifiable	&	Unidentifiable	&	Brier Score	\\ \\
\citet{yoffe-etal-2025-debunc}	&	Unidentifiable	&	Unidentifiable	&	AUROC	\\ \\
\citet{zhang-etal-2025-atomic}	&	LLM judge, Human validation	&	FactScore, Human validation	&	AUROC, Brier, ECE	\\ \\
\citet{vashurin-etal-2025-uncertainty}	&	Semantic similarity	&	COMET, XComet-XXL, MetricX-XXL, AlignScore	&	PR-curve,  Prediction-Rejection Ratio (PRR)	\\ \\
\citet{khanmohammadi-etal-2025-calibrating}	&	LLM judge	&	LLM (GPT-4o-mini)	&	ECE, Brier, AUROC, AUCPR, ACC	\\ \\
\citet{zhang-etal-2025-roads}	&	Unidentifiable	&	Unidentifiable	&	AUROC, ECE, Brier	\\ \\
\citet{shelmanov-etal-2025-head}	&	LLM judge	&	LLM judge (GPT-4o)	&	PR-AUC	\\ \\
\citet{xue-etal-2025-ualign}	&	Exact/fuzzy matching, Human validation	&	Response in reference or vice versa, Human validation	&	Precision, Truthfulness, AUROC	\\ \\
\citet{he-etal-2025-mmboundary}	&	Semantic similarity, Human validation 	&	pairwise similarity scores with threshold, Human validation	&	ECE, MECE, AUROC	\\ \\
\citet{li-etal-2025-towards}	&	LLM judge, Overlap based	&	Rouge-L>0.7 + LLM judge (GPT-turbo-3.5-061)	&	AUROC, ECE	\\ \\
\citet{bakman-etal-2025-reconsidering}	&	LLM judge	&	LLM judge (GPT-4o-mini)	&	Prediction Rejection Ratio, Average Recall Error	\\ \\
\citet{flores-etal-2025-improving}	&	Overlap based	&	ROUGE-L, BLEU, F1	&	Spearman correlation 	\\ \\
\citet{mora-cross-calderon-ramirez-2024-uncertainty}	&	Semantic similarity	&	BERTScore	&	ECE	\\ \\
\citet{tao-etal-2024-trust}	&	LLM judge	&	LLM judge (GPT-4)	&	ECE, Pearson Correlation, Spearman Correlation	\\ \\
\citet{amayuelas-etal-2024-knowledge}	&	Exact/fuzzy matching, Semantic similarity	&	SimCSE > tau or fuzzy matching	&	AUROC curves, Equal Error Rate (EER) 	\\ \\
\citet{zhao-etal-2024-fact}	&	Exact/fuzzy matching	&	Exact match after normalization	&	ECE, MacroECE	\\ \\
\citet{fadeeva-etal-2024-fact}	&	Human labelling, LLM judge	&	Human labelling, FactScore (retrieval+ChatGPT)	&	AUROC, AUPRC	\\ \\
\citet{srinivasan-etal-2024-selective}	&	LLM judge	&	LLM judge (LAVEGPT-3.5)	&	risk, coverage, Effective Reliability, Selective Prediction Recall	\\ \\
\citet{wang-etal-2024-conu}	&	Semantic similarity	&	Sentence similarity (DistillRoberta)) > 0.7	&	AUROC	\\ \\
\citet{huang-etal-2024-uncertainty}	&	Overlap based, semantic similarity, LLM judge	&	Rouge-L, BERT similarity, LLM judge (ChatGPT)	&	Rank calibration error	\\ \\
\citet{zhang-etal-2024-luq}	&	LLM judge, Human validation	&	FACTSCORE, Human validation	&	Pearson Correlation Coefficient (PCC), Spearman Cor- relation Coefficient (SCC)	\\ \\
\citet{xu-etal-2024-sayself}	&	Exact/fuzzy matching	&	annotated answers must be present within the responses	&	ECE, AUROC	\\ \\
\citet{lin-etal-2024-contextualized}	&	LLM judge, Human validation	&	LLMs as judges ( LLaMA2-70B + gpt-3.5-turbo), Human validation	&	AUROC, AURAC	\\ \\
\citet{liu-etal-2024-enhancing-language}	&	Overlap based, LLM judge	&	ROUGE > 0.3, LLM judge (GPT4)	&	ECE, Brier	\\ \\
\citet{liu-etal-2024-llms-learn-uncertainty}	&	Overlap based	&	Rouge-L > 0.5	&	AUROC	\\ \\
\citet{shen-etal-2024-smartcal}	&	Exact/fuzzy matching	&	Exact Match	&	ECE	\\ \\
\citet{cheng-vlachos-2024-measuring}	&	Semantic similarity	&	COMETKiwi	&	Spearman ($\rho$) and Pearson (r) correlations	\\ \\
\citet{gao-etal-2024-spuq}	&	Overlap based	&	F1	&	ECE, Pearson Correlation between confidence and accuracy	\\ \\
\citet{mahaut-etal-2024-factual}	&	Unidentifiable	&	Unidentifiable	&	AUPRC	\\ \\
\citet{duan-etal-2024-shifting}	&	Overlap based, Semantic similarity, Human validation	&	Rouge-L > 0.5 + embed. similarity (Distilroberta) > 0.5, Human validation	&	AUROC	\\ \\
\citet{chen-mueller-2024-quantifying}	&	LLM judge, Human validation	&	LLM judge (GPT- 4), Human validation	&	AUROC	\\ \\
\citet{bakman-etal-2024-mars}	&	LLM judge	&	LLM (GPT-3.5-turbo)	&	AUROC	\\ \\
\citet{zablotskaia-etal-2023-uncertainty}	&	Overlap based	&	ROUGE-1,2,L	&	ECE, AUROC, Quality vs Abstention Curves	\\ \\
\citet{chen-etal-2023-adaptation}	&	Overlap based, Human validation	&	Rouge-L, Human validation	&	AUROC, AUACC	\\ \\
\citet{tian-etal-2023-just}	&	LLM judge, Human validation	&	LLM (GPT4, 3.5), Human validation	&	AUC, ECE, Brier Score	\\ \\
\citet{fadeeva-etal-2023-lm}	&	Overlap based, Semantic similarity	&	Rouge-L, BERTScore	&	Prediction Rejection Ratio (PRR)	\\ \\
\citet{mielke-etal-2022-reducing}	&	Human labelling, Exact/fuzzy matching, Semantic similarity	&	Human labelling, Match based, Bert based trained classifier	&	ECE, MCE, NLL	\\ \\
\citet{dhuliawala-etal-2022-calibration}	&	Unidentifiable	&	Unidentifiable	&	ECE, AURC, AURCC	\\ \\
\citet{si-etal-2022-examining}	&	Exact/fuzzy matching	&	Exact match	&	ECE, MacroCE	\\ \\
\citet{lu-etal-2022-learning}	&	Human labelling	&	Human labelling	&	Pearson’s correlation, AUROC, AUPR	\\ \\
\citet{zhang-etal-2021-knowing}	&	Exact/fuzzy matching	&	exact match	&	AUROC, Cov@Acc	\\ \\
\citet{xiao-wang-2021-hallucination}	&	Exact/fuzzy matching, Human validation	&	Exact match (with synonyms of references), BLEU, Human validation	&	Risk - performance curves	\\ \\
\citet{wang-etal-2020-inference}	&	Overlap based	&	Translation Edit Rate (TER)	&	ECE	\\ \\
\end{longtable}
\end{center}

\twocolumn

\begin{table}[ht]
\centering
\begin{tabular}{lcc}
\hline
\textbf{UQ Eval. metric} & \textbf{\# Papers} & \textbf{Proportion} \\
\hline
AUROC & 40 & 59.7\% \\
ECE  & 26 & 38.8\% \\
AUCPR  & 12 & 17.9\% \\
AURAC  & 11 & 16.4\% \\
Other  & 11 & 16.4\% \\
Brier  & 10 & 14.9\% \\
PRR  & 8 & 11.9\% \\
Correlation scores  & 8 & 11.9\% \\
 
\hline
\end{tabular}
\caption{UQ evaluation metrics used in retrieved papers.}
\label{tab:evaluation_metrics}
\end{table}

\section{\label{sec:bda}Bayesian Data Analysis}

In Section \ref{sec:bda-judges}, we compare automated judges to one or more human judges in terms of their adequacy judgements. 
In Section \ref{sec:bda-quantifiers}, we learn to map a diverse collection of uncertainty quantifiers to a comparable notion of confidence, which we use in order to analyse the impact of automated judgements on metrics of selective prediction. 
In both cases, we employ Bayesian data analysis techniques \citep[BDA;][]{gelman1995bayesian}.

We implement our hierarchical models using numpyro \citep{phan2025numpyro}.   
For inference, we condition on the available data and draw posterior samples for latent parameters via Markov chain Monte Carlo (MCMC), using numpyro's implementation of the \emph{no U-turn sampler} \citep[NUTS;][]{JMLR:v15:hoffman14a}. We use standard convergence diagnostics \citep{brooks1998general}, such as the r-hat statistic, by comparing multiple Markov chains. In all cases, we follow the common approach of experimenting with various choice of priors (as well as tweaks to the model's hierarchical structure) in order to verify that inferences are stable across a range of sensible choices. In development, we use posterior predictive checks, such as label reconstruction, as a tool for further diagnosing design choices as well as MCMC convergence. These practices are common in BDA and they are widely documented in reference texts  \citep{gelman1995bayesian,mcelreath2018statistical,Bayesian-Workflow:2026}. 
Our code and notebooks are available for full reproducibility here: \url{https://github.com/probabll/buqeval}.

In describing our BDA tools, we will refer to a prompt-response pair as an \emph{item}, each such item is classified by various (human and machine) \emph{judges}  who independently assess whether the response is adequate to the prompt (`the item is positive') or not (`the item is negative'). Judges (sometimes called annotators) can be human or machine.

\subsection{\label{sec:bda-judges}Judges}

We have a collection of $I$ items, each judged by $\mathrm{J}$ judges, for a total of $N = I \times J$ observations.\footnote{For simplicity, we describe the model as if every judge had annotated every item once, but our model in fact supports missing data as well as multiple judgements.} 
We gather all such observations in an $N$-dimensional vector $\mathbf y$ of binary adequacy labels and use two other $N$-dimensional vectors, namely $\mathbf j$ and $\mathbf i$, to indicate which judge and item each observation in $\mathbf y$ is about: for some $n \in [N]$, $y_\mathsf{n}$ is 1 (resp., 0) if  judge $j_n$ regards item $i_n$ as positive (resp., negative). We model $\mathbf y$ as a draw from a hierarchical model
\begin{subequations}
\begin{align}
    \text{for } \mathsf{i} \in [I] & \nonumber  \\
    \theta_{\mathsf{i}} &\sim \mathrm{Beta}(\alpha + \mathbf 1_\mathsf{i}, \beta + \mathbf 0_\mathsf{i}) \label{eq:judge-item-prior}\\
    \text{for } \mathsf{j} \in [J] & \nonumber \\
    p_{0,\mathsf{j}} &\sim \mathrm{Beta}(a_0, b_0) \label{eq:judge-specificity} \\
    p_{1,\mathsf{j}} &\sim \mathrm{Beta}(a_1, b_1) \label{eq:judge-sensitivity} \\
    \text{for } n \in [N] & \nonumber  \\    
    y_n|i_n=\mathsf{i}, j_n=\mathsf{j} &\sim \mathrm{Bernoulli}(\phi_{n}) \label{eq:judge-marginal}\\    
    \phi_{n} &= \theta_{\mathsf{i}}p_{1,\mathsf{j}} + (1-\theta_{\mathsf{i}})(1-p_{0,\mathsf{j}}) \nonumber
\end{align}    
\end{subequations}    
described next.
For each item $\mathsf{i} \in [I]$, the parameter $\theta_{\mathsf{i}}$ captures the probability that the item is positive. 
We draw this parameter from a strongly-informed Beta distribution (\ref{eq:judge-item-prior}), where $\mathbf 1_\mathsf{i}$ (resp.,  $\mathbf 0_\mathsf{i}$) counts the number of times our \emph{human judges} labelled the item as positive (resp., negative), and $\alpha=\beta=0.5$.\footnote{In effect, we draw $\theta_\mathsf{i}$ from the posterior distribution of a Beta-Binomial model of the observed human annotation, with a sparse and symmetric $\mathrm{Beta}(0.5, 0.5)$ prior.} 
For each judge $\mathsf{j} \in [J]$: (\ref{eq:judge-sensitivity}) $p_{1,\mathsf{j}}$ models the judge's \emph{sensitivity} (used to infer the judge's true positive rate), that is the probability 
of a positive judgement given a positive item, and (\ref{eq:judge-specificity}) $p_{0,\mathsf{j}}$ models the judge's \emph{specificity} (used to infer the judge's true negative rate), that is the probability 
of a negative judgement given a negative item. For these parameters we use uninformed, flat $\mathrm{Beta}(1, 1)$ priors.\footnote{Except if we are modelling a human judge, for example in order to study how humans differ as judges, in which case we use a weakly informed $\mathrm{Beta}(1, 0.5)$ prior that skews both sensitivity and specificity towards 1 to encode the knowledge that we trust the annotator.} 
Then, given the latent parameters, the probability that the $n$th observation is positive (\ref{eq:judge-marginal}) can be obtained by marginalising the  unknown adequacy label of the corresponding item ($i_n$) under the sensitivity and specificity of the corresponding judge ($j_n$).

\paragraph{Inferences.} 
For each judge $\mathsf{j} \in [J]$, we can then infer the following posterior classification metrics 
\begin{subequations}
    \begin{align}
        \mathrm{TPR}_\mathsf{j} &= \mathbb E\left[ \frac{1}{I} \sum_{i=1}^I \theta_i p_{1,\mathsf{j}} \right]\\
        \mathrm{TNR}_\mathsf{j} &= \mathbb E\left[ \frac{1}{I} \sum_{i=1}^I (1-\theta_i) p_{0,\mathsf{j}} \right]\\
        \mathrm{FPR}_\mathsf{j} &= \mathbb E\left[ \frac{1}{I} \sum_{i=1}^I (1-\theta_i) (1- p_{0,\mathsf{j}}) \right]\\
        \mathrm{FNR}_\mathsf{j} &= \mathbb E\left[ \frac{1}{I} \sum_{i=1}^I \theta_i (1-p_{1,\mathsf{j}}) \right]
    \end{align}
\end{subequations}
by estimating the expected value using samples from $p(\boldsymbol{\theta}, \mathbf p_{0}, \mathbf p_{1}|\mathbf y, \mathbf j, \mathbf i)$. These metrics can then be used to express posterior accuracy and posterior F1. 

\subsection{\label{sec:bda-quantifiers}Uncertainty Quantifiers}

We have a collection of $I$ items, each judged by $\mathrm{J}$ human judges and qualified by $\mathrm{Q}$ uncertainty quantifiers, for a total of $N = I \times J \times Q$ observations. 
We gather all such observations in a collection of $N$-dimensional vectors ($\mathbf y, \mathbf x, \mathbf i, \mathbf j, \mathbf q)$: $y_\mathsf{n}$ is 1 (resp., 0) if  judge $j_n$ regards item $i_n$ as positive (resp., negative), while $x_n$ is the uncertainty (or confidence) score assigned to $i_n$ by UQ $q_n$. We model $\mathbf y|\mathbf x$ as a draw from a hierarchical model described next.
\begin{subequations}
    \begin{align}
        \beta &\sim \mathcal N(0, 2^2) \label{eq:quantifier-top1}\\
        \omega &\sim \mathcal N(0, 2^2) \label{eq:quantifier-top2}\\
        \text{for }\mathsf{j} \in [J] & \nonumber \\
         b_{\mathsf{j}} &\sim \mathcal N(\beta, 2^2) \label{eq:quantifier-judges1}\\
         w_{\mathsf{j}} &\sim \mathcal N(\omega, 2^2) \label{eq:quantifier-judges2} \\
         \text{for }\mathsf{q} \in [Q] & \nonumber \\
         c_{\mathsf{q}} &\sim \mathcal N(0, 2^2) \label{eq:quantifier-q1} \\
         m_{\mathsf{q}} &\sim \mathcal N(0, 2^2) \label{eq:quantifier-q2} \\
         \text{for }\mathsf{n} \in [N] & \nonumber \\
        y_{n} | x_n, j_n=\mathsf{j}, i_n=\mathsf{i} &\sim \mathrm{Bernoulli}(\phi_n)\label{eq:quantifier-obs} \\        
        \phi_n &= \mathrm{sigmoid}(u_n + v_n) \nonumber \\
        u_{n} &=b_{\mathsf{j}} + w_{\mathsf{j}} x_{n}  \label{eq:quantifier-j-predictor}\\
        v_{n} &=c_{\mathsf{q}} + m_{\mathsf{q}} x_{n} \label{eq:quantifier-q-predictor} 
    \end{align}
\end{subequations}
The model is an instance of item-response theory model, with offset and slopes for judges and quantifiers, all of which are modelled by Gaussians. We group the parameters of the human judges (\ref{eq:quantifier-judges1}) and (\ref{eq:quantifier-judges2}) under shared priors  (\ref{eq:quantifier-top1}) and (\ref{eq:quantifier-top2}), but keep the quantifiers' parameters separate (\ref{eq:quantifier-q1}) and (\ref{eq:quantifier-q2}). Finally, given the latent parameters, the probability that the $n$th observation is positive (\ref{eq:quantifier-obs}) is obtained by sigmoid-transforming the sum of the judge-specific linear predictor (\ref{eq:quantifier-j-predictor}) and the UQ-specific linear predictor (\ref{eq:quantifier-q-predictor}).

\paragraph{Inferences.} Having obtained samples from $p(\beta, \omega, \mathbf b, \mathbf w, \mathbf c, \mathbf m|\mathbf y, \mathbf x, \mathbf q, \mathbf j, \mathbf i)$, we can compute posterior confidences $\phi_1, \ldots, \phi_N$, from which we can obtain a point estimate for the posterior confidence $p_{\mathsf{q},\mathsf{i}}$ under $\mathsf{q} \in [Q]$ with which item $\mathsf{i} \in [I]$ is positive: we use the posterior median of all samples $\phi_n$ for which $i_n = \mathsf{i}$ and $q_n = \mathsf{q}$.
We use the posterior median across samples 

\section{Implementation details}

\label{sec:appendix:implement_details}

\subsection{Uncertainty quantifiers}

We present below additional information and implementation details for each of the uncertainty quantifiers used in our experiments. 

\paragraph{Log-Probability}
We compute the log-probability of a response $Y=y$ by decomposing it in tokens and computing their corresponding sum of log-probabilities.

\begin{equation}
\label{equation:log_prob}
\log P(Y=y \mid X = x)
=
-\sum_{i=1}^{N}
\log p(t_i \mid x, t_{<i})
\end{equation}

where $y= t_1,...,t_N$.

\paragraph{Normalised Log-Probability}
We interpret this quantifier either as the average token log-probability, or the length-normalised log probability of the response $y$ (\emph{i.e.} perplexity). It simply is the log-probability from \Cref{equation:log_prob}, divided by the number of tokens in $y$.

\begin{equation}
\label{equation:norm_log_prob}
perplexity= - \frac{1}{n} \log P(Y=y|X = x) 
\end{equation}
\paragraph{Predictive entropy}
The entropy over complete responses $Y$ given a prompt $X=x$, is given by the formula below:
\begin{equation}
H(Y|X = x) = -\sum_{y \in \mathcal Y}p(y|x) \log p(y|x)~,
\end{equation}

where the summation is over the entire (countably infinite) space $\mathcal Y$ of possible responses. We estimate this quantity via Monte Carlo; MC. We sample 10 responses, obtain the probability of each via relative frequency and compute the entropy with the formula above. 

\paragraph{Semantic entropy}
To compute semantic entropy \citep{kuhn2023semantic}, sampled responses are mapped to semantic clusters and entropy is estimated over those. We use \citet{aichberger2025improving}'s implementation:
\begin{equation}
SE(Y|X=x) \approx -\sum_{c_j} p(c_j|x)\log p(c_j|x),
\end{equation}
where $c_1, \ldots, c_J$ are clusters formed using a natural language inference (NLI) model to assess semantic equivalence via bi-directional entailment. $p(c_j|x)$ equal the number of responses, incl. repetitions, in cluster $c_j$ divided by sample size). We use 10 unbiased sampled responses and \texttt{deberta-large-mnli} as the NLI model.

\paragraph{P(True)}
Following \citet{kadavath2022language} we prompt a model to generate true or false regarding the truthfulness of its own previously sampled response to a prompt. 

\textbf{Prompt:}

\texttt{Question: '<question>'}

\texttt{Here are some brainstormed ideas: '<responses>'}

\texttt{Possible answer: '<prediction>'}

\texttt{Is the possible answer: (a) true (b) false}

\texttt{Generate a or b. The answer is:}

We assess the log-probabilities assigned by the
models to the sequences P(‘a’|prompt) and
P(‘b’|prompt), using the model’s output
layer’s logits. We normalise to obtain the probability assigned to P(‘a’) as a proxy to P(True).

\paragraph{Semantic Confidence}
Inspired by \citet{yona-etal-2024-large}, we compute semantic confidence in a response as the relative frequency that a response is not contradicted by sampled responses. Contradiction is assessed by an NLI model.

\begin{equation}
SC(Y=y|X=x) = 1 - \frac{\sum_{y_s \in [y_1,y_n]} 1\{ y \not= y_s\}}{n},
\end{equation}

Two responses are deemed to be contradicted if the NLI model assigns higher probability to contradiction than the sum of the probability of entailment and neutrality. We use the 10 unbiased sampled responses and \texttt{deberta-large-mnli} as the NLI model.

\paragraph{Verbalised Confidence}

We prompt the model for a confidence value in its own previously generated response. The prompt used is found below:

\texttt{You are a helpful assistant that generates faithful numerical confidence values in an answer.}

\texttt{Question: <question>}

\texttt{Answer previously generated: <prediction>}

\texttt{Generate ONLY your numerical confidence in the answer (which is 0-1), up to 2 decimal places.}

\texttt{Confidence:}

\subsection{Correctness Criteria}

We proceed to explain further how we implemented the various correctness criteria.

\paragraph{Exact matching}
We strip and lowercase the response string and assess whether it matches exactly any stripped and lowercased reference responses.

\paragraph{Token F1}
For each response and reference pair (if multiple references exist), we lowercase and split them in words and retrieve the sum of common ones. 
Precision is defined as the sum of common words divided by the number of words in the response and recall is defined as the sum of common words divided by the number of words in the reference. Using these we compute an F1 score. 
We use the maximum F1 score across references and assess whether it surpasses a pre-defined threshold.

\paragraph{BLEU Score}
We use the \texttt{sentence\_bleu} from the \texttt{nltk.translate.bleu\_score} library to
calculate the BLEU score for the response against the reference sentences. 

\paragraph{Rouge-L}
We employ Rouge-L, which measures the longest common subsequence between a response and a reference, via the \texttt{rouge\_scorer.RougeScorer} function from the \texttt{rouge\_score} library. We obtain the max score across references and assess whether it surpasses a pre-defined threshold.

\paragraph{Embedding similarity}
We embed the response and the references. We compute the consine similarity of each response-reference pair. We take the maximum value and compare it to a  threshold. The embedding model we use is \texttt{all-MiniLM-L6-v2}.

\paragraph{BERT score}
From \texttt{bert\_score} we use the \texttt{score} function to compute the semantic similarity between the response and references. The maximum value is used against a pre defined threshold. 

\paragraph{LLM as judge}

We prompt a model using the prompt, response and reference answers for its correctness. The prompt is presented below:

\texttt{You are a judge that evaluates the correctness of an answer given a question and a list of reference answers. Even if one of the reference answers is semantically equivalent to the answer, we can consider it as correct. Generate a True or False answer.}

\texttt{"Question: <question>}

\texttt{Reference answers: <references>}

\texttt{Answer: <answer>}

\texttt{Is the answer correct?}

We employ 2 judges: an open source model, \texttt{Llama-3.1-8B-Instruct}, and a closed source API model, \texttt{gpt-5.4-mini}.

\section{Results}
\label{sec:appendix:results}

\subsection{Extensive results}
\label{sec:appendix:extensive}

\begin{figure*}[t]
    \centering

    \begin{subfigure}[t]{0.48\textwidth}
        \centering
        \includegraphics[width=\linewidth]{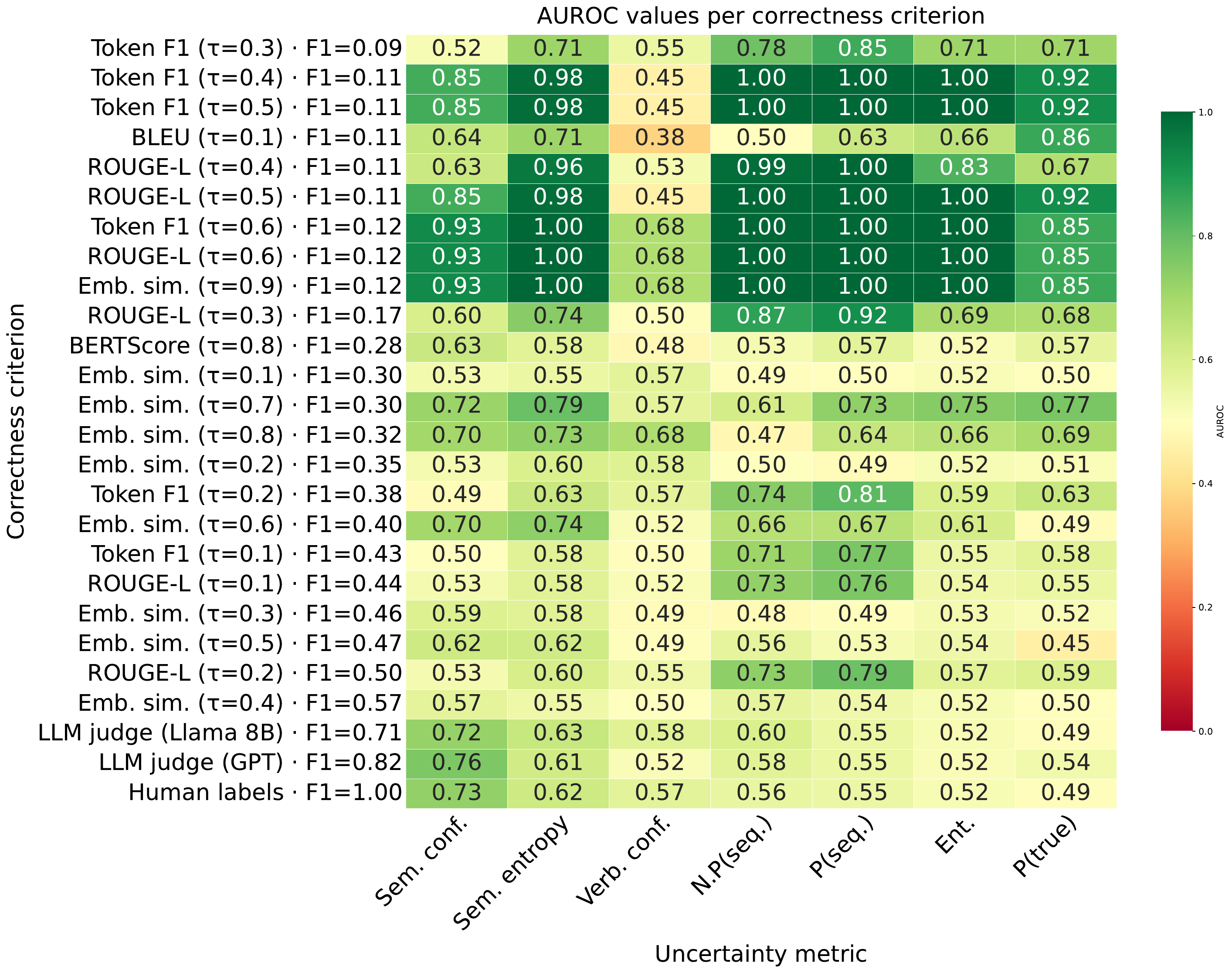}
    \end{subfigure}
    \hfill
    \begin{subfigure}[t]{0.48\textwidth}
        \centering
        \includegraphics[width=\linewidth]{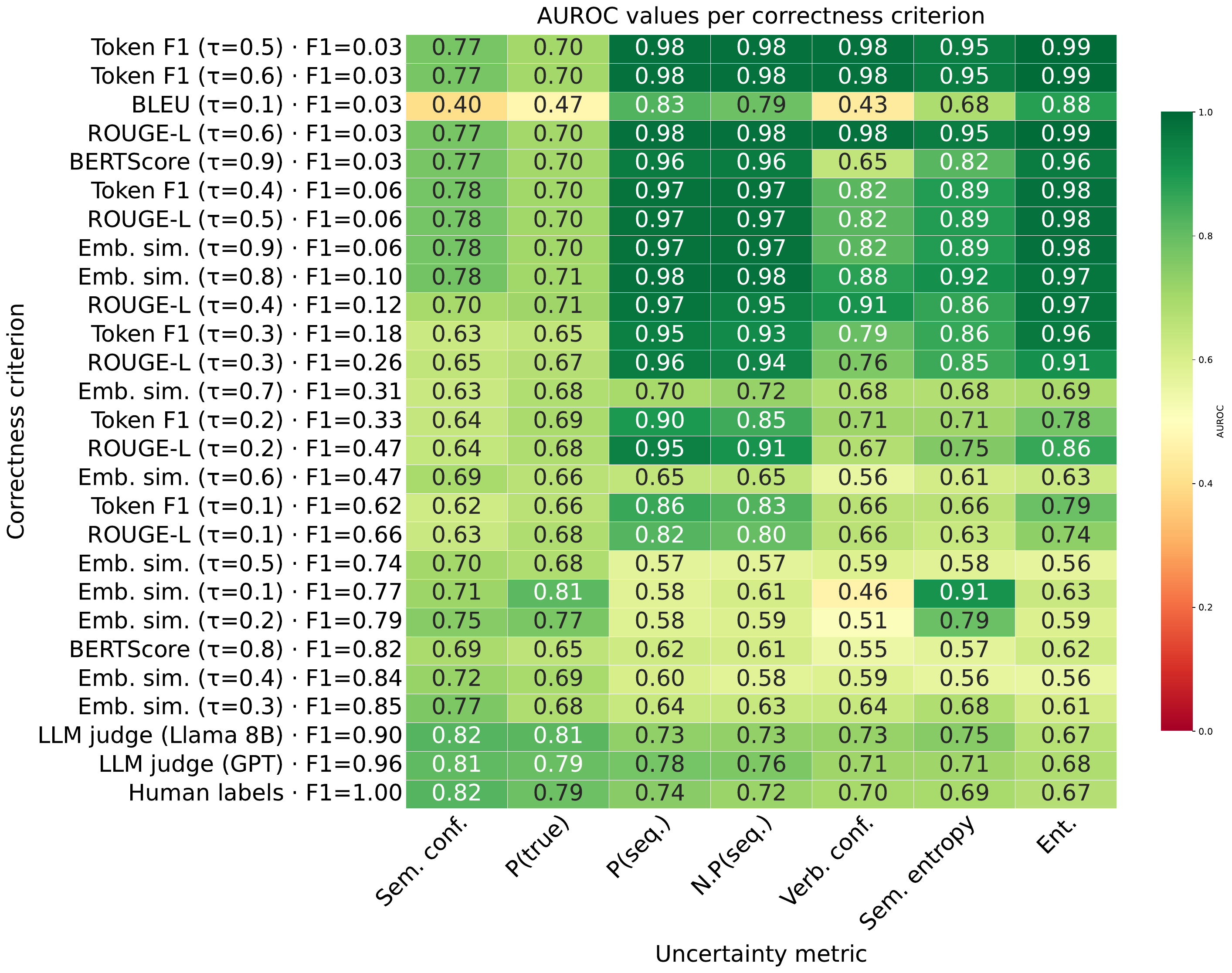}
    \end{subfigure}

    \vspace{0.75em}

    \begin{subfigure}[t]{0.48\textwidth}
        \centering
        \includegraphics[width=\linewidth]{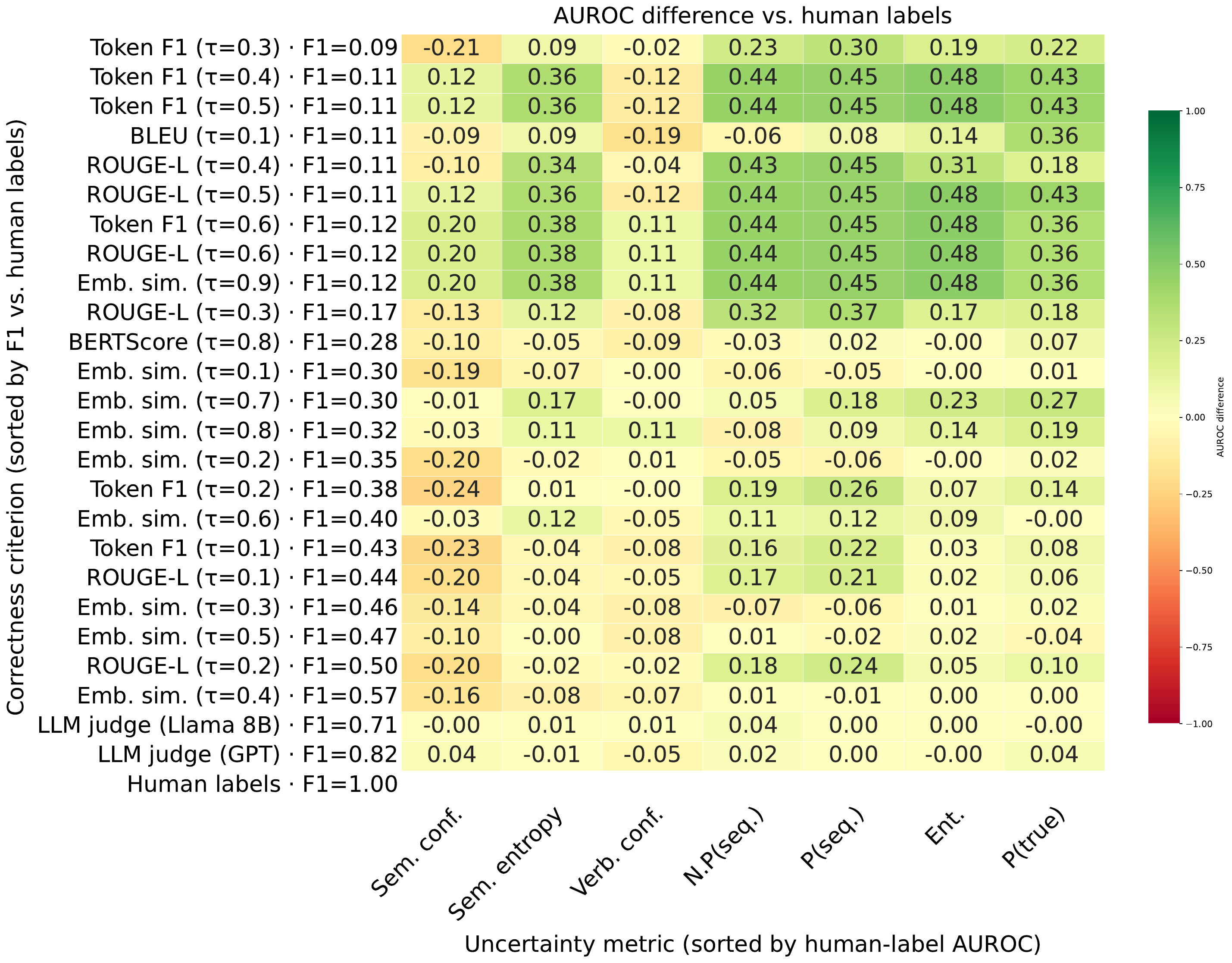}
    \end{subfigure}
    \hfill
    \begin{subfigure}[t]{0.48\textwidth}
        \centering
        \includegraphics[width=\linewidth]{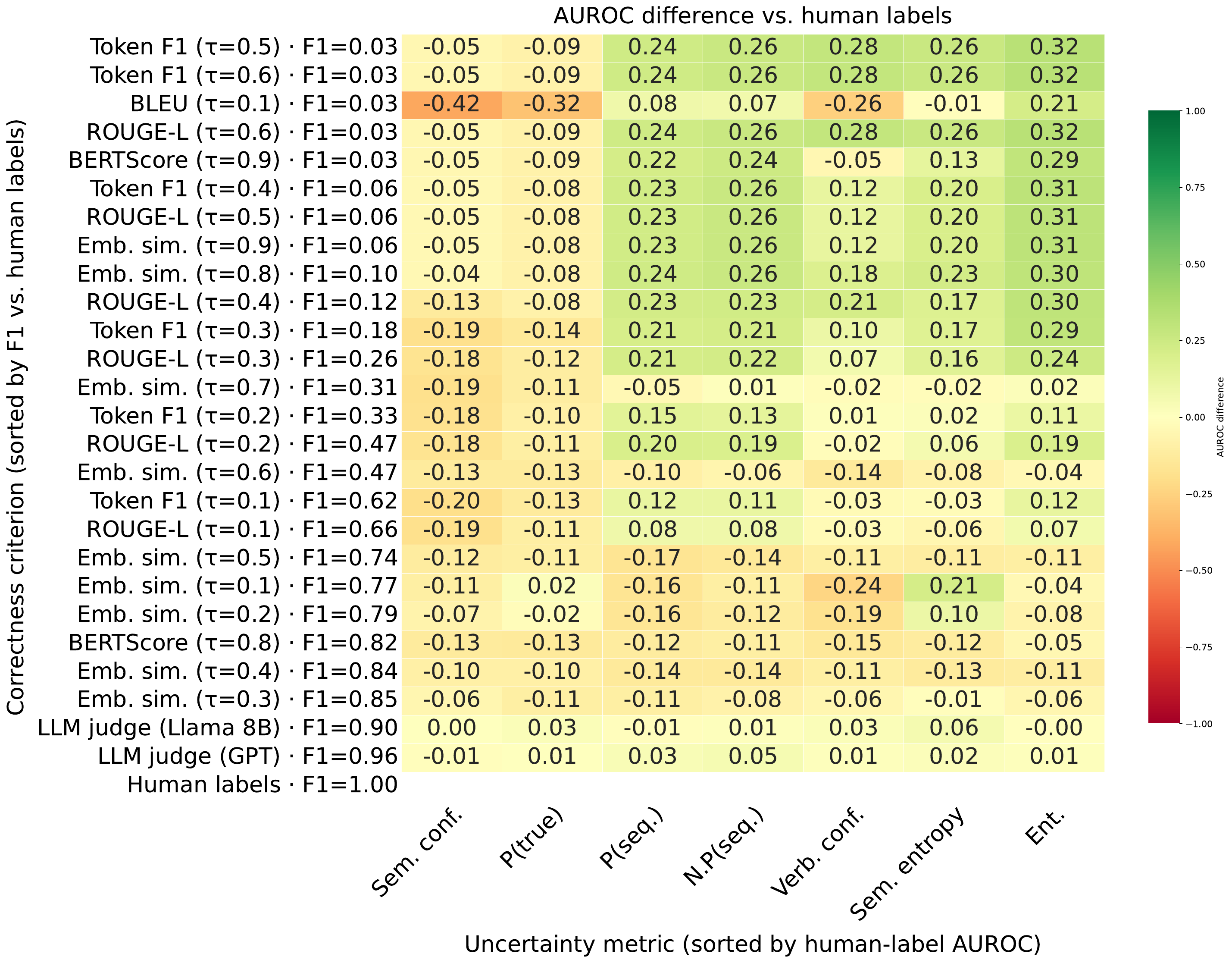}
    \end{subfigure}

    \begin{subfigure}[t]{0.48\textwidth}
        \centering
        \includegraphics[width=\linewidth]{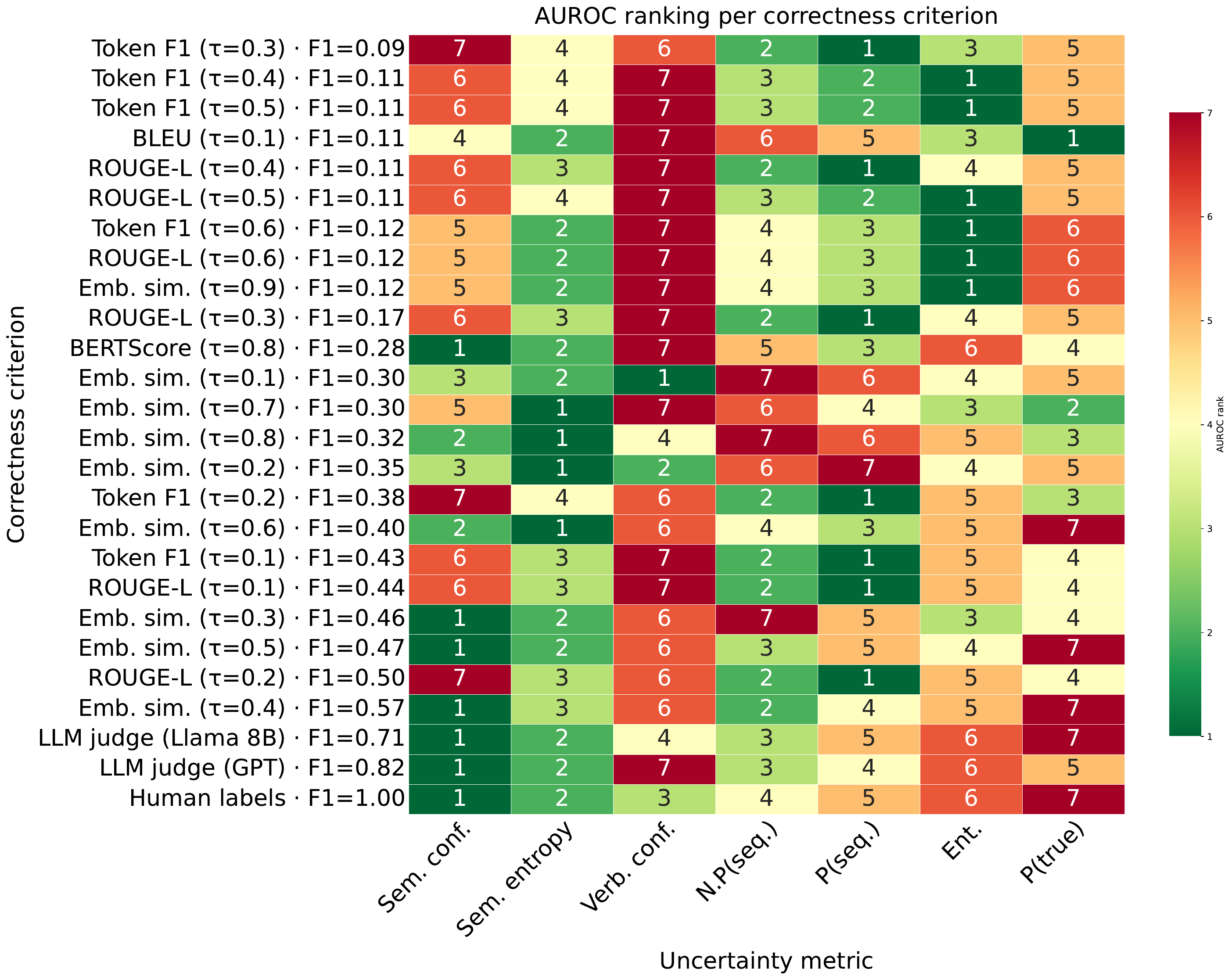}
        \caption{Qwen 0.5B, TriviaQA}
    \end{subfigure}
    \hfill
    \begin{subfigure}[t]{0.48\textwidth}
        \centering
        \includegraphics[width=\linewidth]{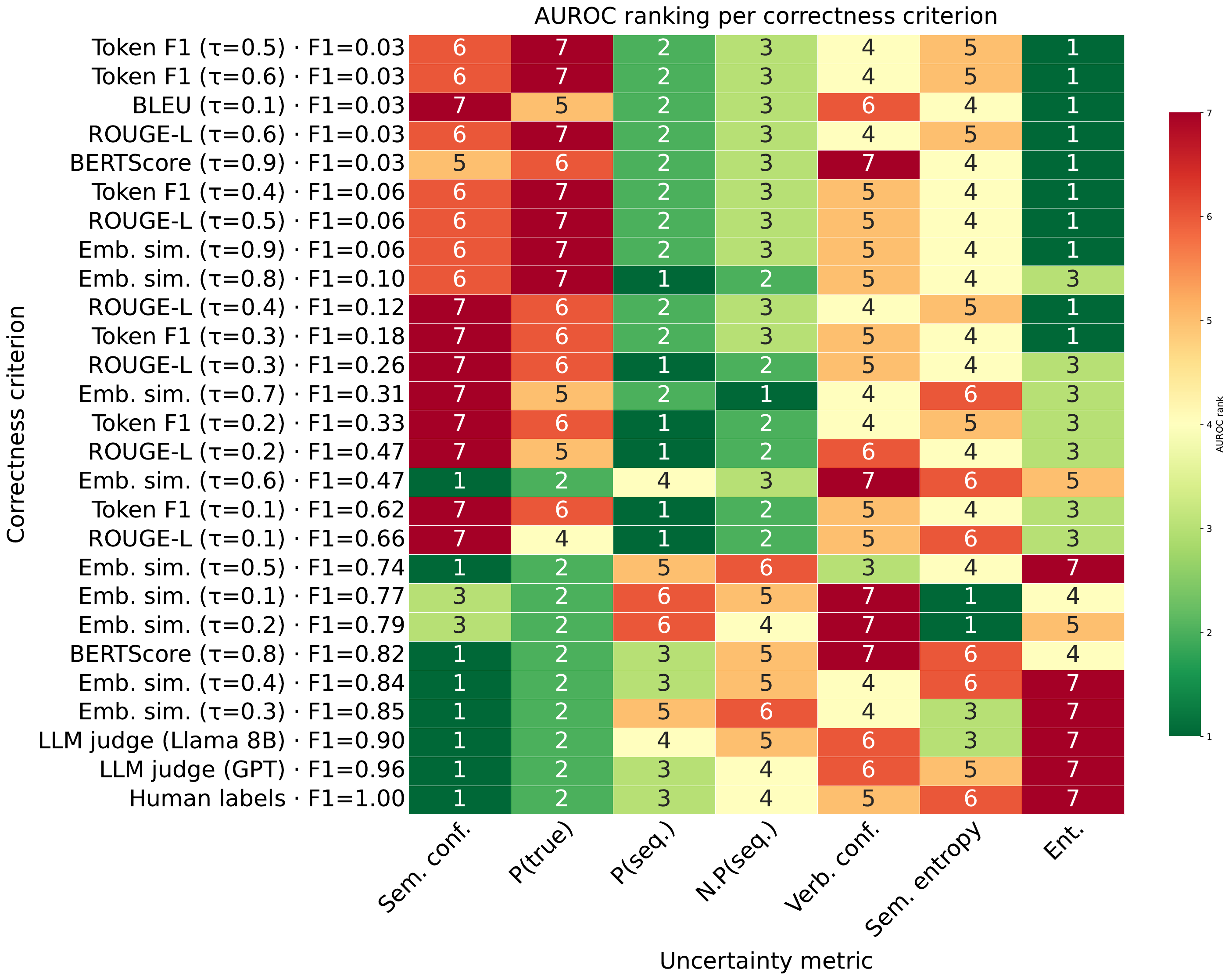}
        \caption{Qwen 7B, TriviaQA}
    \end{subfigure}

    \caption{Top row shows AUROC values, middle row shows differences between oracle and system AUROC values and bottom row rankings for various uncertainty quantifiers (ordered from left to right in a descending order of performance) and correctness functions (ordered from top to bottom in an increasing order of quality). }
    \label{fig:qwen-triviaqa}
\end{figure*}

\begin{figure*}[t]
    \centering

    \begin{subfigure}[t]{0.48\textwidth}
        \centering
        \includegraphics[width=\linewidth]{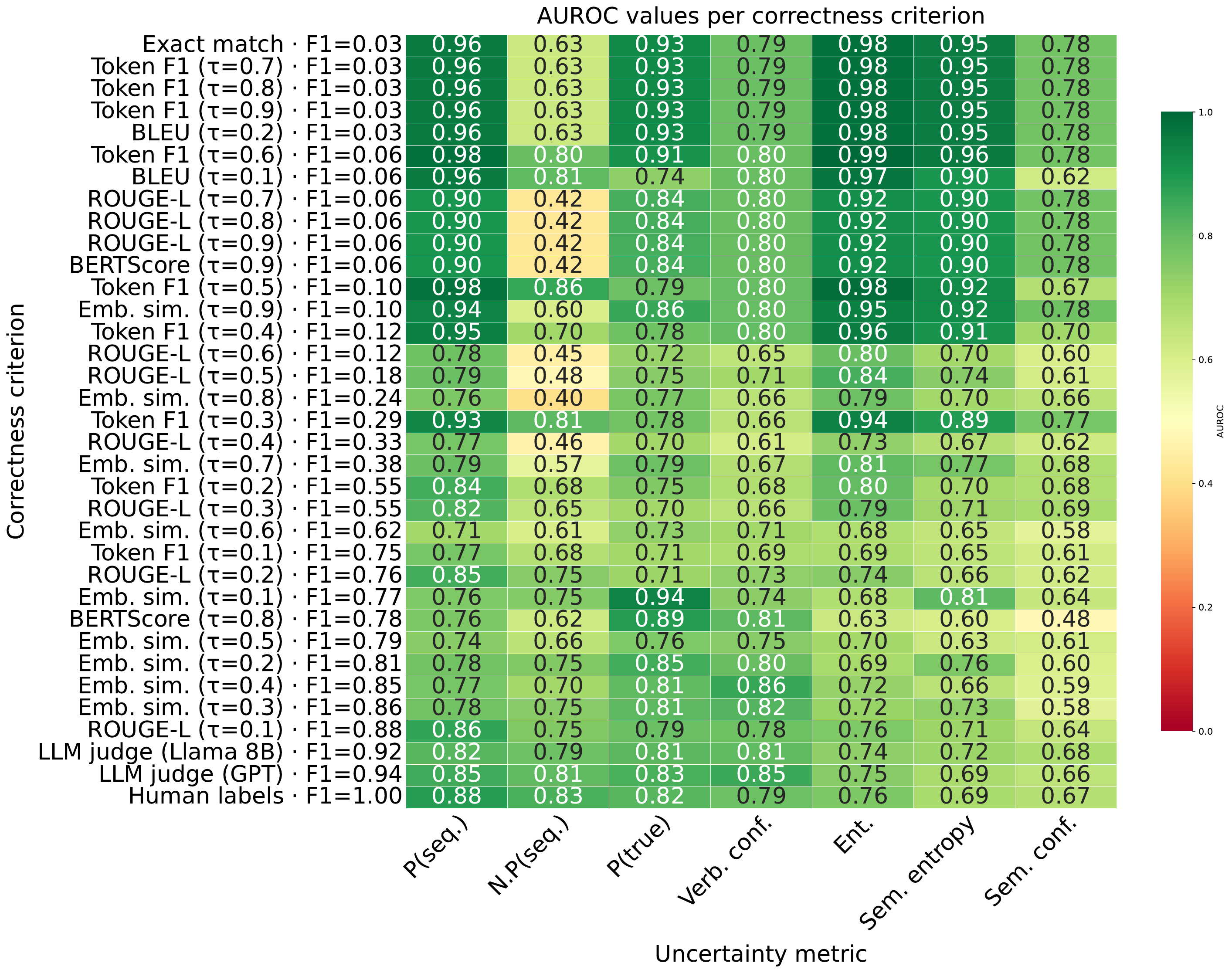}
    \end{subfigure}
    \hfill
    \begin{subfigure}[t]{0.48\textwidth}
        \centering
        \includegraphics[width=\linewidth]{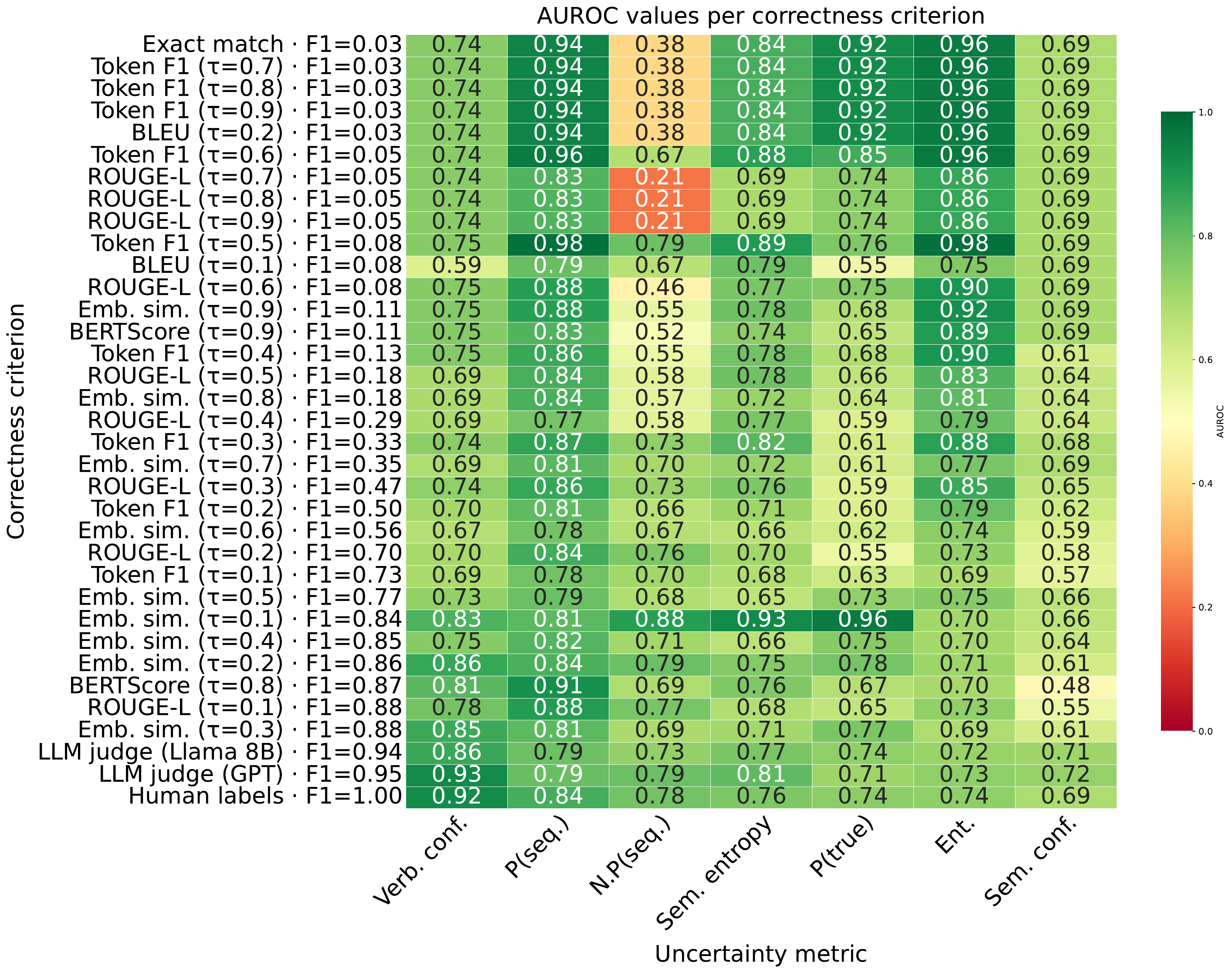}
    \end{subfigure}

    \vspace{0.75em}

    \begin{subfigure}[t]{0.48\textwidth}
        \centering
        \includegraphics[width=\linewidth]{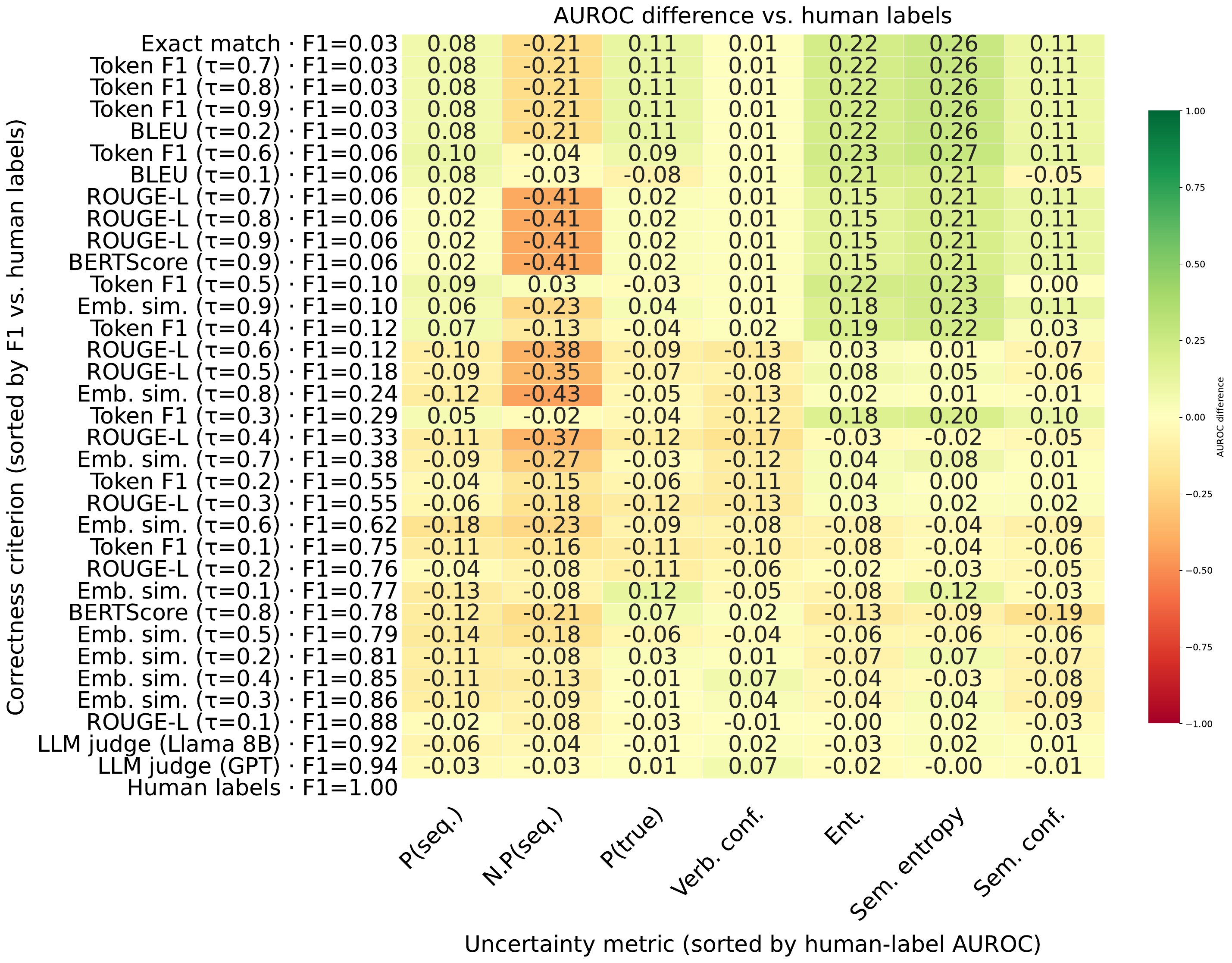}
    \end{subfigure}
    \hfill
    \begin{subfigure}[t]{0.48\textwidth}
        \centering
        \includegraphics[width=\linewidth]{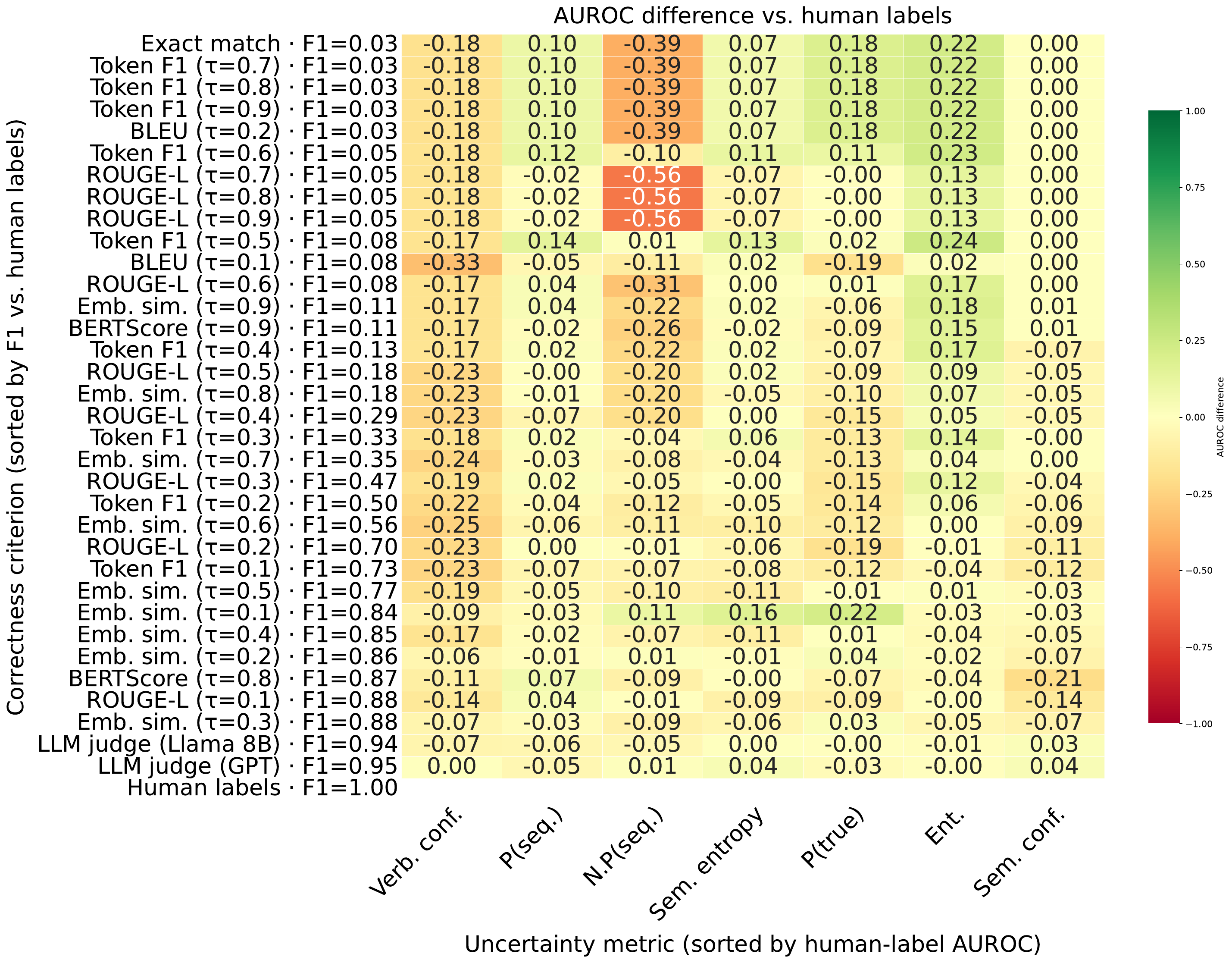}
    \end{subfigure}

    \begin{subfigure}[t]{0.48\textwidth}
        \centering
        \includegraphics[width=\linewidth]{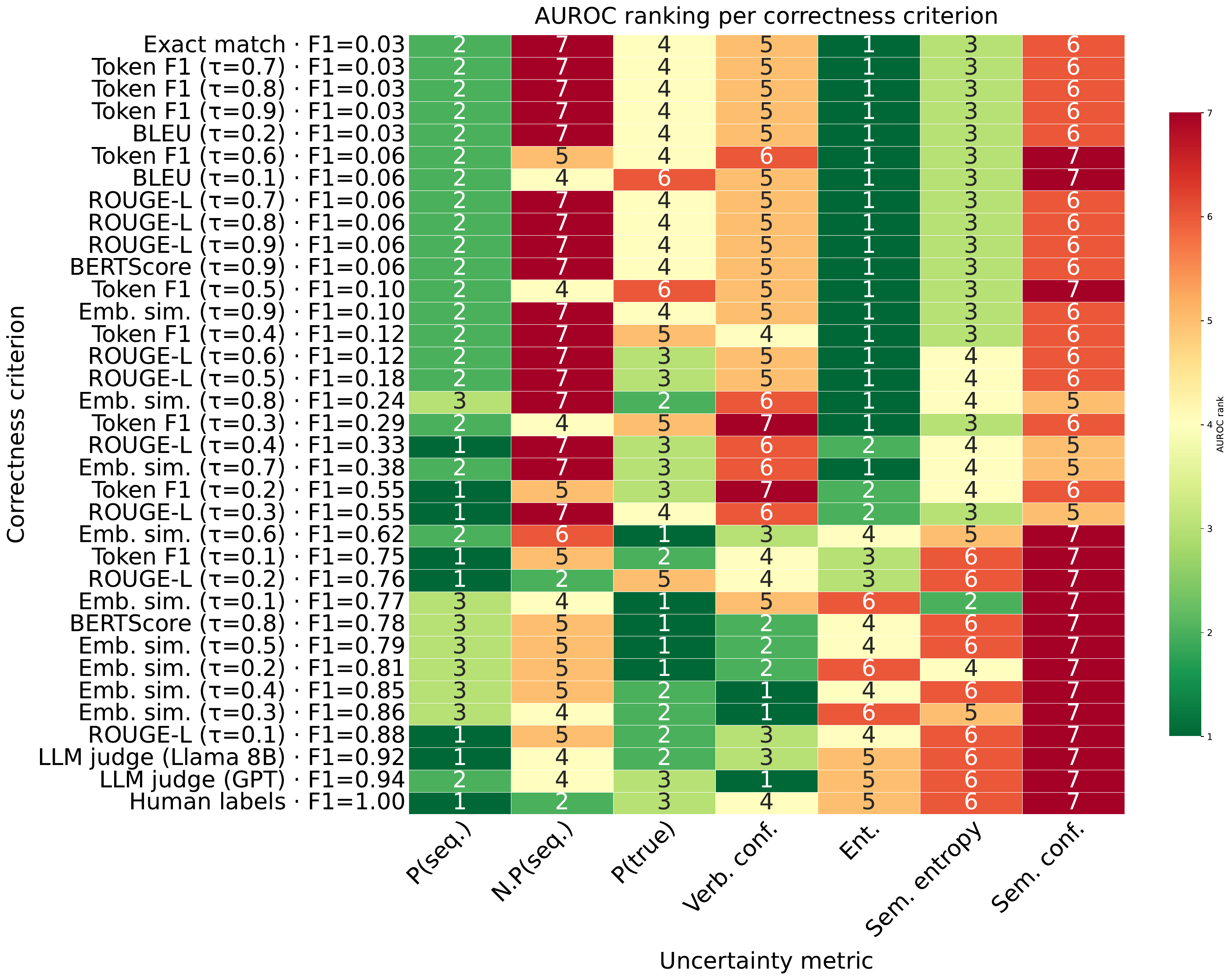}
        \caption{Llama 3B, TriviaQA}
    \end{subfigure}
    \hfill
    \begin{subfigure}[t]{0.48\textwidth}
        \centering
        \includegraphics[width=\linewidth]{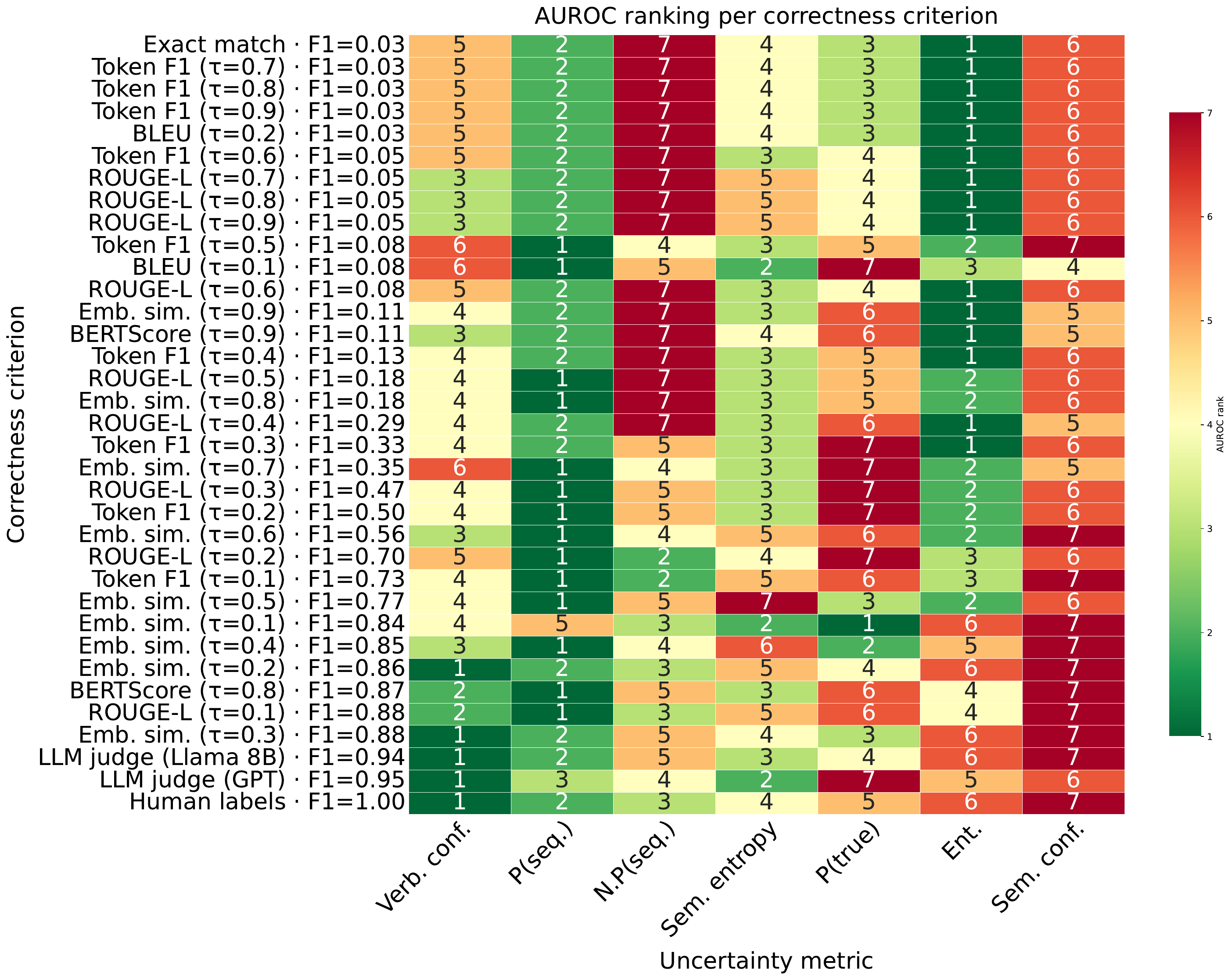}
        \caption{Llama 8B, TriviaQA}
    \end{subfigure}

    \caption{Top row shows AUROC values, middle row shows differences between oracle and system AUROC values and bottom row rankings for various uncertainty quantifiers (ordered from left to right in a descending order of performance) and correctness functions (ordered from top to bottom in an increasing order of quality). }
    \label{fig:llama-triviaqa}
\end{figure*}

\begin{figure*}[t]
    \centering

    \begin{subfigure}[t]{0.48\textwidth}
        \centering
        \includegraphics[width=\linewidth]{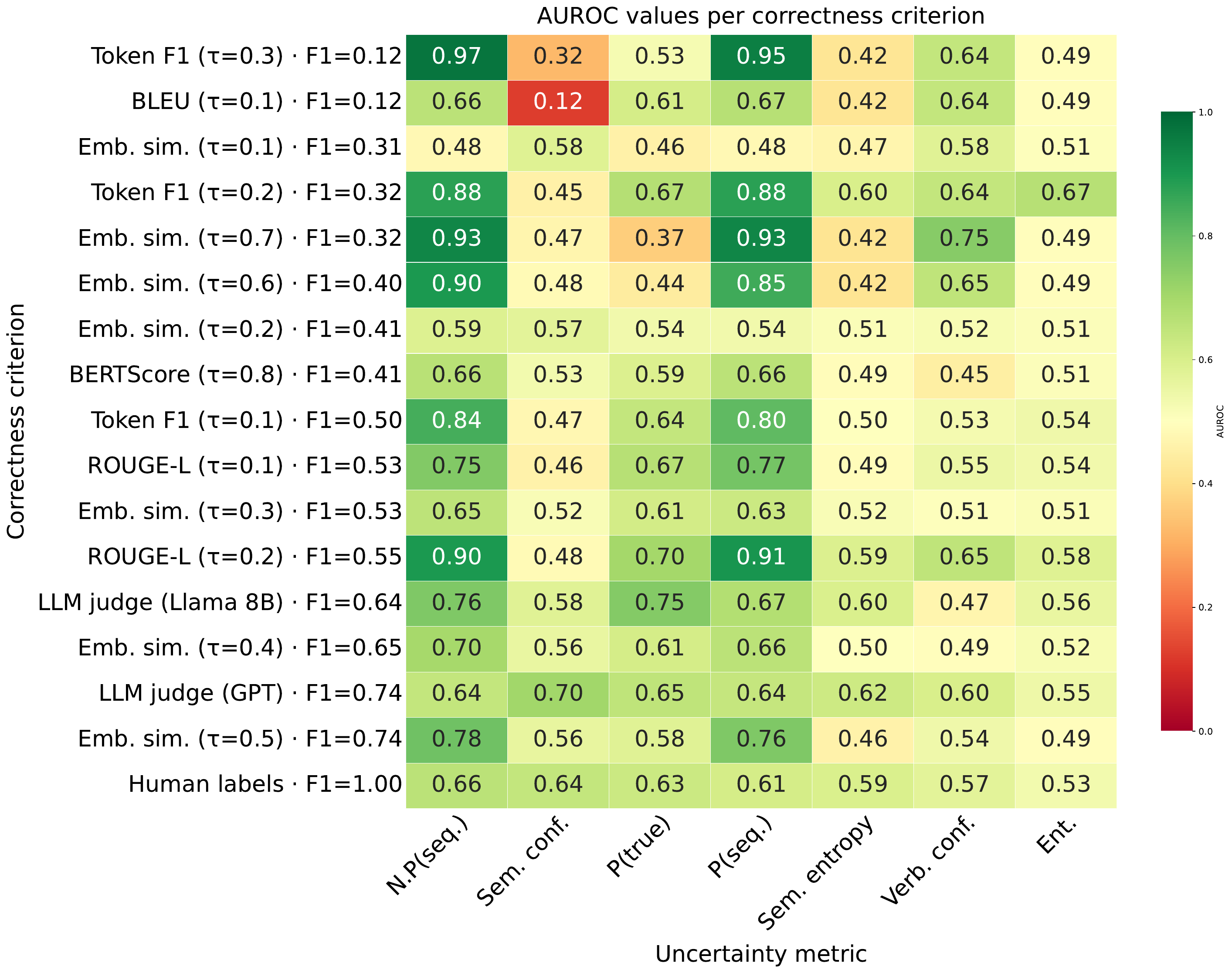}
    \end{subfigure}
    \hfill
    \begin{subfigure}[t]{0.48\textwidth}
        \centering
        \includegraphics[width=\linewidth]{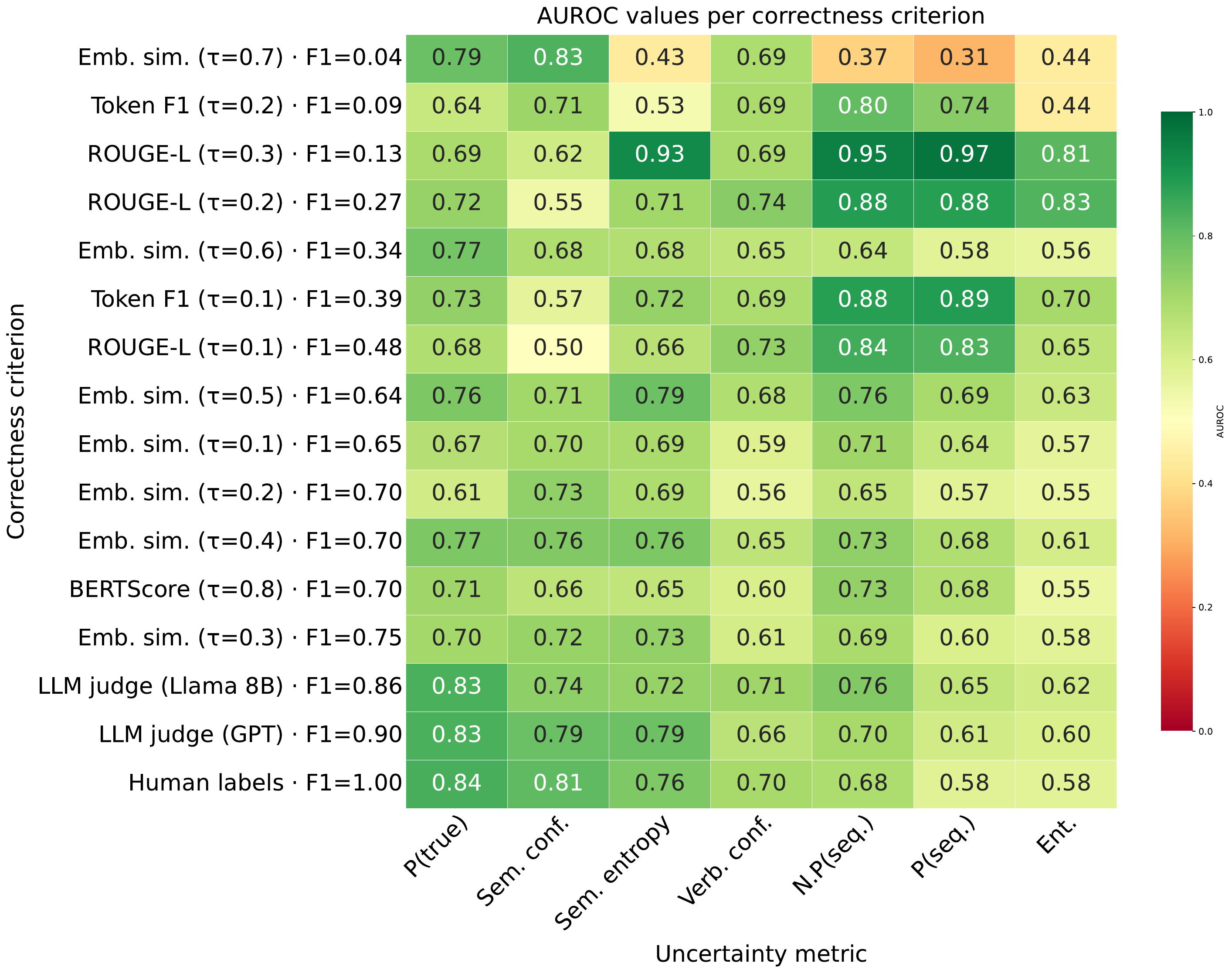}
    \end{subfigure}

    \vspace{0.75em}

    \begin{subfigure}[t]{0.48\textwidth}
        \centering
        \includegraphics[width=\linewidth]{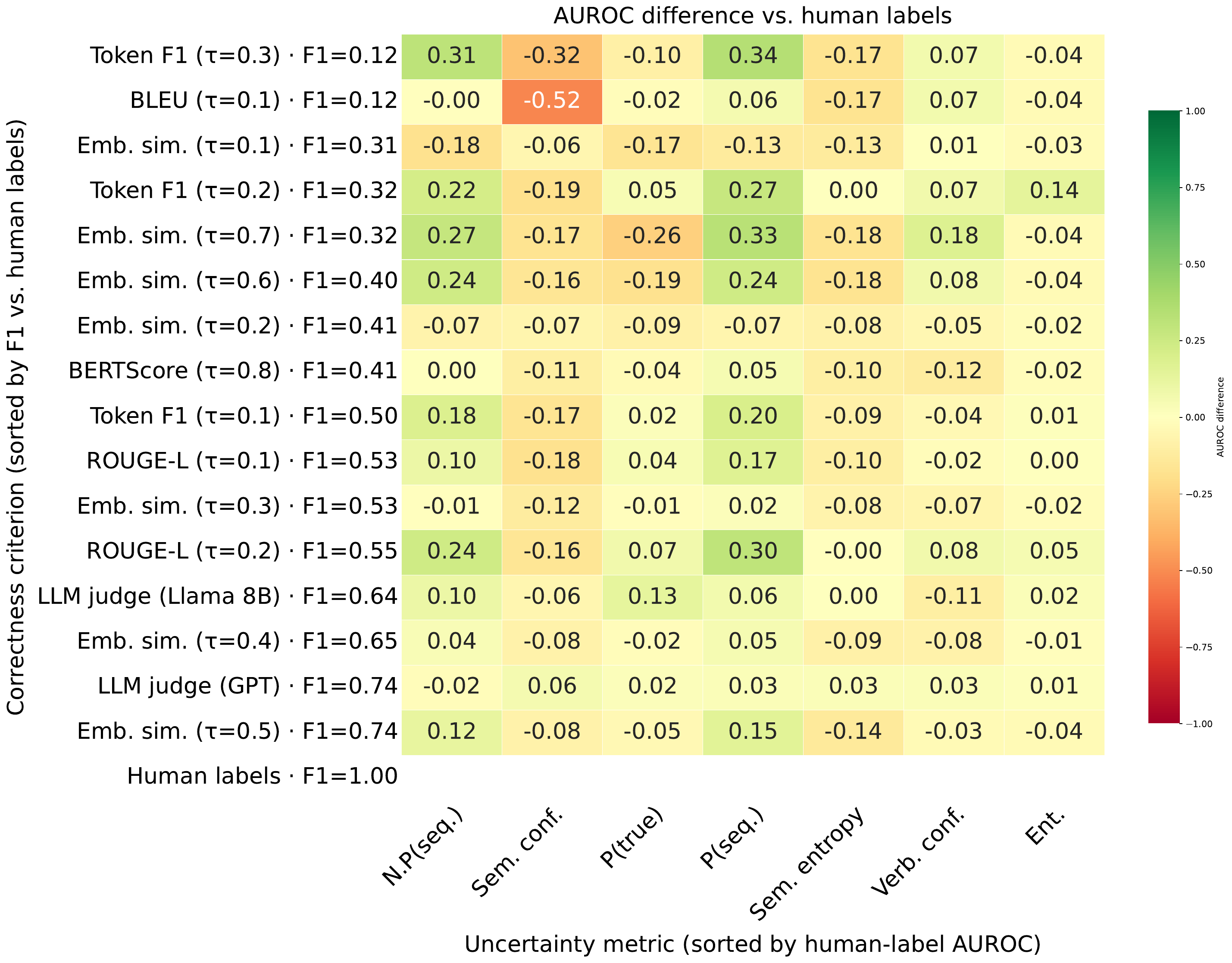}
    \end{subfigure}
    \hfill
    \begin{subfigure}[t]{0.48\textwidth}
        \centering
        \includegraphics[width=\linewidth]{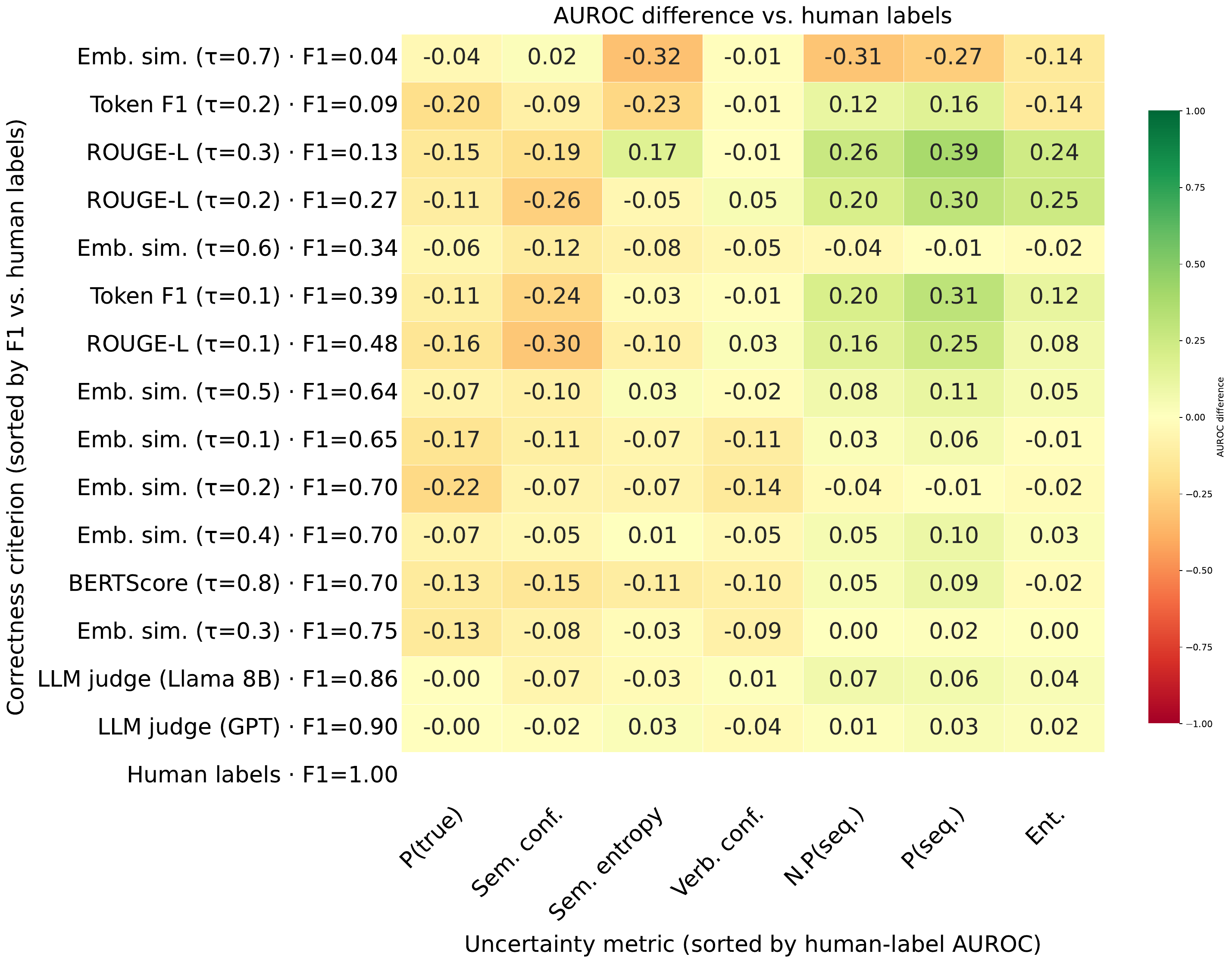}
    \end{subfigure}

    \begin{subfigure}[t]{0.48\textwidth}
        \centering
        \includegraphics[width=\linewidth]{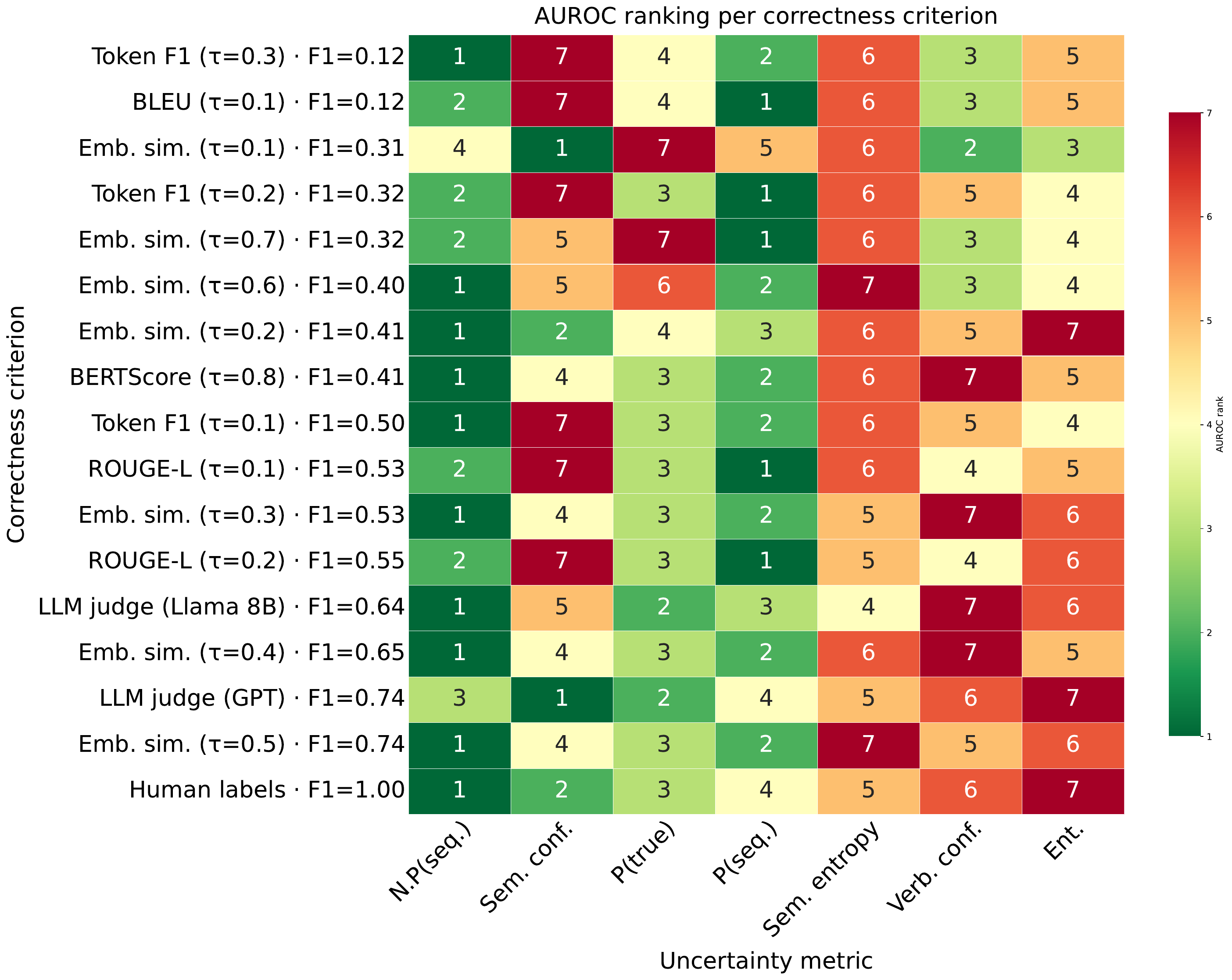}
        \caption{Qwen 0.5B, AmbigQA (unambiguous)}
    \end{subfigure}
    \hfill
    \begin{subfigure}[t]{0.48\textwidth}
        \centering
        \includegraphics[width=\linewidth]{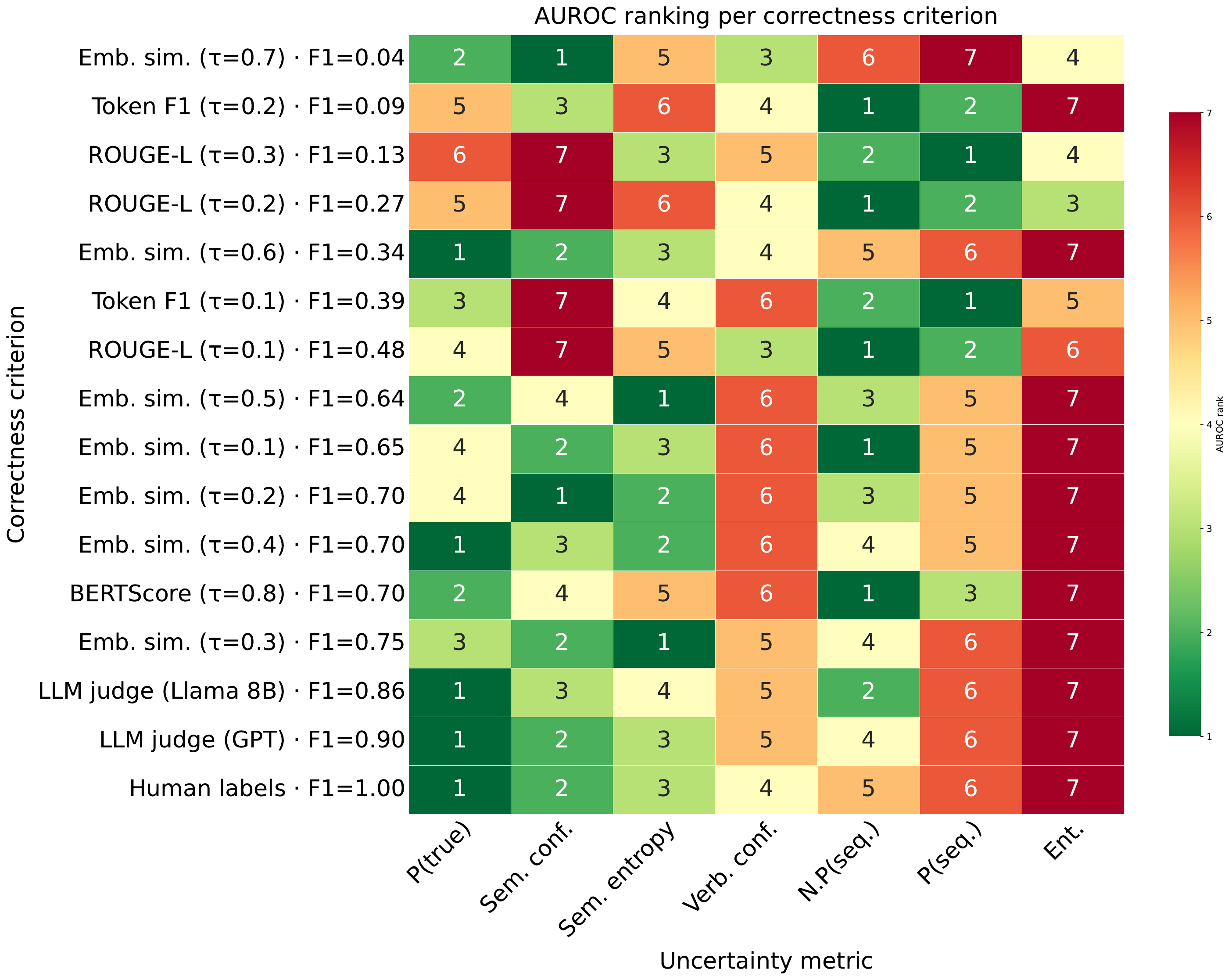}
        \caption{Qwen 7B, AmbigQA (unambiguous)}
    \end{subfigure}

    \caption{Top row shows AUROC values, middle row shows differences between oracle and system AUROC values and bottom row rankings for various uncertainty quantifiers (ordered from left to right in a descending order of performance) and correctness functions (ordered from top to bottom in an increasing order of quality). }
    \label{fig:qwen-ambigqa-sing}
\end{figure*}

\begin{figure*}[t]
    \centering

    \begin{subfigure}[t]{0.48\textwidth}
        \centering
        \includegraphics[width=\linewidth]{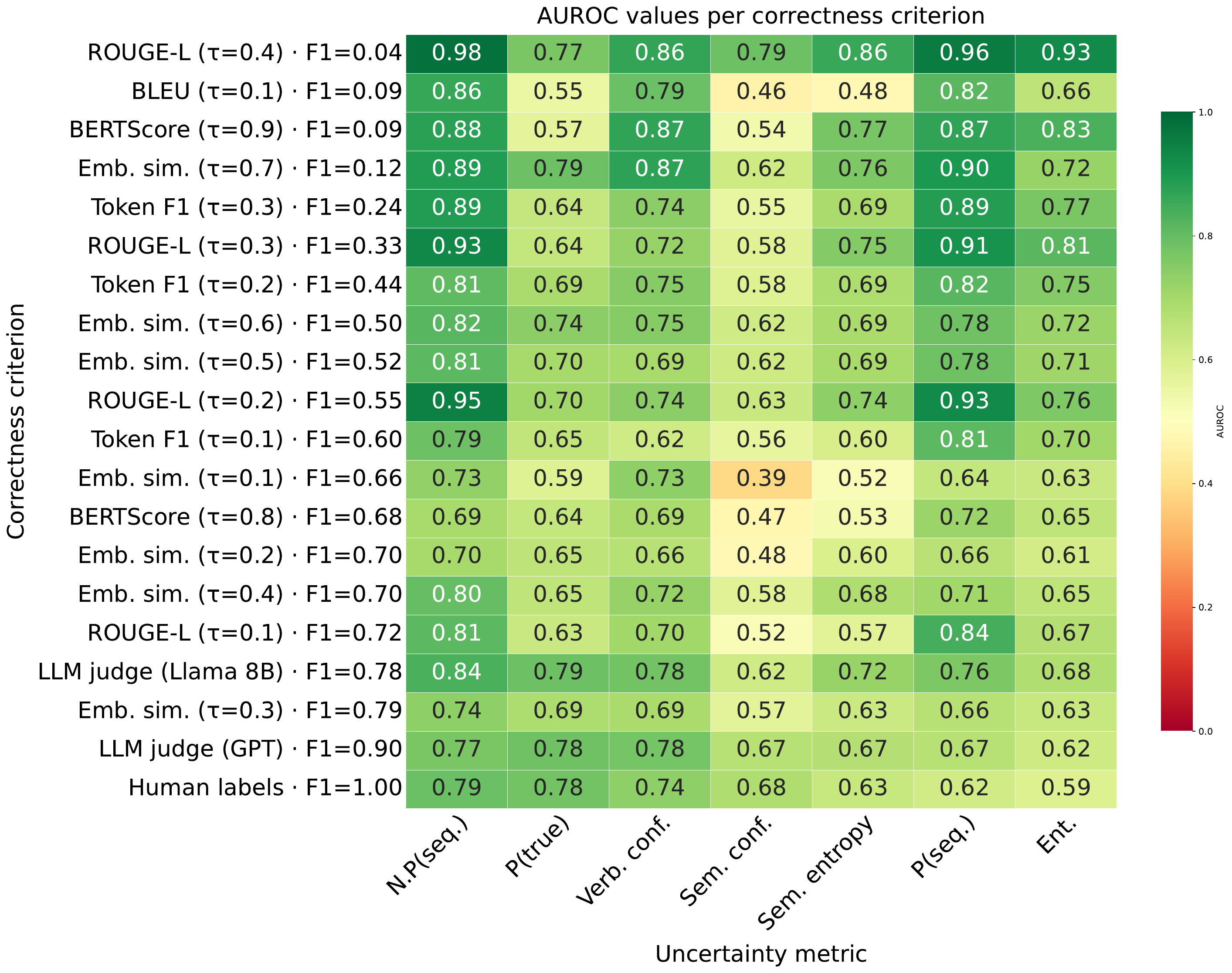}
    \end{subfigure}
    \hfill
    \begin{subfigure}[t]{0.48\textwidth}
        \centering
        \includegraphics[width=\linewidth]{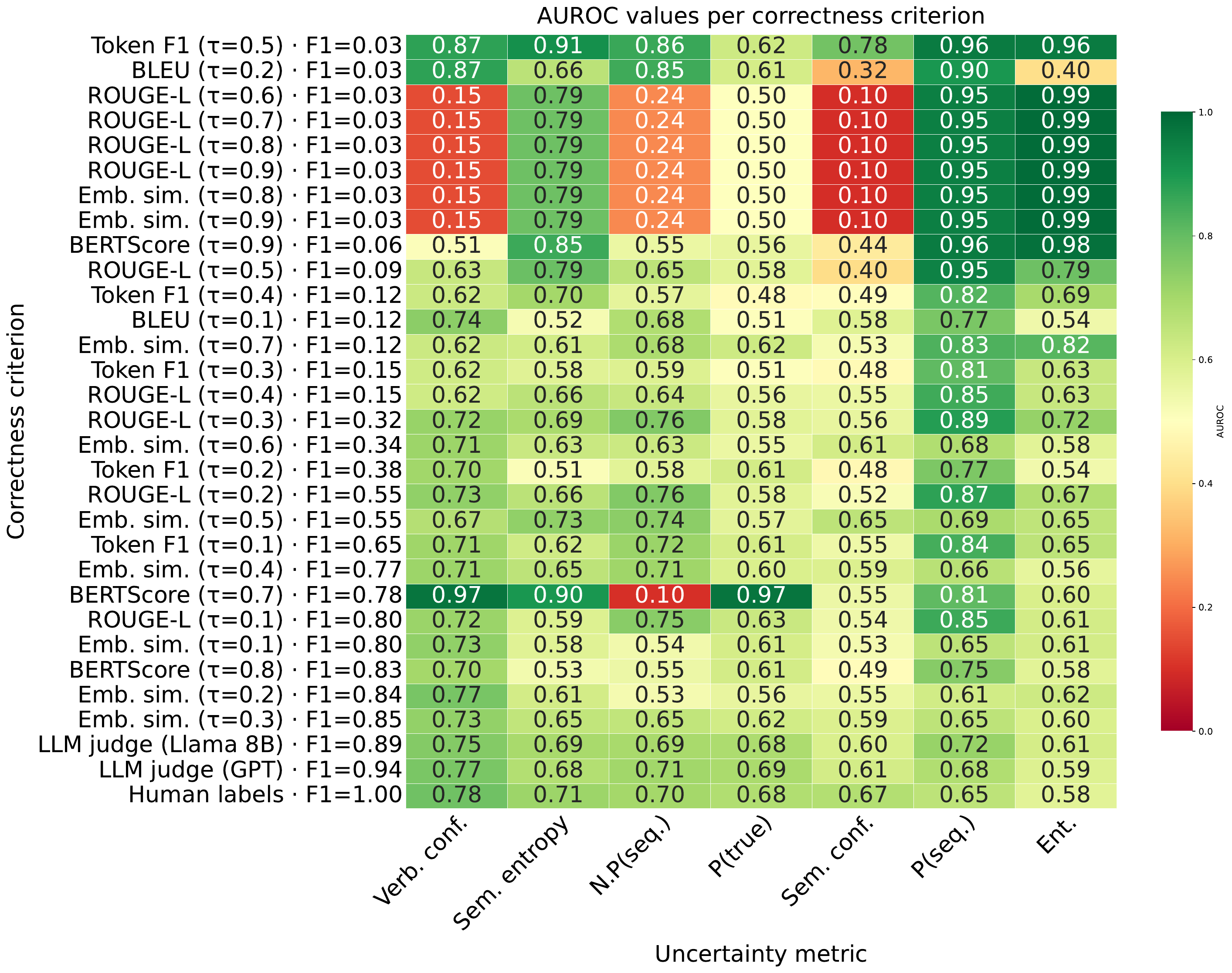}
    \end{subfigure}

    \vspace{0.75em}

    \begin{subfigure}[t]{0.48\textwidth}
        \centering
        \includegraphics[width=\linewidth]{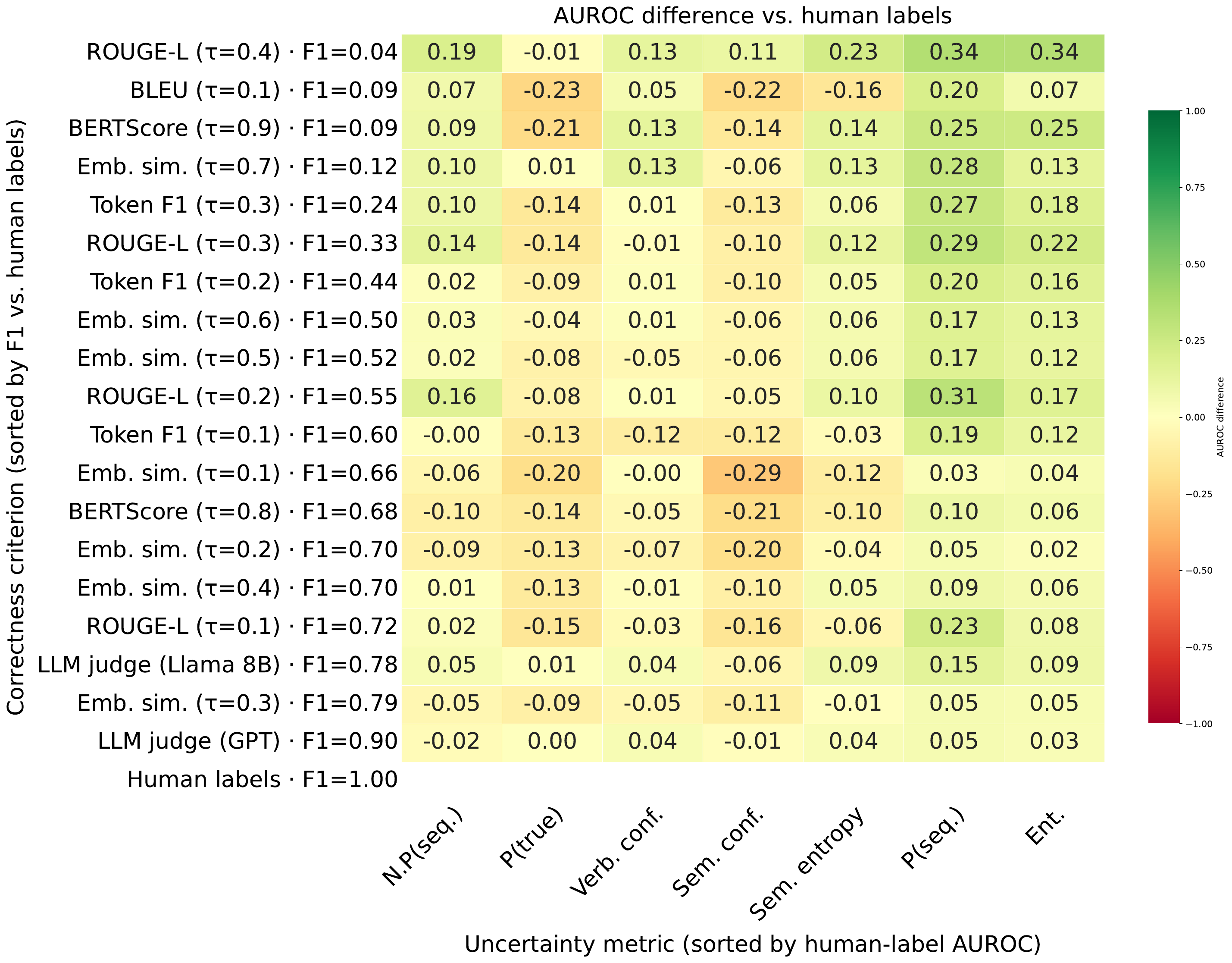}
    \end{subfigure}
    \hfill
    \begin{subfigure}[t]{0.48\textwidth}
        \centering
        \includegraphics[width=\linewidth]{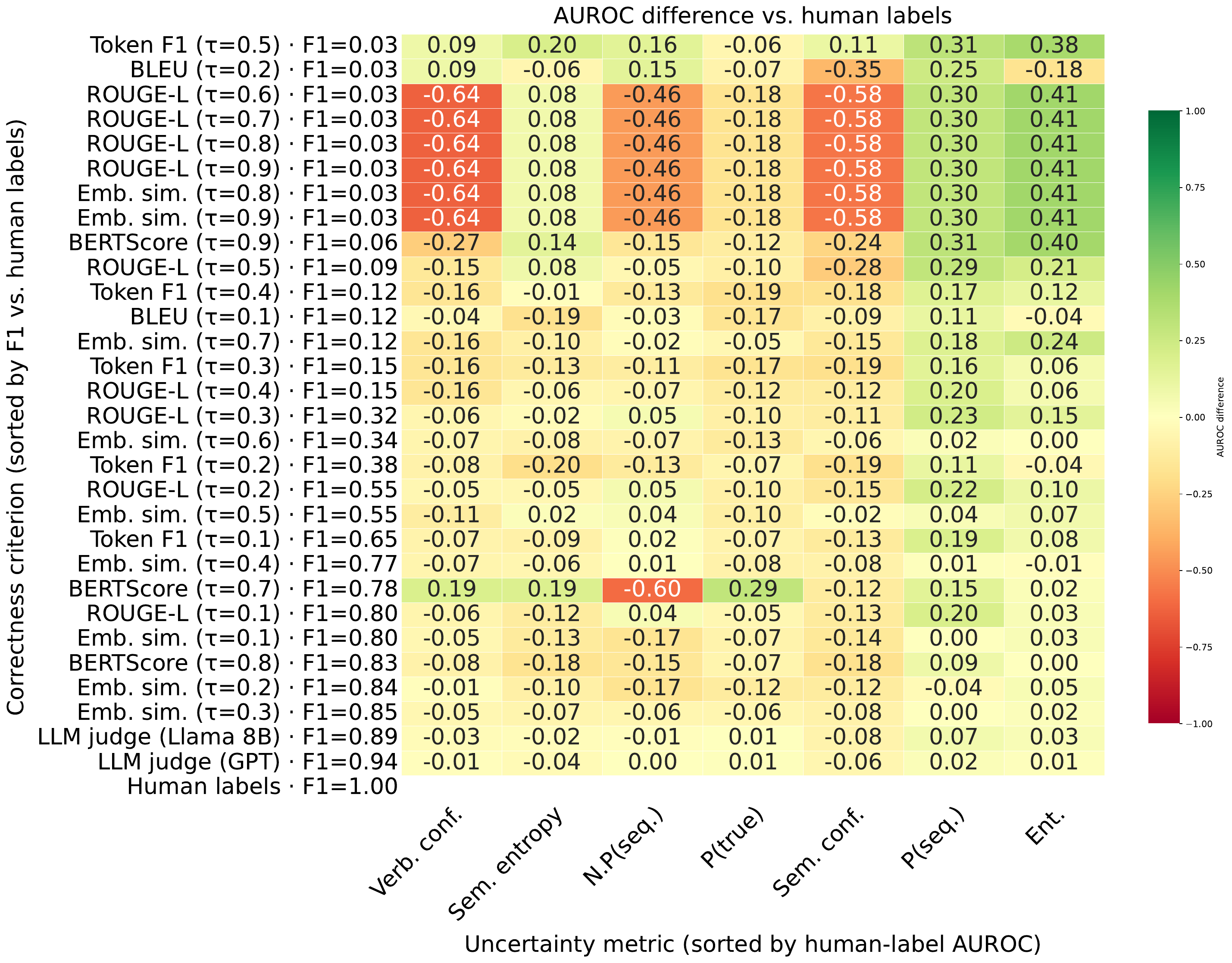}
    \end{subfigure}

    \begin{subfigure}[t]{0.48\textwidth}
        \centering
        \includegraphics[width=\linewidth]{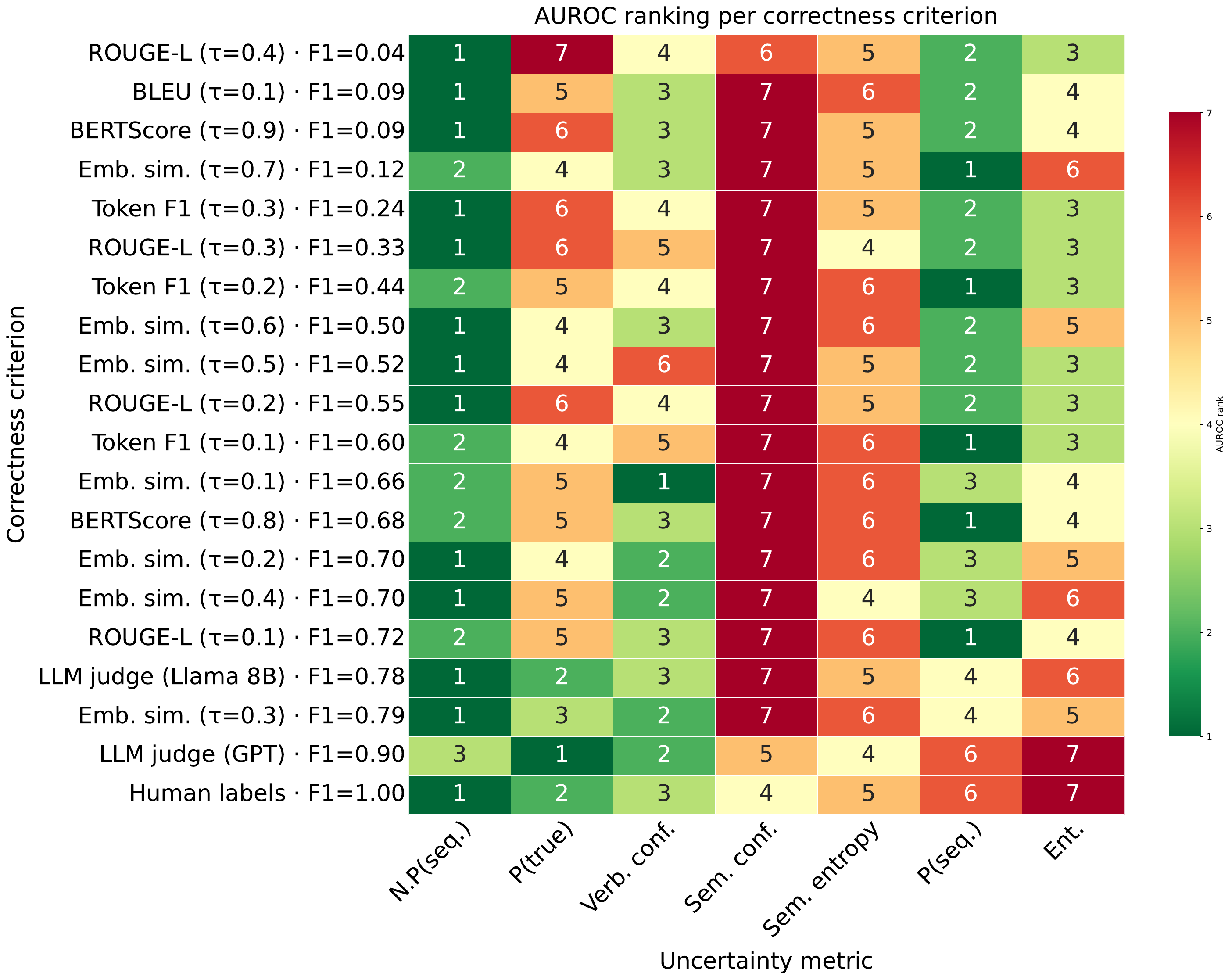}
        \caption{Llama 3B, AmbigQA (unambiguous)}
    \end{subfigure}
    \hfill
    \begin{subfigure}[t]{0.48\textwidth}
        \centering
        \includegraphics[width=\linewidth]{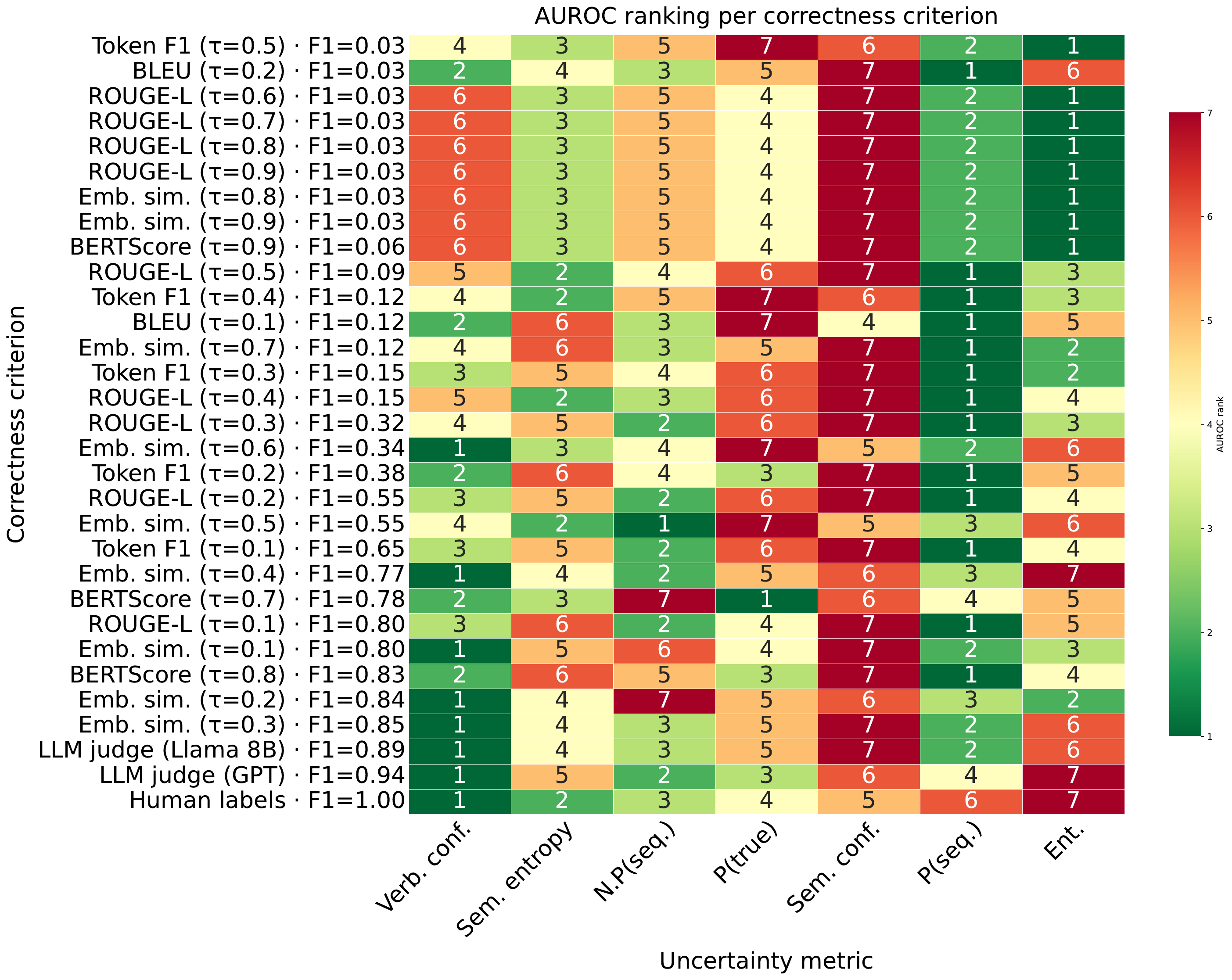}
        \caption{Llama 8B, AmbigQA (unambiguous)}
    \end{subfigure}

    \caption{Top row shows AUROC values, middle row shows differences between oracle and system AUROC values and bottom row rankings for various uncertainty quantifiers (ordered from left to right in a descending order of performance) and correctness functions (ordered from top to bottom in an increasing order of quality). }
    \label{fig:llama-ambigqa-sing}
\end{figure*}

\begin{figure*}[t]
    \centering

    \begin{subfigure}[t]{0.48\textwidth}
        \centering
        \includegraphics[width=\linewidth]{figs/new/auroc_values_per_correctness_ambig_qa_single_answer_qwen-0.5b-instruct_100_0.pdf}
    \end{subfigure}
    \hfill
    \begin{subfigure}[t]{0.48\textwidth}
        \centering
        \includegraphics[width=\linewidth]{figs/new/auroc_values_per_correctness_ambig_qa_single_answer_qwen-7b-instruct_100_0.pdf}
    \end{subfigure}

    \vspace{0.75em}

    \begin{subfigure}[t]{0.48\textwidth}
        \centering
        \includegraphics[width=\linewidth]{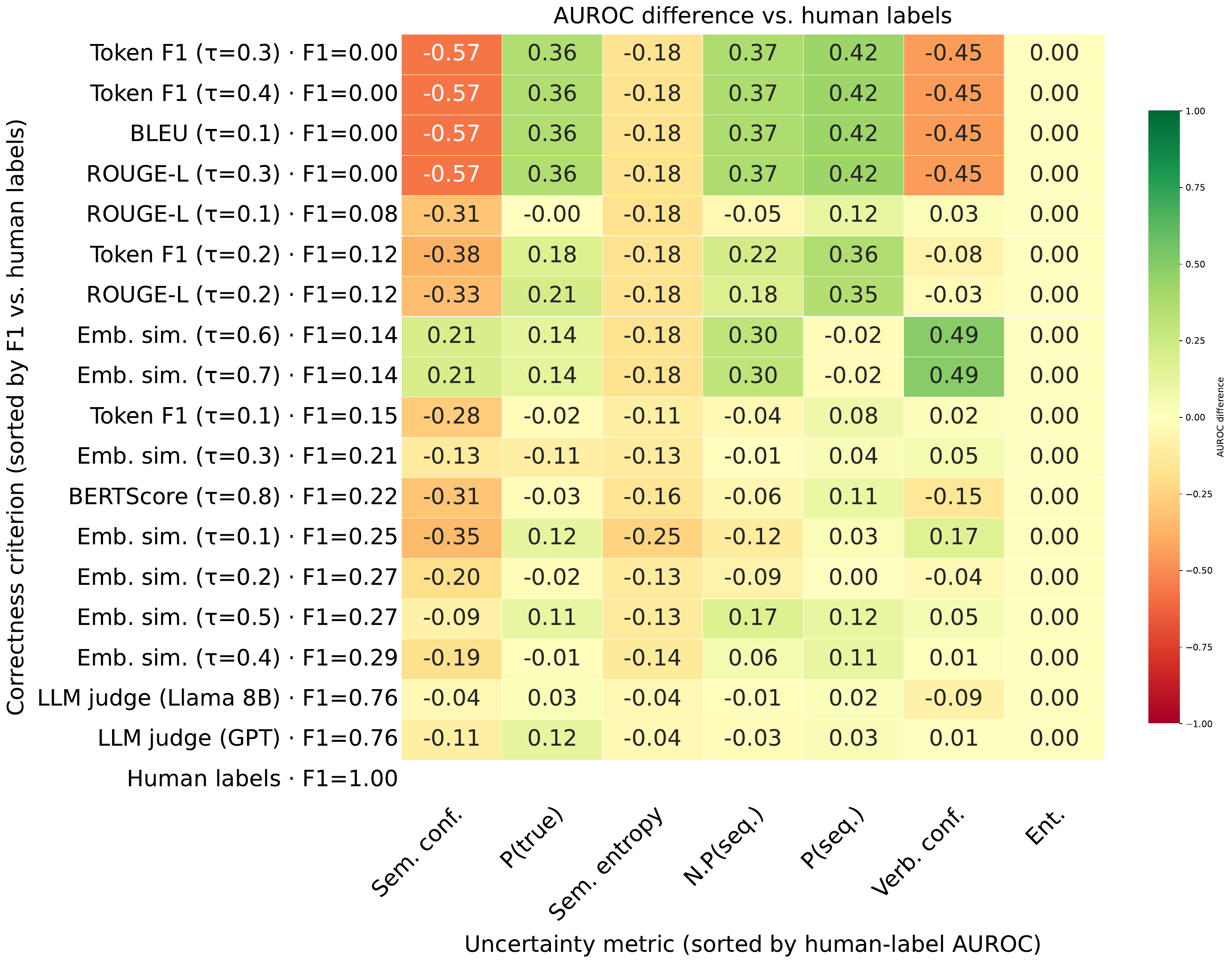}
    \end{subfigure}
    \hfill
    \begin{subfigure}[t]{0.48\textwidth}
        \centering
        \includegraphics[width=\linewidth]{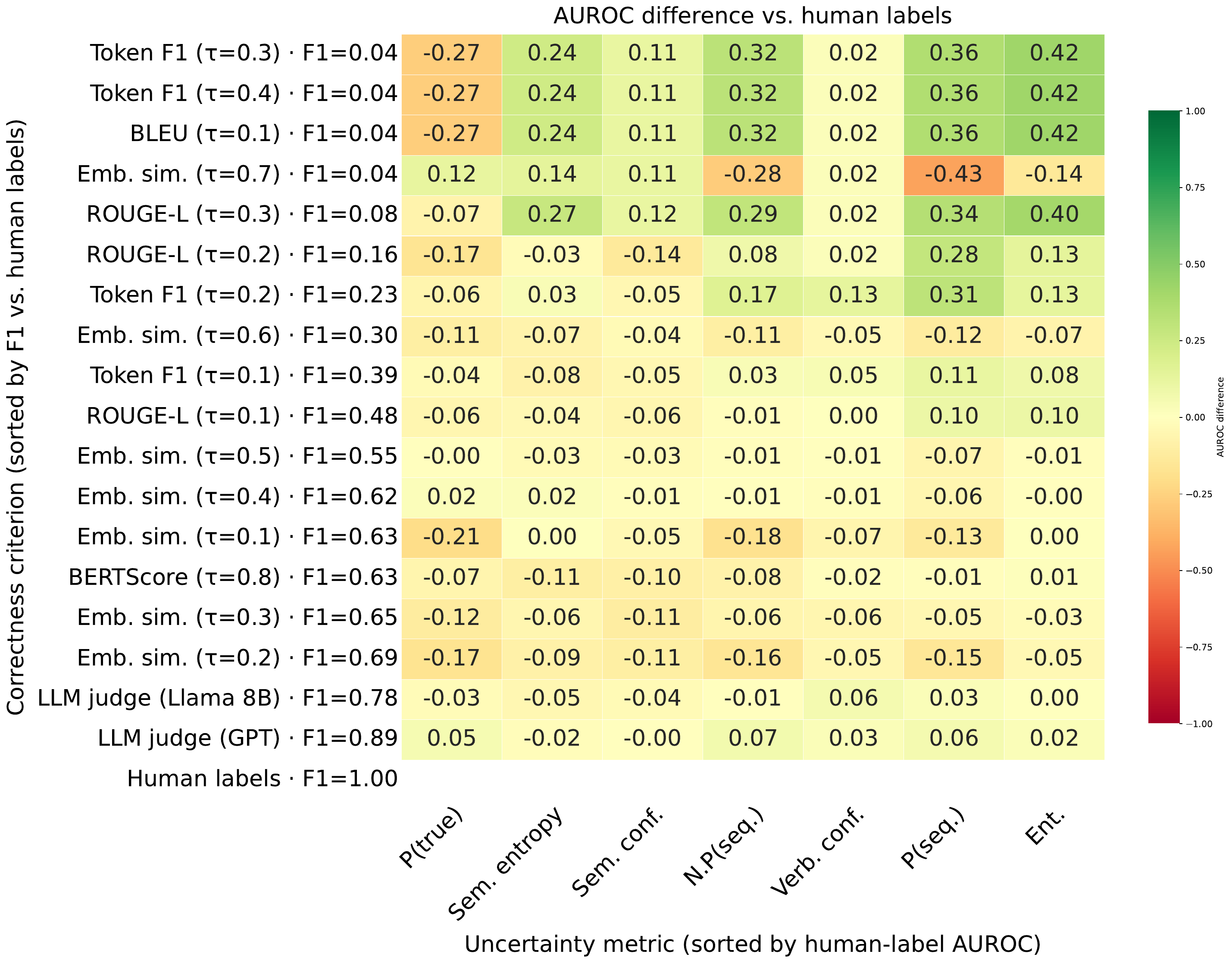}
    \end{subfigure}

    \begin{subfigure}[t]{0.48\textwidth}
        \centering
        \includegraphics[width=\linewidth]{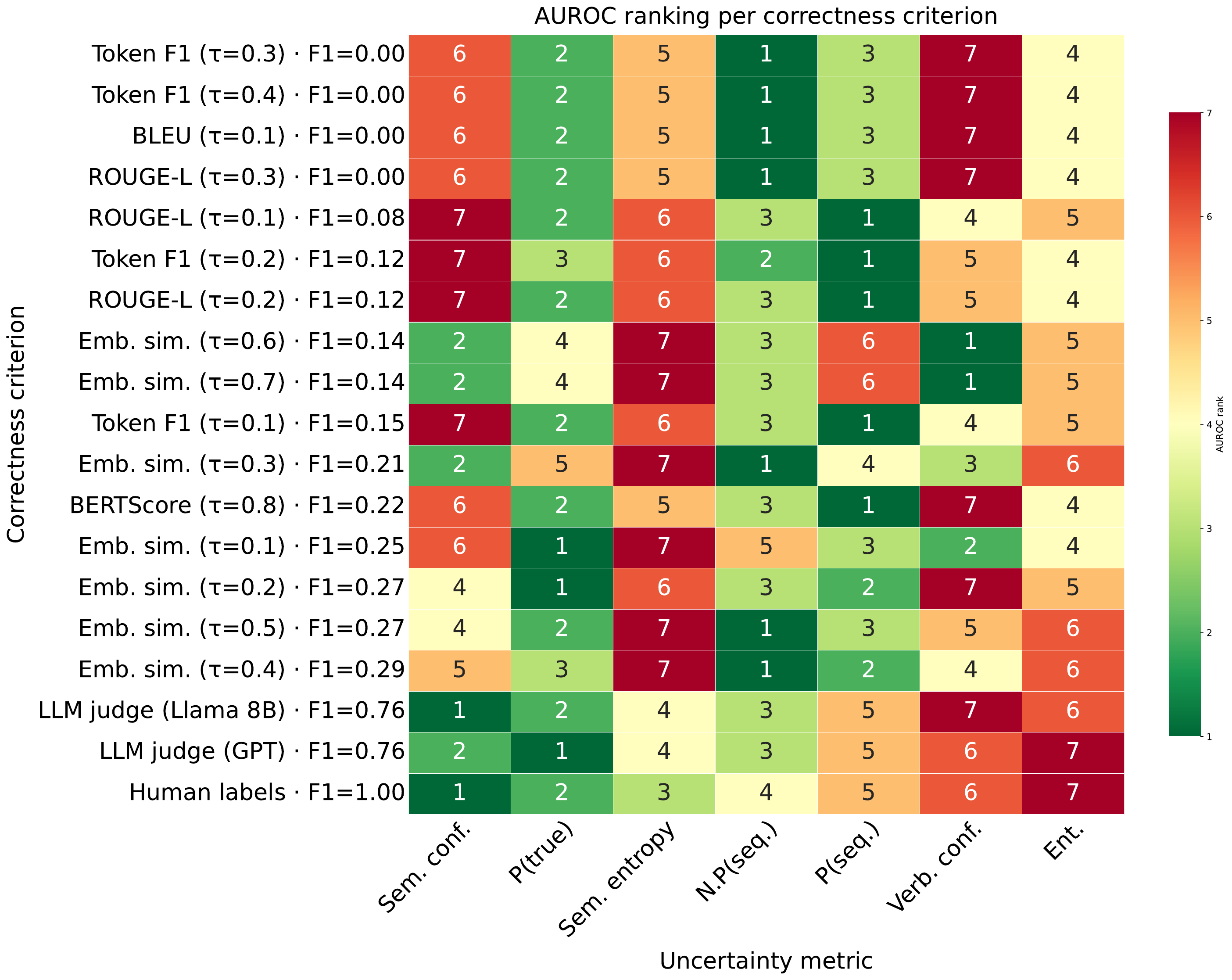}
        \caption{Qwen 0.5B, AmbigQA (ambiguous)}
    \end{subfigure}
    \hfill
    \begin{subfigure}[t]{0.48\textwidth}
        \centering
        \includegraphics[width=\linewidth]{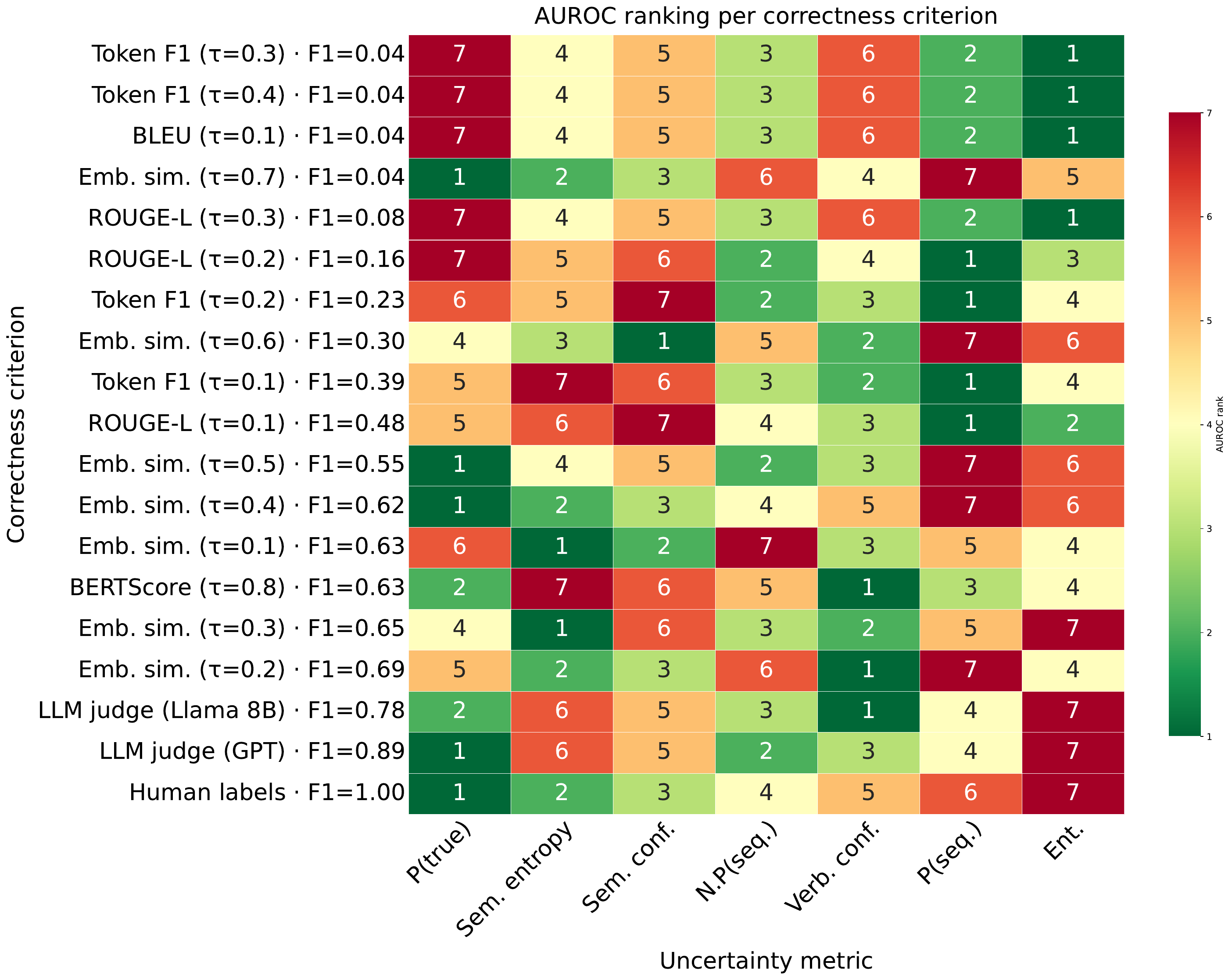}
        \caption{Qwen 7B, AmbigQA (ambiguous)}
    \end{subfigure}

    \caption{Top row shows AUROC values, middle row shows differences between oracle and system AUROC values and bottom row rankings for various uncertainty quantifiers (ordered from left to right in a descending order of performance) and correctness functions (ordered from top to bottom in an increasing order of quality). }
    \label{fig:qwen-ambigqa-mult}
\end{figure*}

\begin{figure*}[t]
    \centering

    \begin{subfigure}[t]{0.48\textwidth}
        \centering
        \includegraphics[width=\linewidth]{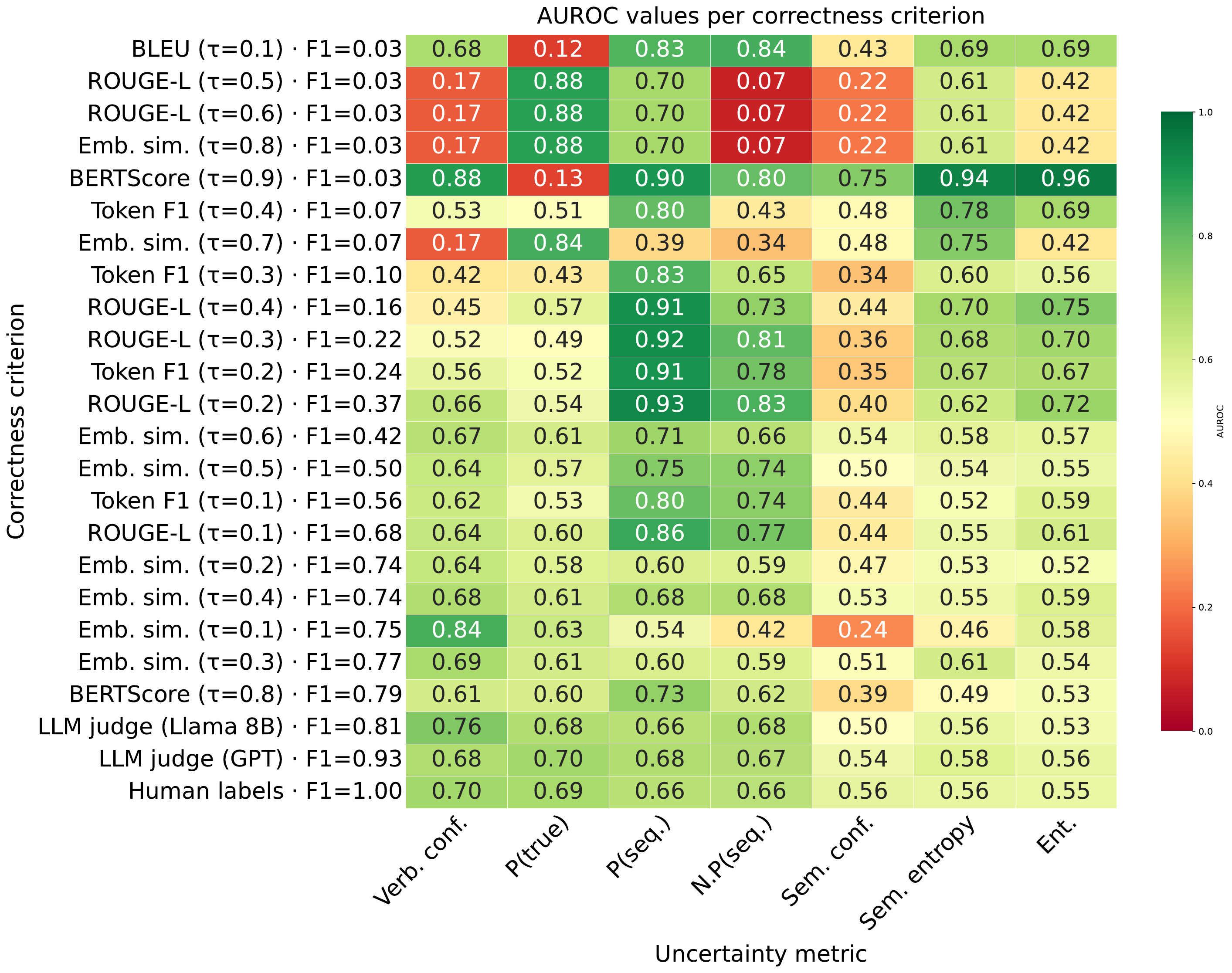}
    \end{subfigure}
    \hfill
    \begin{subfigure}[t]{0.48\textwidth}
        \centering
        \includegraphics[width=\linewidth]{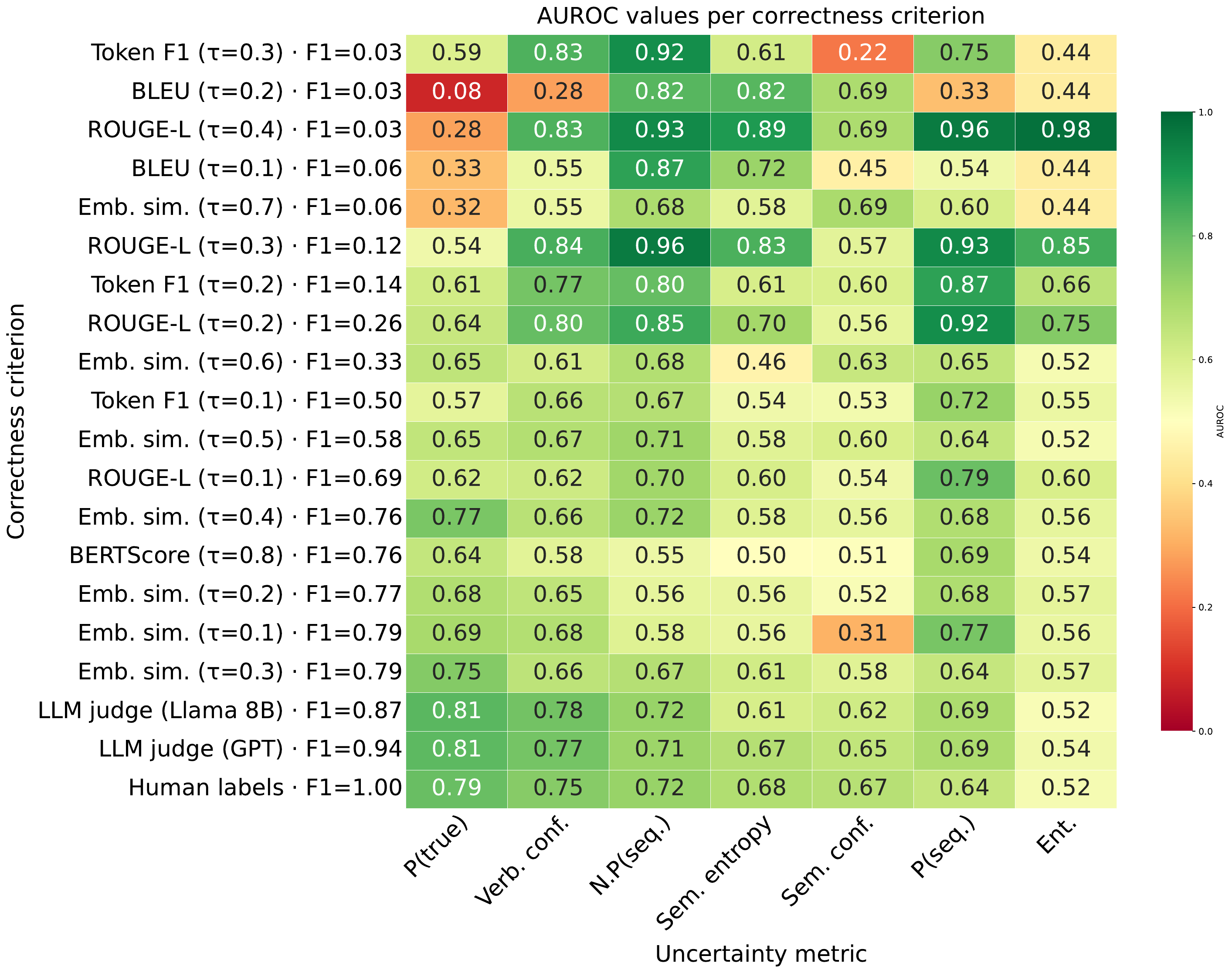}
    \end{subfigure}

    \vspace{0.75em}

    \begin{subfigure}[t]{0.48\textwidth}
        \centering
        \includegraphics[width=\linewidth]{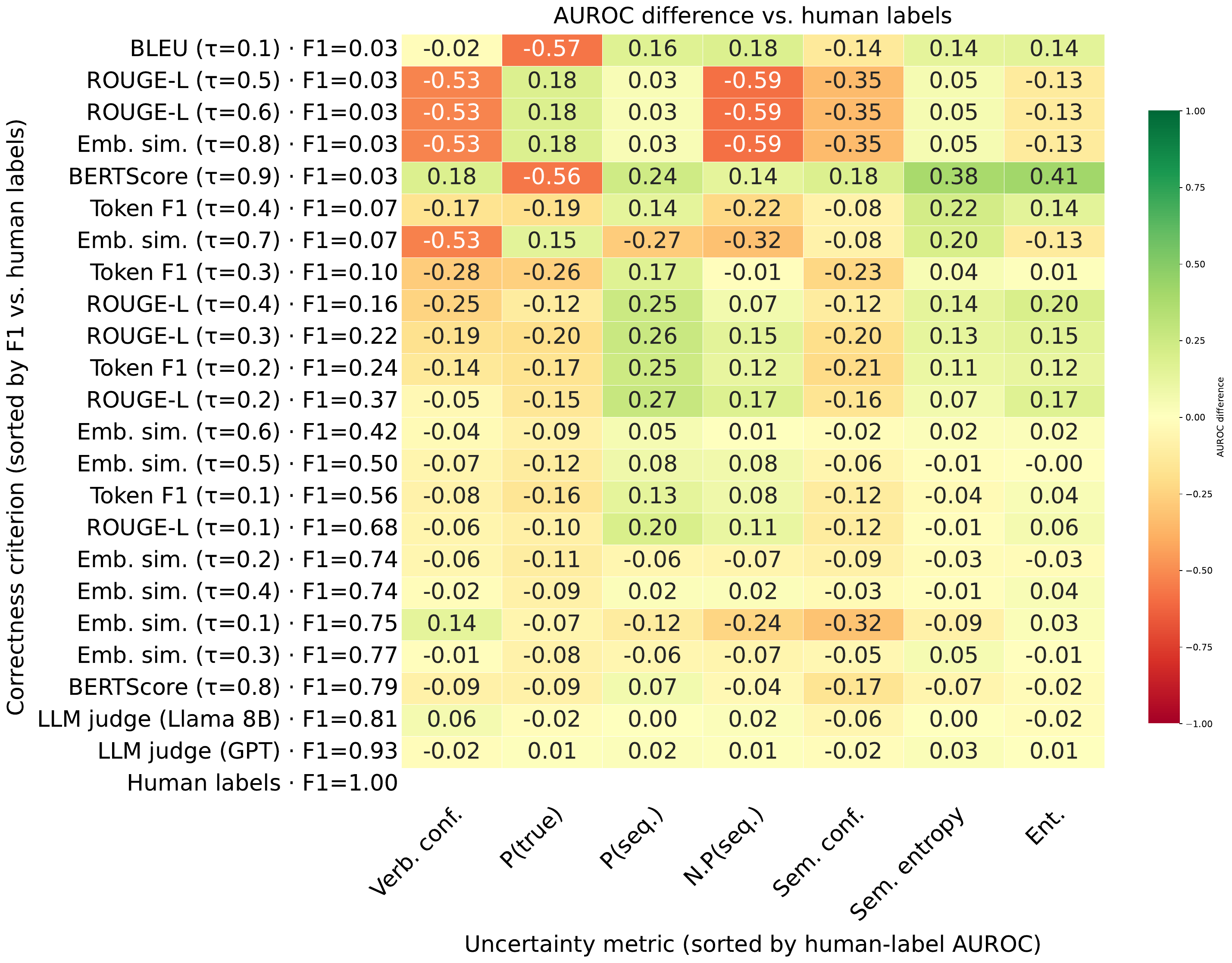}
    \end{subfigure}
    \hfill
    \begin{subfigure}[t]{0.48\textwidth}
        \centering
        \includegraphics[width=\linewidth]{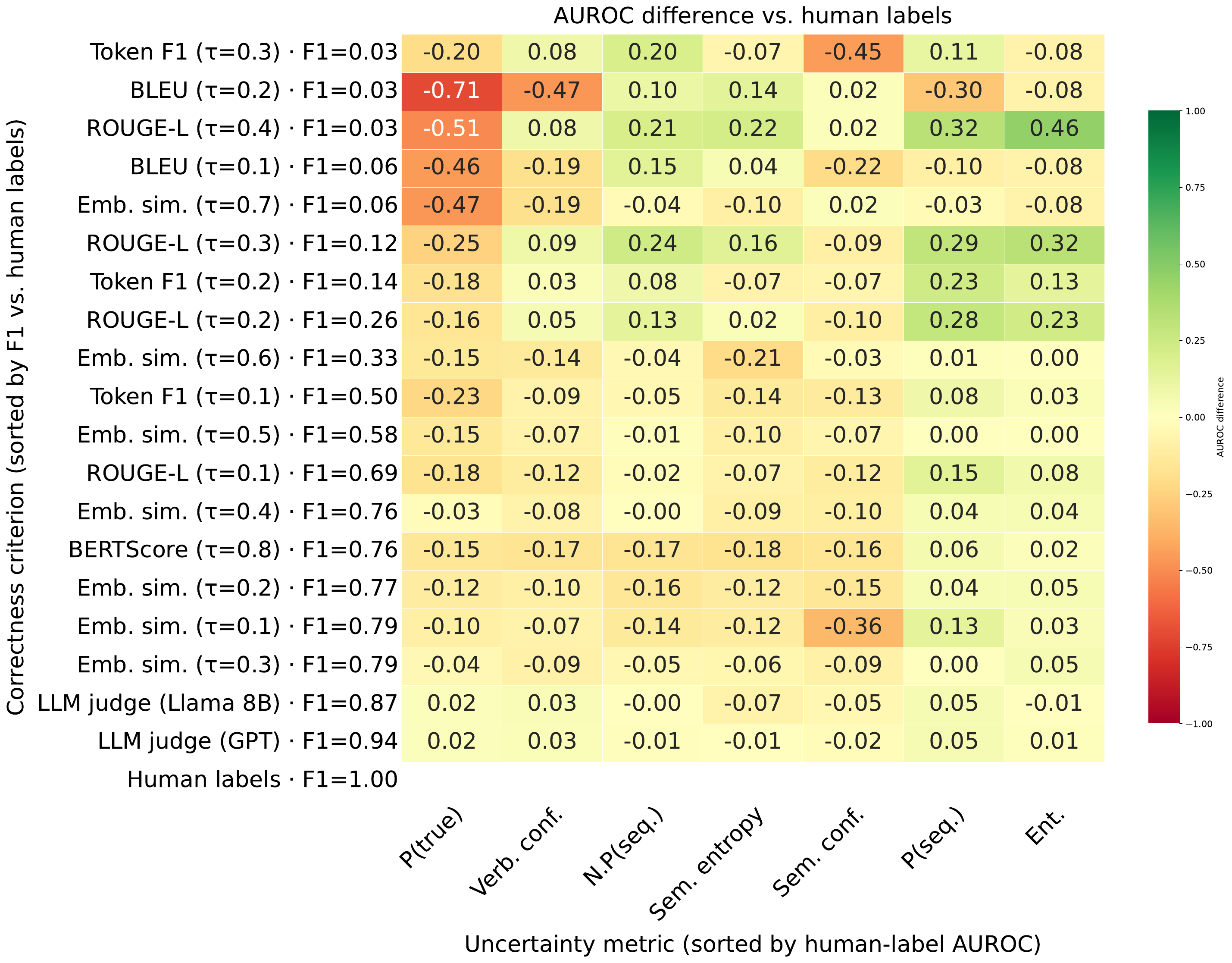}
    \end{subfigure}

    \begin{subfigure}[t]{0.48\textwidth}
        \centering
        \includegraphics[width=\linewidth]{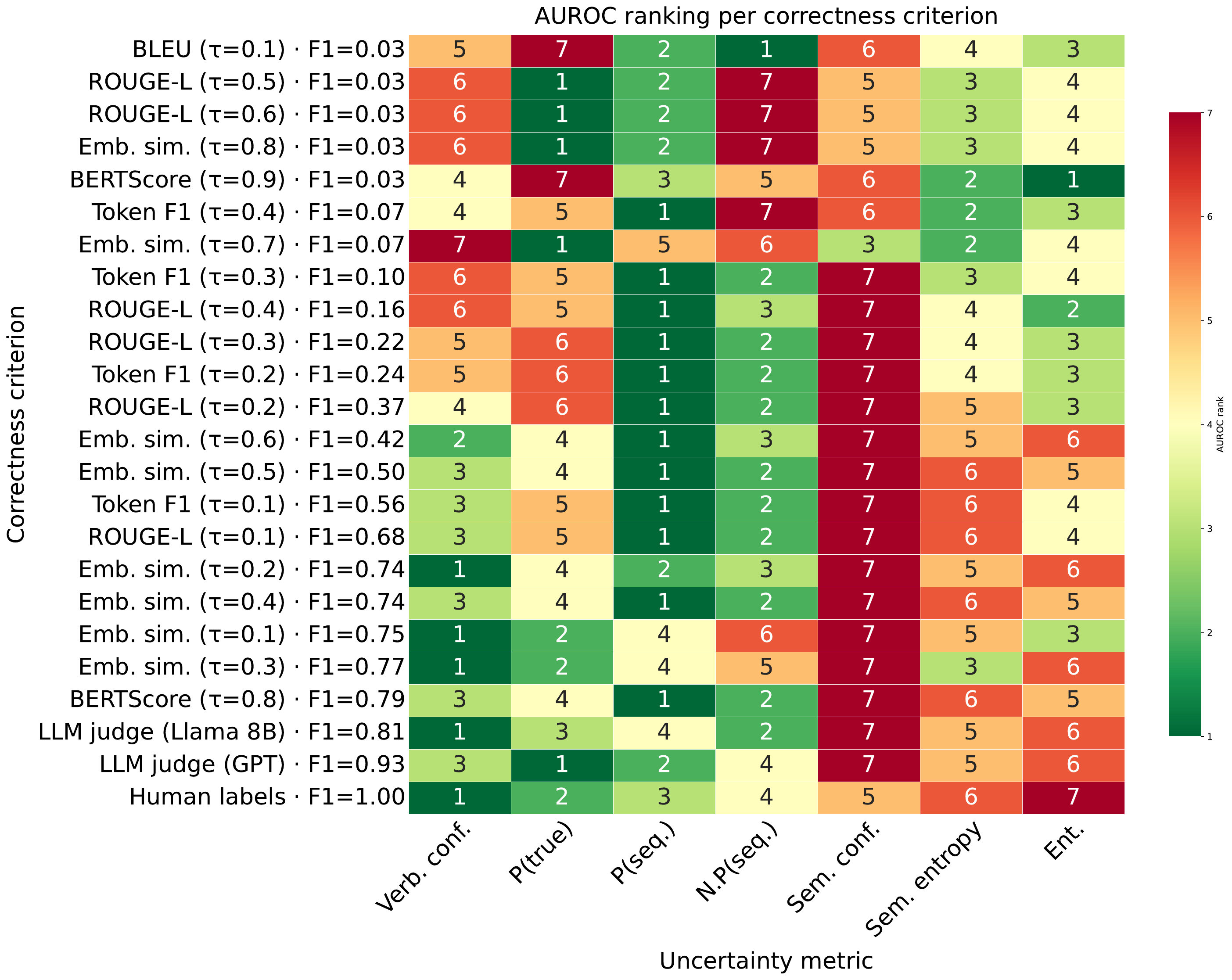}
        \caption{Llama 3B, AmbigQA (ambiguous)}
    \end{subfigure}
    \hfill
    \begin{subfigure}[t]{0.48\textwidth}
        \centering
        \includegraphics[width=\linewidth]{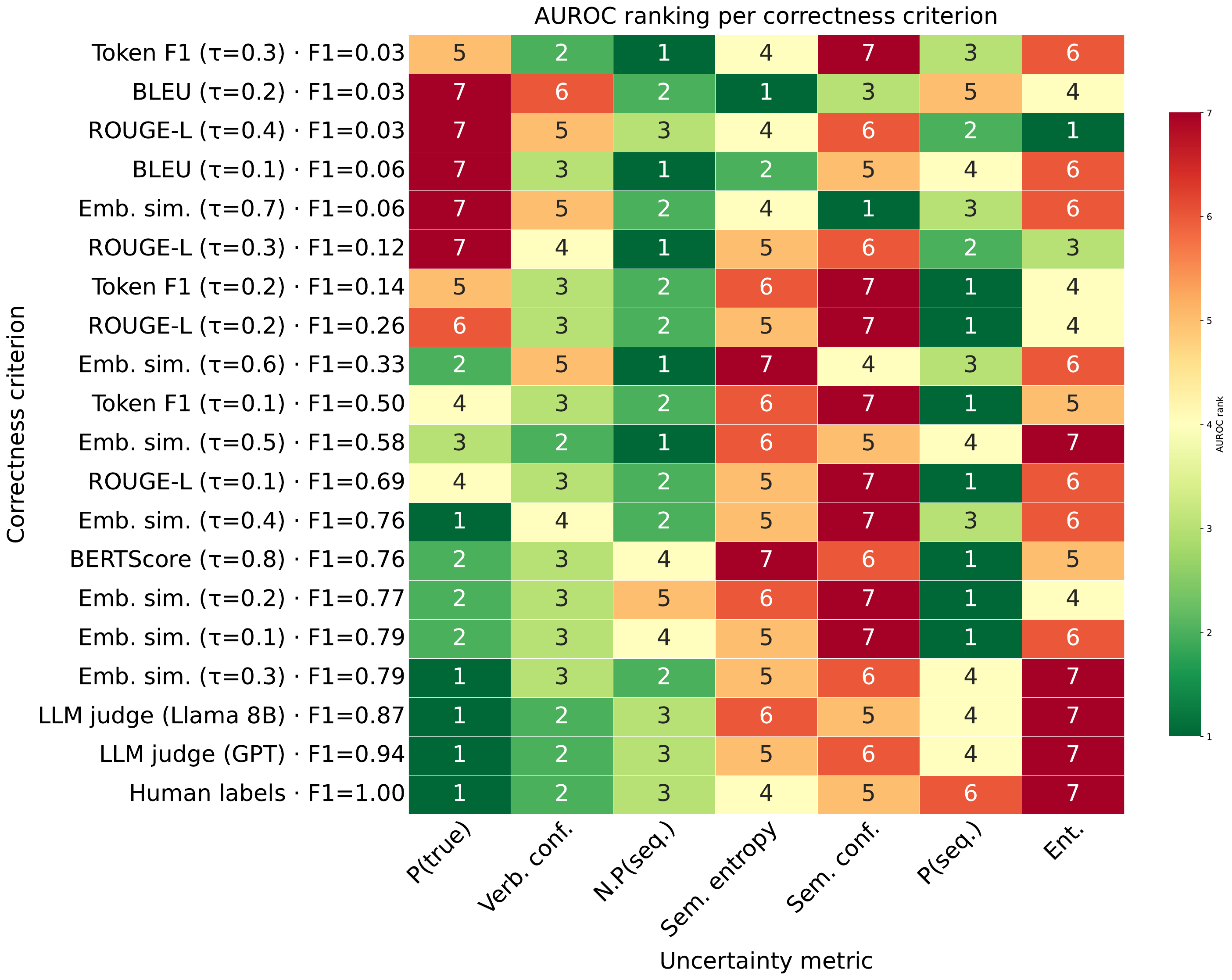}
        \caption{Llama 8B, AmbigQA (ambiguous)}
    \end{subfigure}

    \caption{Top row shows AUROC values, middle row shows differences between oracle and system AUROC values and bottom row rankings for various uncertainty quantifiers (ordered from left to right in a descending order of performance) and correctness functions (ordered from top to bottom in an increasing order of quality). }
    \label{fig:llama-ambigqa-mult}
\end{figure*}

For transparency, we present the complete results for all dataset and generator pairs in Figures \ref{fig:qwen-triviaqa} - \ref{fig:llama-ambigqa-mult}.

\subsection{Inter-Annotator agreement analysis}
\label{sec:appendix:annotators}

As manual annotation is expensive, we sampled 100 prompts for each dataset-generator pair. These were annotated by an annotator (one of the authors). We use the following guidelines:

\begin{enumerate}
    \item Identify whether the response is correct or incorrect, given the set of references. 
    \item To deem a response as correct, it should contain at least one of the reference answers. 
    \item In the absence of a definitive answer under any circumstances --- e.g. the model refuses to answer, claims not to know the answer etc. --- mark this as incorrect. 
    \item If the answer is nonsensical (e.g. repetition of identical strings, non grammatical text etc.), mark this as incorrect.
    \item If the response contains the correct answer, as well as other complementary claims or facts, you do not have to verify their correctness in order to mark the response as correct --- except if the accompanying claims are irrational and nonsensical to the extent that the correctness of the entire response has to be deemed incorrect as per your judgement. 
\end{enumerate}

\begin{figure*}[t]
    \centering

    \begin{subfigure}[t]{0.48\textwidth}
        \centering
        \includegraphics[width=\linewidth]{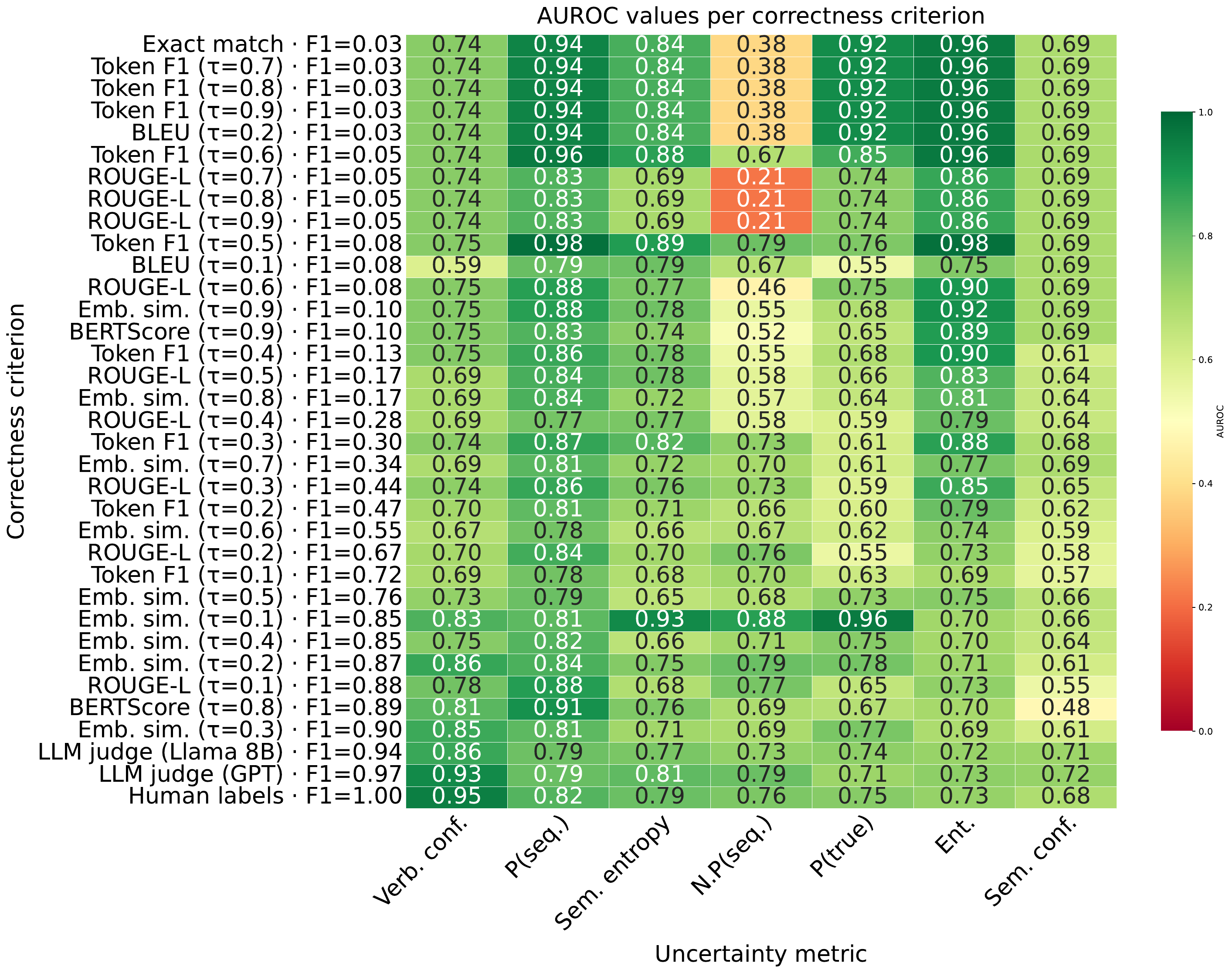}
    \end{subfigure}
    \hfill
    \begin{subfigure}[t]{0.48\textwidth}
        \centering
        \includegraphics[width=\linewidth]{figs/new/auroc_values_per_correctness_trivia_qa_llama-8b-instruct.pdf}
    \end{subfigure}

    \vspace{0.75em}

    \begin{subfigure}[t]{0.48\textwidth}
        \centering
        \includegraphics[width=\linewidth]{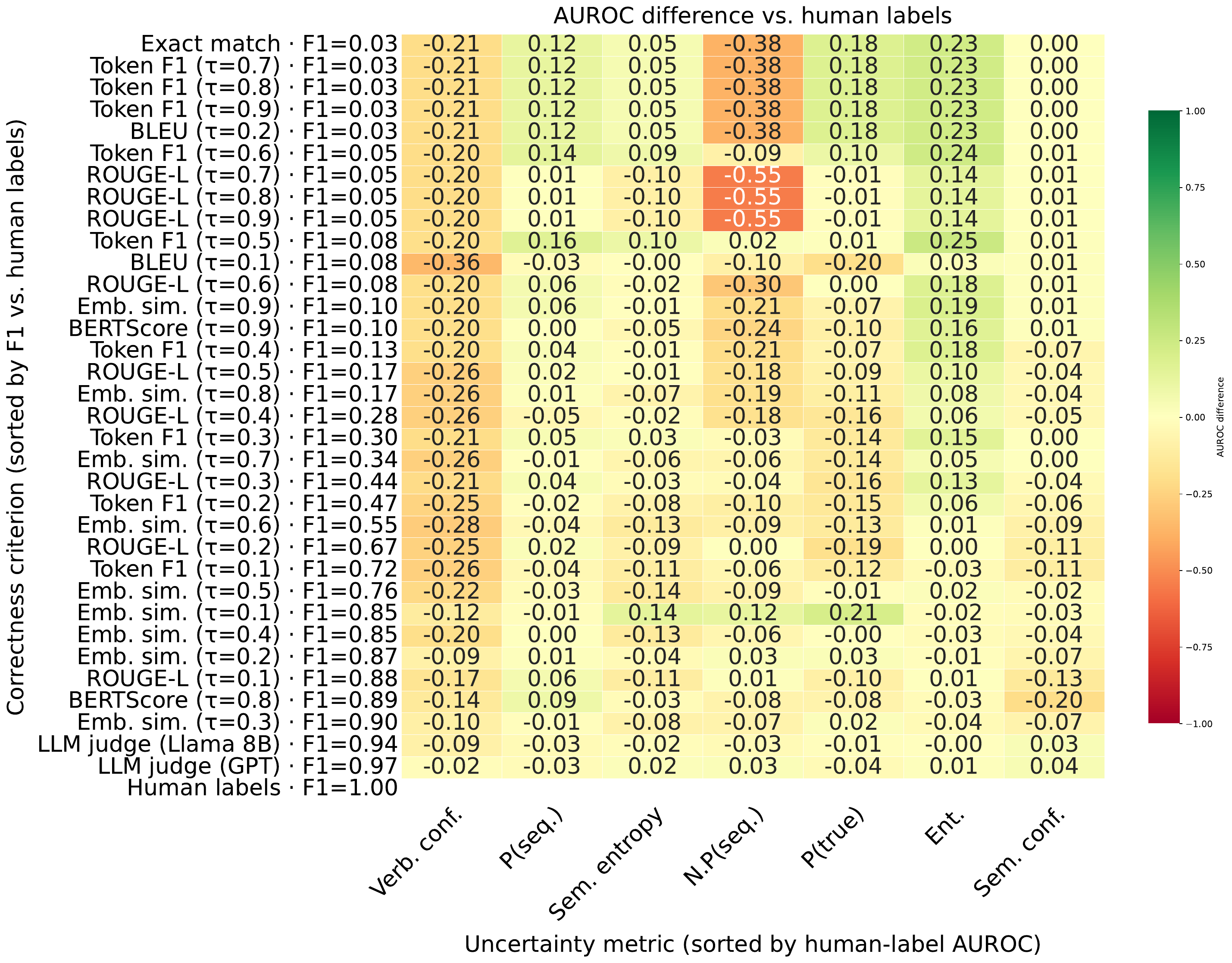}
    \end{subfigure}
    \hfill
    \begin{subfigure}[t]{0.48\textwidth}
        \centering
        \includegraphics[width=\linewidth]{figs/new/auroc_diff_per_correctness_trivia_qa_llama-8b-instruct.pdf}
    \end{subfigure}

    \begin{subfigure}[t]{0.48\textwidth}
        \centering
        \includegraphics[width=\linewidth]{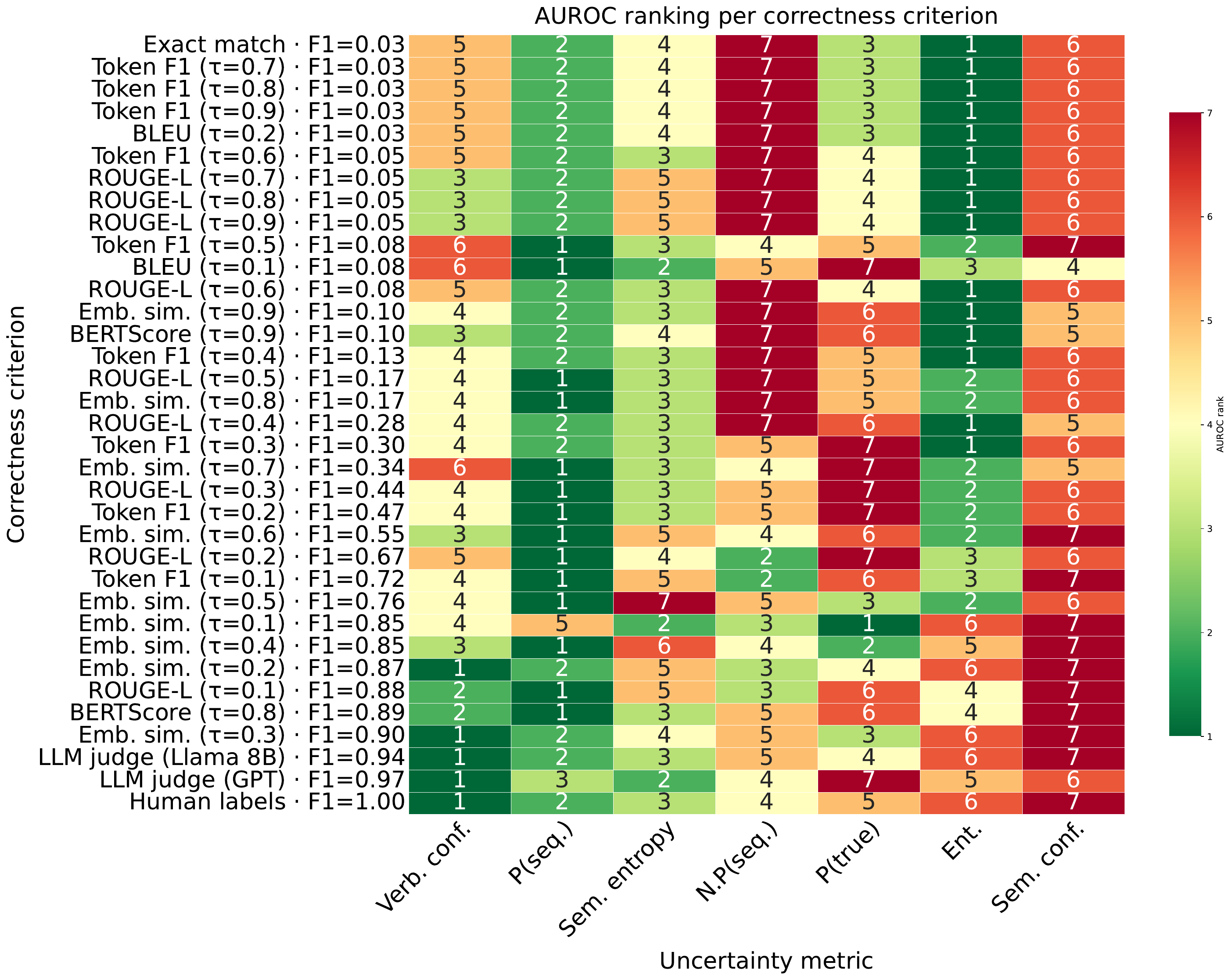}
        \caption{Llama 8B, TriviaQA, New annotator}
    \end{subfigure}
    \hfill
    \begin{subfigure}[t]{0.48\textwidth}
        \centering
        \includegraphics[width=\linewidth]{figs/new/auroc_ranking_per_correctness_trivia_qa_llama-8b-instruct.pdf}
        \caption{Llama 8B, TriviaQA, Original annotator}
    \end{subfigure}

    \caption{Left column shows results given manual labels obtained from new annotator, and right column shows results from original annotator.}
    \label{fig:llama-triviaqa-interannotator}
\end{figure*}

\begin{figure*}[t]
    \centering

    \begin{subfigure}[t]{0.48\textwidth}
        \centering
        \includegraphics[width=\linewidth]{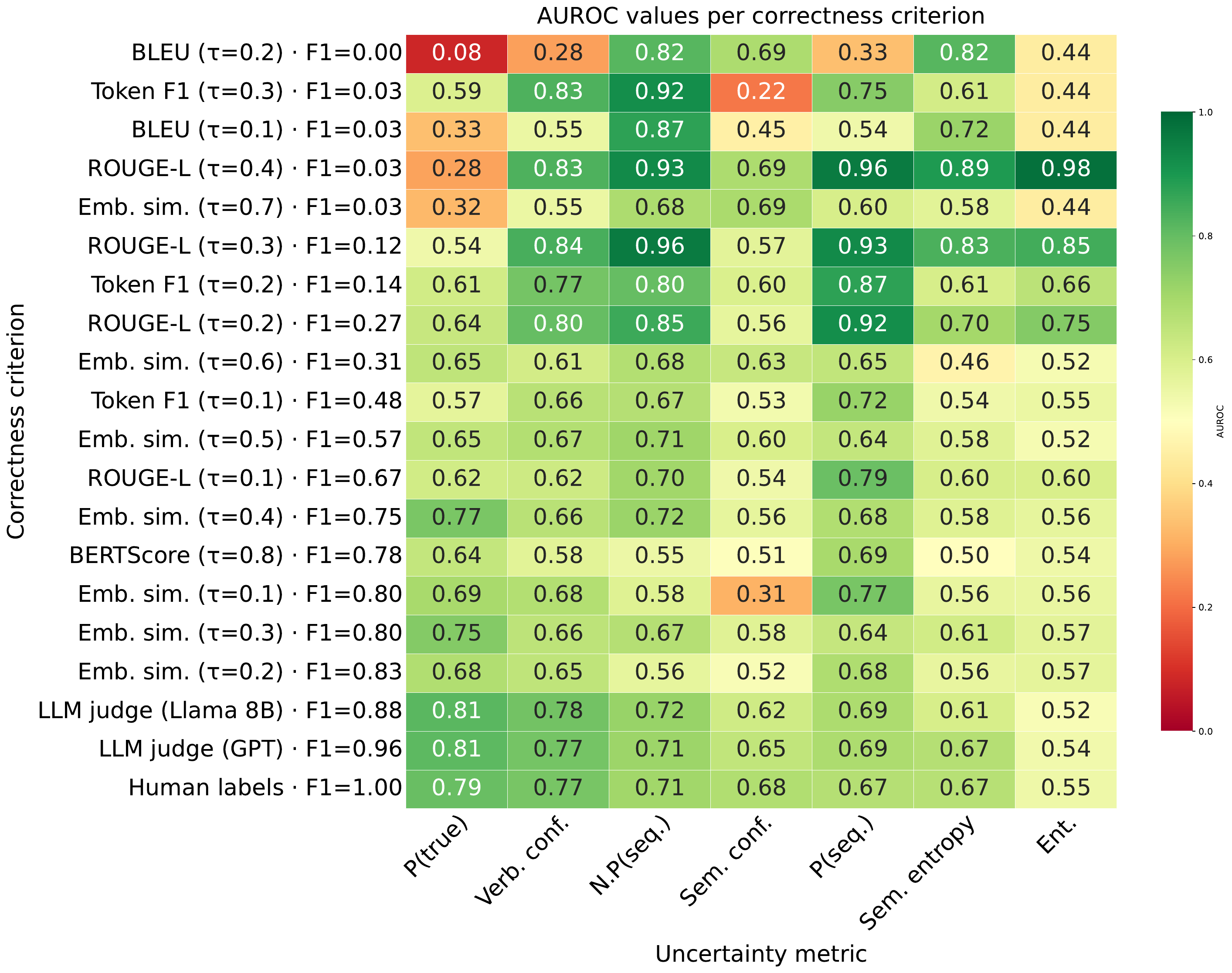}
    \end{subfigure}
    \hfill
    \begin{subfigure}[t]{0.48\textwidth}
        \centering
        \includegraphics[width=\linewidth]{figs/new/auroc_values_per_correctness_ambig_qa_multiple_qas_llama-8b-instruct_100_0.pdf}
    \end{subfigure}

    \vspace{0.75em}

    \begin{subfigure}[t]{0.48\textwidth}
        \centering
        \includegraphics[width=\linewidth]{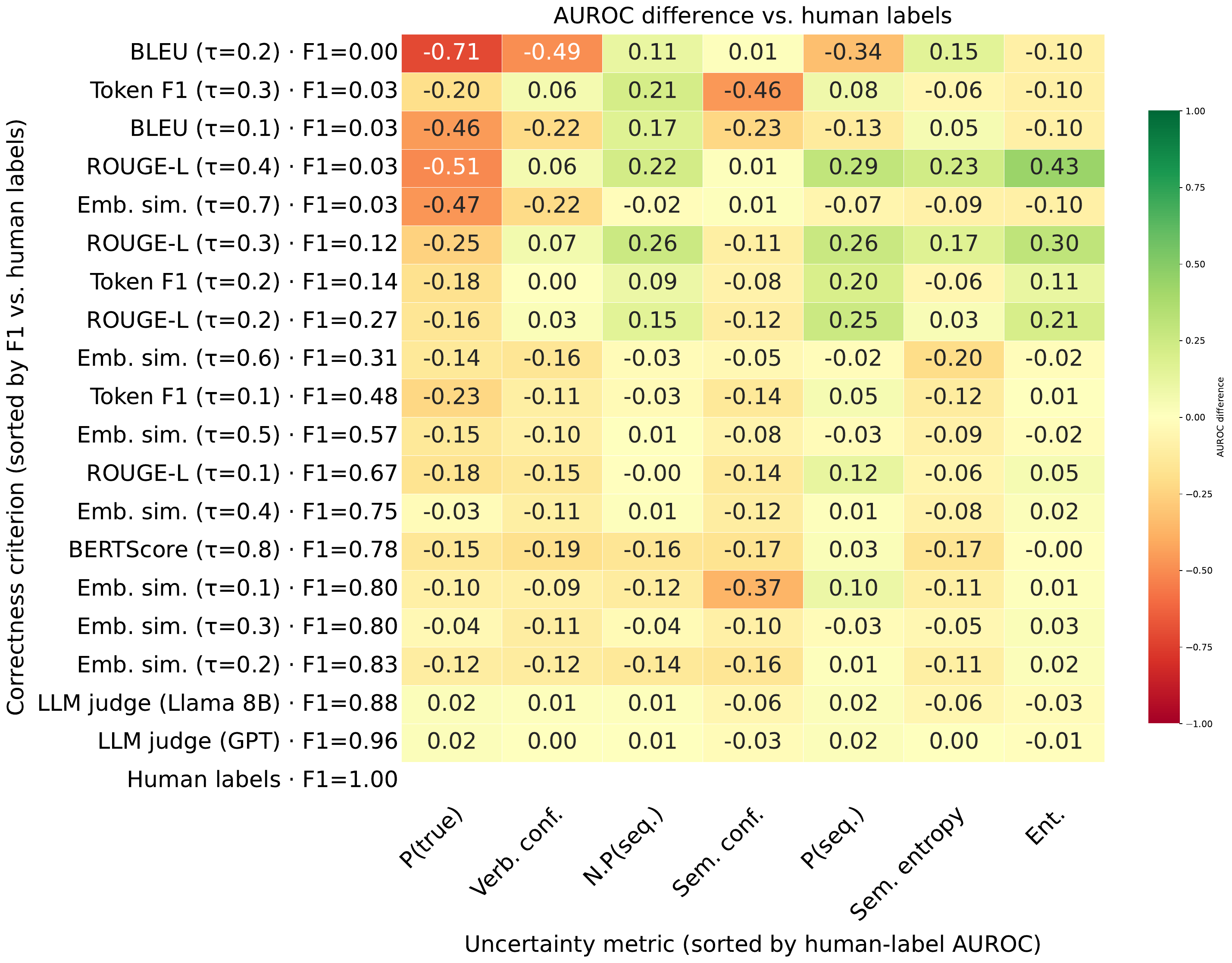}
    \end{subfigure}
    \hfill
    \begin{subfigure}[t]{0.48\textwidth}
        \centering
        \includegraphics[width=\linewidth]{figs/new/auroc_diff_per_correctness_ambig_qa_multiple_qas_llama-8b-instruct_100_0.pdf}
    \end{subfigure}

    \begin{subfigure}[t]{0.48\textwidth}
        \centering
        \includegraphics[width=\linewidth]{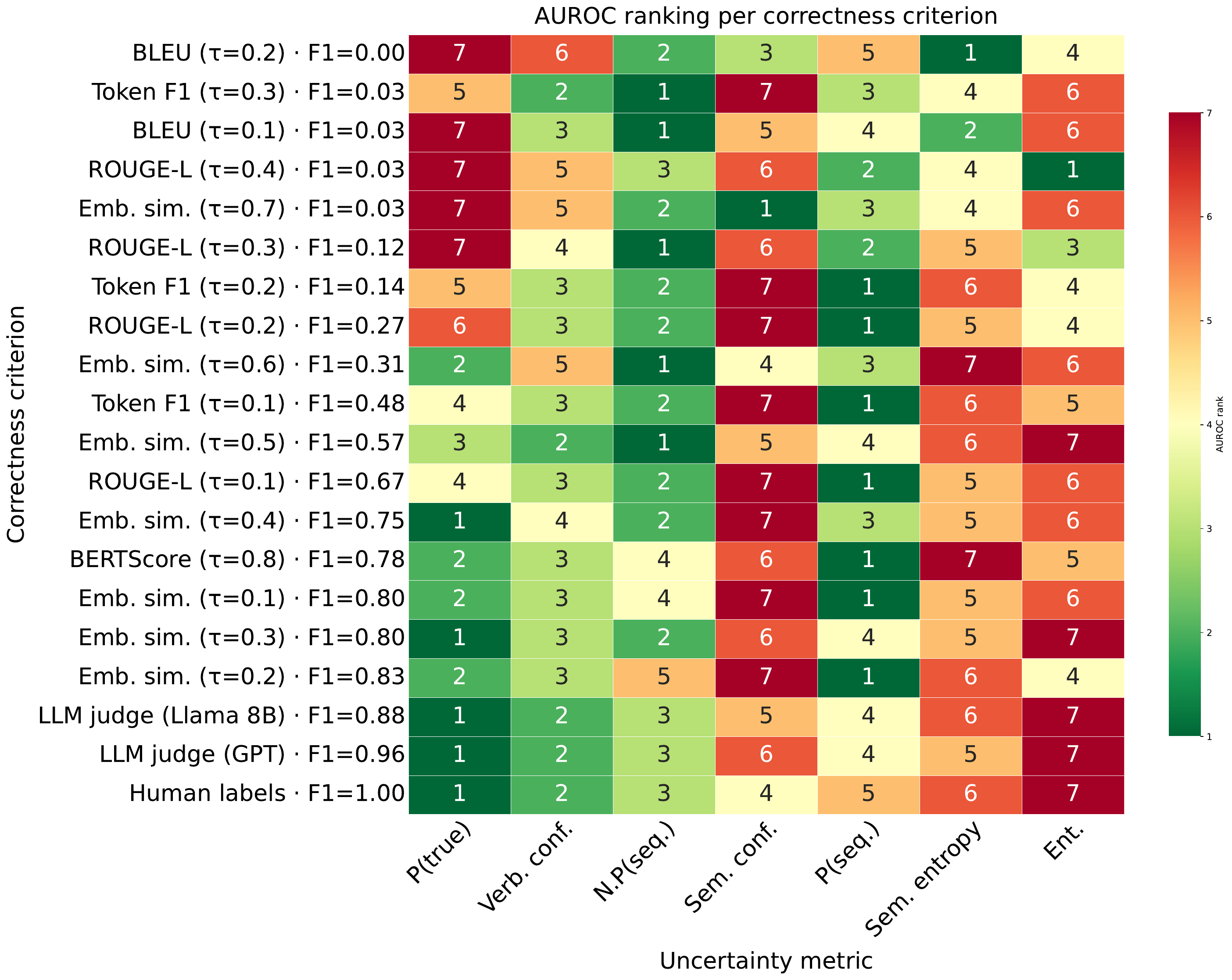}
        \caption{Llama 3B, AmbigQA (ambiguous), New annotator}
    \end{subfigure}
    \hfill
    \begin{subfigure}[t]{0.48\textwidth}
        \centering
        \includegraphics[width=\linewidth]{figs/new/auroc_ranking_per_correctness_ambig_qa_multiple_qas_llama-8b-instruct_100_0.pdf}
        \caption{Llama 8B, AmbigQA (ambiguous), Original annotator}
    \end{subfigure}

    \caption{Left column shows results given manual labels obtained from new annotator, and right column shows results from original annotator.}
    \label{fig:llama-ambigqa-mult-interannotator}
\end{figure*}

To verify that our results are not sensitive to the choice of judge, we employ a new annotator (another one of the authors) to annotate two sets of samples from TriviaQA-Llama 8B and AmbigQA (ambiguous)-Llama 8B. For the former, Cohen's Kappa score between annotators 0.88 and for the latter 0.75. We present result figures, side by side, for both annotators (see Figures \ref{fig:llama-triviaqa-interannotator} and \ref{fig:llama-ambigqa-mult-interannotator}).

\subsection{Bootstrapping analysis}
\label{sec:appendix:bootstrat}
\begin{figure*}[t]
    \centering

    \includegraphics[width=\linewidth]{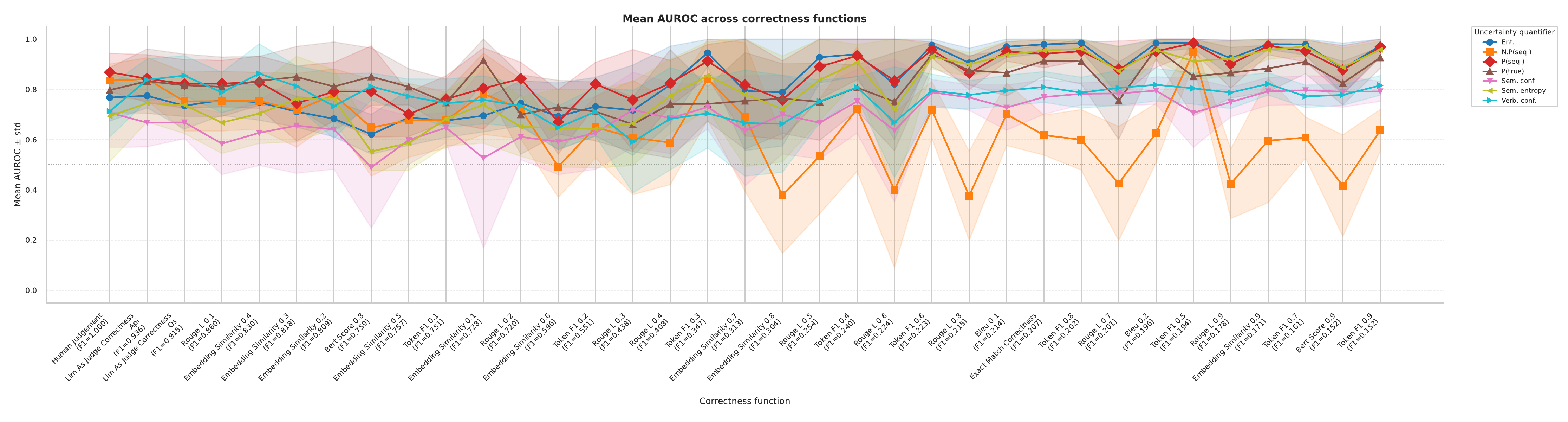}

    \includegraphics[width=\linewidth]{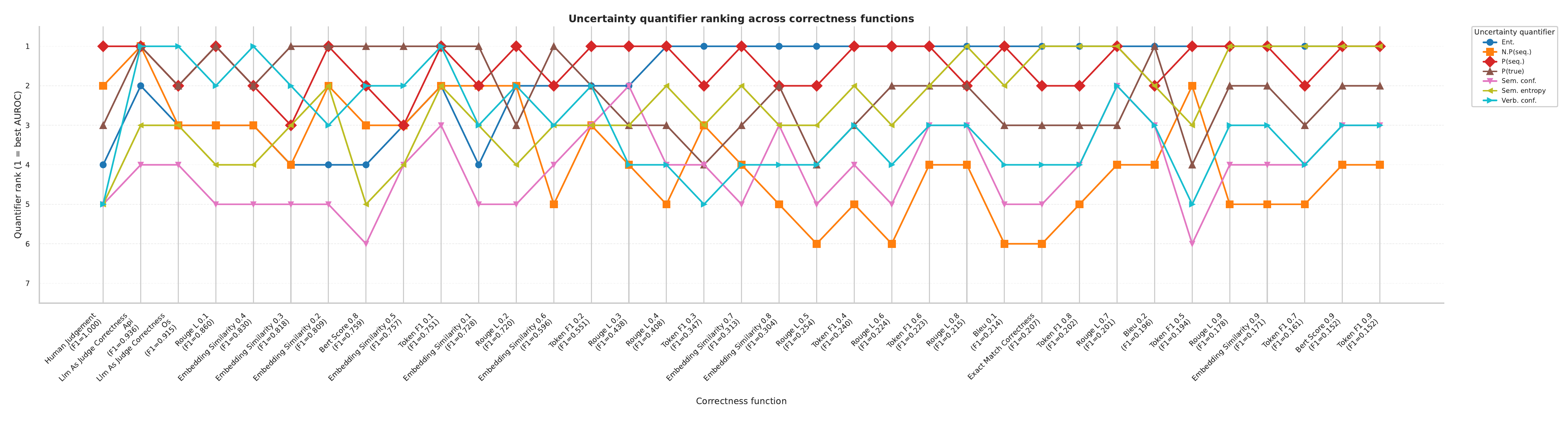}

    
    \includegraphics[width=\linewidth]{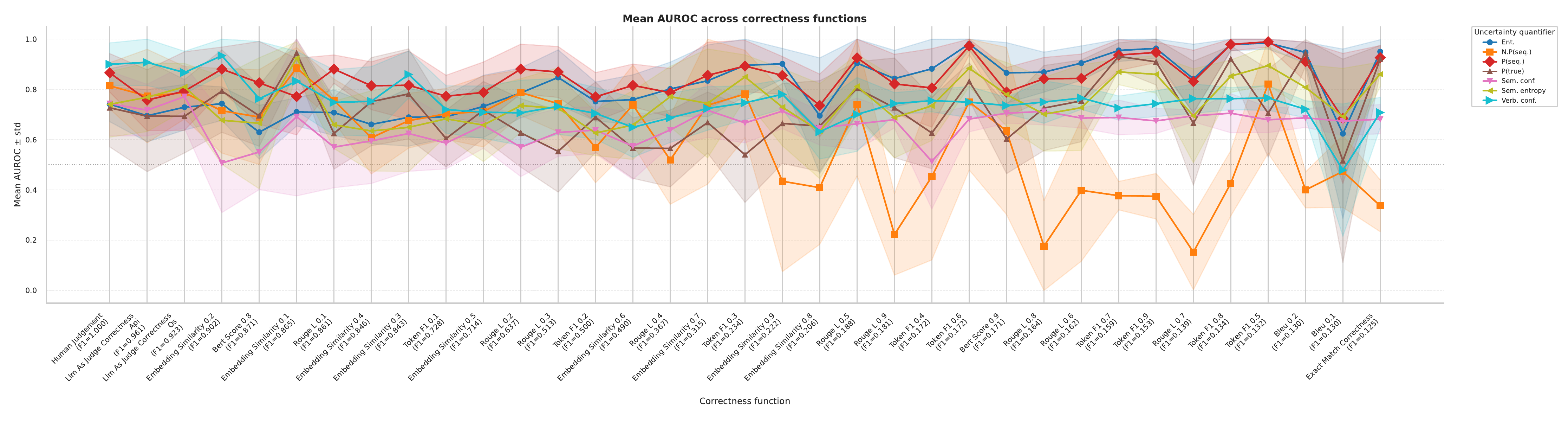}

    \includegraphics[width=\linewidth]{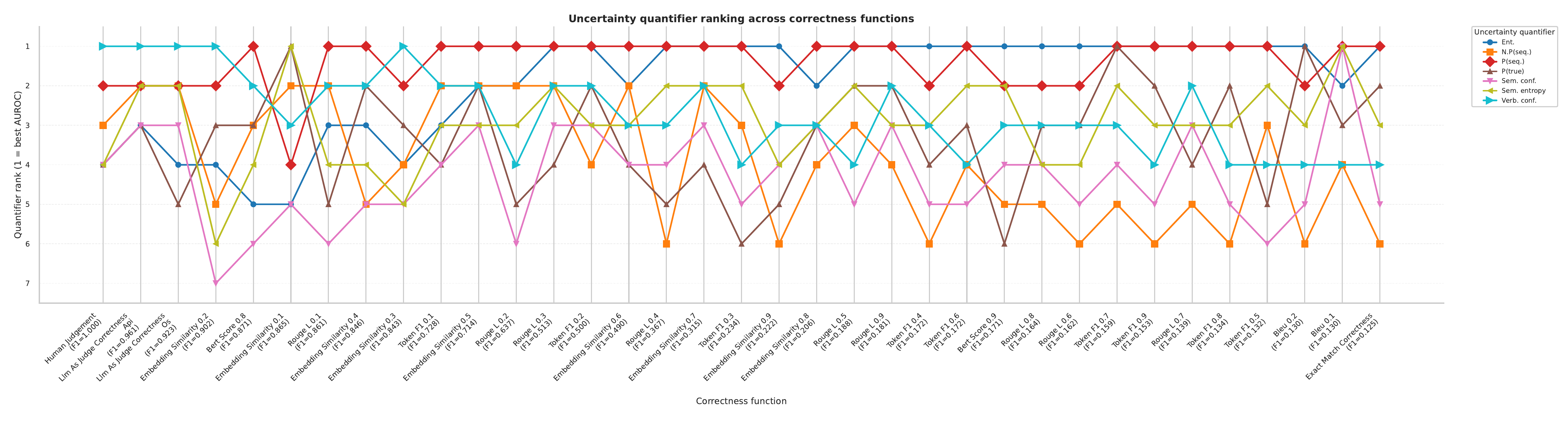}

   \caption{Top two plots correspond to Llama 3B, TriviaQA. Bottom two plots correspond to Llama8b, TriviaQA.}
    \label{fig:bootstrap-llama-triviaqa}
\end{figure*}

\begin{figure*}[t]
    \centering

    \includegraphics[width=\linewidth]{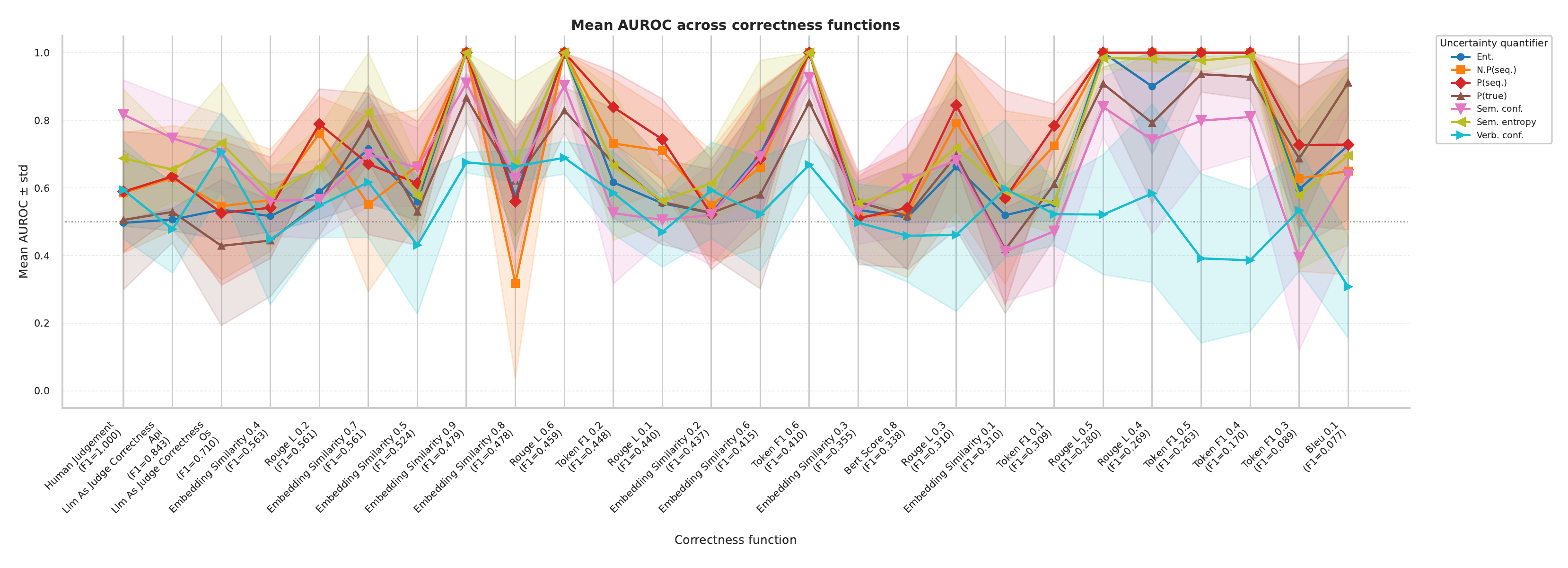}

    \includegraphics[width=\linewidth]{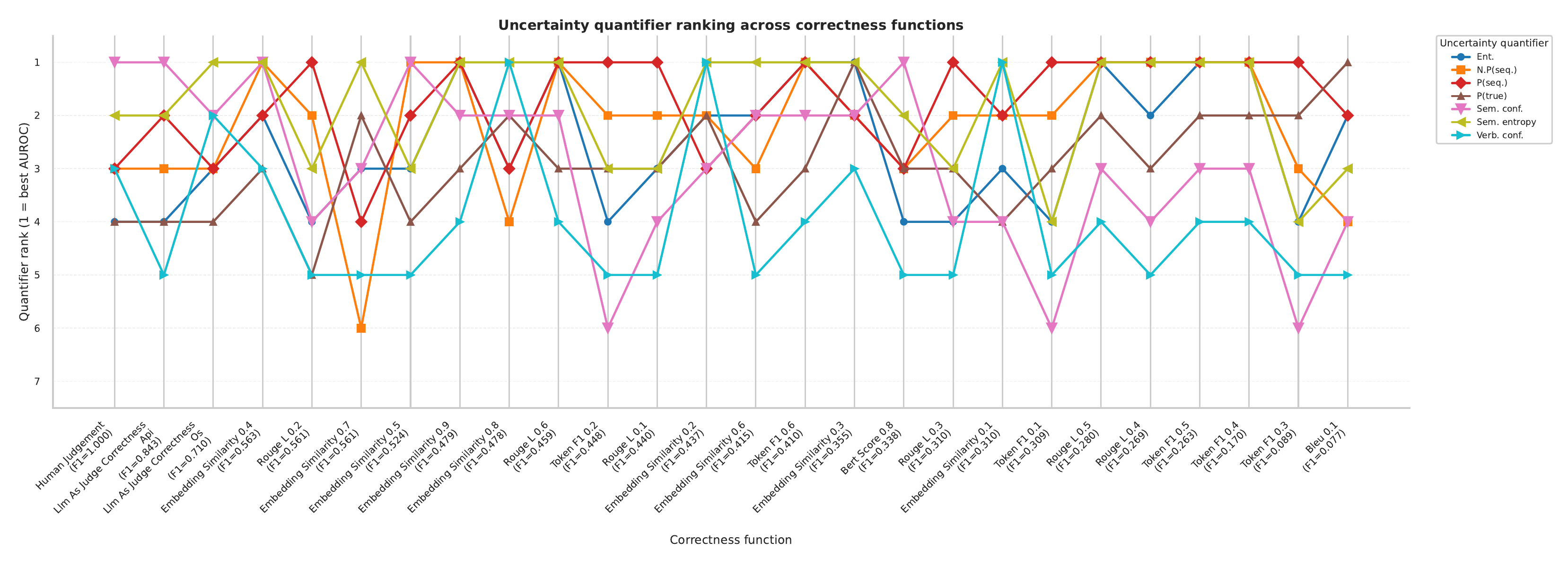}

    
    \includegraphics[width=\linewidth]{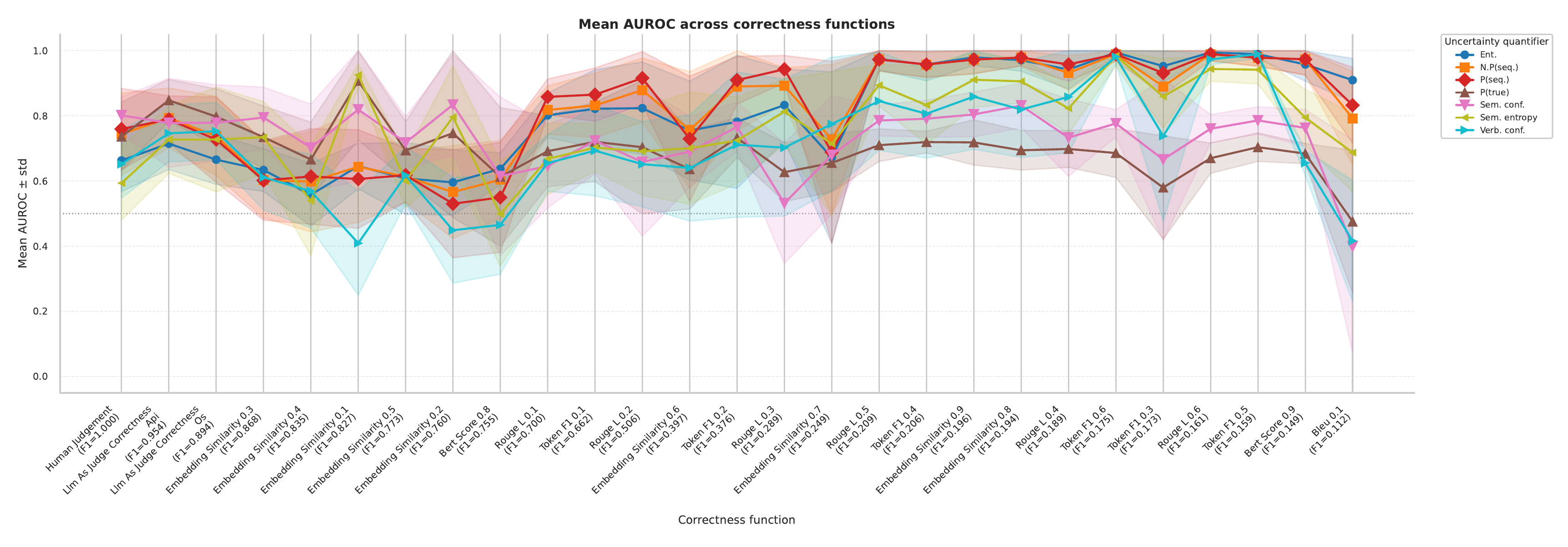}

    \includegraphics[width=\linewidth]{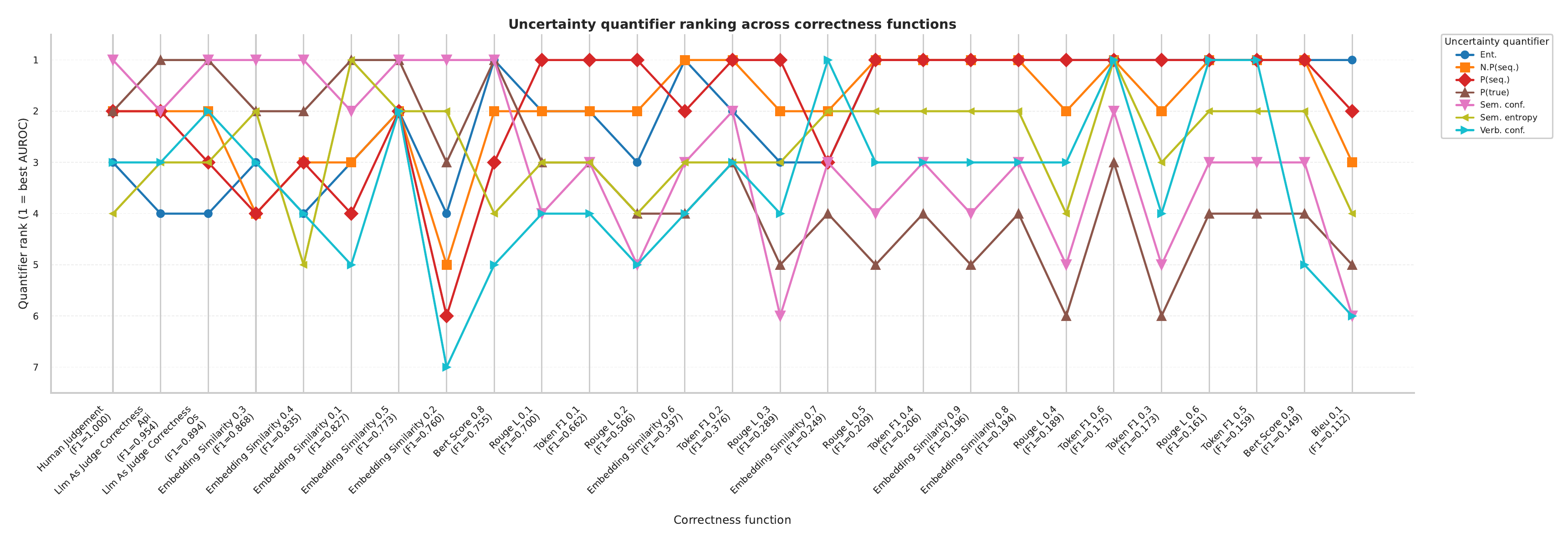}

    \caption{Top two plots correspond to Qwen 0.5B, TriviaQA. Bottom two plots correspond to Qwen 7B, TriviaQA.}
    \label{fig:bootstrap-qwen-triviaqa}
\end{figure*}

\begin{figure*}[t]
    \centering

    \includegraphics[width=0.8\linewidth]{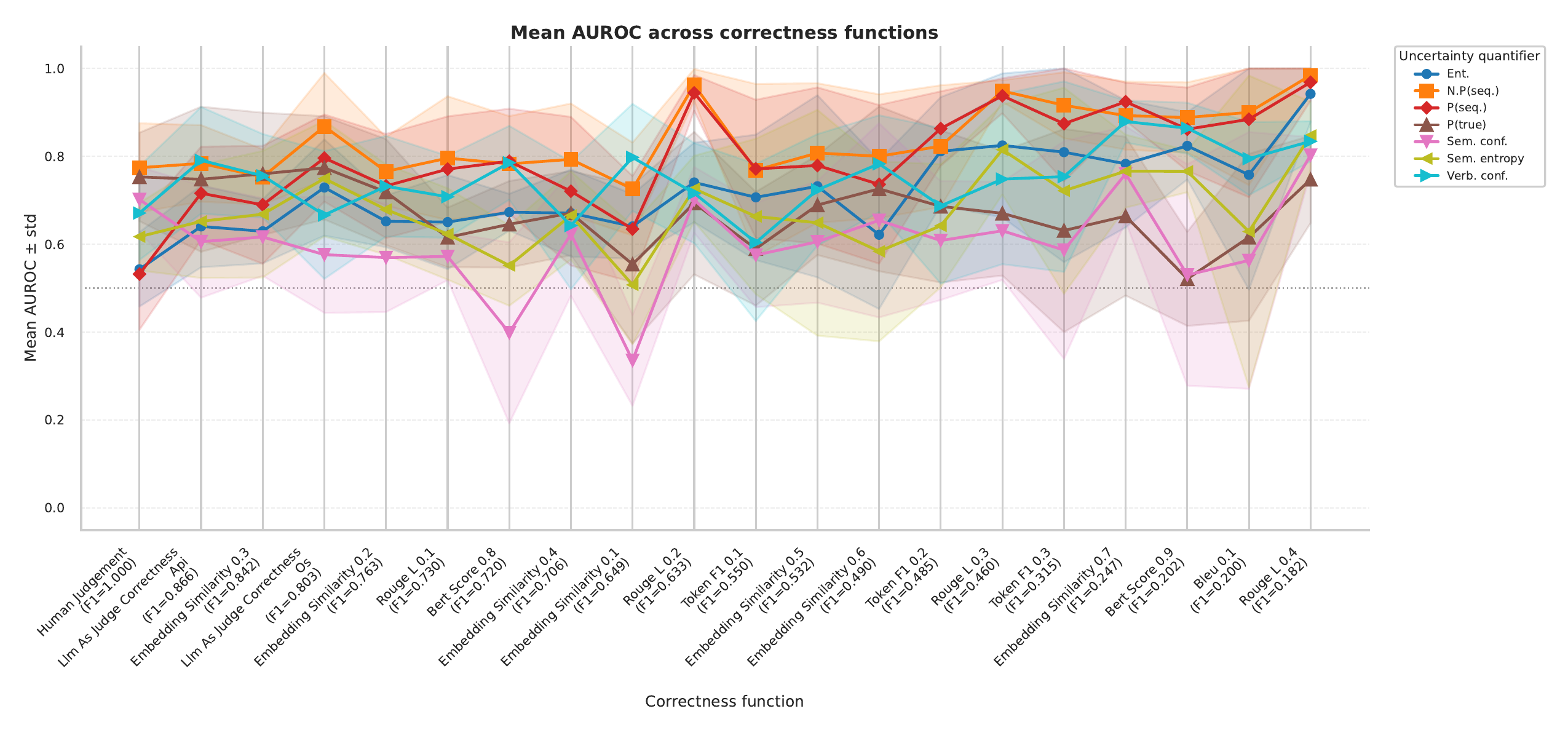}

    \includegraphics[width=0.8\linewidth]{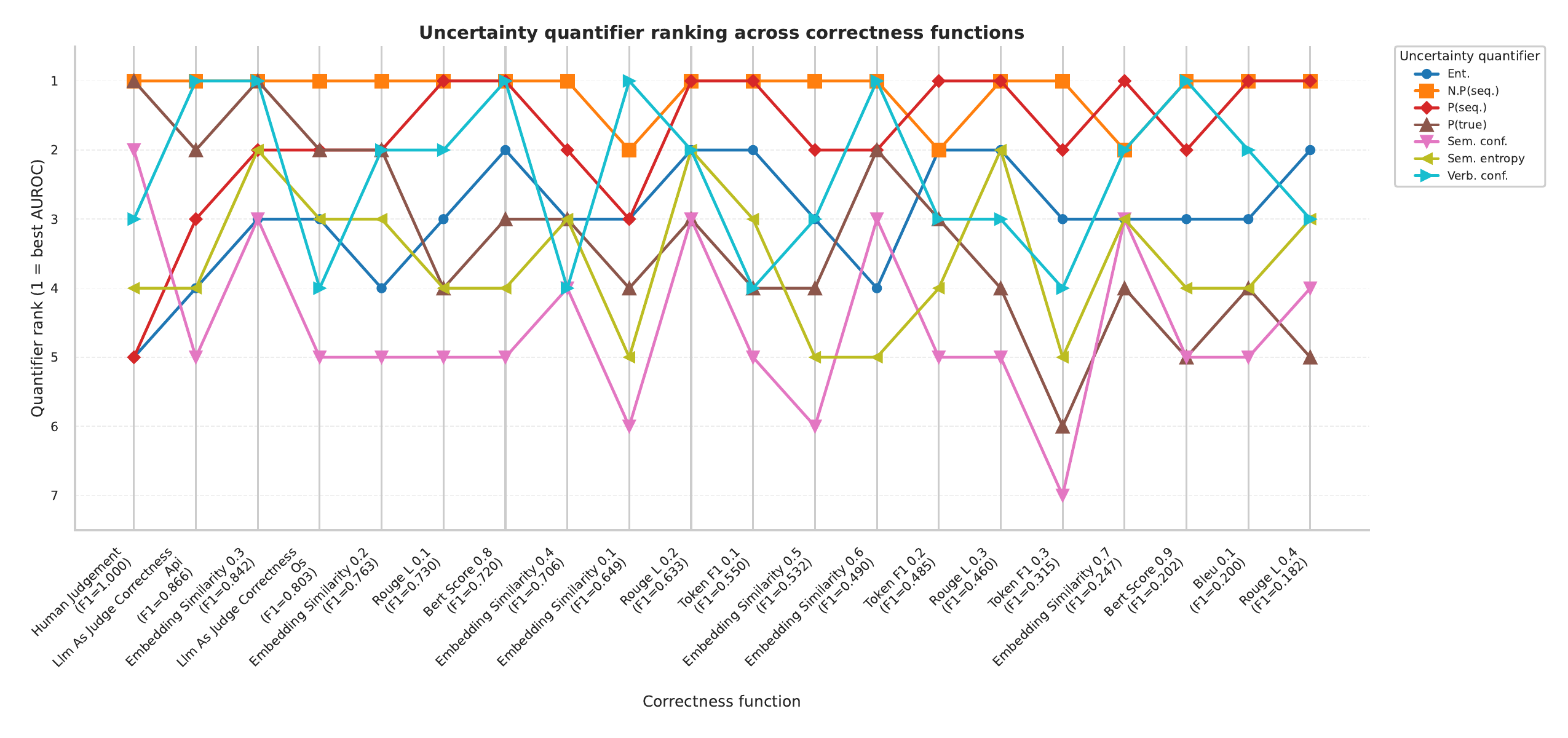}

    
    \includegraphics[width=\linewidth]{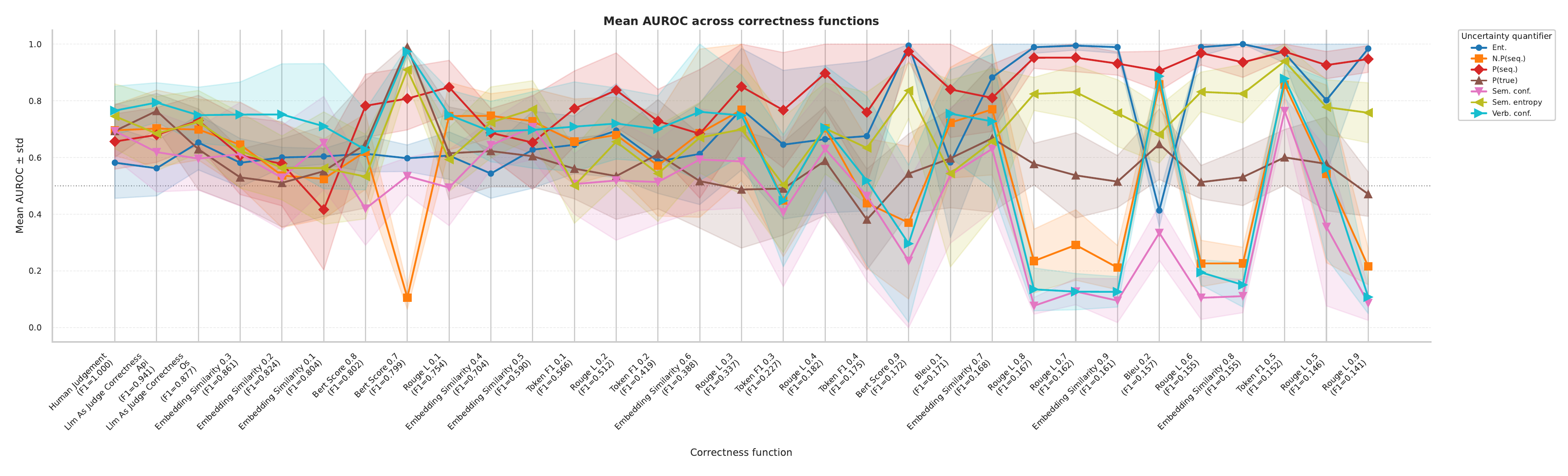}

    \includegraphics[width=\linewidth]{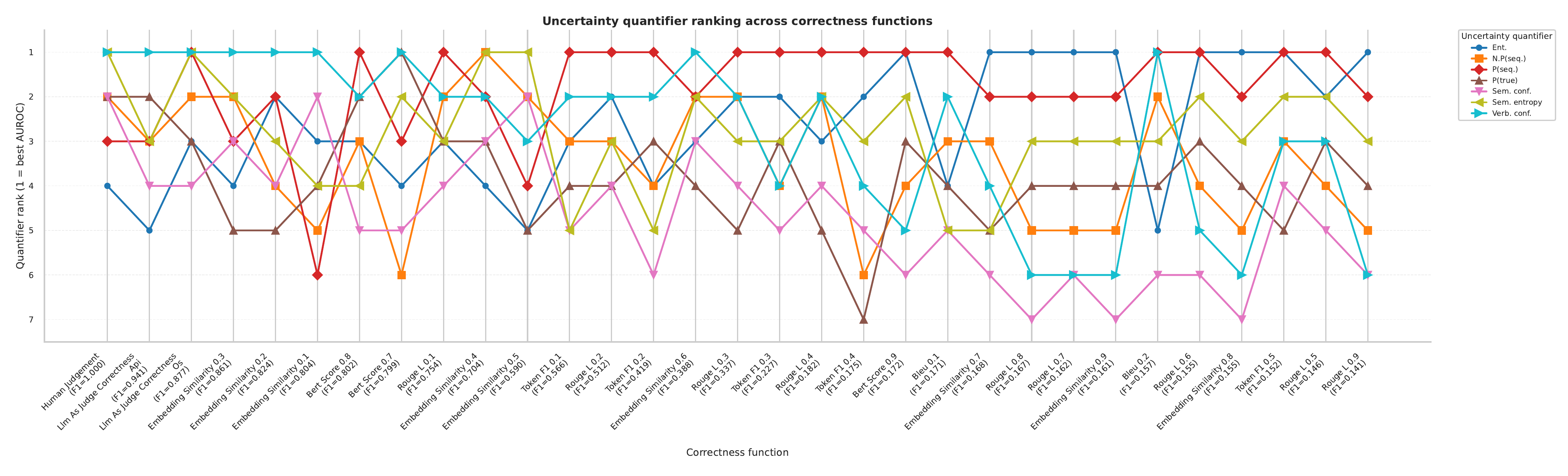}

    \caption{Top two plots correspond to Llama 3B, AmbigQA (unambiguous). Bottom two plots correspond to Llama 8B, AmbigQA (unambiguous).}
    \label{fig:bootstrap-llama-ambigqa_sing}
\end{figure*}

\begin{figure*}[t]
    \centering

    \includegraphics[width=0.65\linewidth]{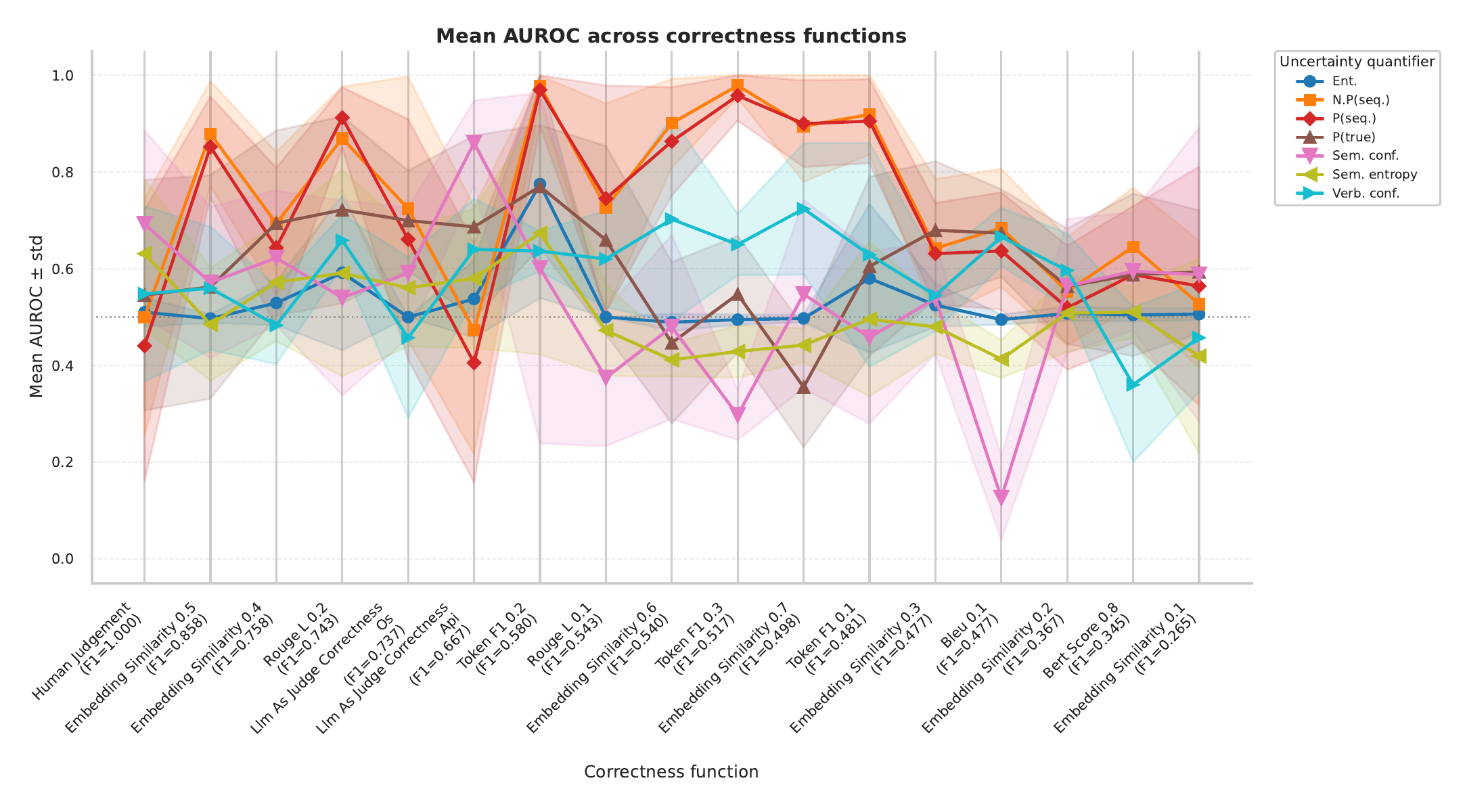}

    \includegraphics[width=0.65\linewidth]{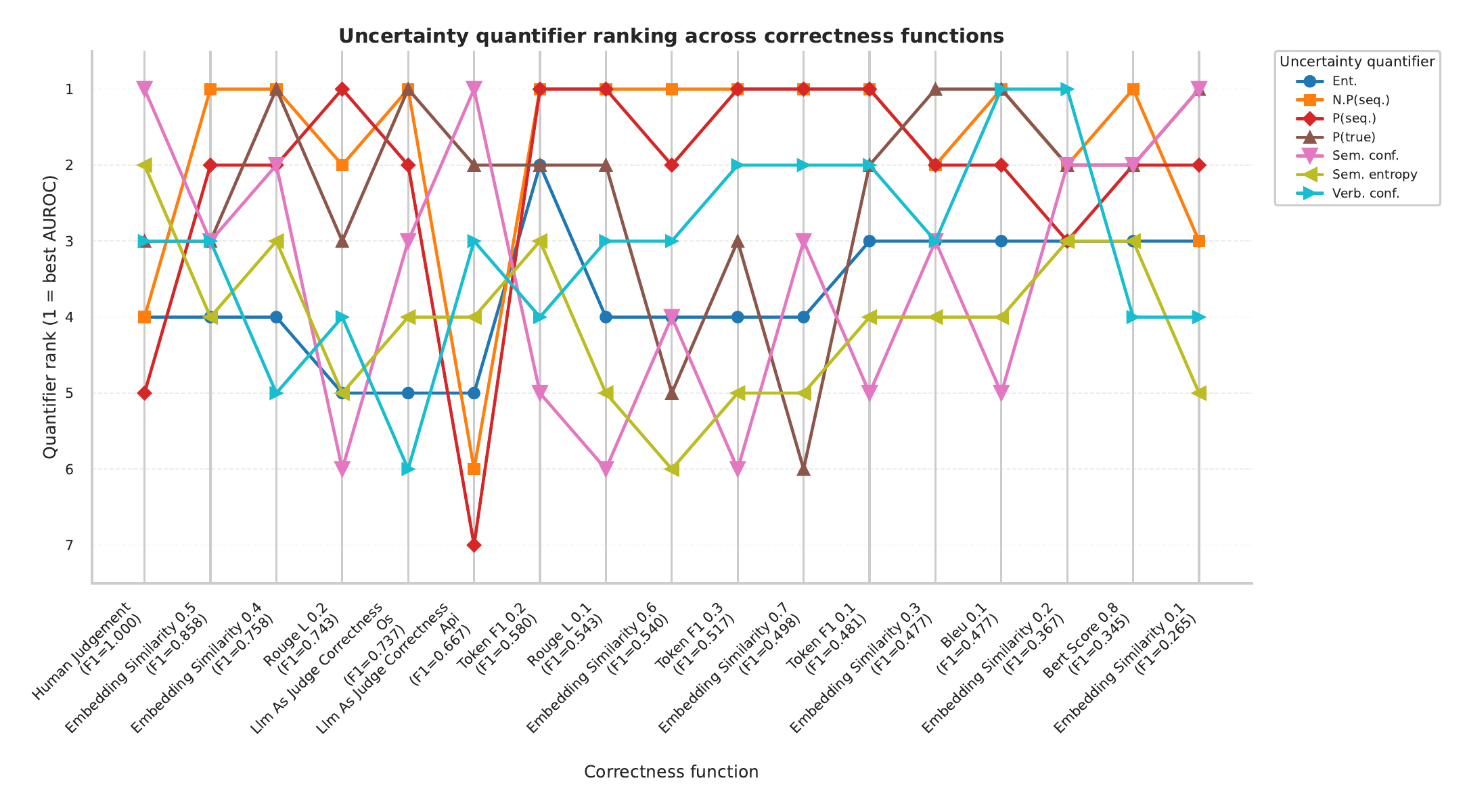}

    
    \includegraphics[width=0.65\linewidth]{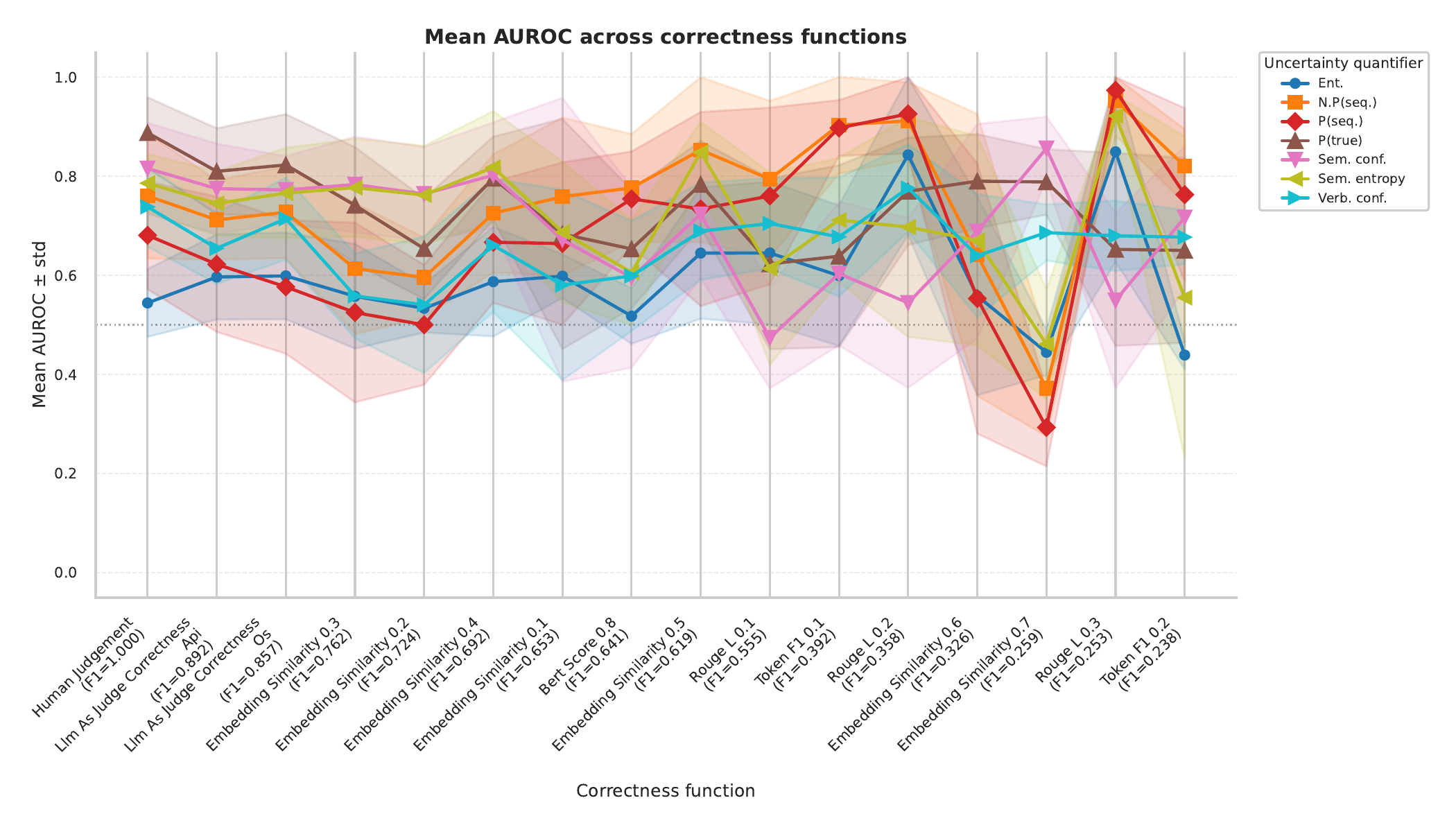}

    \includegraphics[width=0.65\linewidth]{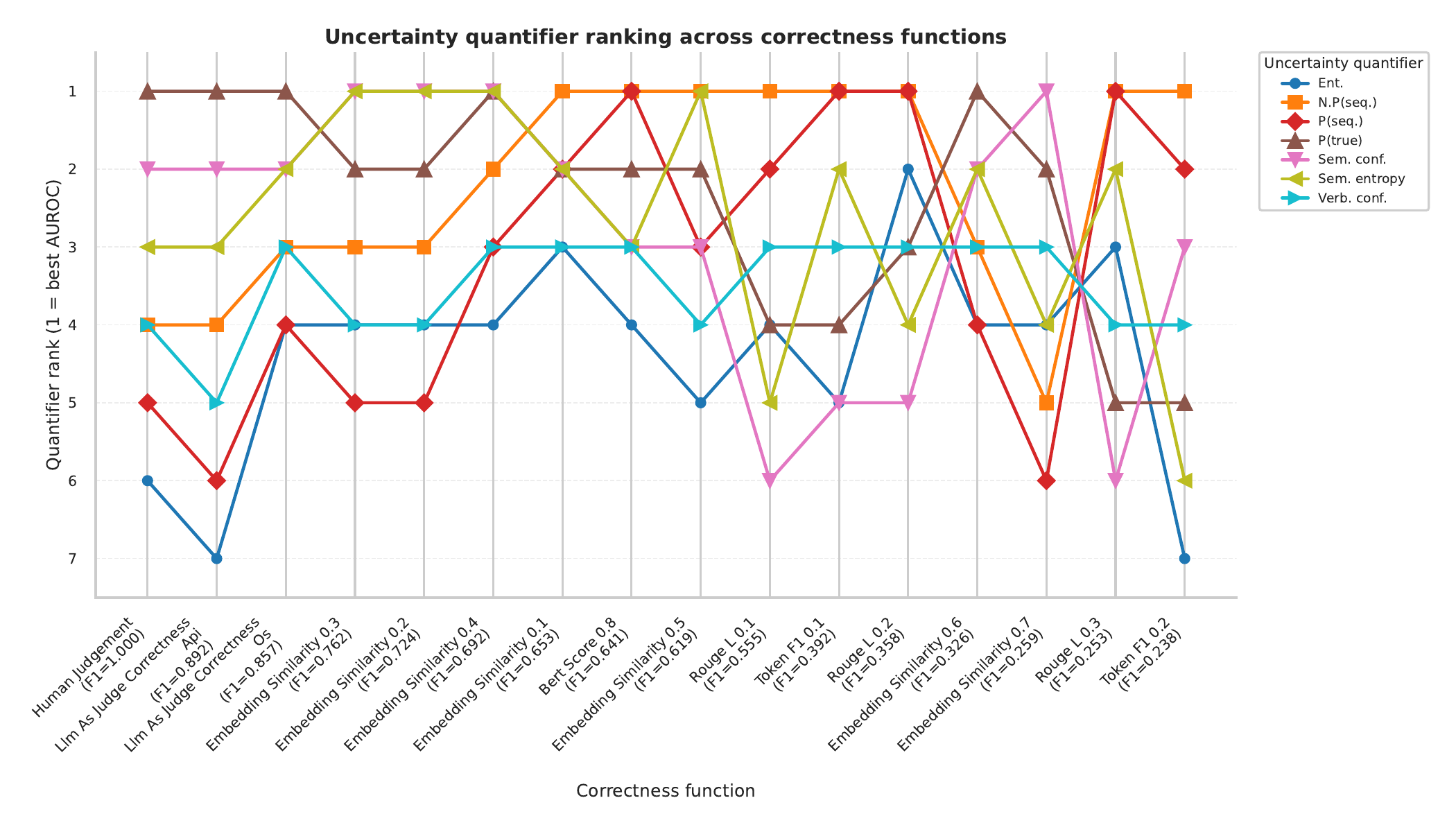}

    \caption{Top two plots correspond to Qwen 0.5B, AmbigQA (unambiguous). Bottom two plots correspond to Qwen 7B, AmbigQA (unambiguous).}
    \label{fig:bootstrap-qwen-ambigqa_sing}
\end{figure*}

\begin{figure*}[t]
    \centering

    \includegraphics[width=0.8\linewidth]{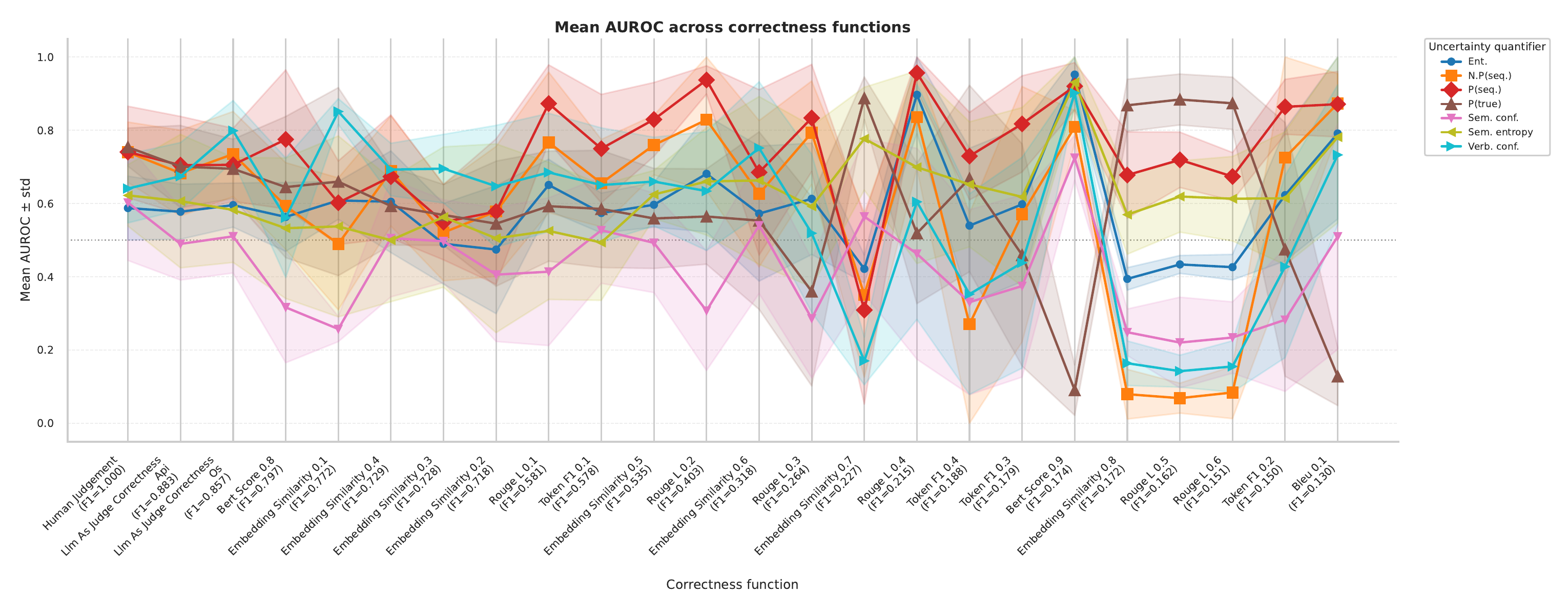}

    \includegraphics[width=0.8\linewidth]{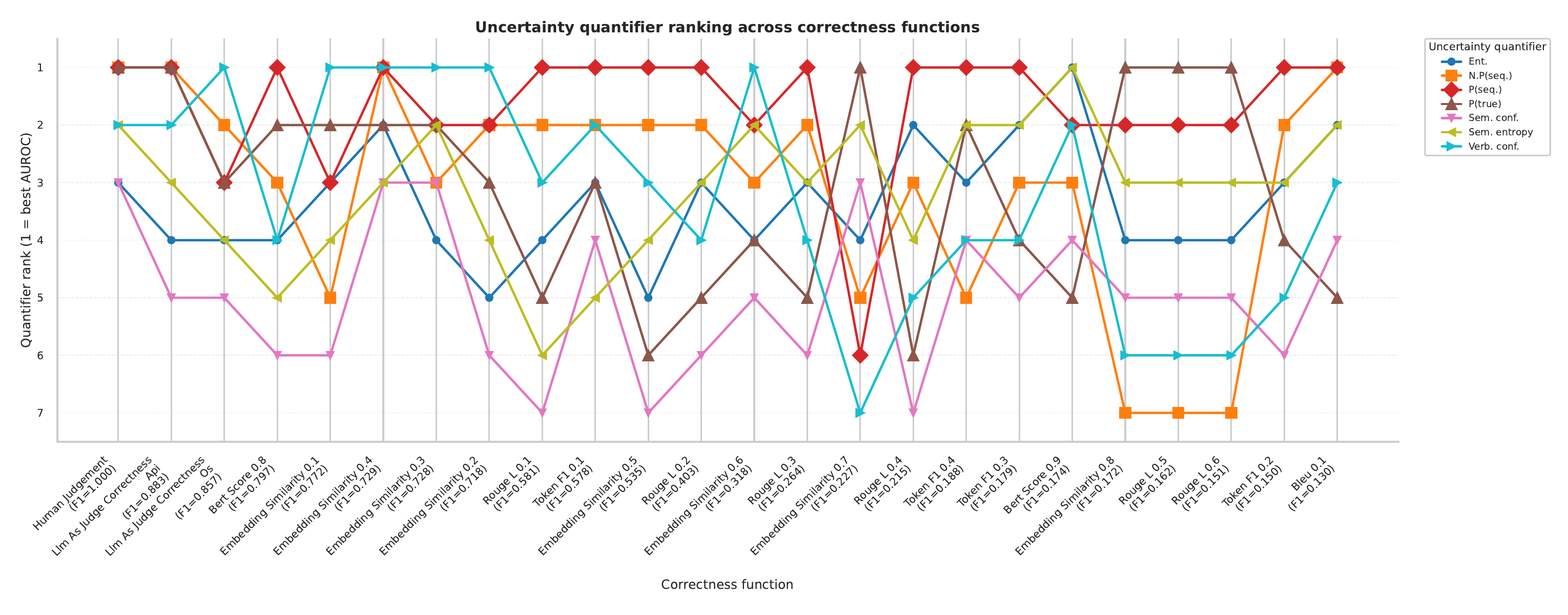}

    
    \includegraphics[width=0.8\linewidth]{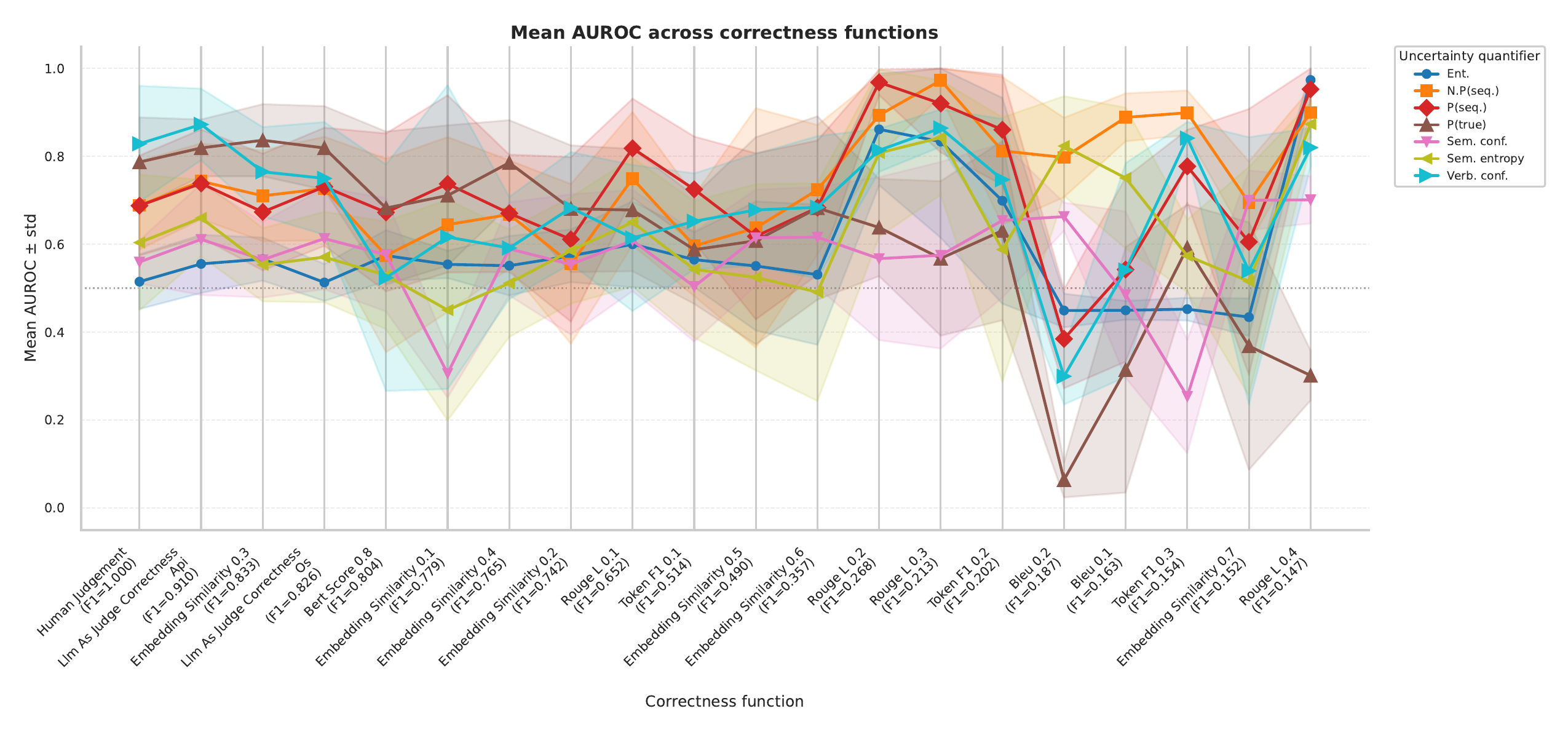}

    \includegraphics[width=0.8\linewidth]{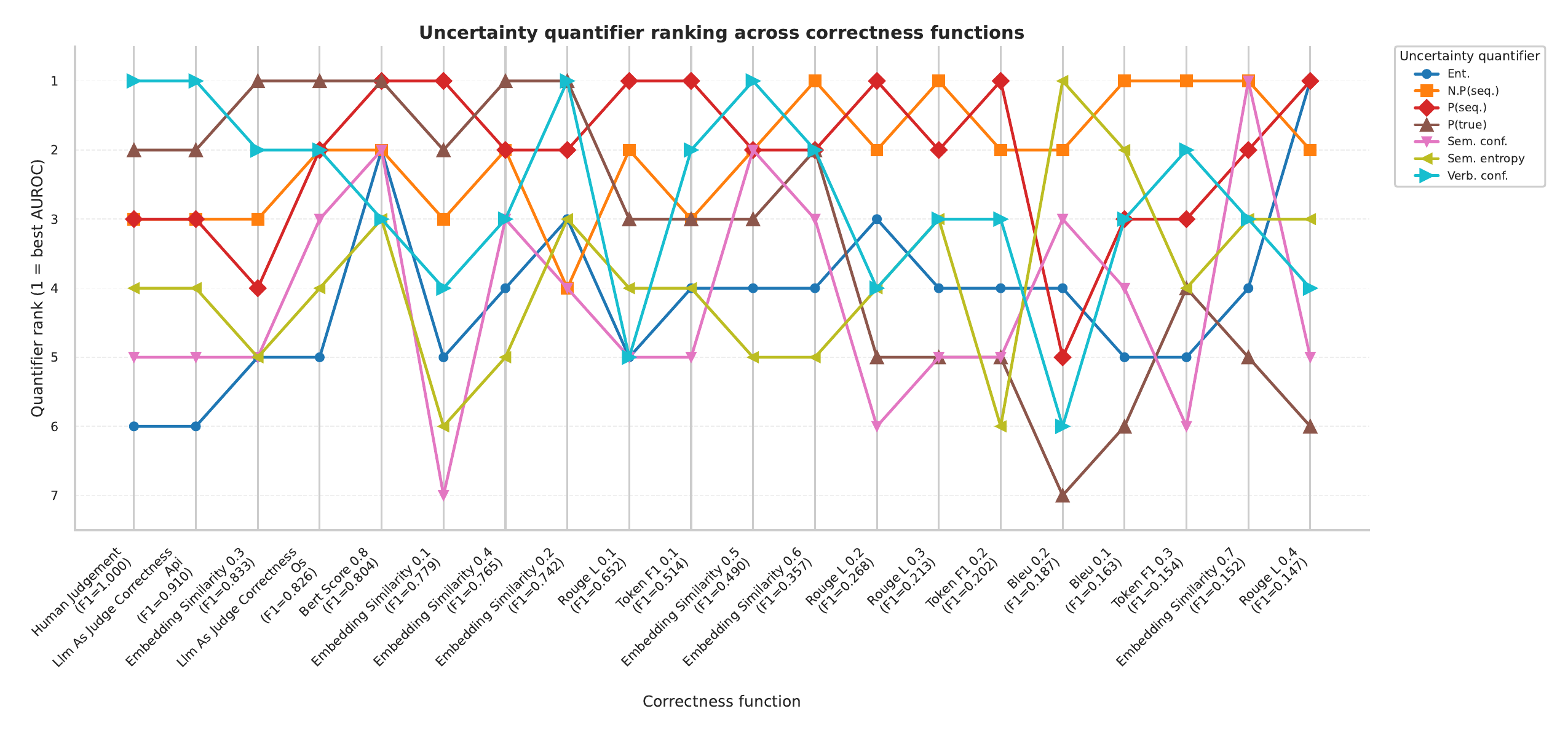}

    \caption{Top two plots correspond to Llama 3B, AmbigQA (ambiguous). Bottom two plots correspond to Llama 8B, AmbigQA (ambiguous).}
    \label{fig:bootstrap-llama-ambigqa_mul}
\end{figure*}

\begin{figure*}[t]
    \centering

    \includegraphics[width=0.7\linewidth]{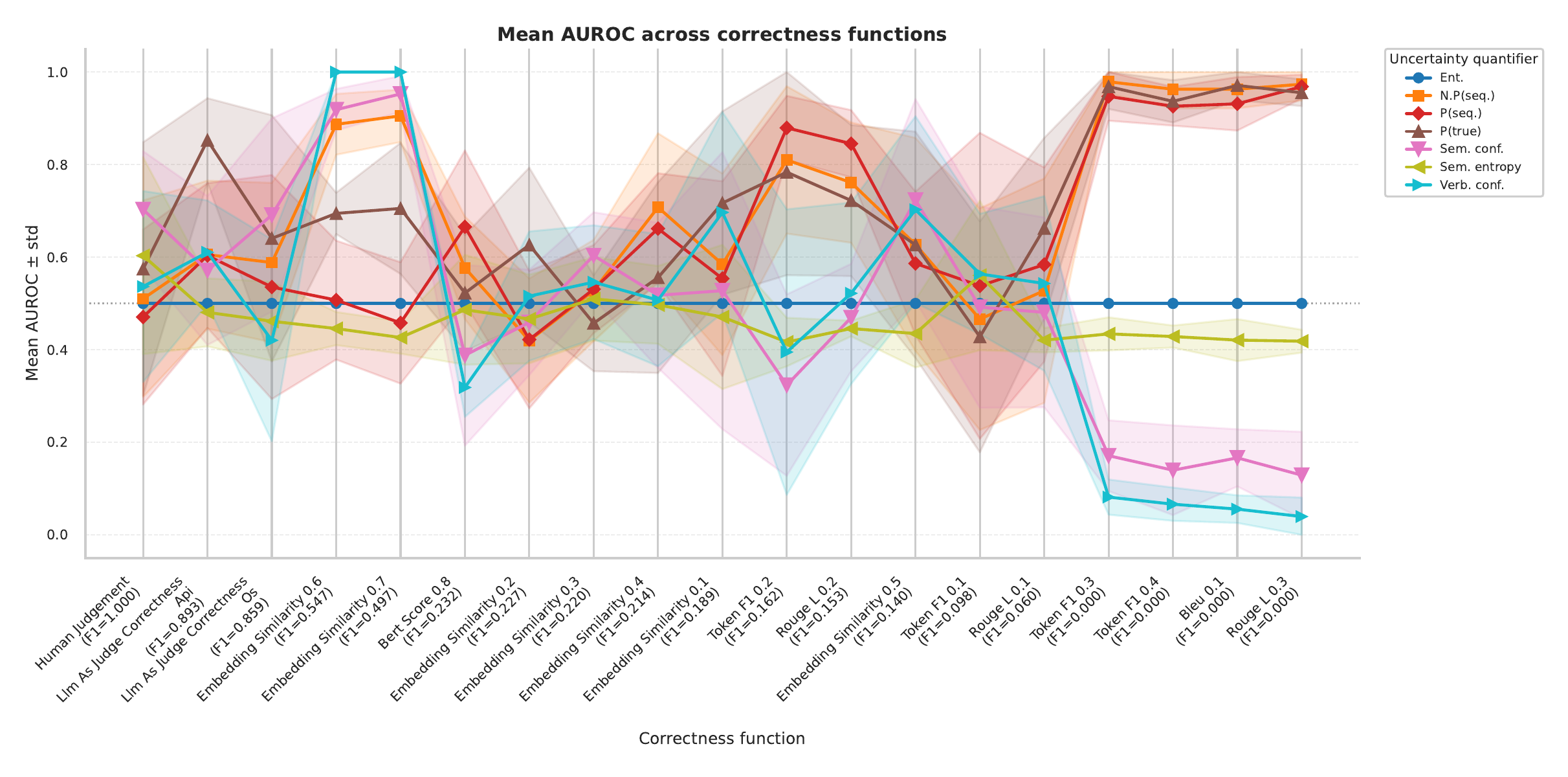}

    \includegraphics[width=0.7\linewidth]{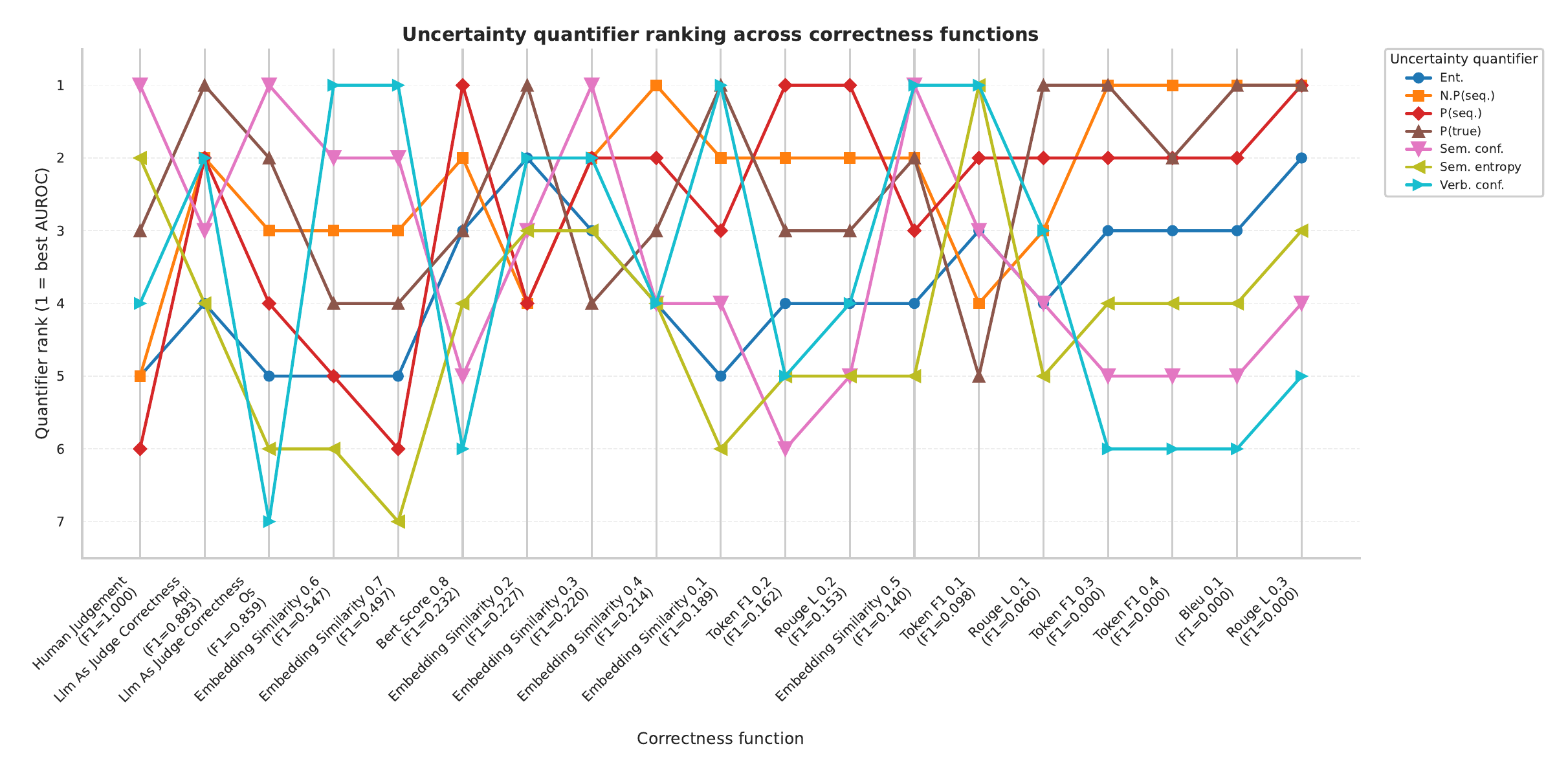}

    
    \includegraphics[width=0.7\linewidth]{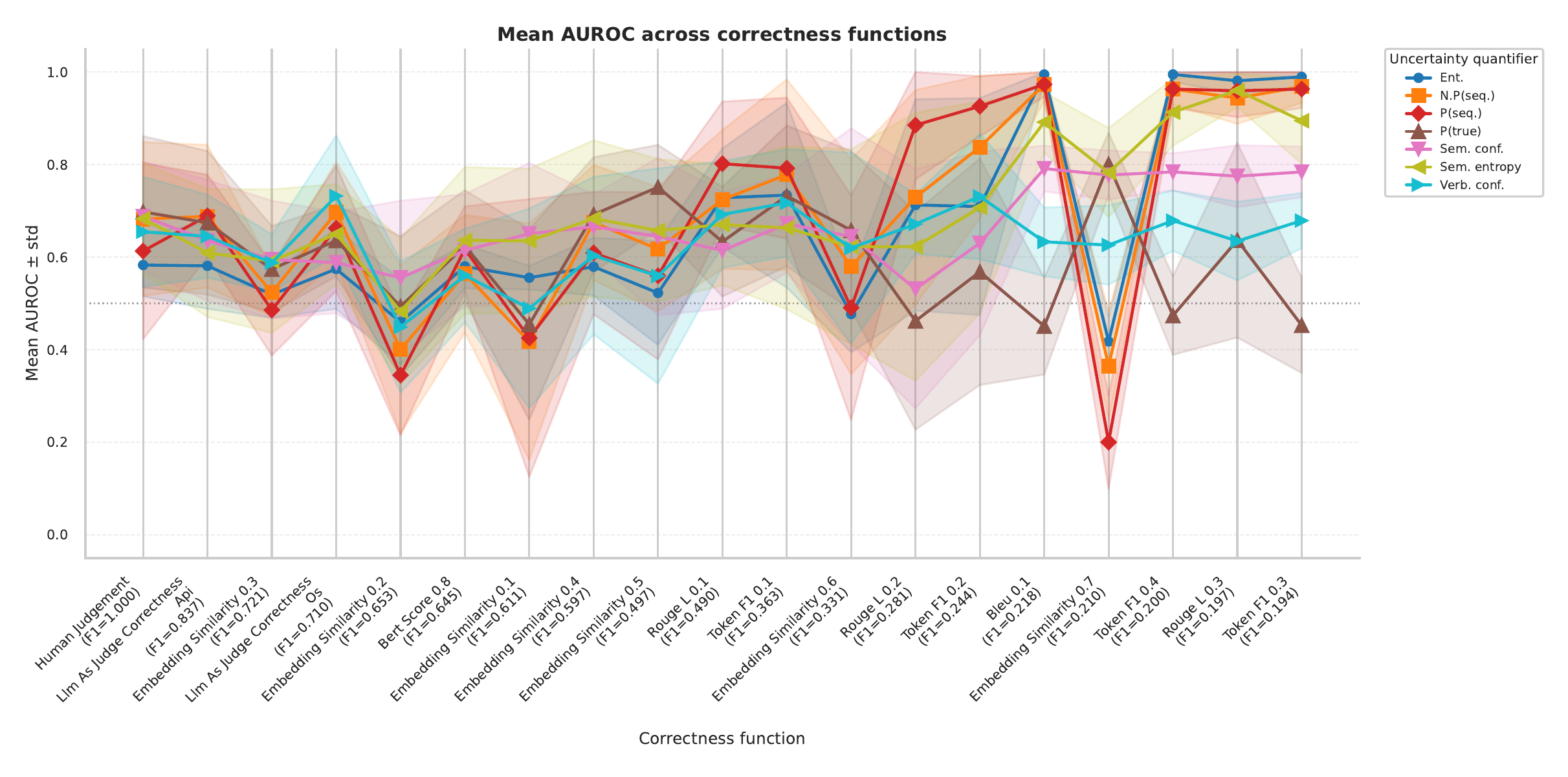}

    \includegraphics[width=0.7\linewidth]{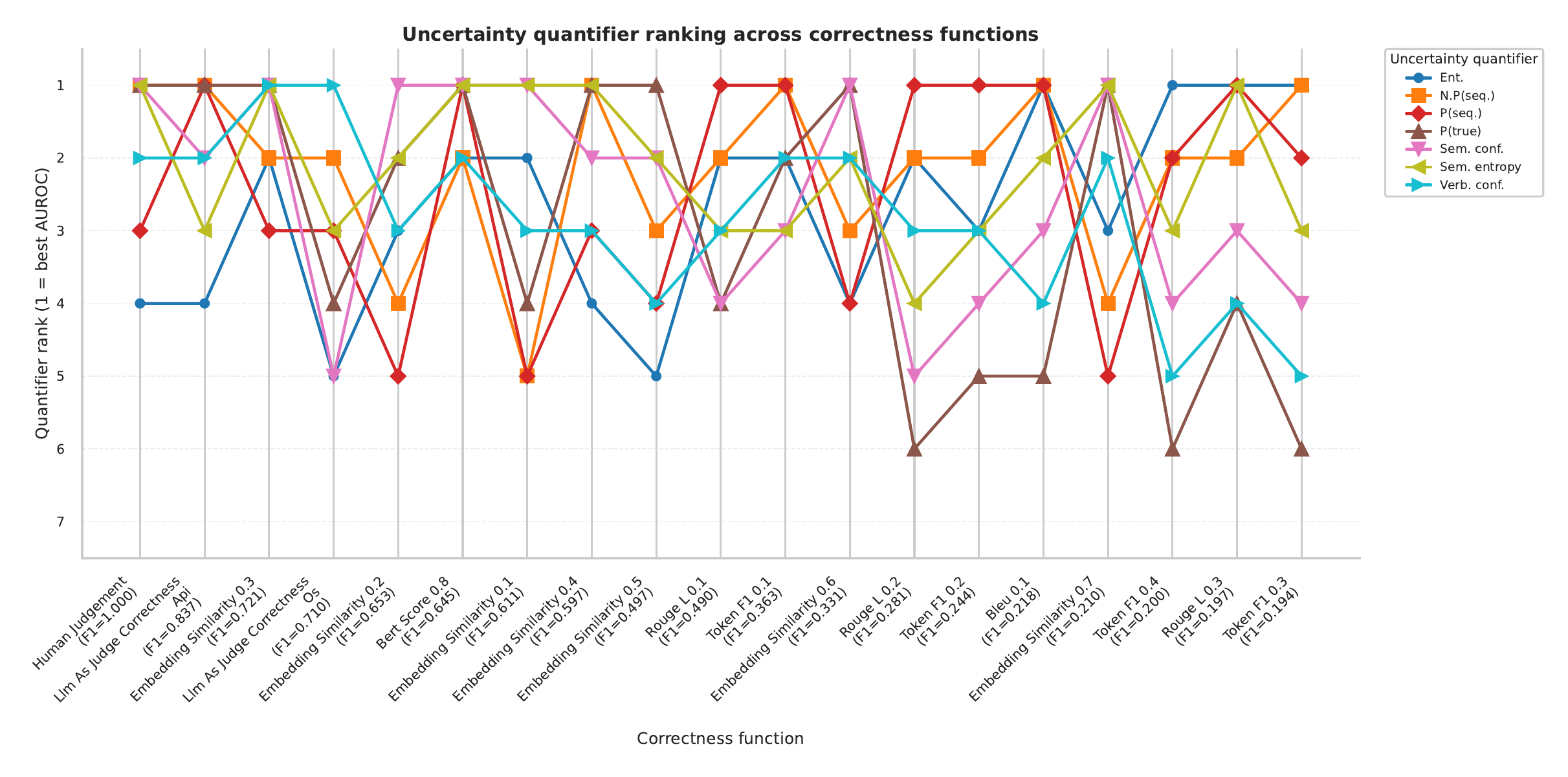}

    \caption{Top two plots correspond to Qwen 0.5B, AmbigQA (ambiguous). Bottom two plots correspond to Qwen 7B, AmbigQA (ambiguous).}
    \label{fig:bootstrap-qwen-ambigqa_mul}
\end{figure*}

To further validate our results, we perform a bootstrapping analysis: sampling 20 prompts 10 times, and computing the average performance metrics across runs. We compute the average AUROC (and its associated standard deviation values) across runs and the ranking of uncertainty quantifiers based on the average AUROC scores. 
To be less sensitive to small deviations of AUROC values when ranking, we assign 2 UQs the same ranking if their AUROCs do not vary by more than 0.025.
Results appear in Figures \ref{fig:bootstrap-llama-triviaqa} - \ref{fig:bootstrap-qwen-ambigqa_mul}.
The quantifiers are ordered by decreasing quality (using F1 score comparing their predictions with human annotations as a proxy), with the leftmost scores showing the actual AUROC values and rankings (which are computed using manual correctness labels). The results present a similar picture with the original outcomes.

\subsection{Alternative metrics}
\label{sec:appendix:other_metrics}

\begin{figure*}[t]
    \centering

    \begin{subfigure}[t]{0.48\textwidth}
        \centering
        \includegraphics[width=\linewidth]{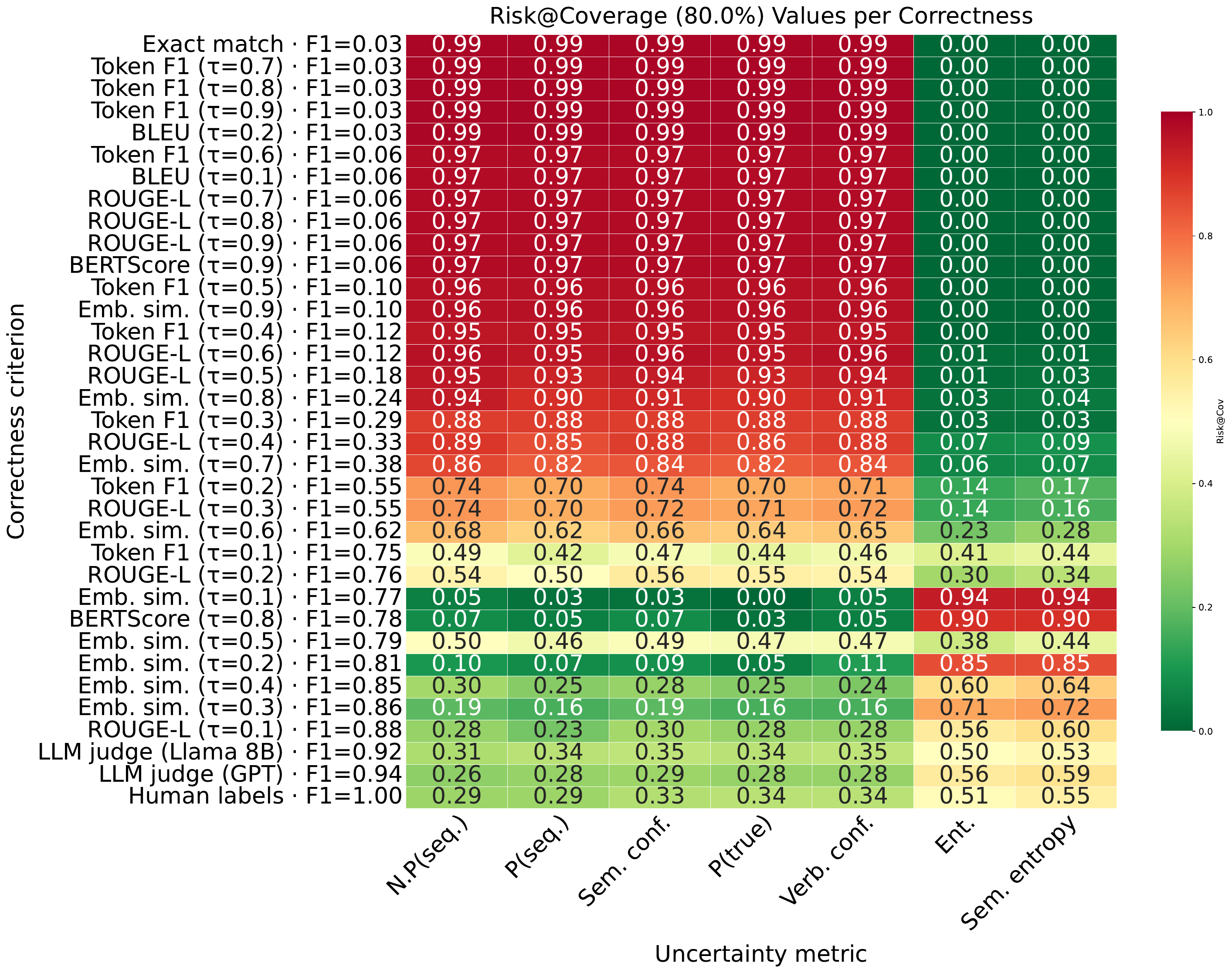}
    \end{subfigure}
    \hfill
    \begin{subfigure}[t]{0.48\textwidth}
        \centering
        \includegraphics[width=\linewidth]{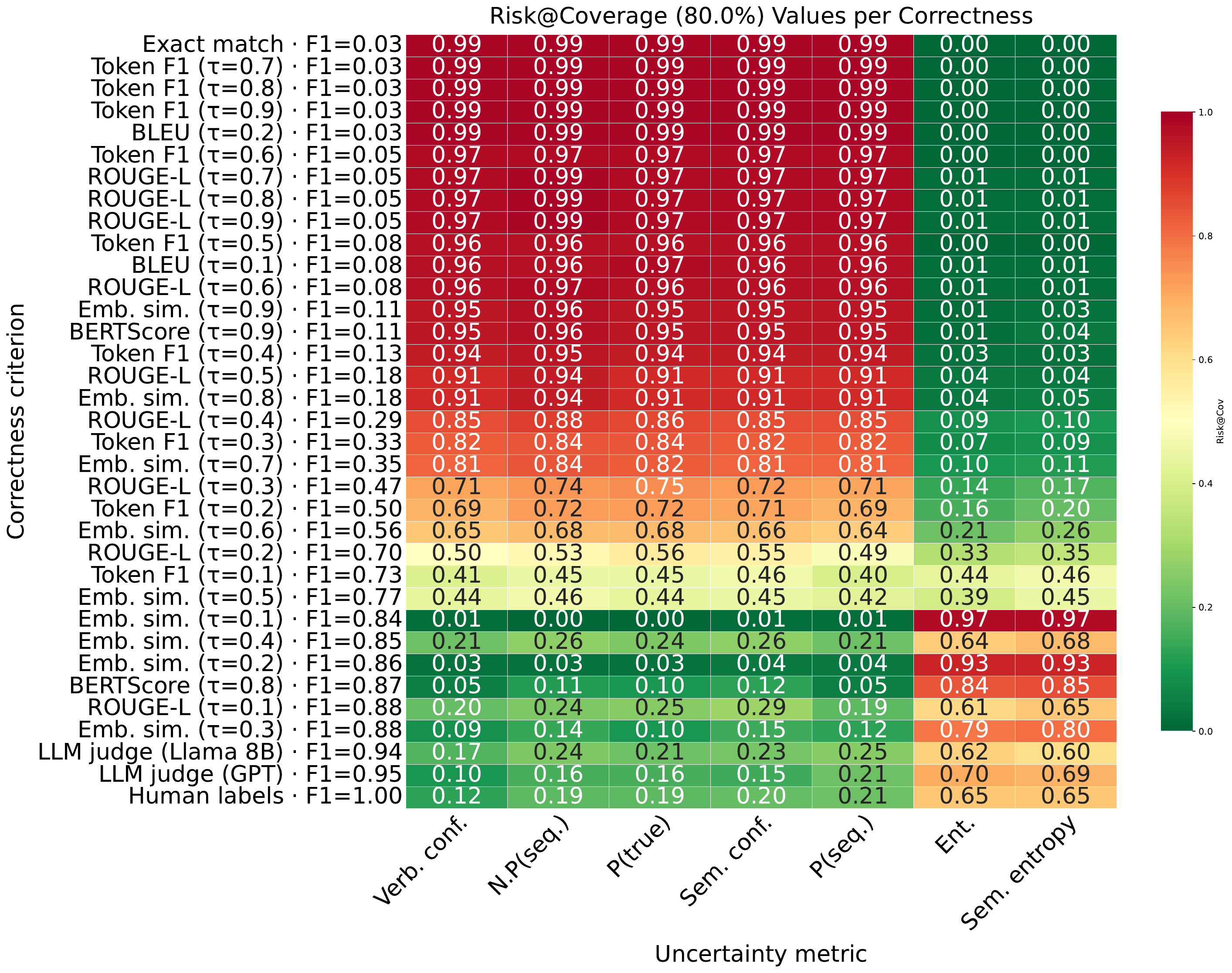}
    \end{subfigure}

    \vspace{0.75em}

    \begin{subfigure}[t]{0.48\textwidth}
        \centering
        \includegraphics[width=\linewidth]{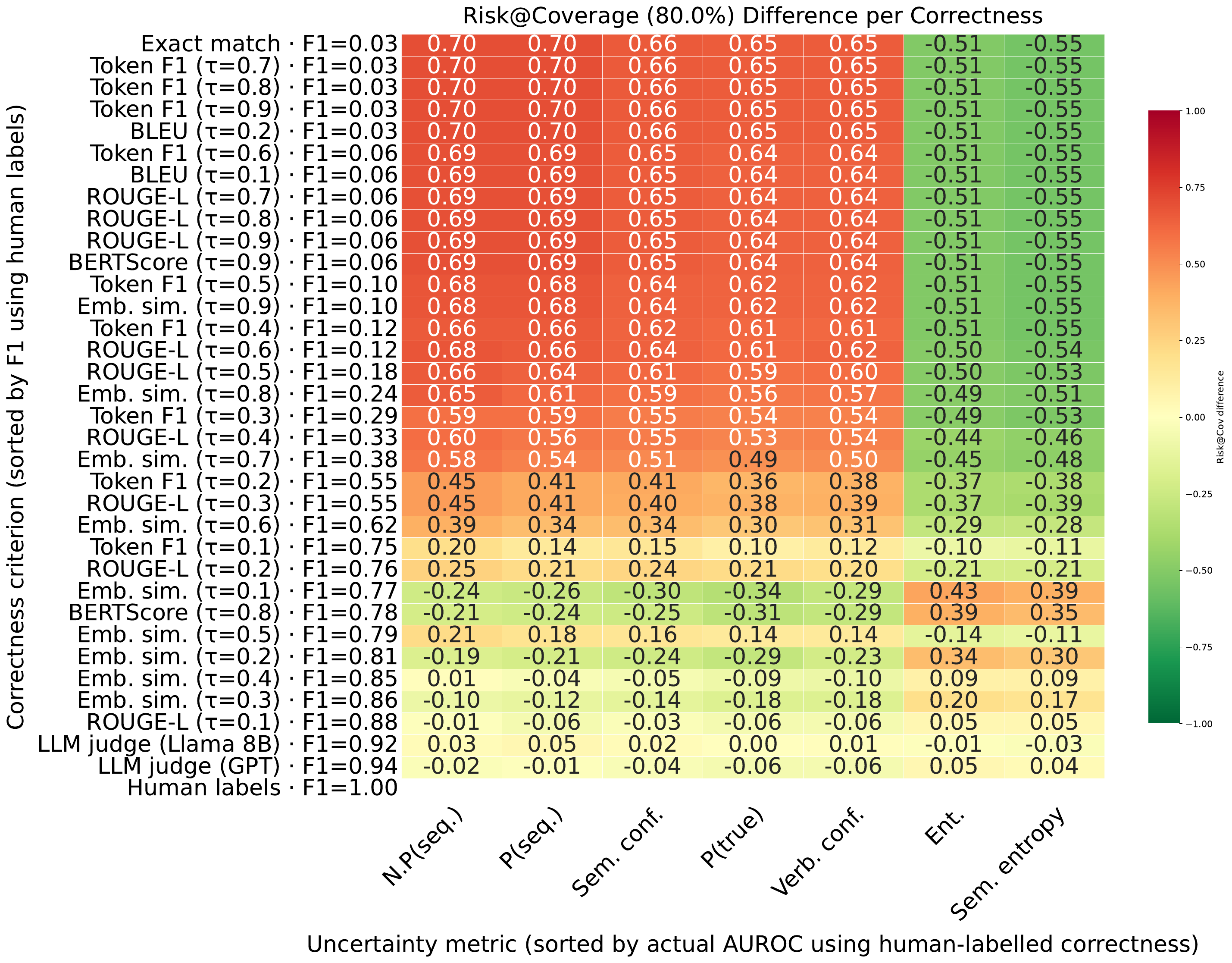}
    \end{subfigure}
    \hfill
    \begin{subfigure}[t]{0.48\textwidth}
        \centering
        \includegraphics[width=\linewidth]{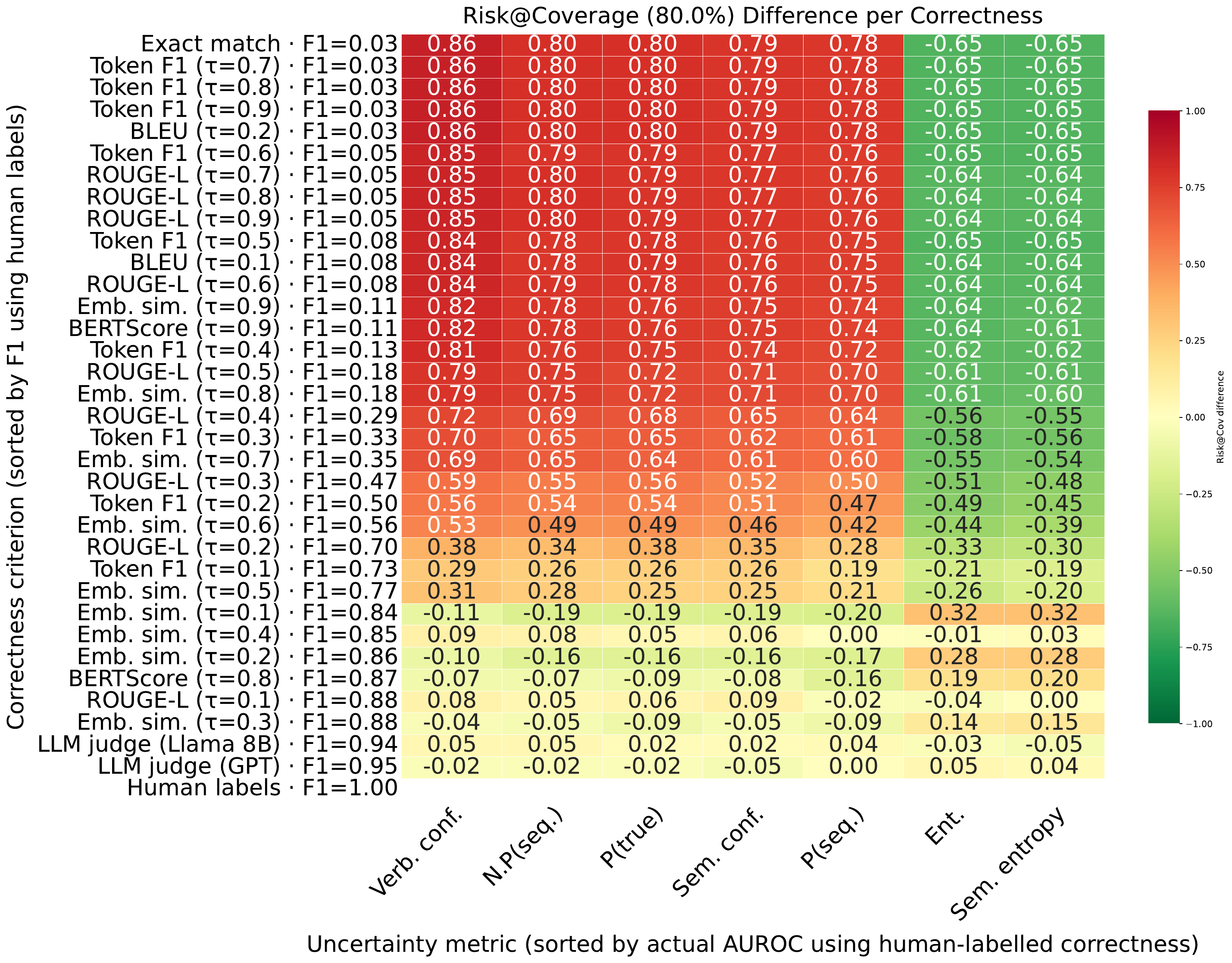}
    \end{subfigure}

    \begin{subfigure}[t]{0.48\textwidth}
        \centering
        \includegraphics[width=\linewidth]{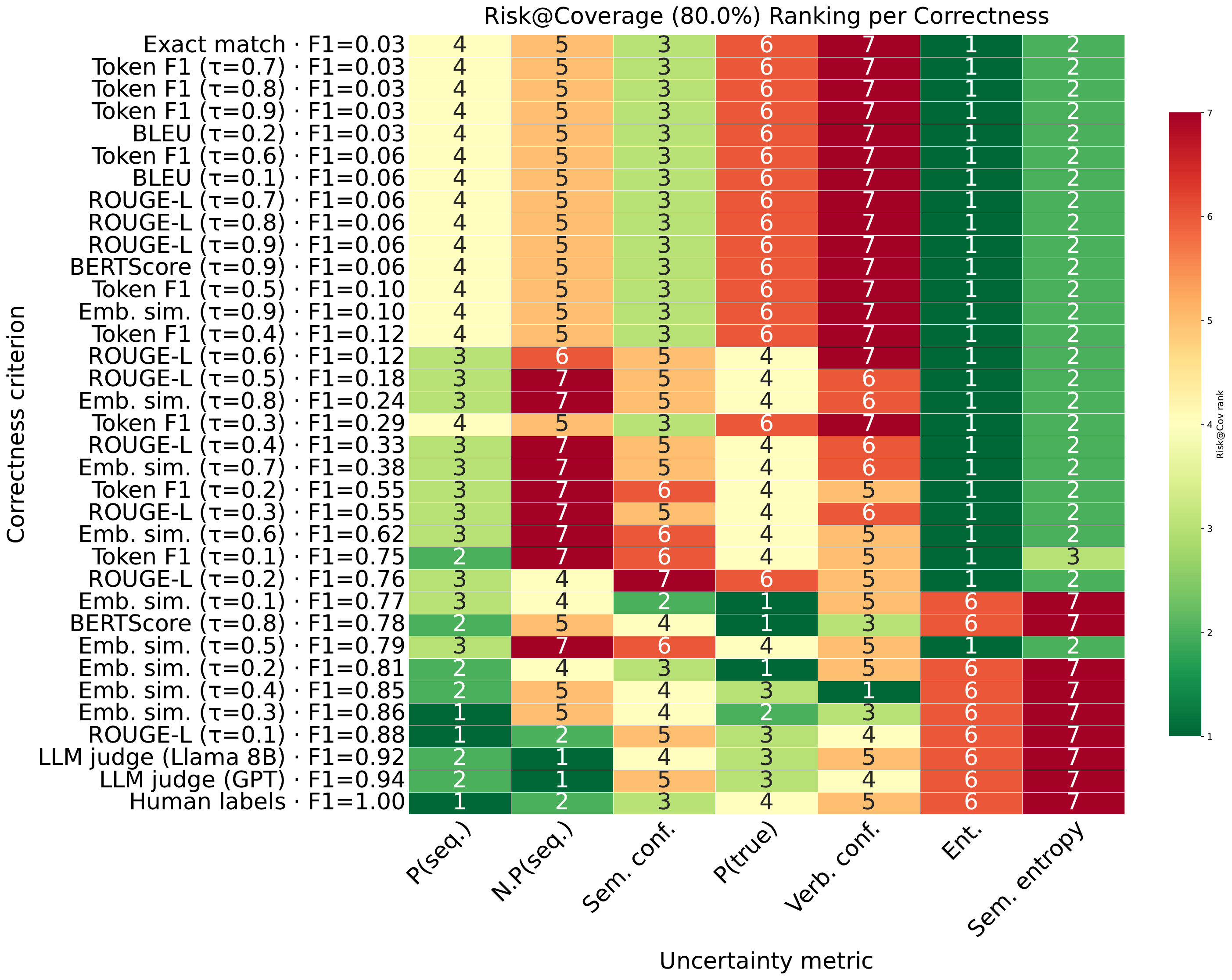}
        \caption{Llama 3B}
    \end{subfigure}
    \hfill
    \begin{subfigure}[t]{0.48\textwidth}
        \centering
        \includegraphics[width=\linewidth]{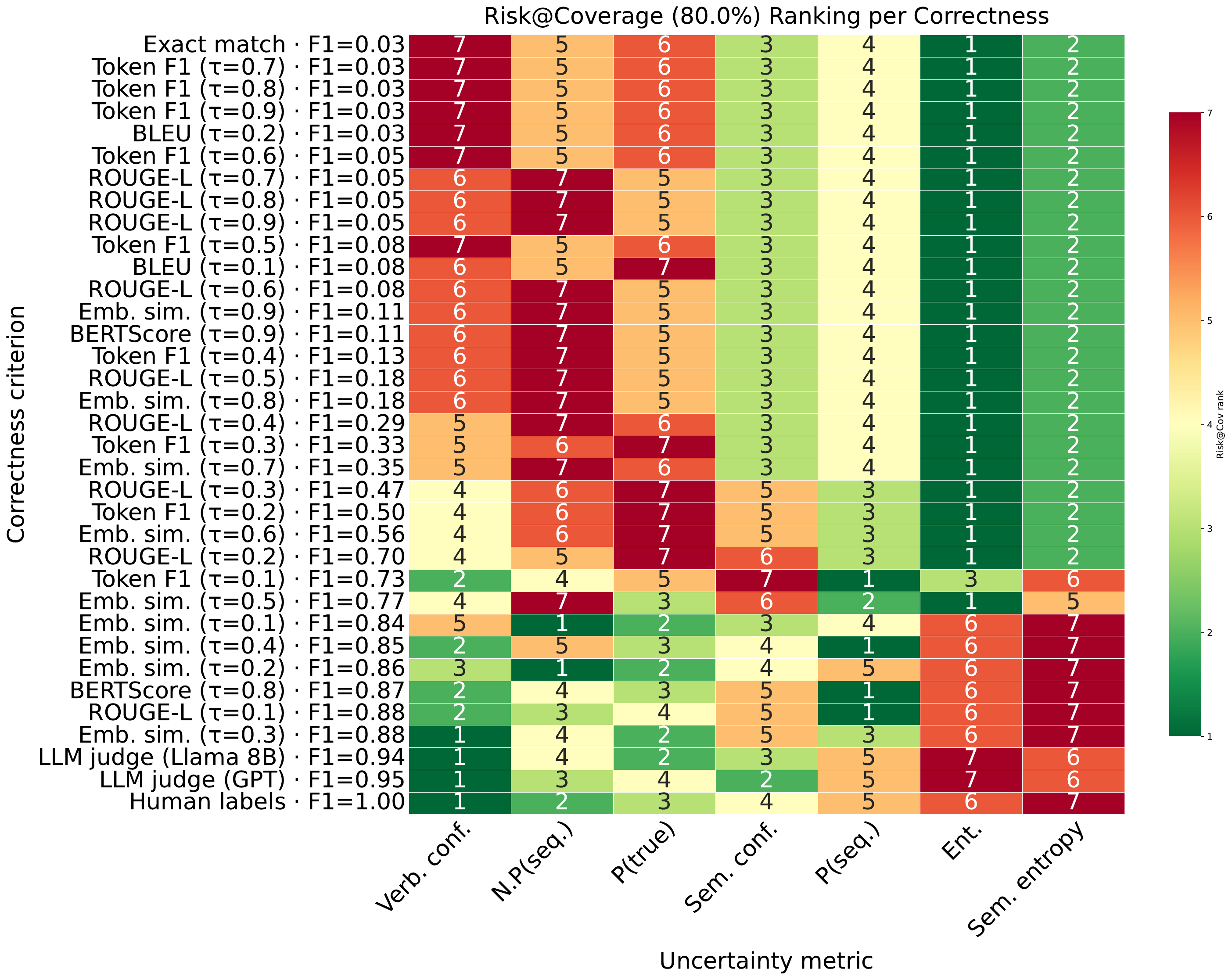}
        \caption{Llama 8B}
    \end{subfigure}

    \caption{Top row shows risk at fixed coverage values, middle row shows differences and bottom row rankings for various uncertainty quantifiers. }
    \label{fig:llama-triviaqa-risk_at_cov}
\end{figure*}

\begin{figure*}[t]
    \centering

    \begin{subfigure}[t]{0.48\textwidth}
        \centering
        \includegraphics[width=\linewidth]{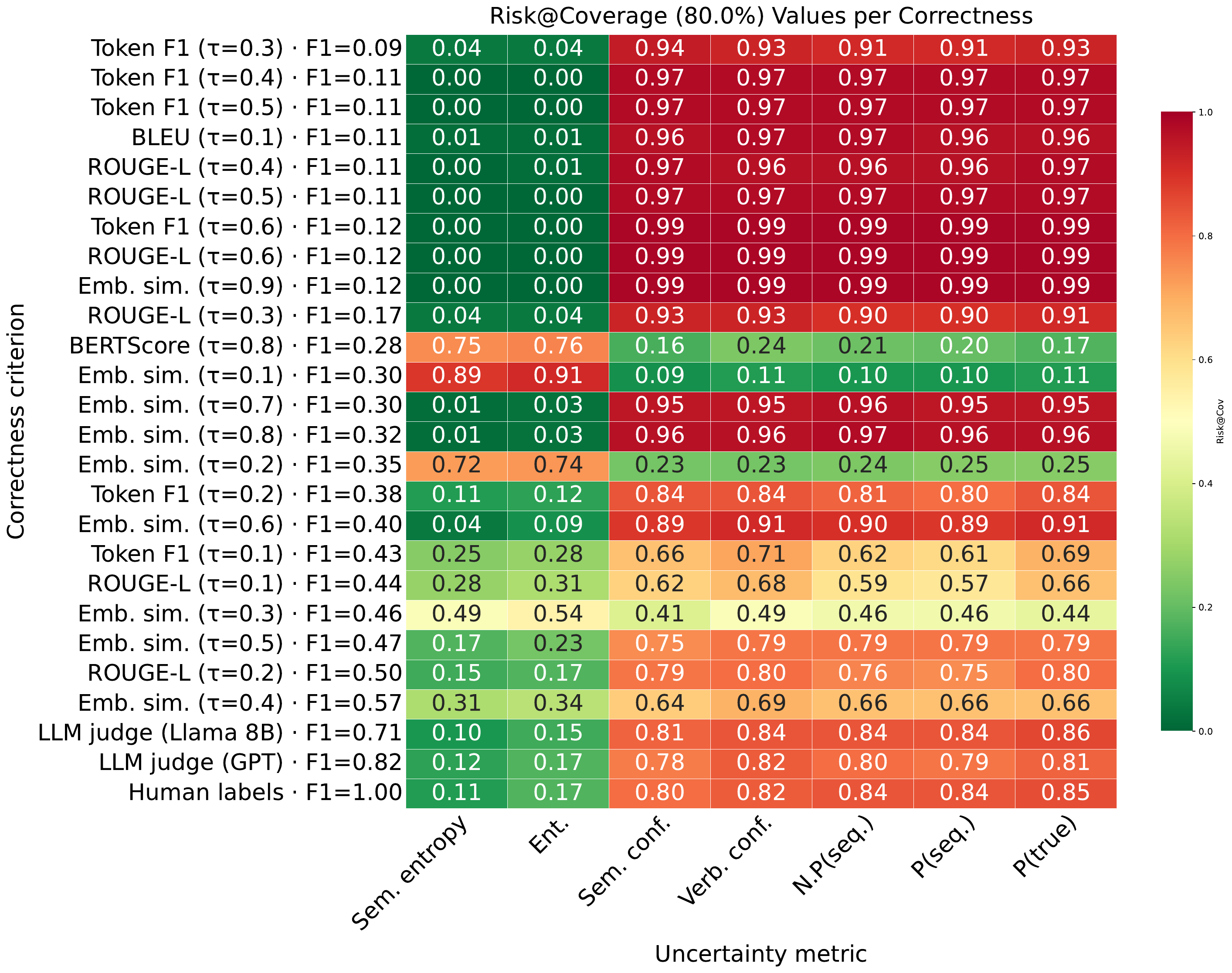}
    \end{subfigure}
    \hfill
    \begin{subfigure}[t]{0.48\textwidth}
        \centering
        \includegraphics[width=\linewidth]{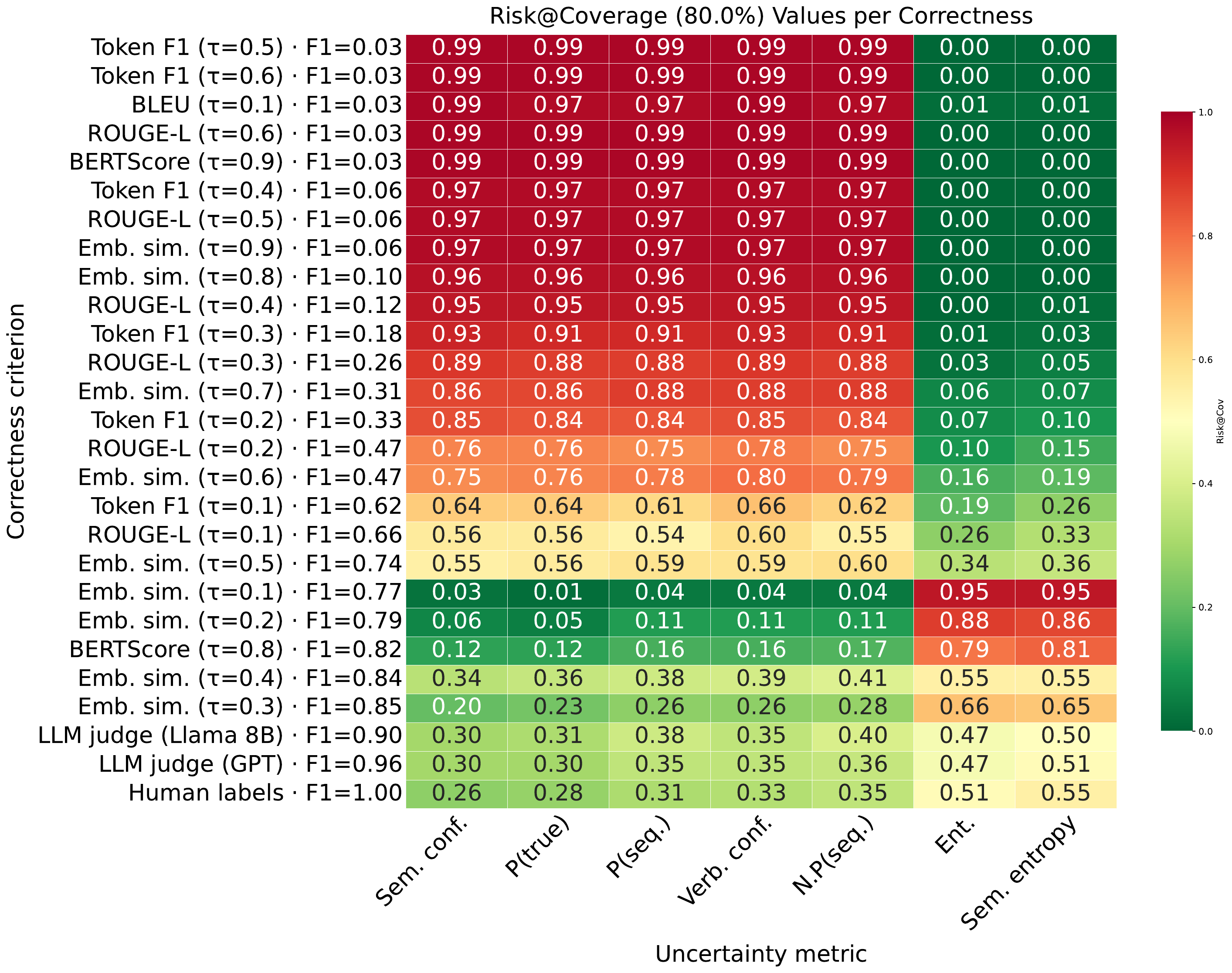}
    \end{subfigure}

    \vspace{0.75em}

    \begin{subfigure}[t]{0.48\textwidth}
        \centering
        \includegraphics[width=\linewidth]{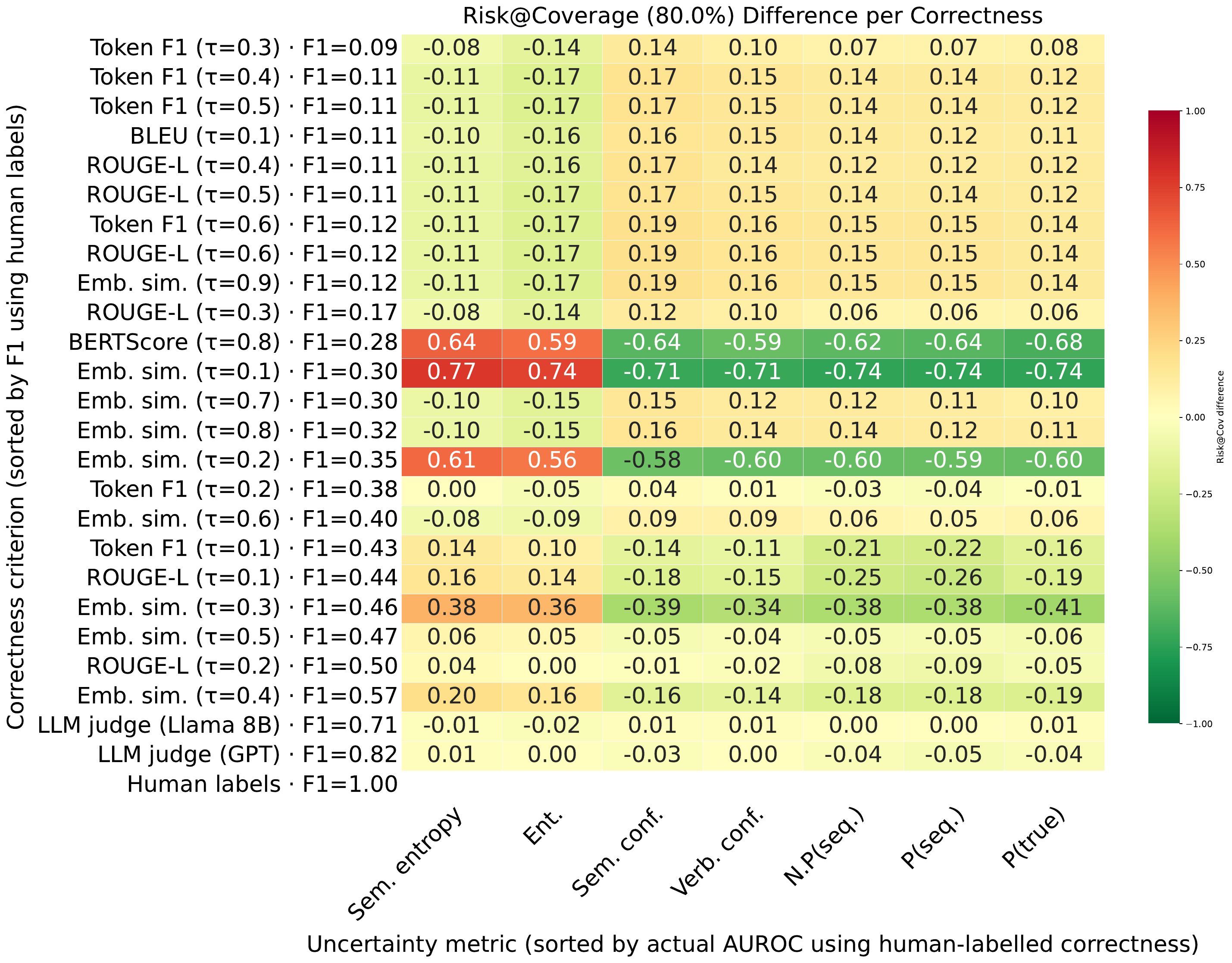}
    \end{subfigure}
    \hfill
    \begin{subfigure}[t]{0.48\textwidth}
        \centering
        \includegraphics[width=\linewidth]{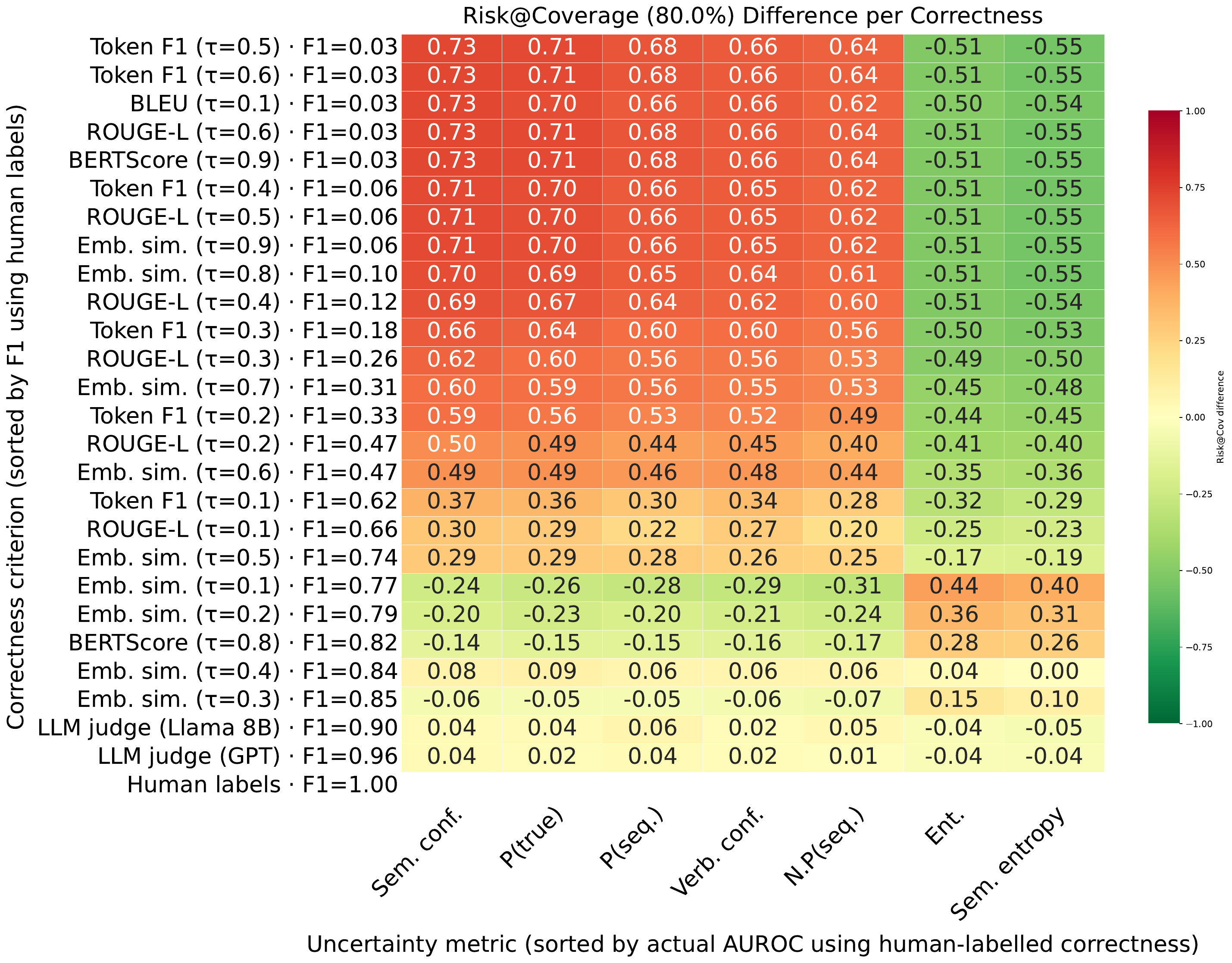}
    \end{subfigure}

    \begin{subfigure}[t]{0.48\textwidth}
        \centering
        \includegraphics[width=\linewidth]{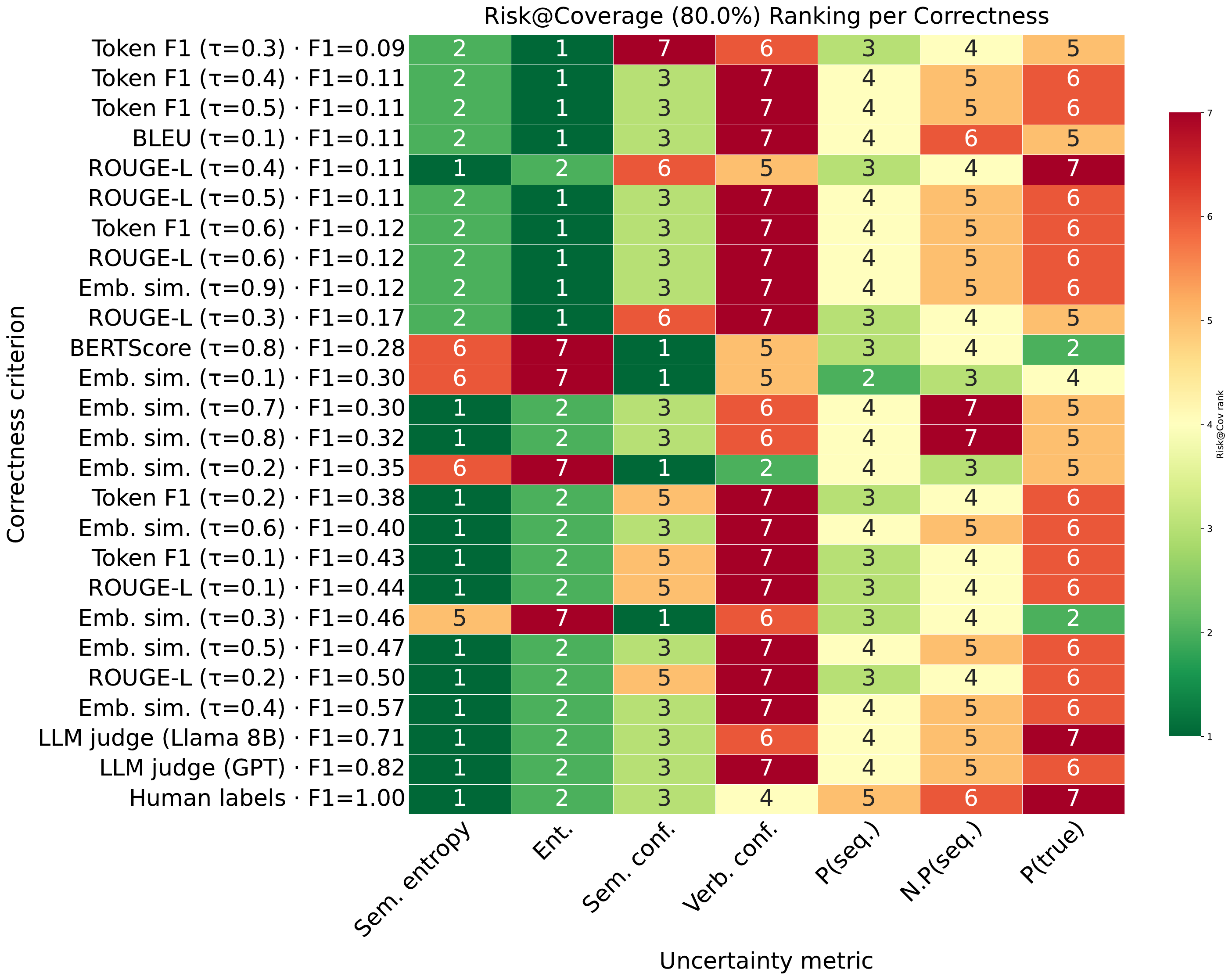}
        \caption{Qwen 0.5B}
    \end{subfigure}
    \hfill
    \begin{subfigure}[t]{0.48\textwidth}
        \centering
        \includegraphics[width=\linewidth]{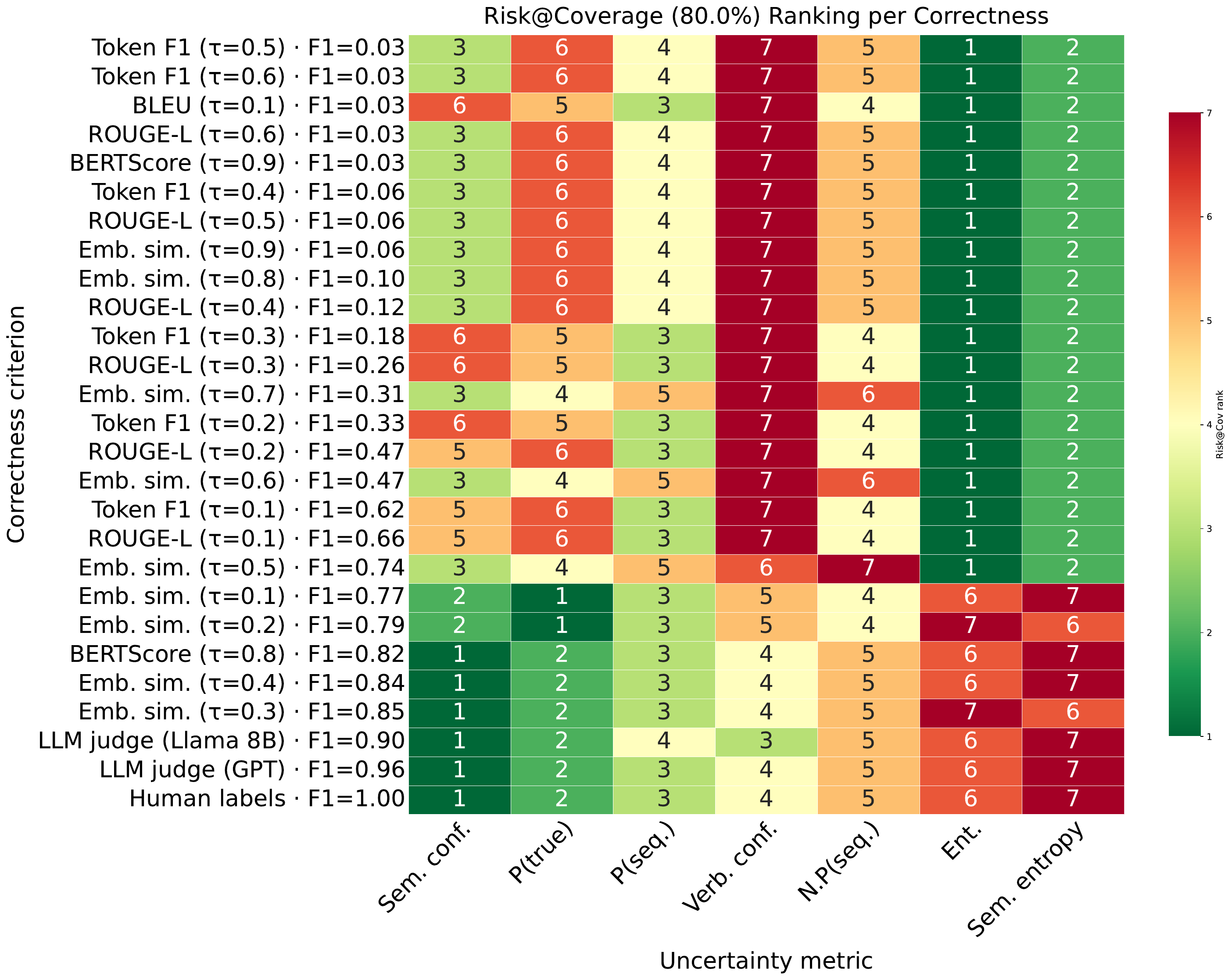}
        \caption{Qwen 7B}
    \end{subfigure}

    \caption{Top row shows risk at fixed coverage values, middle row shows differences and bottom row rankings for various uncertainty quantifiers. }
    \label{fig:qwen-triviaqa-risk_at_cov}
\end{figure*}

\begin{figure*}[t]
    \centering

    \begin{subfigure}[t]{0.48\textwidth}
        \centering
        \includegraphics[width=\linewidth]{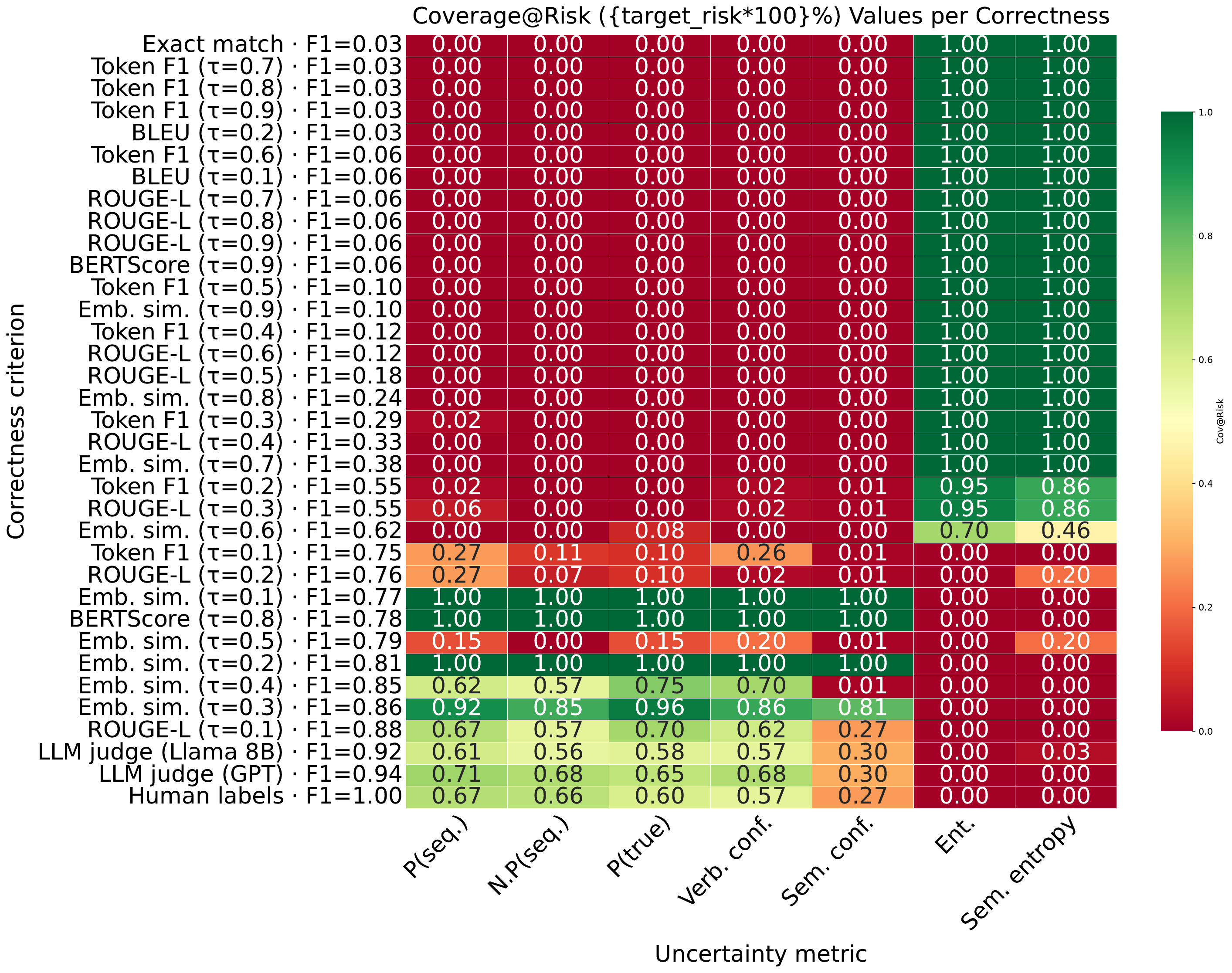}
    \end{subfigure}
    \hfill
    \begin{subfigure}[t]{0.48\textwidth}
        \centering
        \includegraphics[width=\linewidth]{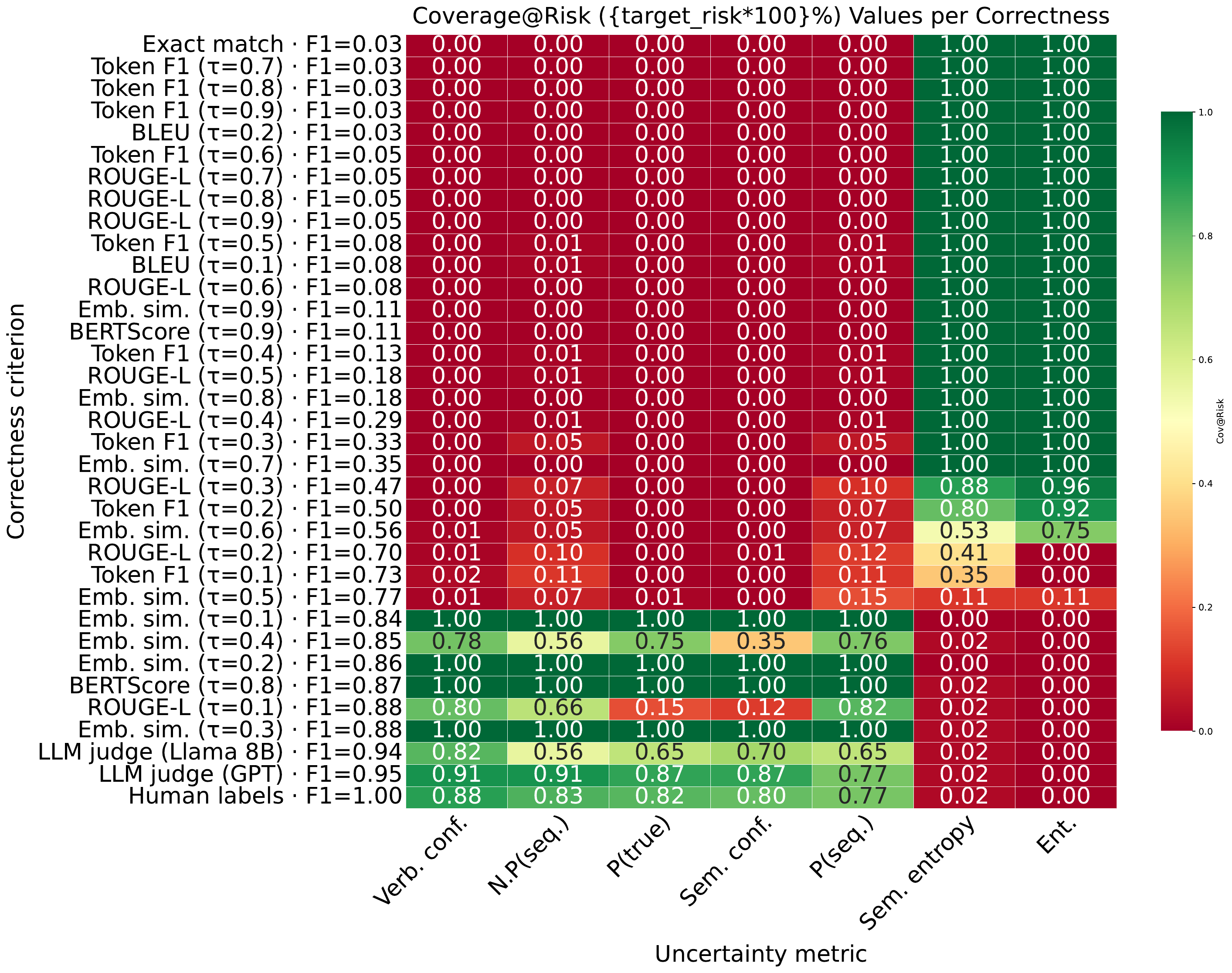}
    \end{subfigure}

    \vspace{0.75em}

    \begin{subfigure}[t]{0.48\textwidth}
        \centering
        \includegraphics[width=\linewidth]{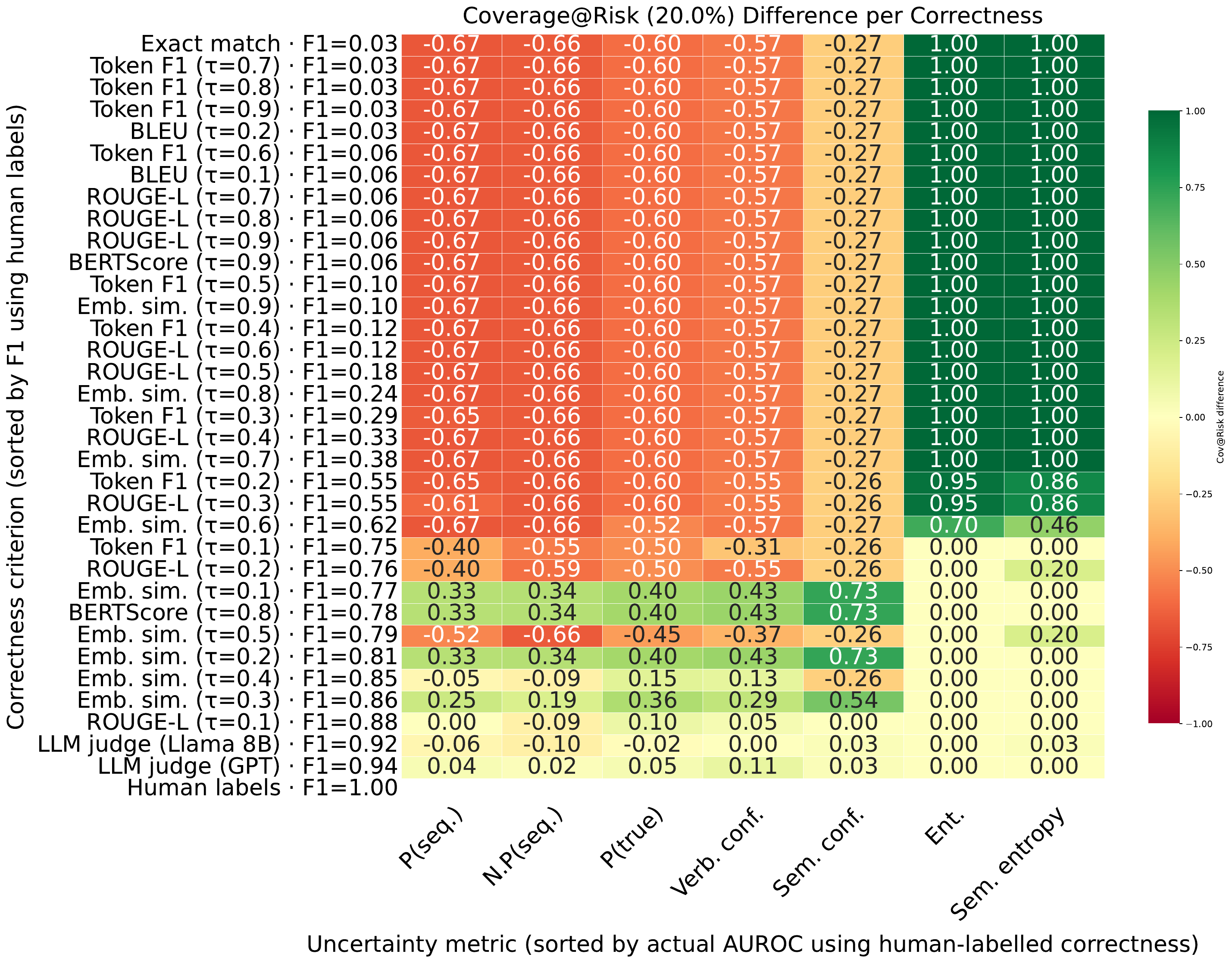}
    \end{subfigure}
    \hfill
    \begin{subfigure}[t]{0.48\textwidth}
        \centering
        \includegraphics[width=\linewidth]{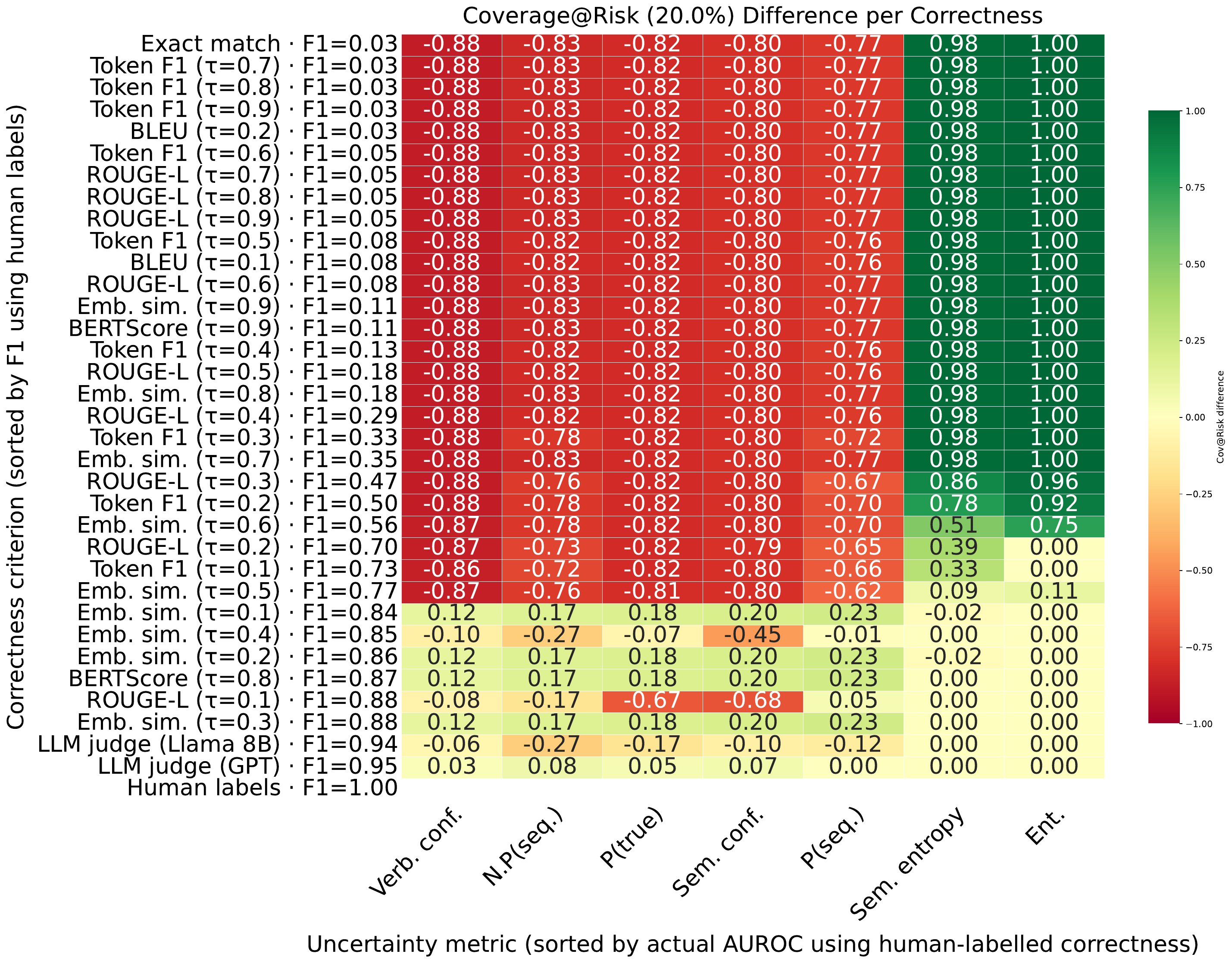}
    \end{subfigure}

    \begin{subfigure}[t]{0.48\textwidth}
        \centering
        \includegraphics[width=\linewidth]{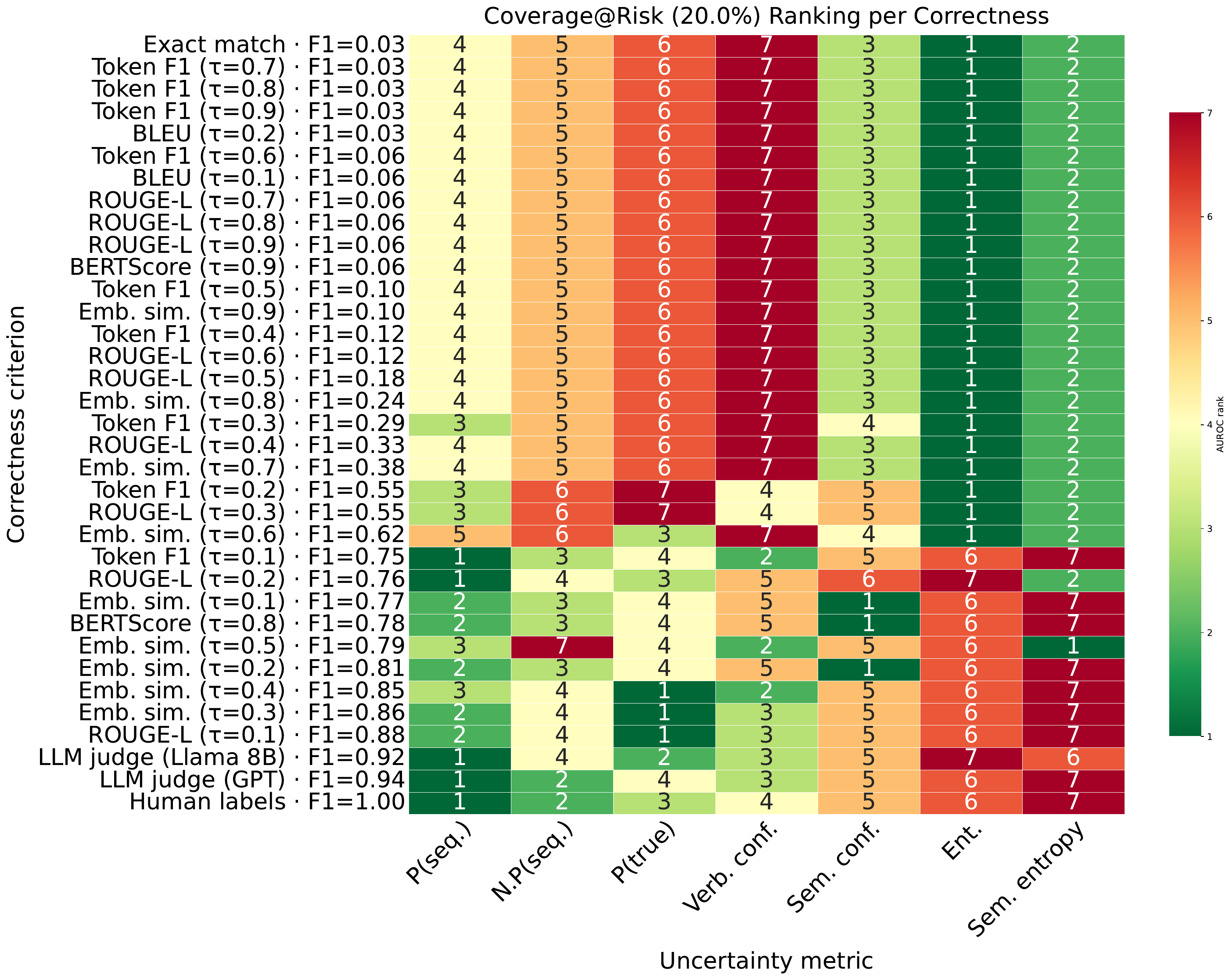}
        \caption{Llama 3B}
    \end{subfigure}
    \hfill
    \begin{subfigure}[t]{0.48\textwidth}
        \centering
        \includegraphics[width=\linewidth]{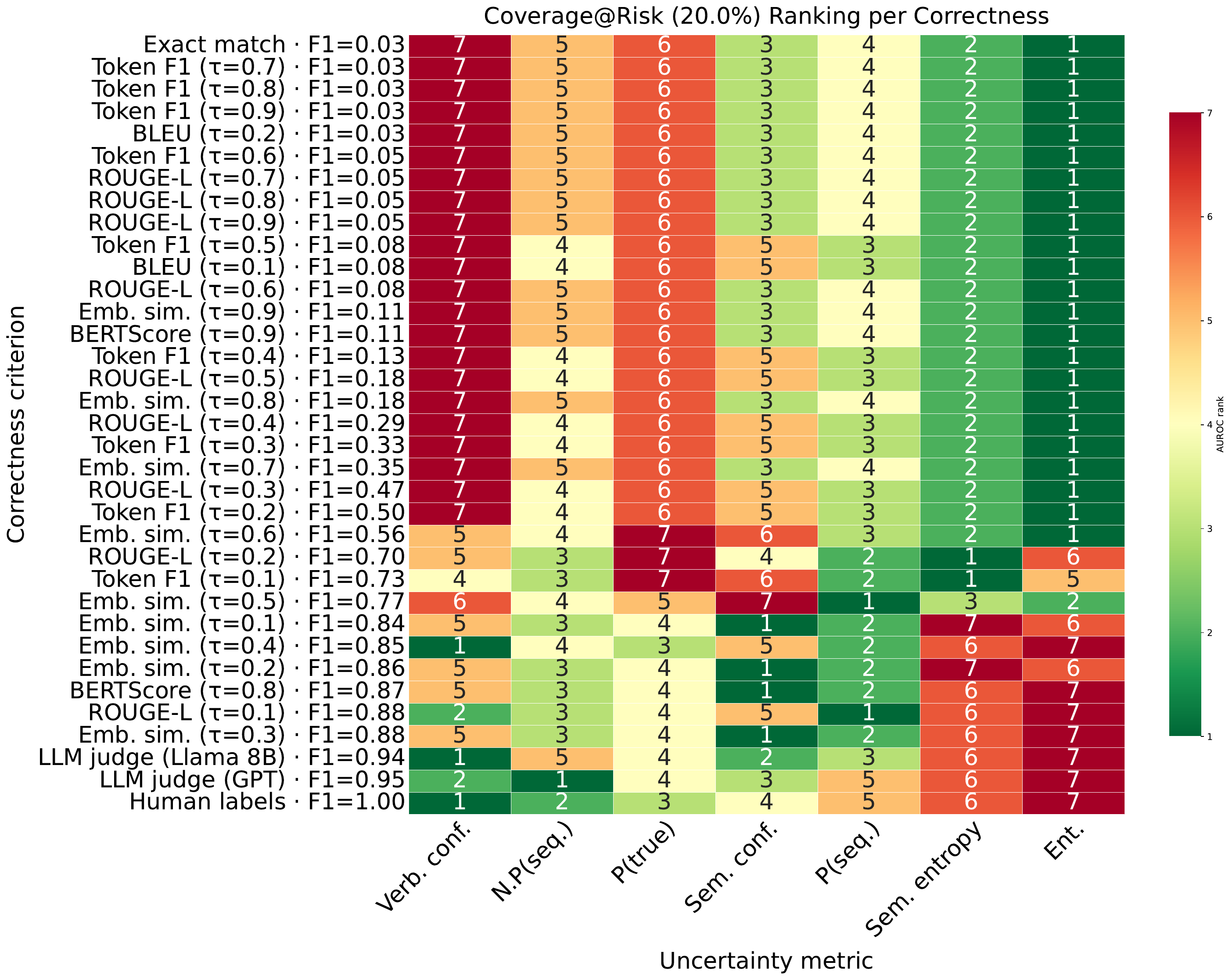}
        \caption{Llama 8B}
    \end{subfigure}

    \caption{Top row shows Coverage at fixed risk values, middle row shows differences and bottom row rankings for various uncertainty quantifiers. }
    \label{fig:llama-triviaqa-cov_at_risk}
\end{figure*}

\begin{figure*}[t]
    \centering

    \begin{subfigure}[t]{0.48\textwidth}
        \centering
        \includegraphics[width=\linewidth]{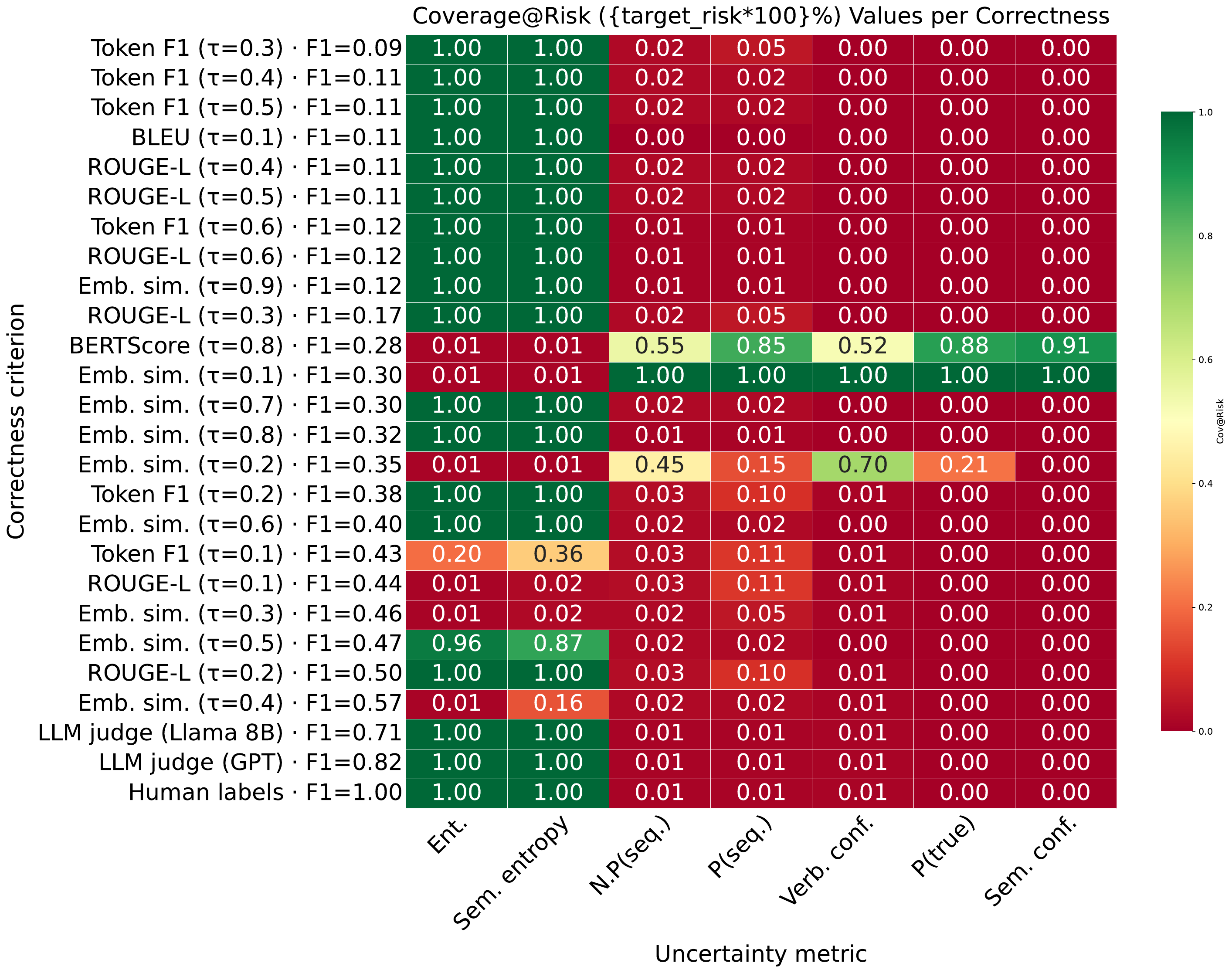}
    \end{subfigure}
    \hfill
    \begin{subfigure}[t]{0.48\textwidth}
        \centering
        \includegraphics[width=\linewidth]{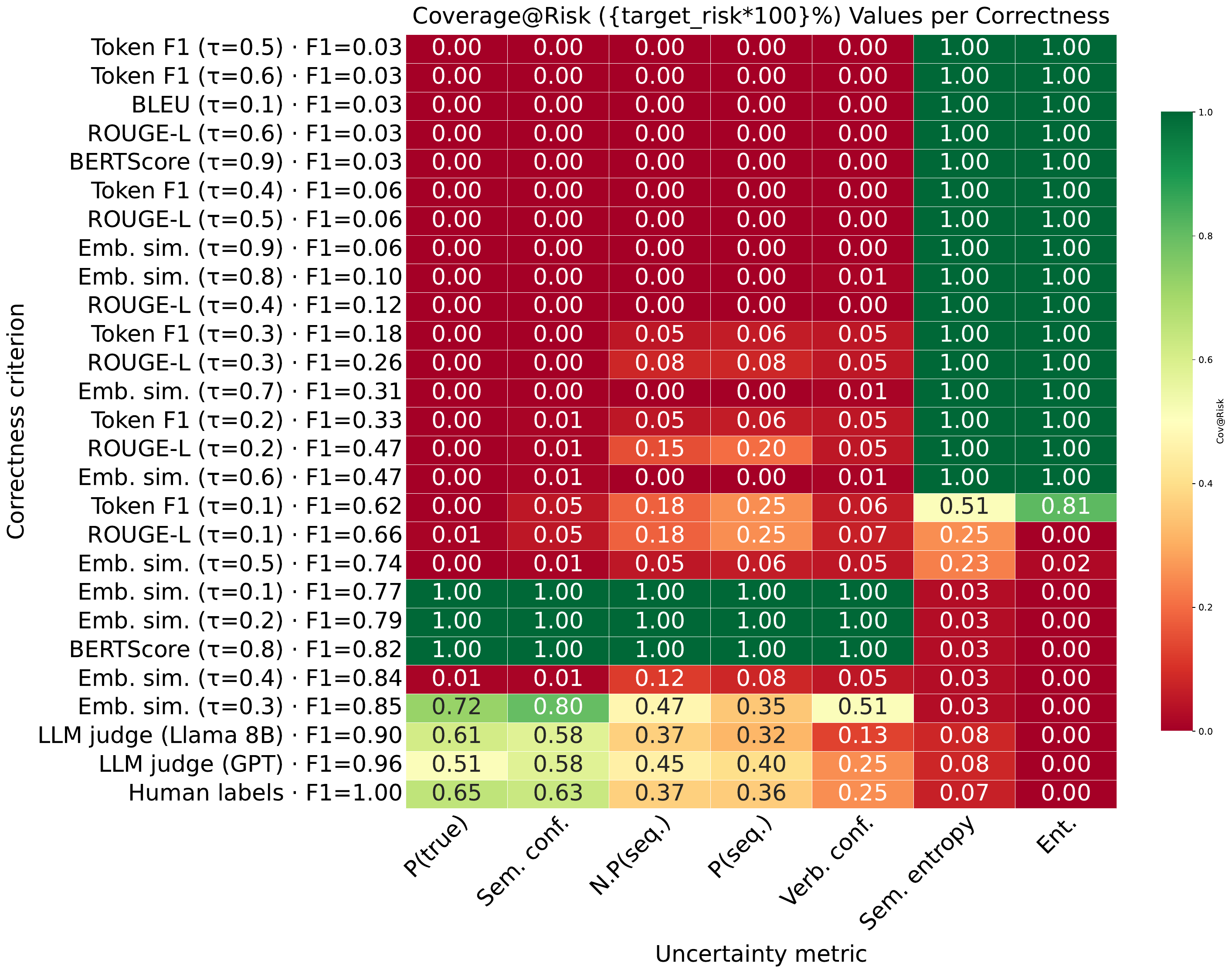}
    \end{subfigure}

    \vspace{0.75em}

    \begin{subfigure}[t]{0.48\textwidth}
        \centering
        \includegraphics[width=\linewidth]{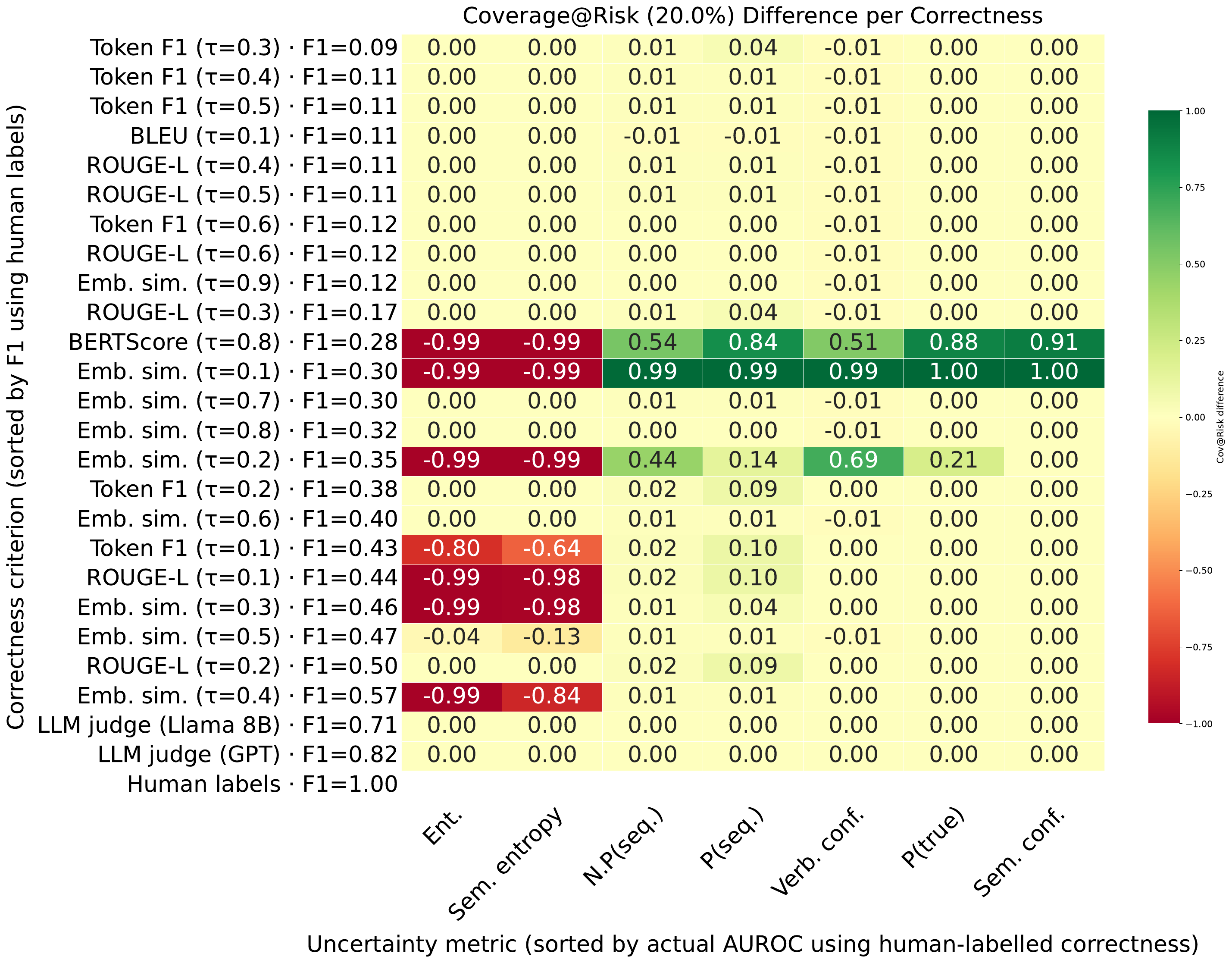}
    \end{subfigure}
    \hfill
    \begin{subfigure}[t]{0.48\textwidth}
        \centering
        \includegraphics[width=\linewidth]{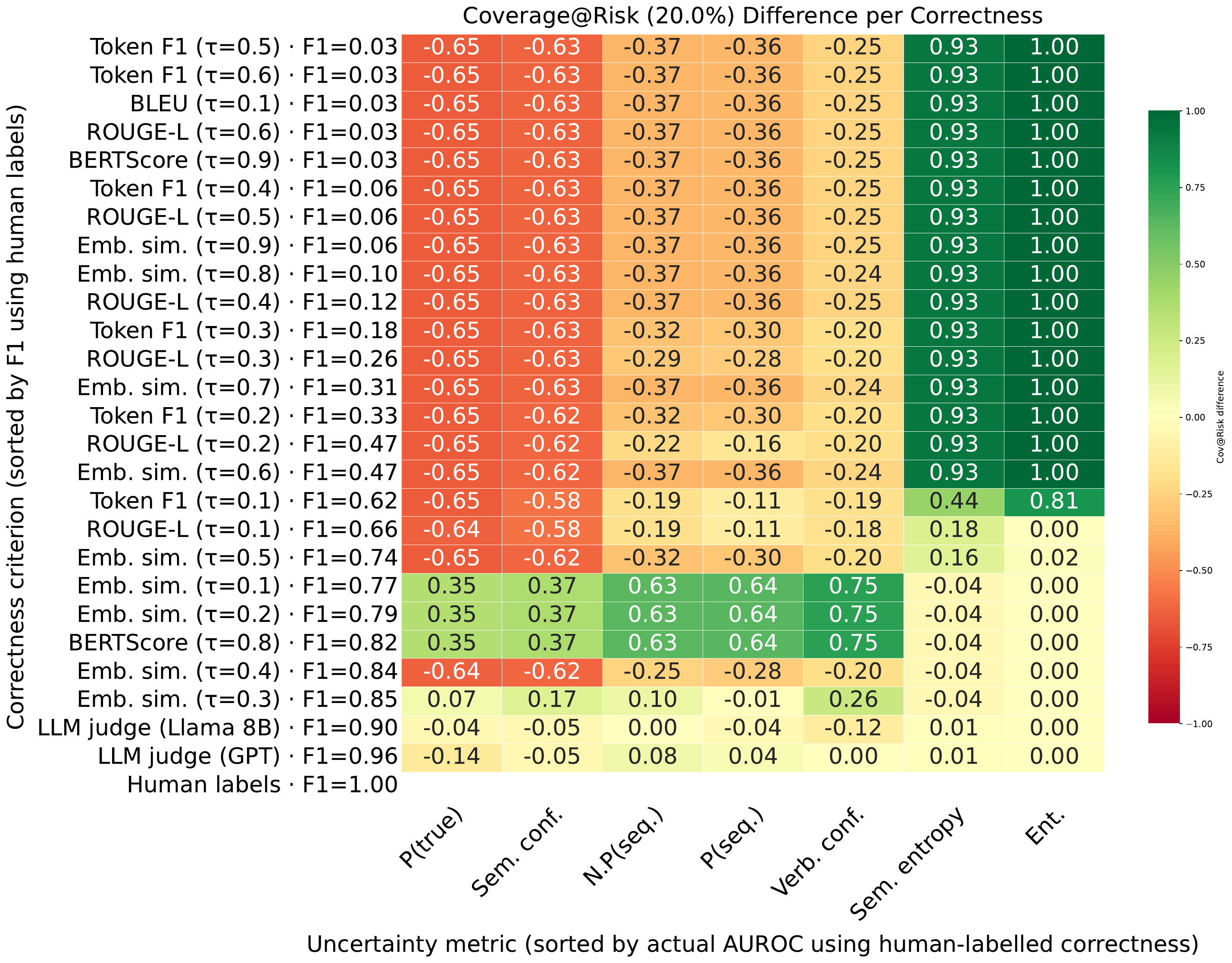}
    \end{subfigure}

    \begin{subfigure}[t]{0.48\textwidth}
        \centering
        \includegraphics[width=\linewidth]{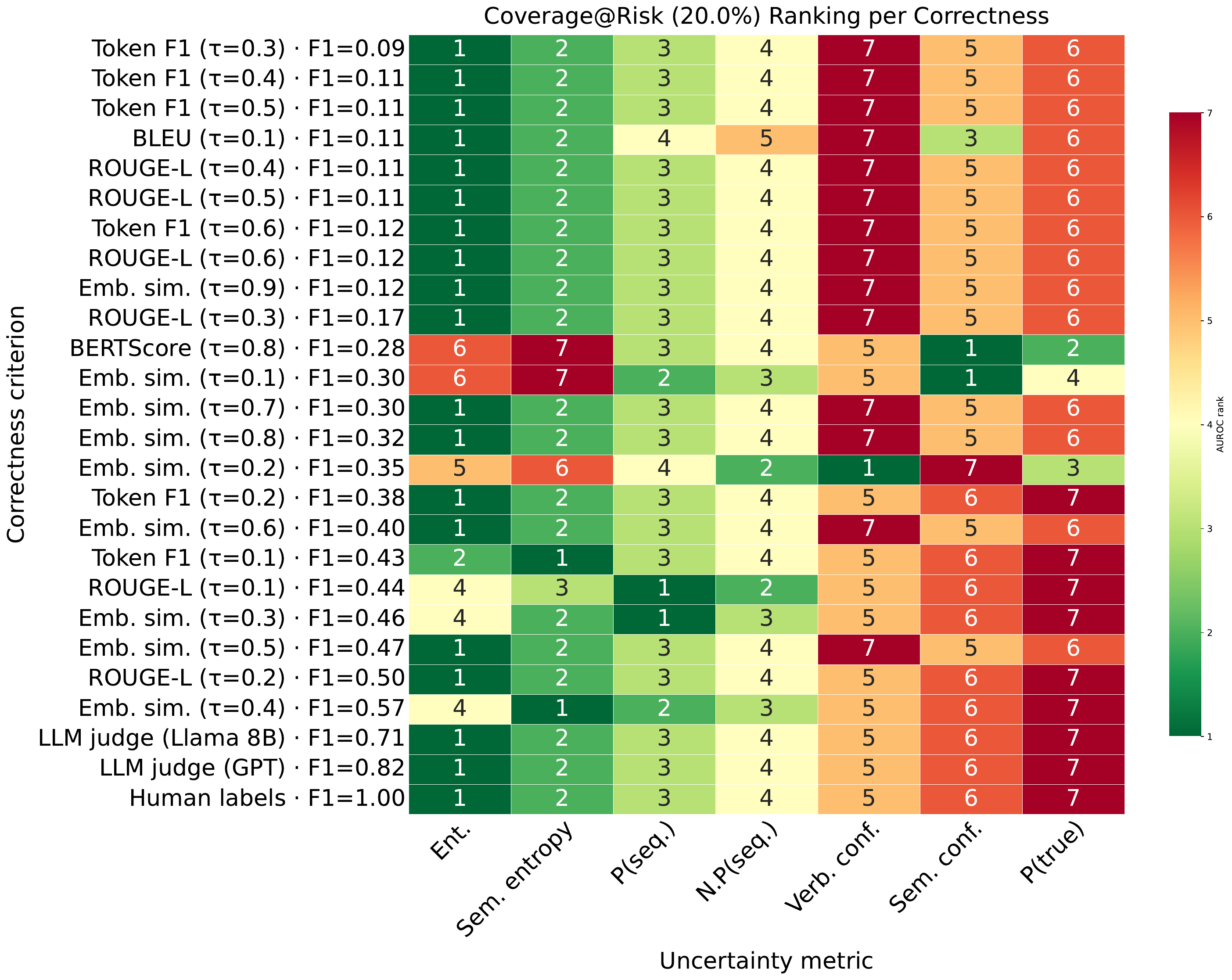}
        \caption{Qwen 0.5B}
    \end{subfigure}
    \hfill
    \begin{subfigure}[t]{0.48\textwidth}
        \centering
        \includegraphics[width=\linewidth]{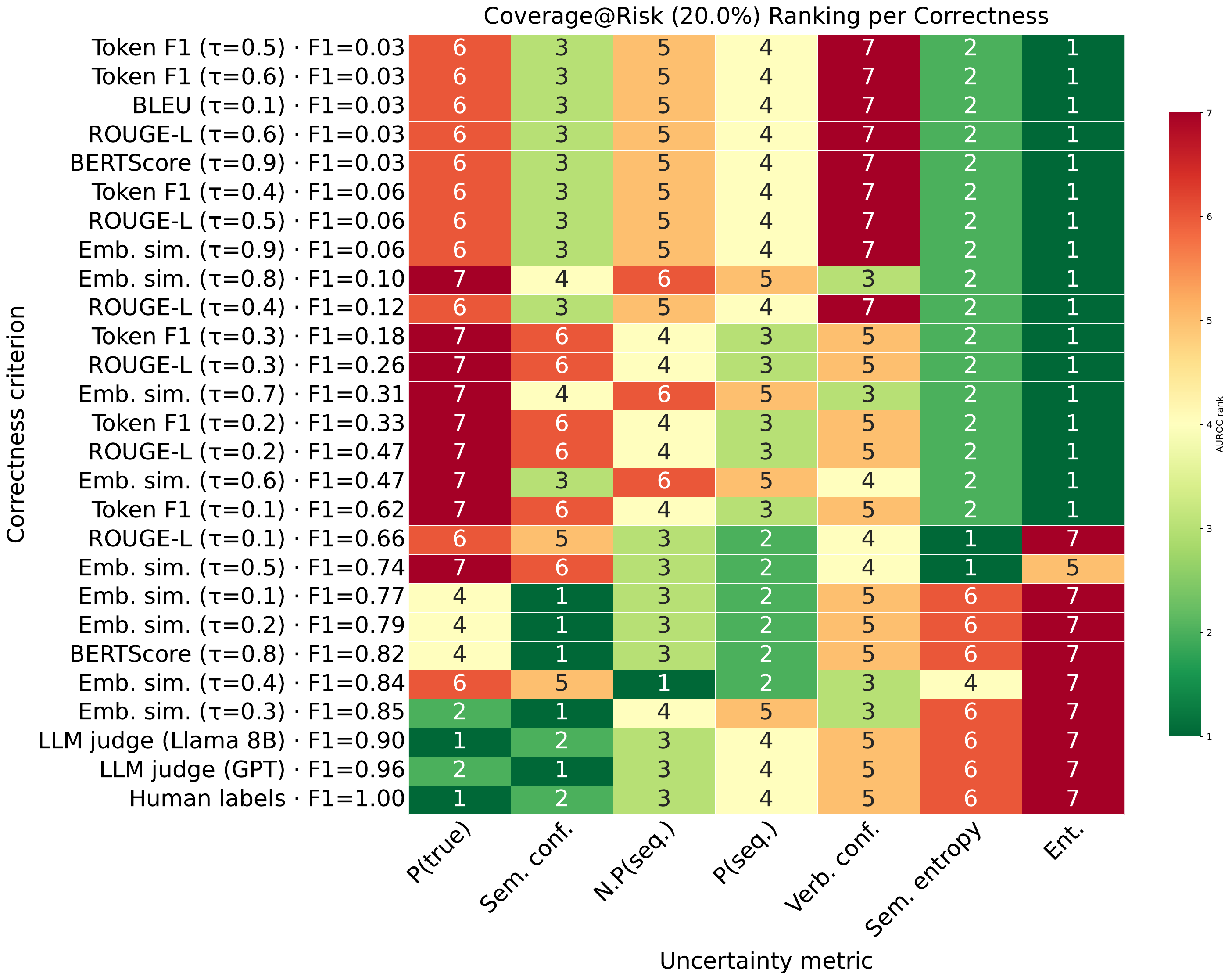}
        \caption{Qwen 7B}
    \end{subfigure}

    \caption{Top row shows Coverage at fixed risk values, middle row shows differences and bottom row rankings for various uncertainty quantifiers. }
    \label{fig:qwen-triviaqa-cov_at_risk}
\end{figure*}

Beyond AUROC, other metrics can measure performance trade-offs in SP. We perform a similar analysis for various automated judges and UQs. We investigate the prompts from TriviaQA and all generators. We assess: coverage at fixed risk ($risk=0.2$) and risk as fixed coverage ($coverage=0.8$). The results are shown in Figures \ref{fig:llama-triviaqa-risk_at_cov} - \ref{fig:qwen-triviaqa-cov_at_risk}. Results generally extend to other SP metrics --- with better quantifiers getting negatively impacted to a higher extent, obscuring conclusions.

\subsection{F1 Distributions}
\label{appendix:f1_dists}
\begin{figure*}[t]
    \centering

    \begin{subfigure}[t]{0.48\textwidth}
        \centering
        \includegraphics[width=\linewidth]{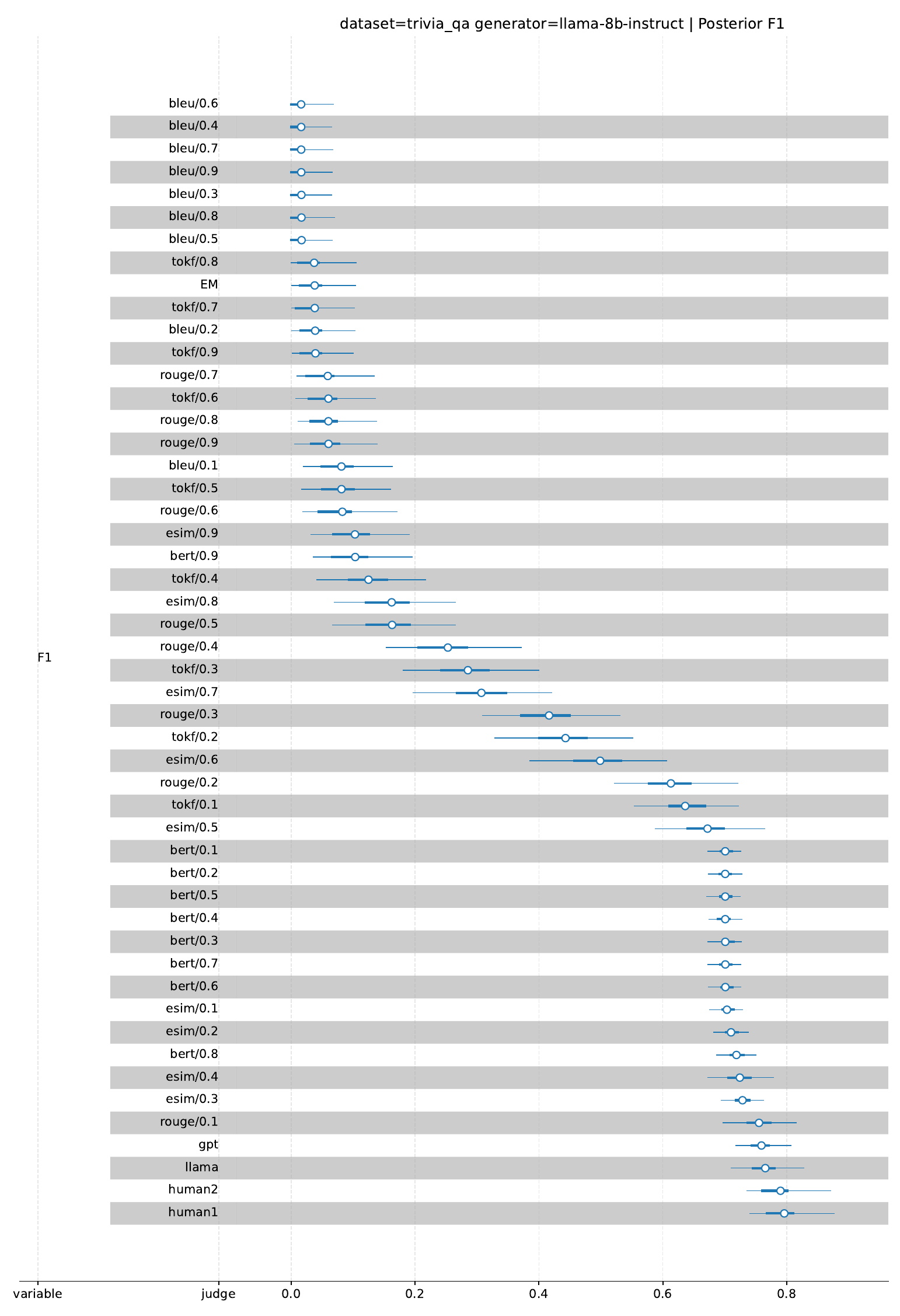}
    \end{subfigure}
    \hfill
    \begin{subfigure}[t]{0.48\textwidth}
        \centering
        \includegraphics[width=\linewidth]{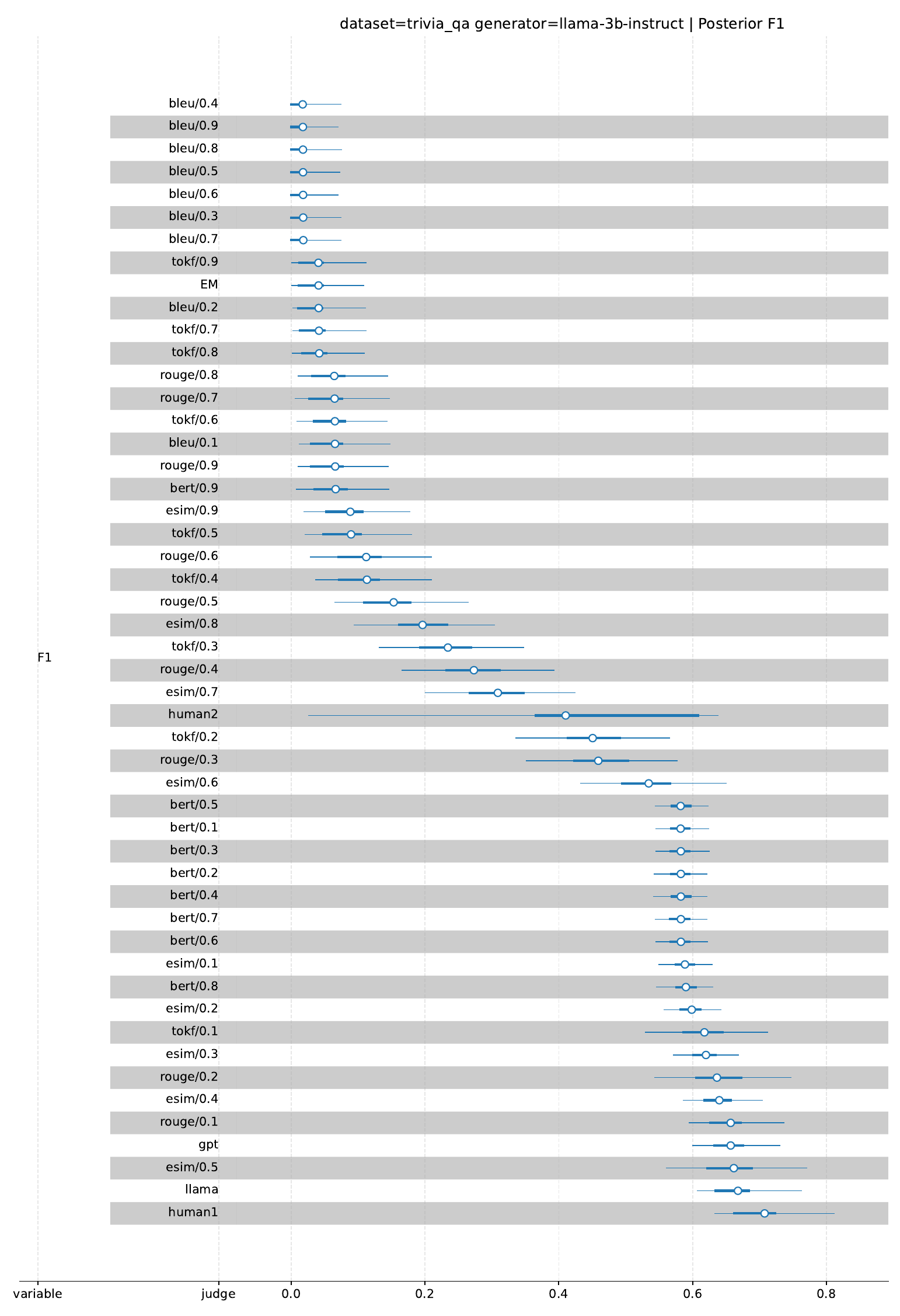}
    \end{subfigure}

    \vspace{0.75em}

    \begin{subfigure}[t]{0.48\textwidth}
        \centering
        \includegraphics[width=\linewidth]{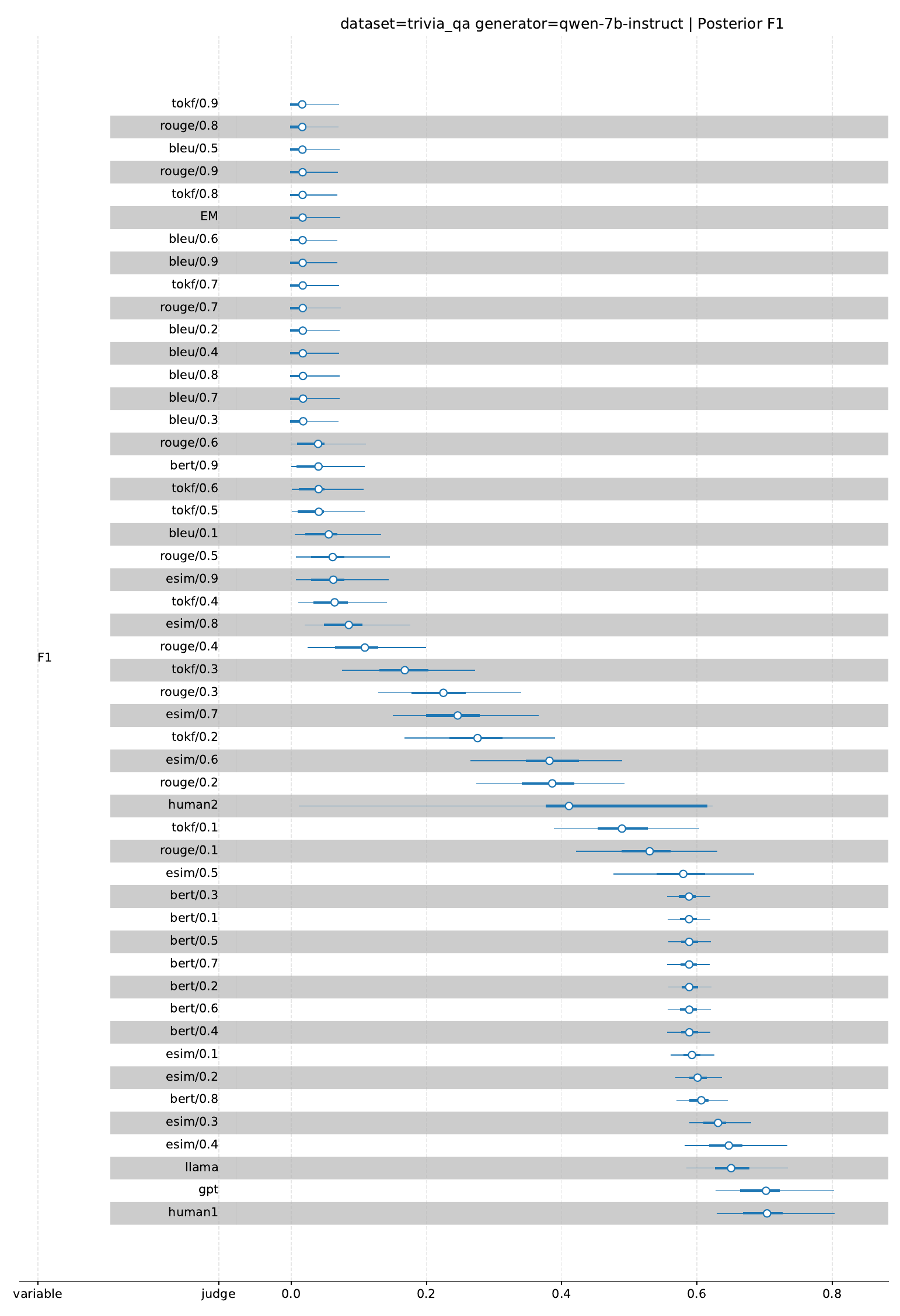}
    \end{subfigure}
    \hfill
    \begin{subfigure}[t]{0.48\textwidth}
        \centering
        \includegraphics[width=\linewidth]{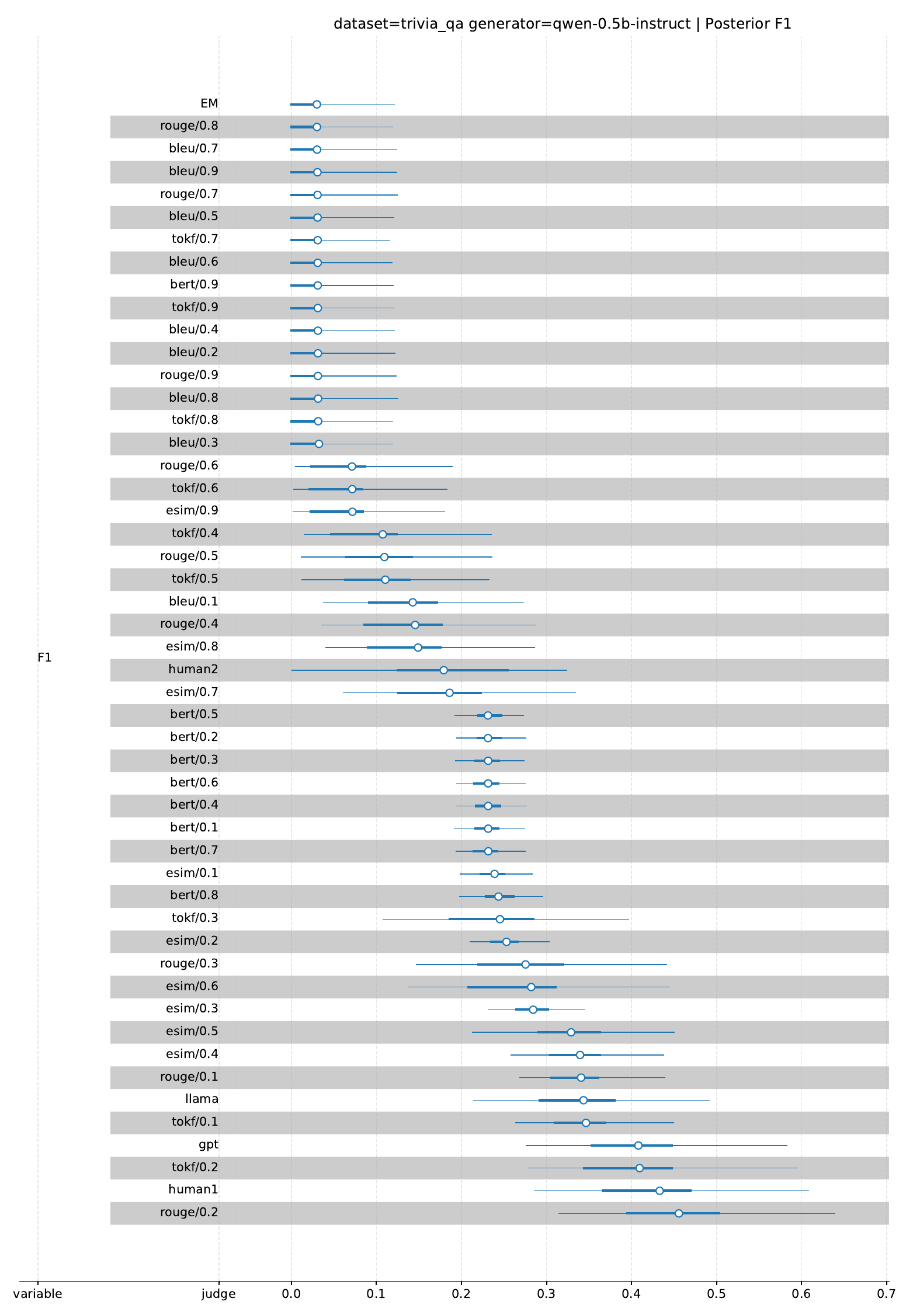}
    \end{subfigure}

    \caption{F1 distributions for TriviaQA dataset and Llama 8B (top left corner), Llama 3B (top right corner), Qwen 7B (bottom left corner) and Qwen 0.5B (bottom right corner).}
    \label{fig:f1_triviaqa}
\end{figure*}

\begin{figure*}[t]
    \centering

    \begin{subfigure}[t]{0.48\textwidth}
        \centering
        \includegraphics[width=\linewidth]{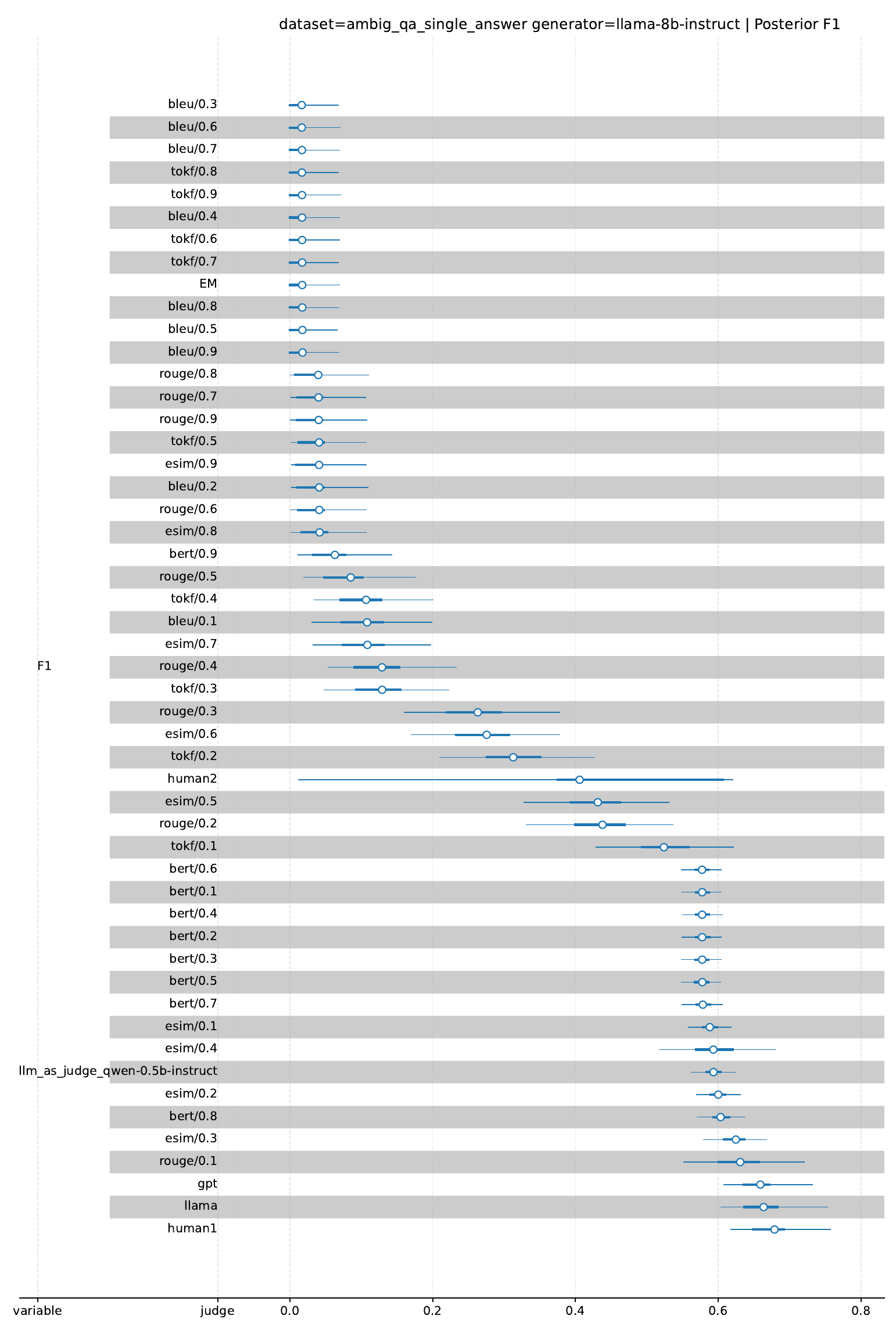}
    \end{subfigure}
    \hfill
    \begin{subfigure}[t]{0.48\textwidth}
        \centering
        \includegraphics[width=\linewidth]{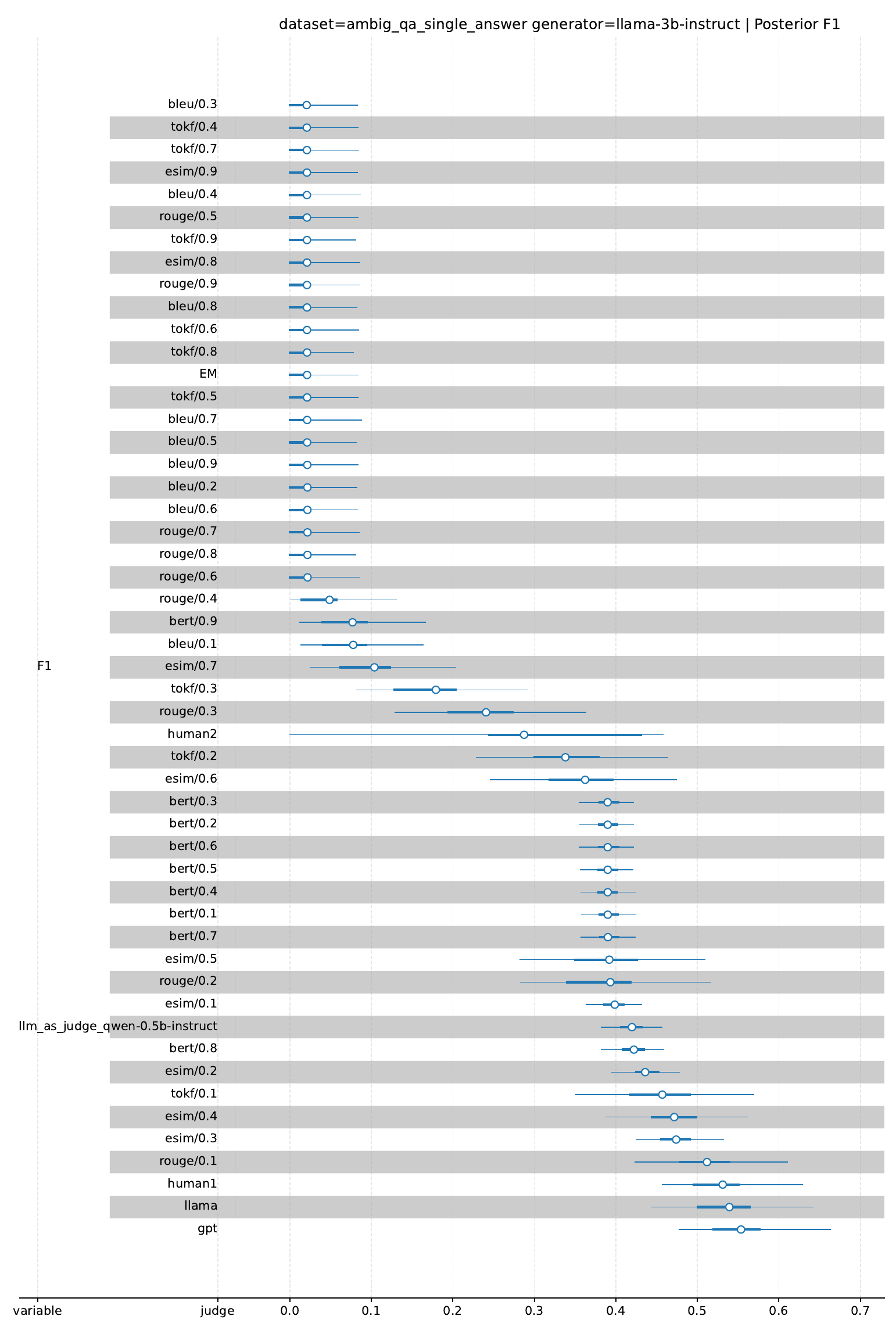}
    \end{subfigure}

    \vspace{0.75em}

    \begin{subfigure}[t]{0.48\textwidth}
        \centering
        \includegraphics[width=\linewidth]{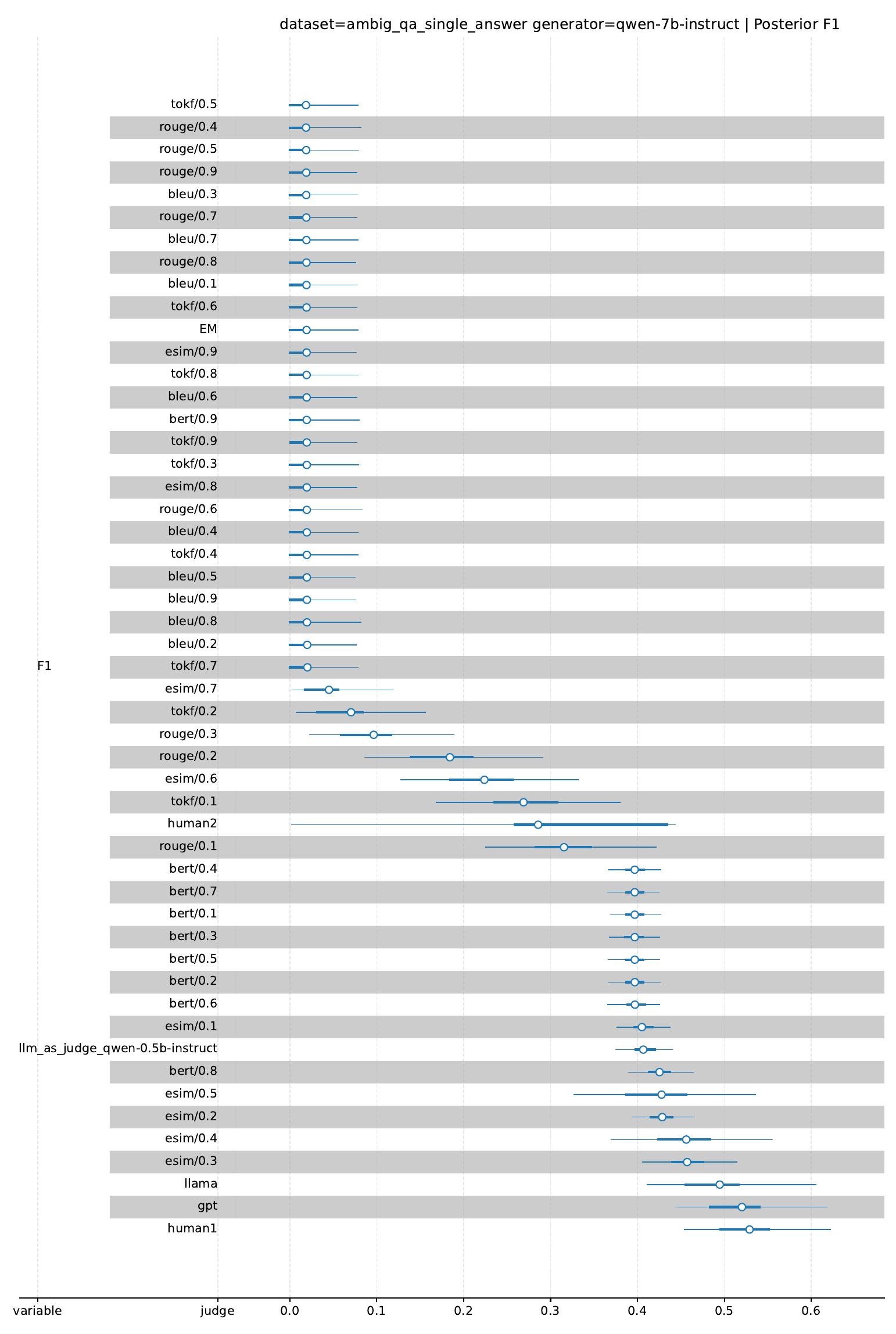}
    \end{subfigure}
    \hfill
    \begin{subfigure}[t]{0.48\textwidth}
        \centering
        \includegraphics[width=\linewidth]{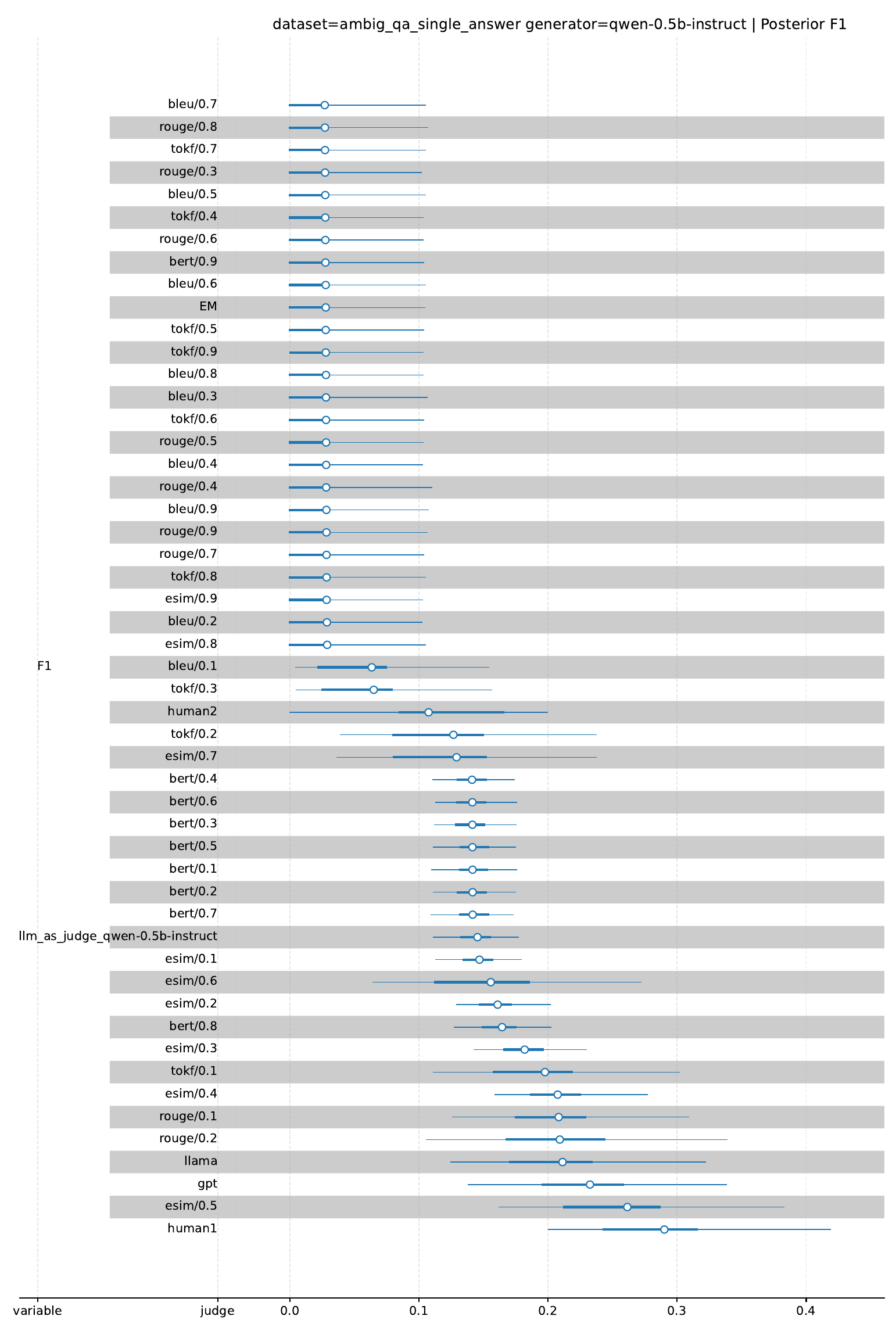}
    \end{subfigure}

    \caption{F1 distributions for AmbigQA (non-ambiguous) dataset and Llama 8B (top left corner), Llama 3B (top right corner), Qwen 7B (bottom left corner) and Qwen 0.5B (bottom right corner).}
    \label{fig:f1_ambigqa_single}
\end{figure*}

\begin{figure*}[t]
    \centering

    \begin{subfigure}[t]{0.48\textwidth}
        \centering
        \includegraphics[width=\linewidth]{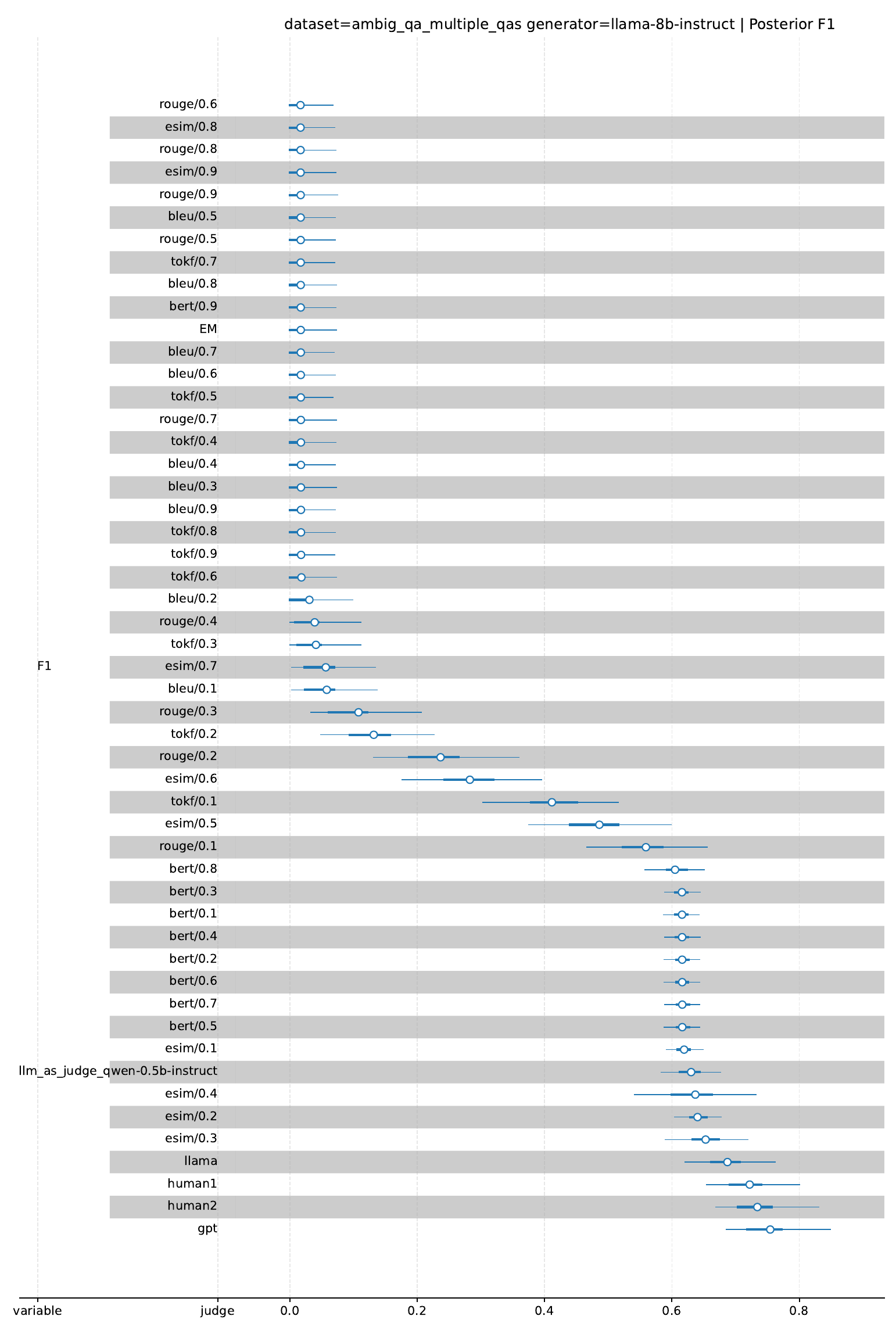}
    \end{subfigure}
    \hfill
    \begin{subfigure}[t]{0.48\textwidth}
        \centering
        \includegraphics[width=\linewidth]{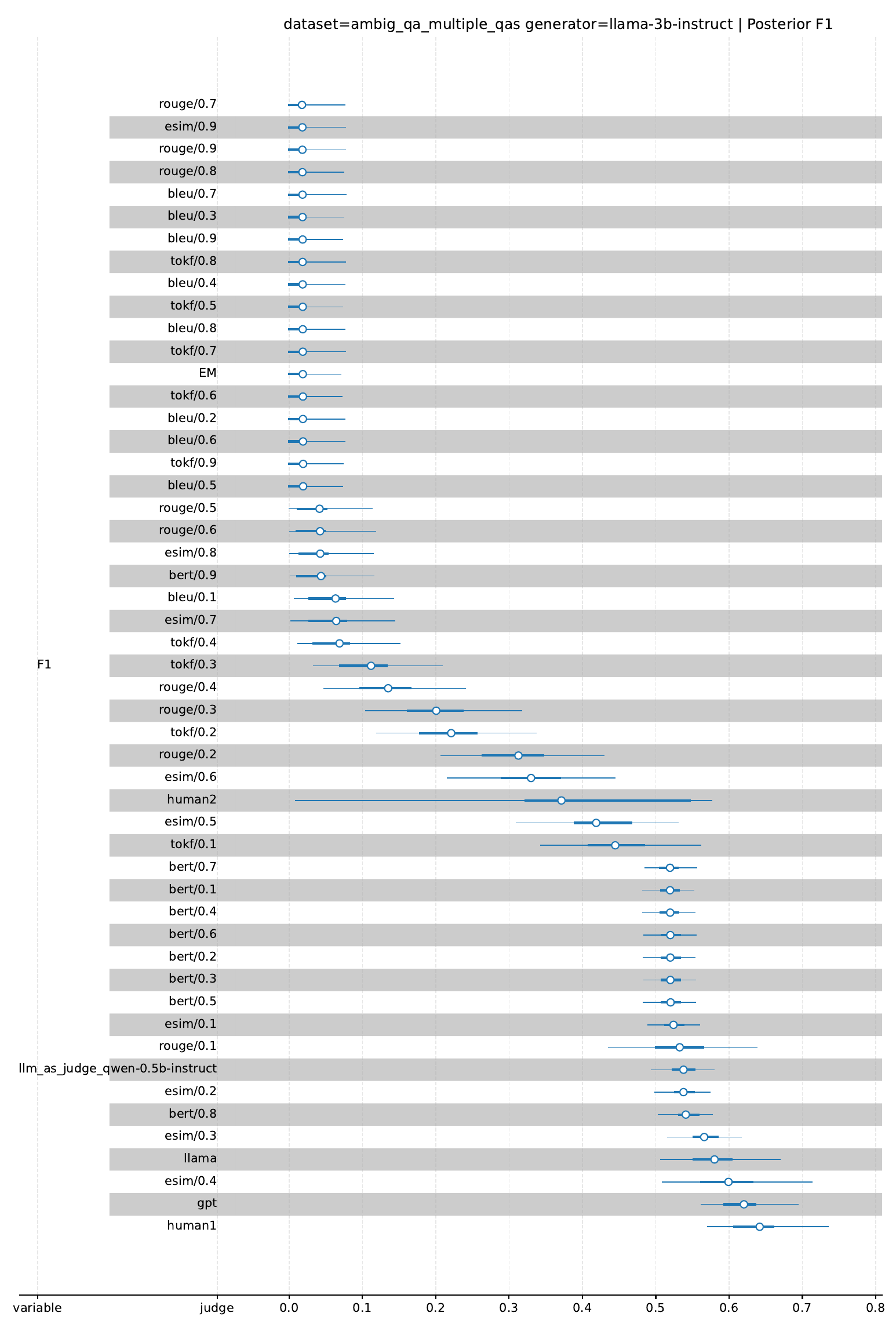}
    \end{subfigure}

    \vspace{0.75em}

    \begin{subfigure}[t]{0.48\textwidth}
        \centering
        \includegraphics[width=\linewidth]{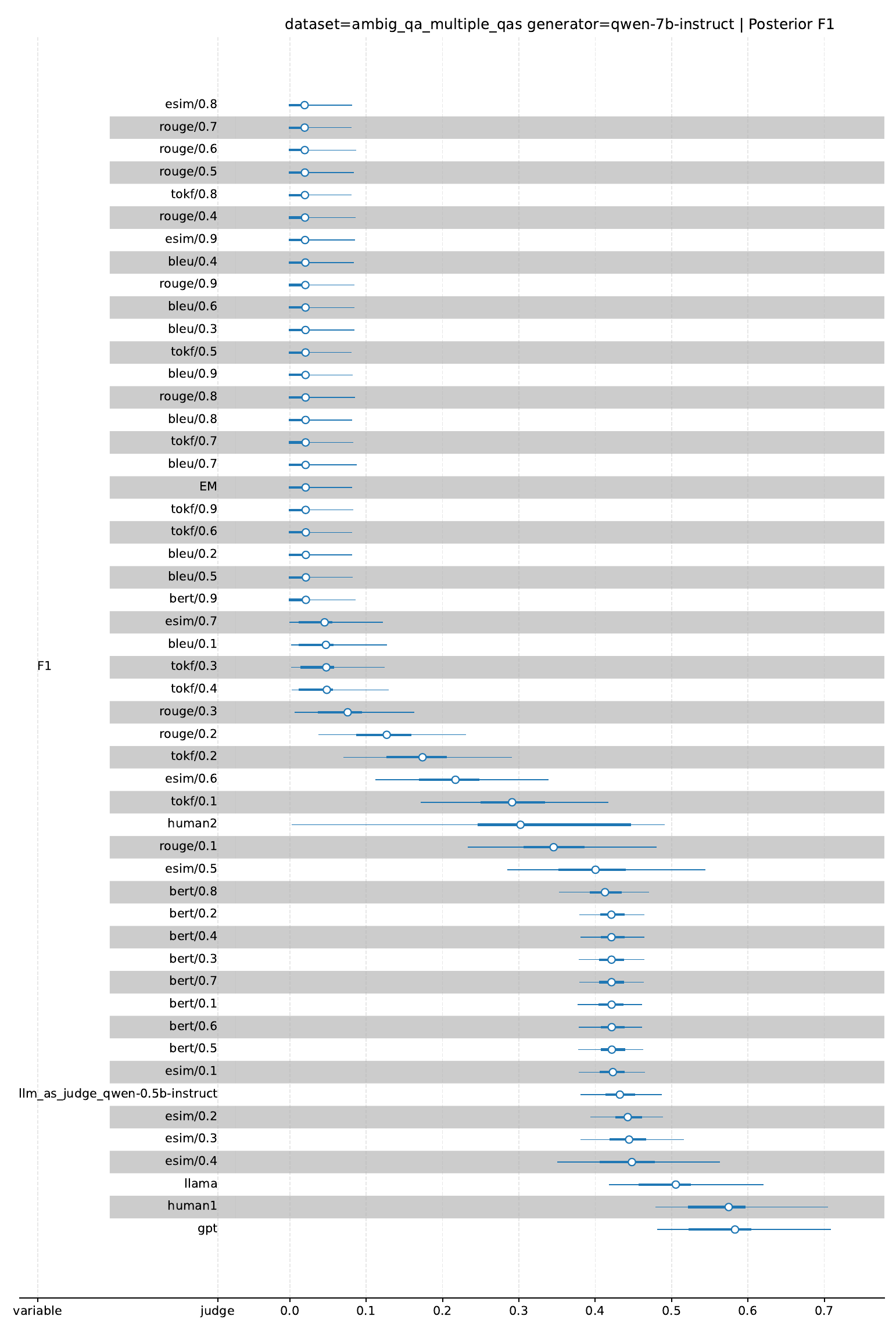}
    \end{subfigure}
    \hfill
    \begin{subfigure}[t]{0.48\textwidth}
        \centering
        \includegraphics[width=\linewidth]{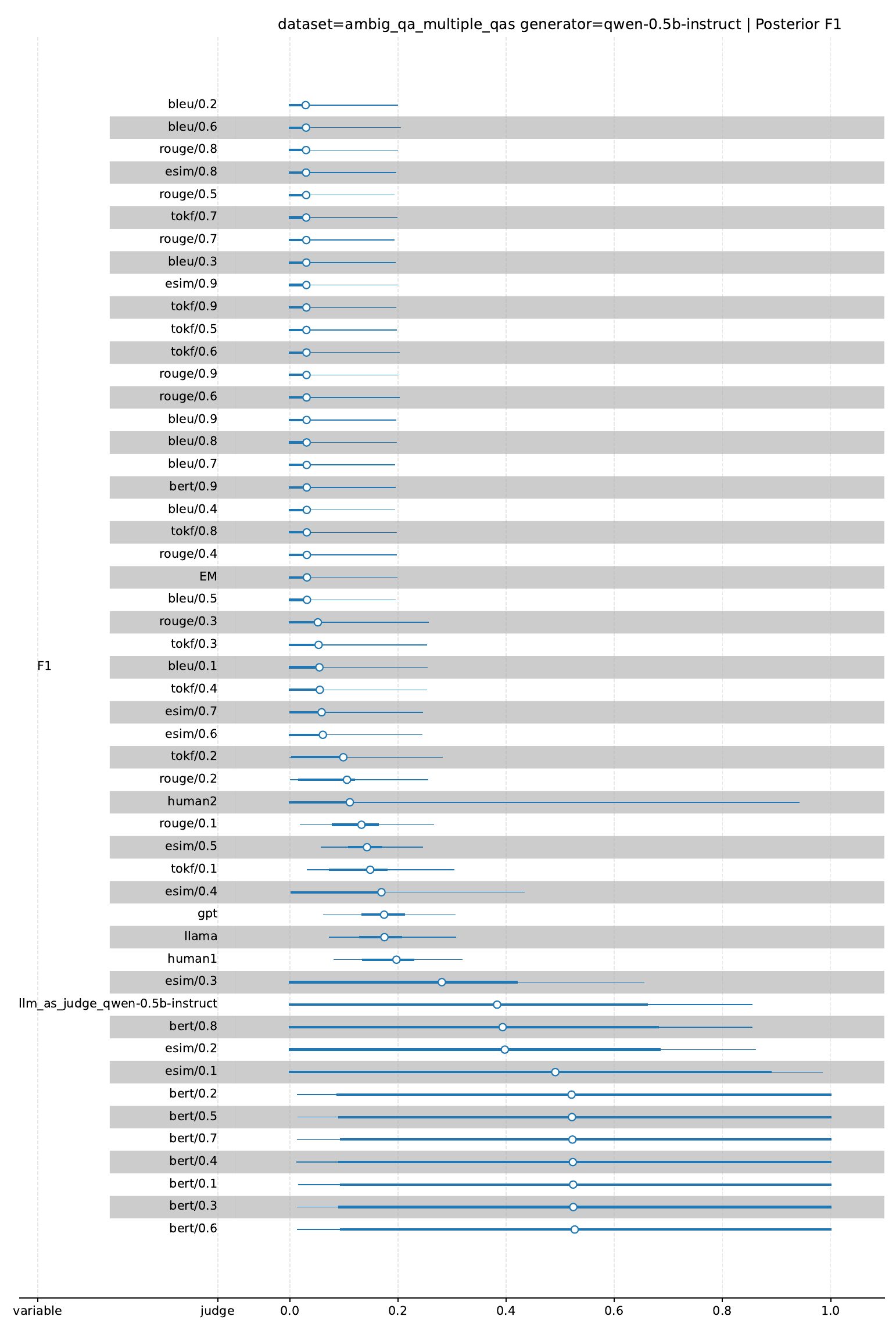}
    \end{subfigure}

    \caption{F1 distributions for AmbigQA (ambiguous) dataset and Llama 8B (top left corner), Llama 3B (top right corner), Qwen 7B (bottom left corner) and Qwen 0.5B (bottom right corner).}
    \label{fig:f1-ambigqa_mul}
\end{figure*}

For transparency, we present all F1 posterior distributions for all dataset-generator pairs. They are shown in Figures \ref{fig:f1_triviaqa} - \ref{fig:f1-ambigqa_mul}.

\subsection{Ranking correlations}
\label{appendix:rbo_plots}
\begin{figure*}[h!]
    \centering
    \includegraphics[width=\linewidth]{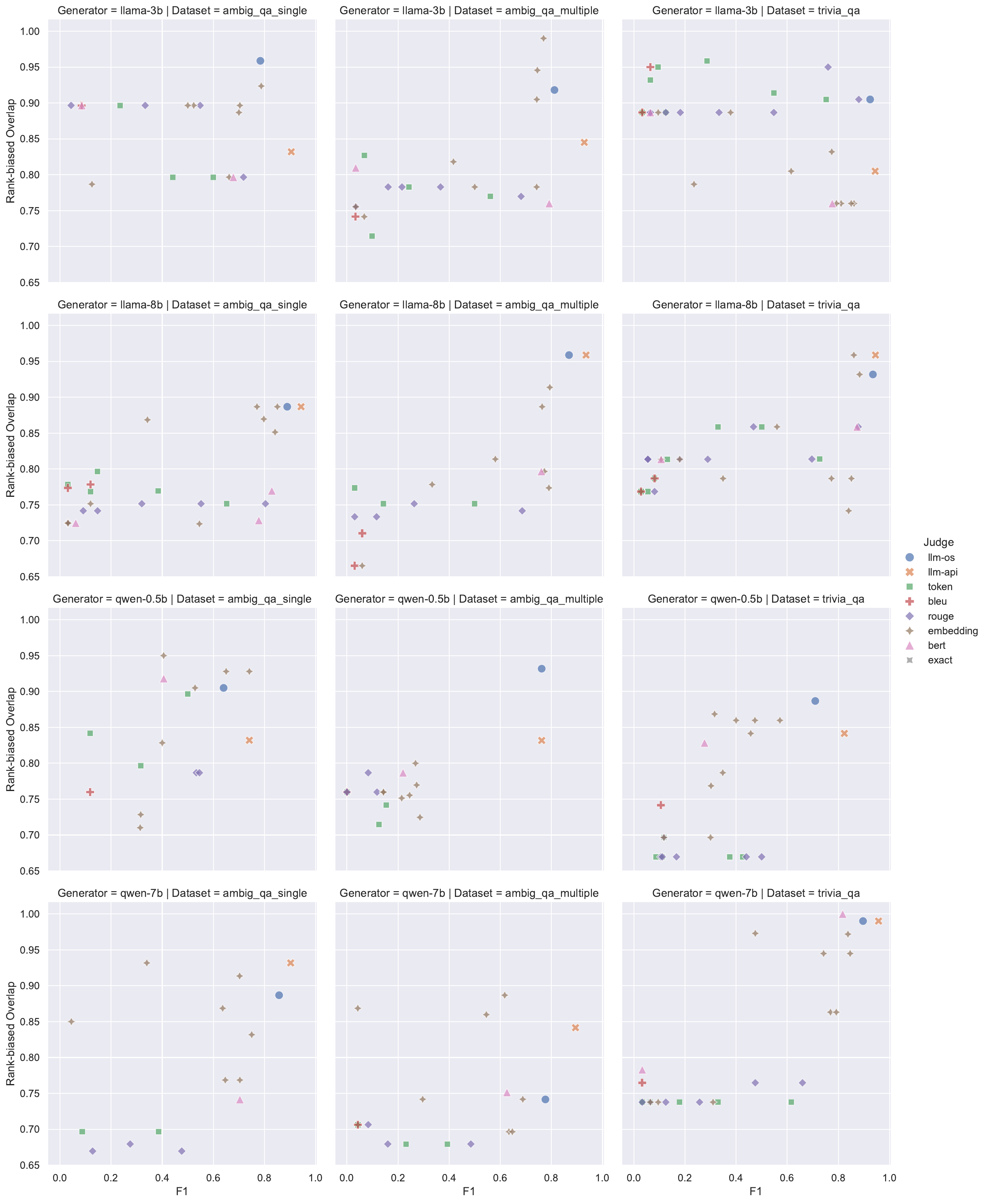}
    \caption{F1 scores of automated judges against RBO scores for all dataset-generator pairs.}
    \label{fig:RBO_all}
\end{figure*}

We show the plots representing F1 scores of automated judges against RBO scores for all dataset-generator pairs in \Cref{fig:RBO_all}

\end{document}